\documentclass{article}

\usepackage[main, final]{neurips_2026}

\usepackage[utf8]{inputenc} % allow utf-8 input
\usepackage[T1]{fontenc}    % use 8-bit T1 fonts
\usepackage{hyperref}       % hyperlinks
\usepackage{url}            % simple URL typesetting
\usepackage{booktabs}       % professional-quality tables
\usepackage{amsfonts}       % blackboard math symbols
\usepackage{nicefrac}       % compact symbols for 1/2, etc.
\usepackage{microtype}      % microtypography
\usepackage{xcolor}         % colors

\usepackage{amsmath}
\usepackage{amssymb}
\usepackage{amsthm}
\usepackage{kotex}
\usepackage{algorithm}
\usepackage{algorithmic}
\usepackage{multirow}
\usepackage{multicol}
\usepackage{booktabs}
\usepackage{multirow}
\usepackage{makecell}
\usepackage{adjustbox}
\usepackage{svg}
\usepackage{wrapfig}
\usepackage[table]{xcolor}
\usepackage{stfloats}
\definecolor{PRKV}{RGB}{255,248,220}
\usepackage{subfigure}
\usepackage{subcaption}

\newtheorem{theorem}{Theorem}[section]

\newtheorem{lemma}[theorem]{Lemma}

\theoremstyle{definition}

\theoremstyle{remark}

\title{Global Importance Estimation for KV Cache Eviction}

\author{%
  Seojin Kim\\
  KAIST \\
  Daejeon, Republic of Korea \\
  \texttt{seojinkim@kaist.ac.kr} \\
  \And
    Noseong Park\thanks{Corresponding author.} \\
    KAIST \\
    Daejeon, Republic of Korea \\
    \texttt{noseong@kaist.ac.kr}
}

\begin{document}

\maketitle

\begin{abstract}
Memory consumption of the key-value (KV) cache remains a central bottleneck for long-context LLM inference. We introduce PRKV, a training-free KV eviction method that estimates multi-hop global token importance by applying Personalized PageRank over attention-induced transition graphs. This enables importance propagation beyond recency or instantaneous attention, without modifying the model. We benchmark PRKV on long-context understanding, sparse retrieval, and chain-of-thought reasoning tasks using multiple LLM families. Under moderate to aggressive compression ratios, including up to 90\%, PRKV achieves competitive task performance against established eviction baselines while reducing peak memory usage with low end-to-end overhead. These results demonstrate that multi-hop global influence estimation yields a favorable accuracy--memory trade-off for scalable long-context inference.
\end{abstract}

\section{Introduction} \label{sec:introduction}

\begin{wrapfigure}[18]{r}{0.4\textwidth}   
\centering
\vspace{-2.5em}
\includegraphics[width=\linewidth]{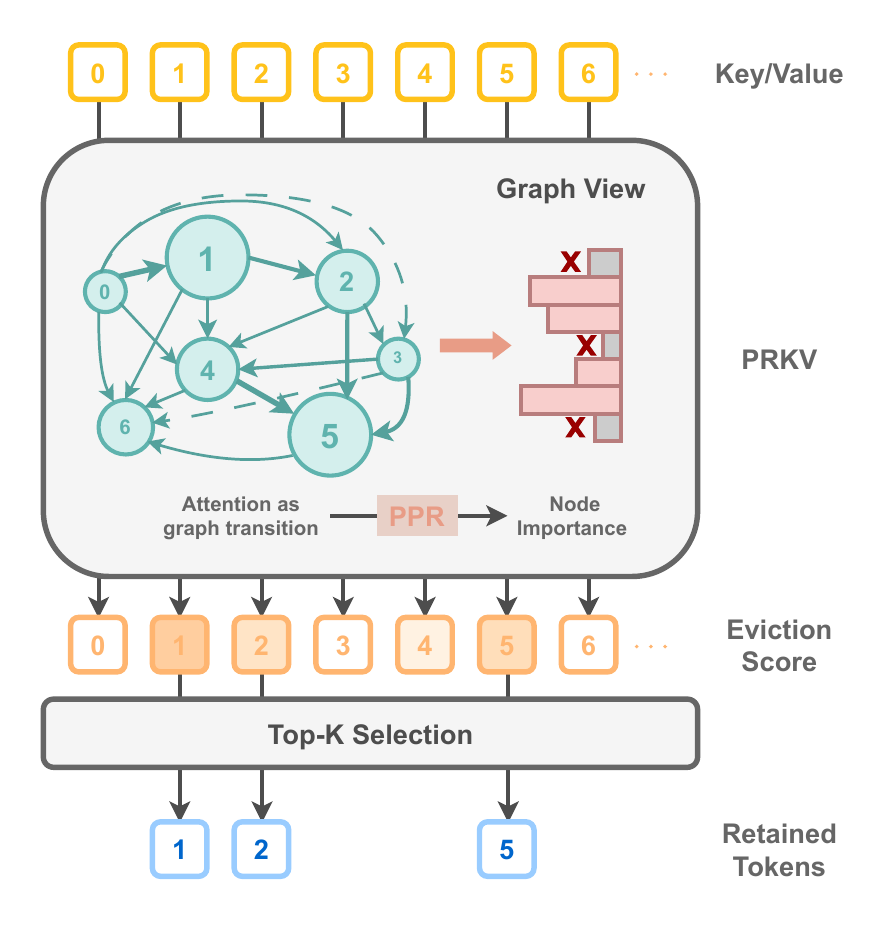}
\vspace{-1.5em}
\caption{
\textbf{Overview of PRKV.}
PRKV propagates token influence over an attention-induced transition graph to select globally important KV states.
}
\label{fig:method}
\end{wrapfigure}

Large language models (LLMs) increasingly rely on long-context inference, where key-value (KV) caching is essential for efficient autoregressive decoding. However, the KV cache grows linearly with sequence length, making memory management a central bottleneck for deployment.

A growing line of research therefore aims to reduce the KV memory footprint. Architectural approaches such as state-space models~\citep{gu2023mamba}, linear RNN variants~\citep{peng2023rwkv}, or modified attention mechanisms including Grouped Query Attention~\citep{ainslie2023gqa} and Multi-head Latent Attention~\citep{liu2024deepseek} improve asymptotic efficiency, but require non-trivial architectural changes and retraining. In contrast, post-training approaches reduce memory at inference time by compressing cached tokens, including eviction or selection~\citep{zhang2023h2o,liu2023scissorhands}, merging~\citep{liu2024minicache}, quantization~\citep{hooper2024kvquant,liu2024kivi}, or low-rank projection~\citep{chang2025palu,sun2024shadowkv}.

Among post-training approaches, eviction is particularly attractive because it can be applied directly to existing models. Existing eviction policies commonly rely on local usage signals such as recency~\citep{xiao2024efficient} or instantaneous attention~\citep{li2024snapkv, oren2024transformers}. These signals are useful and often temporally persistent~\citep{liu2023scissorhands}, but they mainly capture direct, one-hop usage. Prefix-level aggregation methods introduce broader context~\citep{chen2024nacl}, yet still summarize direct usage statistics rather than propagating influence among tokens.

This motivates viewing KV eviction as global influence estimation. Important tokens are not necessarily those with the largest immediate attention, but those that affect future computation through multi-step attention paths, consistent with prior work on attention diffusion over token graphs~\citep{feng2023diffuser}. Since full-graph influence computation is impractical in long-context inference, the key challenge is to approximate such multi-hop importance from local or recently observed attention while remaining training-free and efficient.
To this end, we propose \textbf{PRKV}, a training-free KV eviction framework that estimates global token importance with Personalized PageRank (PPR). PRKV uses two scalable transition operators: a landmark-augmented node graph for sharper relevance paths, and a global co-attention graph for smoother propagation over all cached keys.

We evaluate PRKV on multiple instruction-tuned LLM families across benchmarks covering long-context understanding (RULER, LongBench), chain-of-thought reasoning (AIME25, MATH-500), and sparse-evidence retrieval (Needle-in-a-Haystack, NIAH). Across moderate to aggressive compression budgets, PRKV variants achieve a favorable accuracy--memory trade-off against strong eviction baselines, while reducing peak KV memory with low end-to-end overhead. 
Our contributions are summarized as follows:
\begin{itemize}
\item We introduce \textbf{PRKV}, a practical, training-free KV eviction method that estimates global token importance by applying PPR to attention-induced transition graphs.
\item We instantiate PRKV with two scalable transition operators, including a landmark-augmented node variant and a global co-attention variant for different long-context regimes.
\item We evaluate PRKV on multiple long-context LLMs and five benchmarks, demonstrating a favorable accuracy--memory trade-off across a wide range of compression budgets.
\item We provide empirical analyses of PPR-based importance propagation, including sensitivity studies, qualitative case analyses, and memory--latency behavior.
\end{itemize}

\section{Background}

\subsection{KV Caching and Eviction}
\label{sec:background_kv_cache}
We describe attention for a single layer and head and omit layer and head indices for notational simplicity. 
Let $h_i \in \mathbb{R}^{1\times d}$ denote the hidden representation of token $i$. Each token is projected to a query, key, and value:
\begin{equation}
  q_i = h_i W_Q^\top, \quad k_i = h_i W_K^\top, \quad v_i = h_i W_V^\top,
\end{equation}
where $W_Q \in \mathbb{R}^{d_k \times d}, \;
       W_K \in \mathbb{R}^{d_k \times d}, \;
       W_V \in \mathbb{R}^{d_v \times d}$,
and $d_k$ and $d_v$ are the key (and query) and value dimensions, respectively.
During autoregressive decoding, at step $t+1$ only the new query, key, and value
$(q_{t+1}, k_{t+1}, v_{t+1})$ are computed, while keys and values from previous
tokens are stored in a KV cache and reused by all future queries.
Let $C_t = (K_t, V_t)$ denote the KV cache after processing tokens
$\{1,\dots,t\}$, where
$K_t = [k_1; \dots; k_t] \in \mathbb{R}^{t \times d_k}, \quad
V_t = [v_1; \dots; v_t] \in \mathbb{R}^{t \times d_v}$.
The attention output at step $t+1$ is computed over the cached keys and values:
\begin{equation} \label{eq:attn}
o_{t+1} = \mathrm{Attn}\!\left(
q_{t+1},
[K_t \,\|\, k_{t+1}],
[V_t \,\|\, v_{t+1}]
\right) = \phi\!\left(
\frac{ q_{t+1} K_{t+1}^\top }{\sqrt{d_k}}
\right)
V_{t+1} = \sum_{i=1}^{t+1} A_{t+1,i}\, v_i,
\end{equation}
where $\phi$ denotes the softmax function applied over keys and $A_{t+1,:} \in \mathbb{R}^{1 \times (t+1)}$ is the row-stochastic attention distribution over keys produced by the softmax.
Once stored, a KV pair $(k_i, v_i)$ may contribute to the attention outputs of
all future decoding steps $j > i$. KV caching reduces decoding latency by reusing past keys and values, but its memory cost grows linearly with sequence length. To bound memory, eviction methods enforce a fixed cache budget $K$, restricting the cache at step $t$ to a subset $\mathcal{S}_t \subseteq \{1,\dots,t\}$ with $|\mathcal{S}_t| \le K$ after each compression. Tokens are assigned importance scores $s_i^{(t)} \ge 0$, and only the top-$K$ tokens are retained for future attention.

\subsection{Attention as a Directed Graph}
Self-attention can be interpreted through a graph-theoretic lens, where tokens correspond to nodes and attention weights define directed, weighted edges among them~\citep{wang2022anti,shi2022revisiting,choi2024graph,wi2025learning}. At decoding step $t{+}1$, applying a row-wise softmax yields the per-step attention distribution $A_{t+1,:} \in \mathbb{R}^{1 \times (t+1)}$, where $A_{t+1,j} \ge 0$,  $\sum_{j=1}^{t+1} A_{t+1,j} = 1.$
Conceptually, $A_{t+1,j}$ specifies the weight of a directed edge from node $(t{+}1)$ to node $j$, and repeated decoding steps reveal a row-stochastic adjacency matrix. This induces a Markov transition structure over tokens, allowing influence to propagate across multiple hops rather than single-step interactions~\citep{feng2023diffuser,lee2025mitigating}. In this regard, long-range token importance is naturally captured by random-walk based centrality measures such as PageRank.

\subsection{PageRank and Importance Aggregation} \label{sec:backgroudn_ppr}
Given an attention-induced Markov chain over tokens, PageRank provides a natural mechanism for quantifying global importance on the resulting directed graph.
Originally introduced to rank web pages via link structure \citep{page1999pagerank}, PageRank characterizes the stationary distribution of a random walk with restarts, capturing the long-term visitation frequency of nodes.
Given a row-stochastic transition matrix $P \in \mathbb{R}^{n \times n}$, the PageRank vector $\pi \in \mathbb{R}^n$ is defined as the stationary distribution satisfying
\begin{equation}
\pi = (1-\alpha)\,\mathbf{u} + \alpha\, \pi P,
\end{equation}
where $\alpha \in (0,1)$ is the damping factor and $\mathbf{u}$ denotes
the uniform distribution. Personalized PageRank (PPR) generalizes this formulation by replacing the uniform teleportation distribution with a personalization vector $u \in \Delta^n$ where $\Delta^n = \{ x \in \mathbb{R}^n_{\ge 0} \mid \sum_i x_i = 1 \}$ ~\citep{haveliwala2002topic}, resulting in the fixed-point equation
\begin{equation}\label{eq:ppr}
\pi(P,u) = (1-\alpha)\, u + \alpha\, \pi(P,u)\, P.
\end{equation}
PPR accumulates influence via repeated transitions, making it well suited for token importance.

\section{Motivational Analyses}
\label{sec:motivation}

\begin{figure*}[t]
    \centering
    \vspace{-1em}
    \subfigure[Top-$K$ Overlap
    \label{fig:motiv_analysis_topk}]{
          \includegraphics[width=0.23\linewidth]{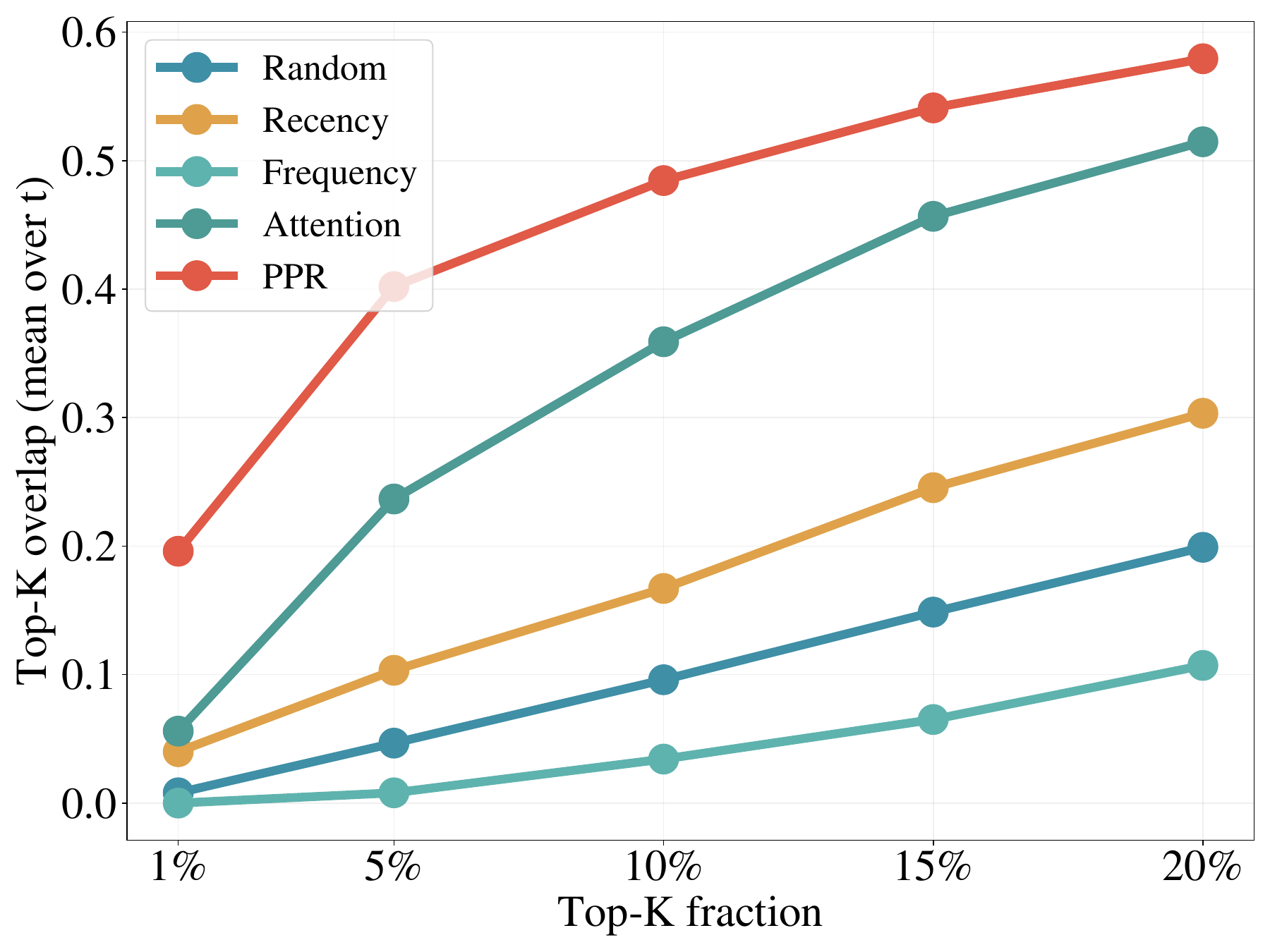}
    }
    \subfigure[Stability\label{fig:motiv_analysis_stability}]{
          \includegraphics[width=0.23\linewidth]{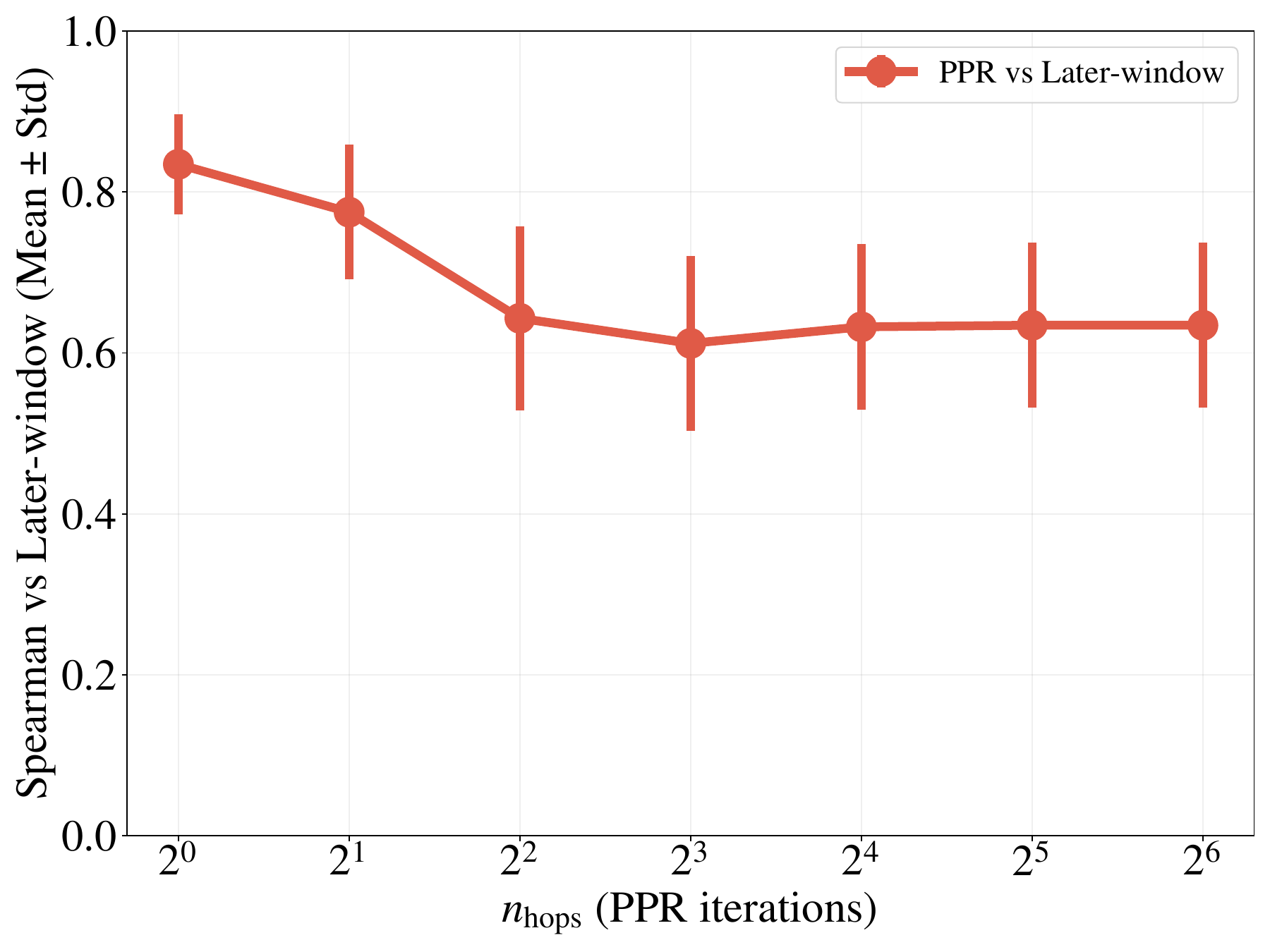}
    }
    \subfigure[Attention Retention AUC\label{fig:motiv_analysis_auc}]{
          \includegraphics[width=0.23\linewidth]{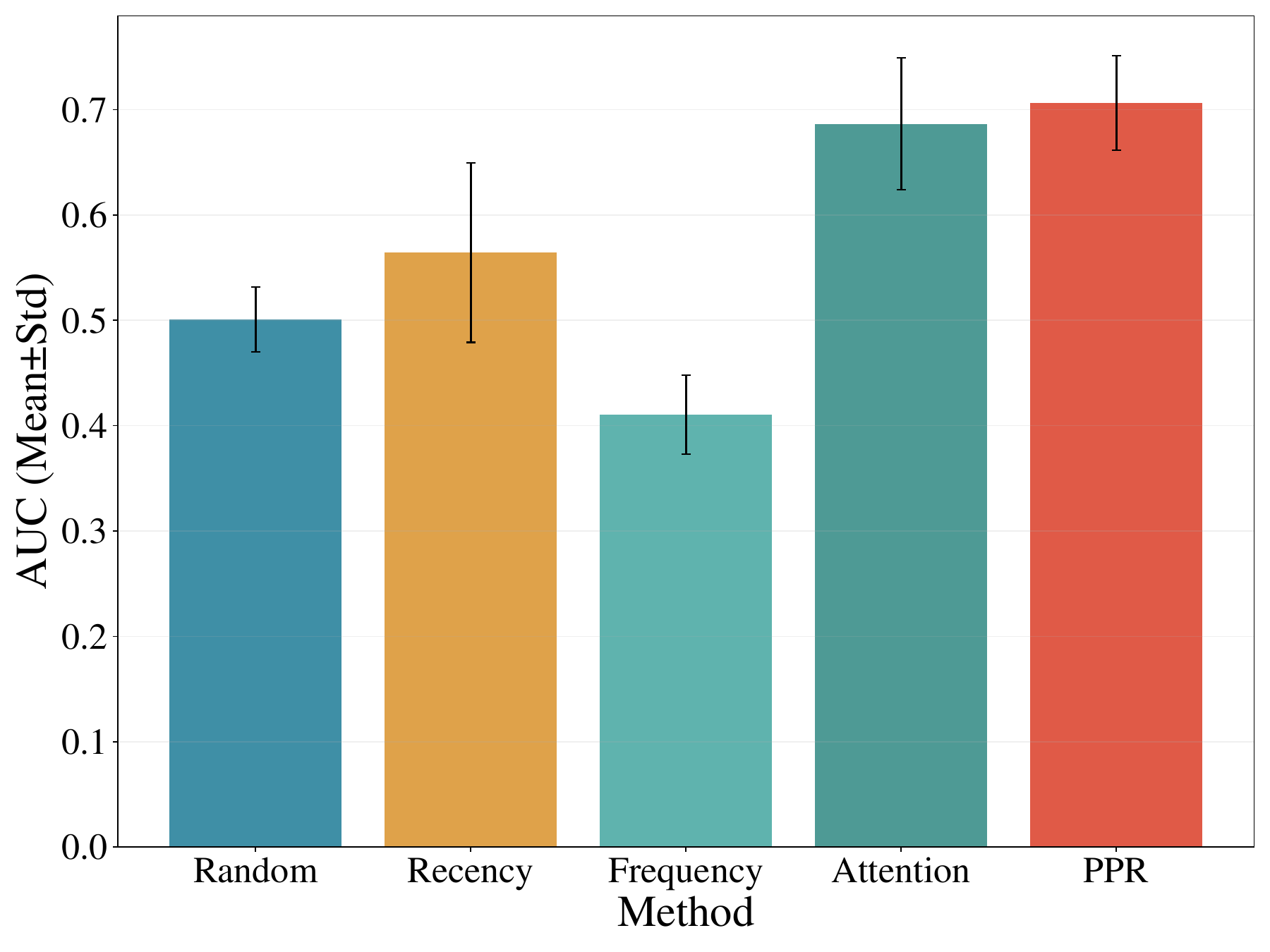}
    }
    \subfigure[KL divergence\label{fig:motiv_analysis_kl}]{
          \includegraphics[width=0.23\linewidth]{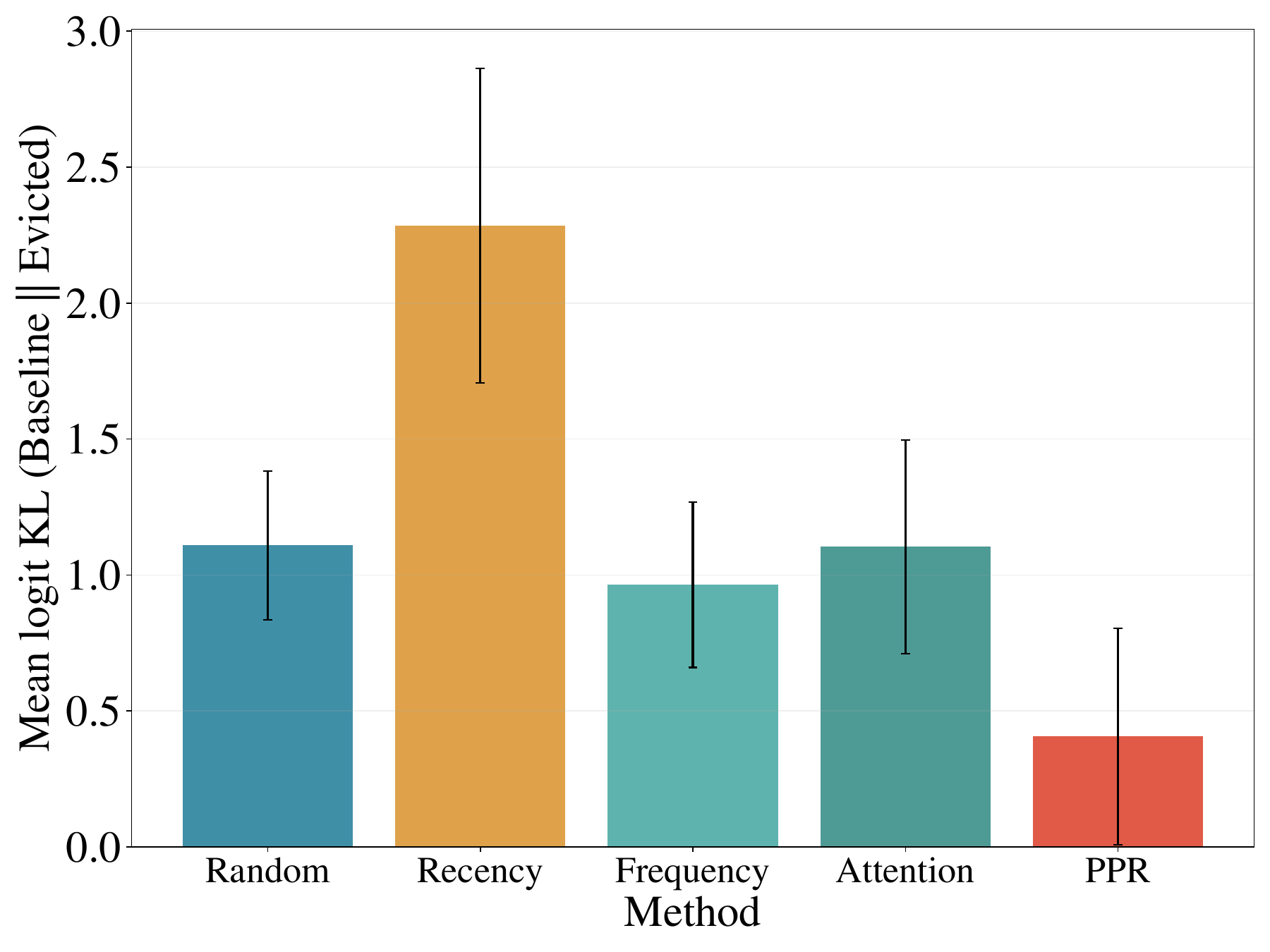}
    }
    \caption{
    \textbf{Motivational analysis of PPR.}
    (a) PPR recovers stable high-importance tokens under budgeted eviction, capturing long-range influence.
    (b) PPR scores rapidly stabilize within a few propagation steps.
    (c) PPR shows stronger alignment with downstream attention than baselines.
    (d) PPR induces the smallest eviction-induced output distortion, preserving long-range influence.
    Results are reported on 25\textsuperscript{th} layer of Llama-3.1-8B.
    }
    \vspace{-1em}
    \label{fig:motivational_analysis}
\end{figure*}

\paragraph{PPR as a global importance proxy}
\label{sec:motivation_pagerank}
In long-context inference, KV eviction should preserve tokens whose removal would impact future computation, requiring importance estimates that go beyond local attention statistics.
Prior work shows that token importance exhibits short-term temporal persistence during decoding, making recent attention a meaningful signal of future relevance~\citep{liu2023scissorhands}, and that multi-hop diffusion on attention graphs recovers longer-range token interactions~\citep{feng2023diffuser}.
Viewing self-attention as inducing a directed token-transition graph, these observations suggest that aggregating local attention signals across hops can better approximate longer-horizon influence.
We adopt PPR to operationalize this view, providing a proxy for sustained token importance.

\paragraph{Empirical validation of PPR importance}
We conduct empirical analyses on the PaulGrahamEssays corpus with Llama-3.1-8B using a multi-anchor evaluation protocol to reduce sensitivity to evaluation position (Figure~\ref{fig:motivational_analysis}). The analyses examine whether importance scores computed at an anchor can predict the subsequent usage of currently cached tokens and support eviction with limited output distortion. We compare against random scoring (Random), positional recency (Recency), token-ID frequency in the observed prefix (Frequency), and direct attention from the anchor (Attention). In this diagnostic, PPR uses the same anchor-attention signal for both initialization and personalization, isolating the effect of graph propagation. Detailed experimental settings and definitions of the baselines and metrics are provided in Appendix~\ref{appendix:motiv_details}.

Figure~\ref{fig:motiv_analysis_topk} shows that PPR achieves higher top-$K$ overlap with the later-window attention ranking than the compared baselines. Under the same retention budget, PPR therefore selects more of the cached tokens that subsequently receive high attention. Its improvement over direct anchor attention suggests that propagating the observed attention signal through the token graph helps predict subsequent usage beyond the anchor's immediate attention pattern.
Figure ~\ref{fig:motiv_analysis_stability} further shows that the Spearman correlation between PPR scores and later-window attention rapidly stabilizes after a small number of iterations, retaining a clear positive correlation. Most of the full-ranking agreement is thus attained early in propagation, and further iterations do not substantially change the ranking.

Figure~\ref{fig:motiv_analysis_auc} complements the top-$K$ comparison by measuring how much later-window attention mass is preserved by the retained tokens. Whereas top-$K$ overlap measures agreement on which tokens are selected, retained attention mass accounts for how much those tokens are subsequently used. PPR preserves more of this mass than the baselines, indicating that its selected tokens collectively account for a larger share of subsequent attention, rather than merely matching more members of the reference top-$K$ set.
Finally, Figure~\ref{fig:motiv_analysis_kl} evaluates whether the selected cache also preserves model outputs. PPR-based eviction yields the smallest KL divergence from the full-KV output distribution among the compared methods. This provides complementary, output-level evidence that the attention-based improvements are accompanied by reduced eviction-induced distortion.

Together, these results show that PPR better identifies tokens with high subsequent attention usage while more closely preserving the full-KV output distribution in the evaluated setting, supporting its use as an importance proxy for budgeted KV eviction.

\section{Method}
\label{sec:method}
\subsection{Overview of PRKV}
\label{sec:method_overview}

PRKV is a training-free eviction method that maintains a fixed-size KV cache by estimating token influence through importance propagation on the attention-induced graph, as illustrated in Figure~\ref{fig:method}.
Concretely, PRKV initializes importance with a recency-biased personalization vector, iteratively propagates importance through attention graph, and evicts the lowest-scoring tokens until the cache meets a target budget. This yields a compact representation of long-range influence robust to transient attention patterns. We have two variants of PRKV that follow this same eviction pipeline introduced in Section~\ref{sec:method_unified}. They differ in how attention observations are aggregated to propagate importance. 

PRKV integrates into inference-time KV management and supports two eviction modes: (i) prefill eviction, which compresses the cache once after processing the full input, and (ii) decoding eviction, which triggers periodic evictions during autoregressive generation to maintain the budget. Both share the same scoring pipeline; the difference between these two modes is when evictions are invoked. The full eviction algorithm is summarized in Appendix~\ref{appendix:algorithm}.

\subsection{Unified PRKV Framework}
\label{sec:method_unified}
We formulate importance scoring in terms of a transition operator $Q$ that specifies how importance propagates between tokens. For a given $Q$, the formulation below applies to both variants, while their respective constructions of the operator are described in the next subsection.

\paragraph{Recency-biased personalization vector}
We use a shared recency-biased personalization vector that assigns uniform base mass and boosts the most recent fraction.
For $n=\max\{1,\lfloor fW\rfloor\}$, define
\begin{equation}
u_i \;\propto\; 1 + (b-1)\,\mathbf{1}\{i>W-n\},
\qquad i=1,\dots,W,
\end{equation}
with normalization $u \leftarrow u/\sum_i u_i$, where $f$ denotes recent fraction and $W$ denotes window size, and $b$ denotes boost constant. This choice reflects the inductive bias of attention mechanisms~\citep{xiao2024efficient} toward recent tokens while remaining computationally lightweight.

\paragraph{Generic PPR iteration}
Given a personalization vector $u$ and a transition operator $Q$, we compute PPR by iterating
\begin{equation}
\label{eq:ppr_iter}
r^{(k+1)} \;=\; \alpha\,r^{(k)} Q + (1-\alpha)\,u.
\end{equation} 
We compute $n_\text{hops}$ iterations and use $\hat{\pi} = r^{(n_\text{hops})}$ as an approximation to the PPR vector $\pi$ in Eq.~\ref{eq:ppr}. This procedure is summarized in Algorithm~\ref{alg:prkv} of Appendix~\ref{appendix:algorithm}. 

\paragraph{Value-aware importance scaling}
All variants produce the approximated PPR vector that reflects how frequently tokens are expected to be accessed under attention-driven PPR.
However, the impact of a token on the residual stream also depends on the magnitude of its value representation~\citep{feng2025identifycriticalkvcache,devoto2025expectedattention}. For a single head, the contribution of token $i$ to the residual at time $t$ is $\Delta h_{t,i} = A_{t,i} v_i W_o$. Then, the norm of token contribution satisfies 
\begin{equation}
\|\Delta h_{t,i}\|_2 = A_{t,i}\,\|v_i W_o\|_2.
\end{equation}
Thus, the value norm acts as a payload term that modulates the effect of each access. We therefore convert $\hat{\pi}$ into an eviction score by scaling with the value norm:
\begin{equation}
s_i = (\hat{\pi}_i + \epsilon)\,\|v_i W_o\|_2,
\end{equation}
omitting $W_o$ and using $\|v_i\|_2$, which is exact when values are already projected into the residual space and serves as a proxy otherwise. This scaling yields a surrogate for the expected residual contribution rather than an ad-hoc heuristic. Scores are then used for Top-$K$ selection within the cache manager and sink tokens are assigned $+\infty$ prior to Top-$K$ selection. 

\subsection{Transition Operators from Local Attention}
\label{sec:method_transitions}

\paragraph{PRKV: Landmark-augmented node transition}
\label{sec:method_landmark}

To summarize long-range context while keeping computation bounded, we form a node graph consisting of (i) a small set of landmark tokens selected from the distant prefix and (ii) all tokens in the recent window. Each node corresponds to a token position, so we reuse the underlying token-level query and key vectors $q_i$ and $k_i$ for node attention. 

We first select a small set of prefix landmarks using cosine diversity. 
We partition the prefix into contiguous chunks of size $C$ and compute a global anchor direction $\mathbf{k}$ as the normalized mean direction of the prefix keys.
For each chunk $\mathcal{C}_c$ we select
$\ell_c \;=\; \arg\min_{i \in \mathcal{C}_c} \langle \tilde{k}_i, \mathbf{k} \rangle,$
where $\tilde{k}_i$ denotes the normalized key. Let $\mathcal{L} = \{\ell_c\}$ denote the resulting set of prefix landmarks. This diversity-driven criterion avoids selecting redundant prefix tokens and suggests that distant semantic regions are represented, which aims to improve coverage under tight memory budgets.

In prefill eviction, where prefix hidden states remain materialized, landmarks are included as nodes together with the recent window:
\begin{equation}
\mathcal{N} = \mathcal{L} \cup \{T-W+1,\dots,T\},
\label{eq:eq9}
\end{equation}
with $N = W + |\mathcal{L}|$ and $|\mathcal{L}| \le \left\lceil \frac{T-W}{C} \right\rceil$. We compute node-to-node attention with the standard scaled dot-product form:
\begin{equation}
A_{\mathrm{node}}[p,j]
= \phi\!\left( \frac{q_p k_j^\top}{\sqrt{d_k}} \right),
\qquad p,j\in\mathcal{N}.
\end{equation}
We apply a causal mask based on the original token positions before softmax, setting $A_{\mathrm{node}}[p,j]=0$ for $j>p$. This yields a row-stochastic transition $Q_{\mathrm{node}}=A_{\mathrm{node}}$ for PPR iteration (Algorithm~\ref{alg:apply-nodes}). In prefill, landmarks participate in propagation and remain subject to the final PPR/value-aware top-$K$ selection.

During decoding, prefix hidden states are not stored, so landmarks cannot participate as nodes. In this case, $\mathcal{L}$ are retained in the KV cache and excluded from eviction, while the node set reduces to the recent window,
\begin{equation}    
\mathcal{N}_t = \{t-W+1,\dots,t\},
\label{eq:eq11}
\end{equation}
allowing PRKV to operate without full-prefix storage.

\paragraph{PRKV-GLB: Global co-attention transition}
\label{sec:method_global}
To approximate global importance over all cached tokens using recently observed attention, we construct a transition operator based on key-key co-attention. Unlike token-level attention, this construction aggregates symmetric affinity between keys based on how similarly they are attended by the window. At each inference step, we observe causal attention over the last $W$ query positions. After applying softmax, we obtain a window attention matrix $A \in \mathbb{R}^{W \times T}$. 
We first define an unnormalized co-attention affinity
\begin{equation}
S := A^\top A \in \mathbb{R}^{T \times T},
\qquad
S_{ij} = \sum_{q=1}^W A_{q i} \, A_{q j},
\end{equation}
which measures how strongly keys $i$ and $j$ are jointly attended by recent queries. Averaging attention over queries summarizes the observed usage of each key with a single score. The co-attention affinity $S_{ij}$ additionally captures how strongly keys $i$ and $j$ are attended by the same queries, providing the pairwise connections through which PPR propagates importance. Let $d := S \mathbf{1}_T = A^\top \mathbf{1}_W$ denote the row-sum vector of $S$, and define $D := \mathrm{diag}(d).$ We then obtain a row-stochastic global transition operator
\begin{equation}    
Q_{\mathrm{glob}} := D^{-1} S,
\qquad
Q_{\mathrm{glob}} \mathbf{1}_T = \mathbf{1}_T,
\end{equation}
where $d>0$, which defines a valid Markov chain over all $T$ keys. Intuitively, $Q_{\mathrm{glob}}$ diffuses importance across keys that tend to co-occur under attention, enabling global influence propagation beyond the observed window. This construction exploits the implicit low-rank structure of the co-attention affinity: since $S = A^\top A$ has rank at most $W$, we can apply $Q_{\mathrm{glob}}$ to a vector $r$ without materializing $S$ via the factorized product
\begin{equation}
r Q_{\mathrm{glob}} = r D^{-1} A^\top A.
\end{equation}

As a result, PPR iterations require only $O(WT)$ time per step, making global influence tractable even for long sequences (Algorithm~\ref{alg:apply-global}).

\paragraph{Variant interpretation.}
The choice of transition operator induces different scoring behavior. PRKV propagates importance over node transition, producing sharper scores from local or intermediate relevance paths. PRKV-GLB instead propagates over a global co-attention graph, aggregating weak evidence across the prefix and producing smoother, globally consistent scores (cf. Appendix~\ref{appendix:score_behavior}). Thus, PRKV-GLB is generally effective when relevance is sparse, distributed, or globally correlated, while PRKV can be preferable when local or intermediate attention paths are cleaner than global co-attention. We use this distinction to interpret both prefill and decoding eviction results in Section~\ref{sec:experiment}.

\section{Experiments} \label{sec:experiment}

\begin{table}
\caption{\textbf{Performance comparison on LongBench.}
We report the average of sixteen tasks from LongBench benchmarks over five compression ratios and three models. 0\% indicates average results without compression. Bold values indicate the best performance within each block.}
\label{tab:longbench-results}
\vspace{0.5em}
\resizebox{\textwidth}{!}{
\begin{tabular}{l|cccccc|cccccc|cccccc}
\toprule
    \multirow{2}{*}{\textbf{Method}} &
    \multicolumn{6}{c|}{\textbf{Llama-3.1-8B}} &
    \multicolumn{6}{c|}{\textbf{Qwen3-8B}} &
    \multicolumn{6}{c}{\textbf{Gemma-3-12B}} \\
    & 0\% & 10\% & 25\% & 50\% & 75\% & 90\% 
    & 0\% & 10\% & 25\% & 50\% & 75\% & 90\% 
    & 0\% & 10\% & 25\% & 50\% & 75\% & 90\% 
    \\
\midrule

SnapKV 
& \textbf{45.35} & 45.20 & 45.03 & 43.69 & 40.84 & 33.23
& \textbf{47.27} & 47.07 & 46.65 & 45.12 & 41.90 & 35.09 
& \textbf{49.79} & 49.45 & 49.36 & 47.37 & \textbf{44.36} & 37.27 \\
PyramidKV 
& \textbf{45.35} & 45.39 & 45.00 & 43.75 & 40.35 & 33.28
& \textbf{47.27} & 46.97 & 46.10 & 42.50 & 39.78 & 34.04 
& \textbf{49.79} & 49.65 & 48.99 & 46.74 & 43.01 & 36.73 \\
EA 
& \textbf{45.35} & \textbf{45.91} & \textbf{47.56} & 47.11 & 44.03 & 35.79
& \textbf{47.27} & \textbf{48.08} & \textbf{48.87} & 48.46 & 45.42 & 37.19
& \textbf{49.79} & 49.48 & 49.06 & 47.58 & 43.08 & 36.81 \\
TOVA 
& \textbf{45.35} & 45.20 & 44.59 & 43.07 & 39.58 & 33.96 
& \textbf{47.27} & 46.95 & 46.87 & 45.35 & 41.91 & 36.60
& \textbf{49.79} & 49.63 & 49.19 & 47.47 & 43.34 & \textbf{38.39} \\
KeyDiff 
& \textbf{45.35} & 45.24 & 47.40 & 45.68 & 41.40 & 34.50
& \textbf{47.27} & 46.83 & 43.85 & 38.57 & 29.17 & 20.63 
& \textbf{49.79} & 48.33 & 46.02 & 40.35 & 31.39 & 22.81 \\
\rowcolor{PRKV}
PRKV
& \textbf{45.35} & 45.72 & 47.50 & \textbf{47.47} & \textbf{44.40} & \textbf{36.88}
& \textbf{47.27} & 47.74 & 48.56 & \textbf{48.92} & \textbf{46.26} & 37.59 
& \textbf{49.79} & 49.57 & 48.89 & 47.54 & 43.26 & 35.82 \\
\rowcolor{PRKV}
PRKV-GLB
& \textbf{45.35} & 45.60 & 46.45 & 46.60 & 44.20 & 36.32
& \textbf{47.27} & 47.52 & 47.86 & 48.81 & 45.84 & \textbf{39.09}
& \textbf{49.79} & \textbf{49.76} & \textbf{49.60} & \textbf{47.89} & 44.17 & 37.55 \\

\bottomrule
\end{tabular}
}
\end{table}

\subsection{Prefill Eviction}
\paragraph{Setup}
We evaluate PRKV on three instruction-tuned LLMs: Llama-3.1-8B~\citep{dubey2024llama}, 
Qwen3-8B~\citep{yang2025qwen3}, and Gemma-3-12B~\citep{team2025gemma}, with the evaluated model set varying by benchmark.
We build our implementation on \texttt{kvpress} library~\citep{devoto2025expectedattention}. Specifically, we compare PRKV to five baselines: SnapKV~\citep{li2024snapkv}, PyramidKV~\citep{cai2025pyramidkvdynamickvcache}, Expected Attention (EA)~\citep{devoto2025expectedattention}, KeyDiff~\citep{park2025keydiff}, and TOVA~\citep{oren2024transformers}.
For long-context evaluation, we use sixteen English tasks from LongBench~\citep{bai2024longbench} and thirteen tasks from RULER~\citep{hsiehruler}, and additionally benchmark NIAH~\citep{liu2024lost,gkamradt2023needle} for sparse eviction. Specifically, LongBench experiments are conducted on all three model families, RULER experiments on Qwen3-8B and Gemma-3-12B, and NIAH experiments on Llama-3.1-8B.
We keep PRKV hyperparameters fixed across all evaluations, using $\alpha = 0.85$, $C = 64$, $f = 0.25$, $b = 2$, and $n_{\text{sink}} = 4$. We use $W=128$ for RULER, and $W=64$ for LongBench. For value normalization, we adopt a small $\epsilon$ for numerical stability. Additional setup details including $\epsilon$, are provided in Appendix~\ref{appendix:experimental_details}.

\begin{table*}[t]
  \centering
  \small
    \caption{\textbf{Performance comparison on RULER.}
    We report the average of thirteen tasks from RULER benchmarks (sequence length 4K, 16K) over five compression ratios and two models. 0\% indicates average results without compression. Bold values indicate the best performance within each block.}
  \label{tab:ruler-results}
  \vspace{-0.5em}
\resizebox{0.9\textwidth}{!}{
  \begin{tabular}{@{} c |c| *{6}{r}| *{6}{r}   @{}}
    \toprule
    \multirow{2}{*}{\textbf{Model}} &
    \multirow{2}{*}{\textbf{Method}} &
    \multicolumn{6}{c|}{\textbf{Ruler 4K}} &
    \multicolumn{6}{c}{\textbf{Ruler 16K}} 
    \\
    & & 0\% & 10\% & 25\% & 50\% & 75\% & 90\% &
        0\% & 10\% & 25\% & 50\% & 75\% & 90\% \\
    \midrule
    % ===== Qwen3-8B =====
    
      \multirow{7}{*}{\cellcolor{white}\textit{Qwen3-8B}} & SnapKV & \textbf{95.37} & 92.38 & 80.24 & 55.23 & 36.19 & 23.35
                     & \textbf{93.02} & 88.67 & 80.54 & 61.57 & 39.80 & 27.47 \\
      & PyramidKV & \textbf{95.37} & 91.55 & 73.21 & 43.72 & 36.26 & 23.47
                     & \textbf{93.02} & 86.88 & 76.22 & 45.51 & 32.33 & 27.48 \\
      & EA & \textbf{95.37} & 95.20 & 93.84 & \textbf{86.68} & 71.00 & 58.85 
                     & \textbf{93.02} & \textbf{93.20} & 90.97 & 83.85 & 67.72 & 56.53 \\ 
      & TOVA    & \textbf{95.37} & 89.15 & 82.44 & 77.66 & 62.60 & 24.78
                     & \textbf{93.02} & 88.30 & 81.68 & 76.25 & 68.66 & 52.36  \\
      & KeyDiff & \textbf{95.37} & 93.76 & 89.34 & 78.72 & 64.29 & 38.08
                     & \textbf{93.02} & 89.02 & 83.29 & 74.44 & 66.89 & 53.36 \\
      \rowcolor{PRKV}
      \cellcolor{white} & PRKV & \textbf{95.37} & \textbf{95.25} & 93.67 & 86.52 & 72.16 & 58.47
                     & \textbf{93.02} & 92.92 & 90.94 & 83.90 & \textbf{68.78} & 57.18 \\
      \rowcolor{PRKV}
      \cellcolor{white} & PRKV-GLB & \textbf{95.37} & 95.24 & \textbf{94.02} & 86.59 & \textbf{72.40} & \textbf{59.18}
                     & \textbf{93.02} & 92.97 & \textbf{91.19} & \textbf{84.23} & 68.74 & \textbf{57.72} \\
    \midrule
    % ===== Gemma-3-12B =====
      \multirow{7}{*}{\cellcolor{white}\textit{Gemma-3-12B}} 
      \cellcolor{white} & SnapKV & \textbf{95.28} & 84.17 & 73.94 & 58.65 & 45.43 & 33.54
                     & \textbf{85.99} & 73.56 & 62.36 & 46.54 & 37.24 & 32.04 \\
      & PyramidKV & \textbf{95.28} & 82.57 & 69.79 & 53.79 & 45.39 & 33.60
                     & \textbf{85.99} & 72.12 & 57.99 & 40.43 & 35.21 & 32.07 \\
      & EA & \textbf{95.28} & \textbf{95.14} & 94.81 & 88.18 & 79.22 & 66.13 
                     & \textbf{85.99} & 82.53 & 81.60 & 75.65 & 64.71 & 48.93 \\
     & TOVA & \textbf{95.28} & 89.65 & 81.11 & 76.60 & 58.20 & 25.18 
                     & \textbf{85.99} & 79.73 & 72.71 & 62.60 & 47.05 & 32.76 \\
     & KeyDiff & \textbf{95.28} & 94.41 & 90.74 & 79.83 & 61.91 & 34.30 
                     & \textbf{85.99} & 81.71 & 78.68 & 72.52 & 58.52 & 37.10 \\
     \rowcolor{PRKV}
     \cellcolor{white} & PRKV & \textbf{95.28}  & 95.13 & 94.80 & 88.29 & 79.42 & 66.02
                     & \textbf{85.99} & \textbf{82.97} & \textbf{81.98} & \textbf{76.27} & \textbf{65.83} & \textbf{51.05} \\
      \rowcolor{PRKV}
      \cellcolor{white} & PRKV-GLB & \textbf{95.28} & \textbf{95.14} & \textbf{94.91} & \textbf{88.55} & \textbf{79.71} & \textbf{67.46} 
                     & \textbf{85.99} & 82.48 & 81.74 & 75.58 & 64.92 & 49.92 \\
    \bottomrule
  \end{tabular}
  }
  \vspace{-.5em}
\end{table*}

\paragraph{RULER and LongBench}
Tables~\ref{tab:longbench-results} and~\ref{tab:ruler-results} report performance on LongBench and RULER, respectively. On LongBench (Table~\ref{tab:longbench-results}), PRKV variants remain competitive across models and compression ratios, with the preferred variant depending on the setting. Llama-3.1-8B consistently favors PRKV, whereas Gemma-3-12B favors PRKV-GLB. For Qwen3-8B, PRKV performs better under mild-to-moderate compression, while PRKV-GLB becomes preferable under aggressive compression. Thus, the benefit of broader co-attention aggregation depends on both the model and the available cache budget.
On RULER (Table~\ref{tab:ruler-results}), PRKV-GLB generally achieves higher scores than PRKV on Qwen3-8B at both context lengths and on Gemma-3-12B at 4K, whereas Gemma-3-12B favors PRKV at 16K. These patterns are consistent with the operators' different scoring biases: global co-attention can aggregate weak or distributed evidence into smoother importance estimates, while node-level propagation can be preferable when local or intermediate relevance paths are reliable. 
Per-task results and results on larger dense and MoE models are in Appendices~\ref{appendix:full_exp} and \ref{app:rb-experiments}, respectively.

\begin{figure*}[]
\centering
    \subfigure[Full KV]{
    \includegraphics[width=0.3\linewidth]{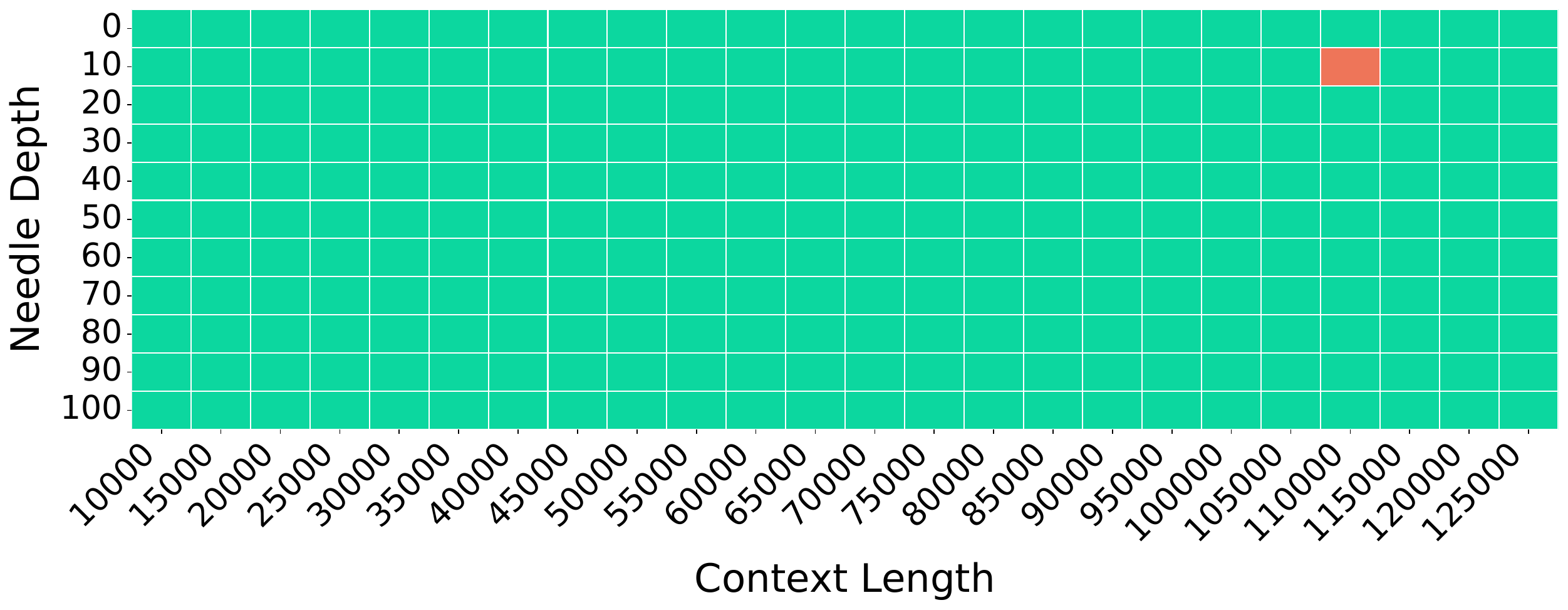}
    }
    \subfigure[SnapKV]{
    \includegraphics[width=0.3\linewidth]{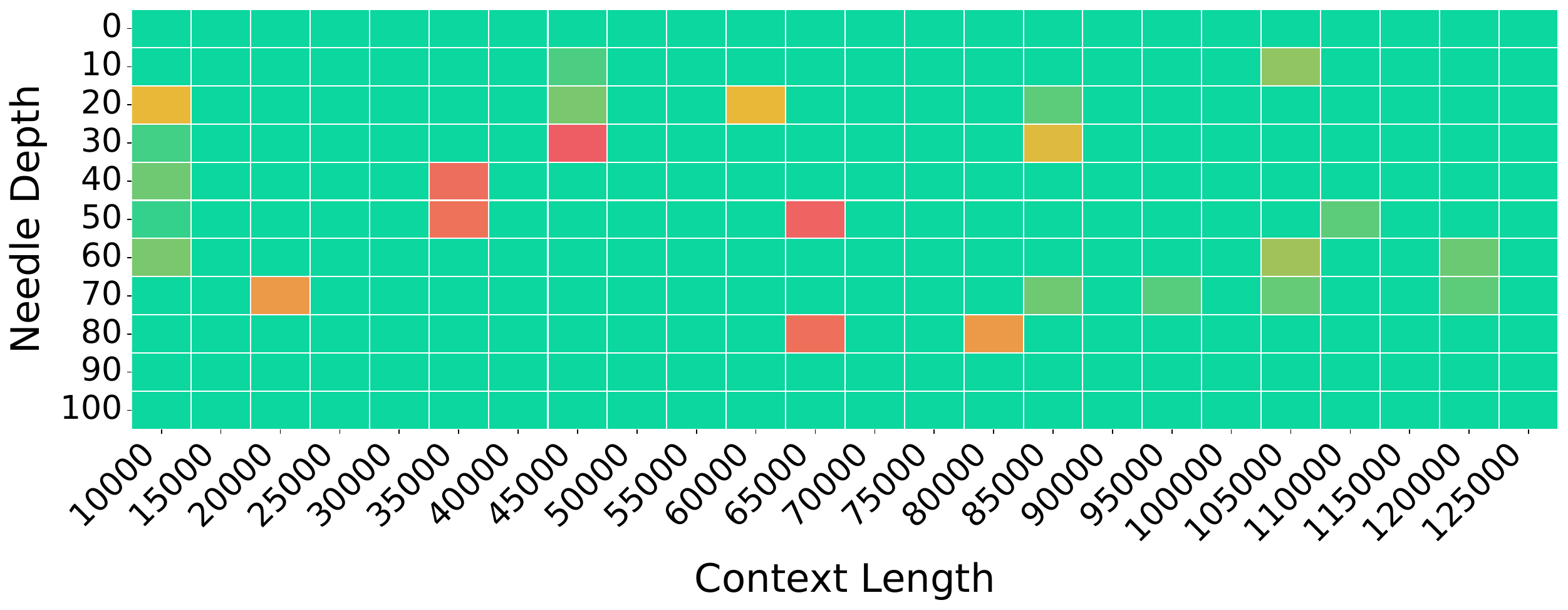}
    }
    \subfigure[PyramidKV]{
    \includegraphics[width=0.3\linewidth]{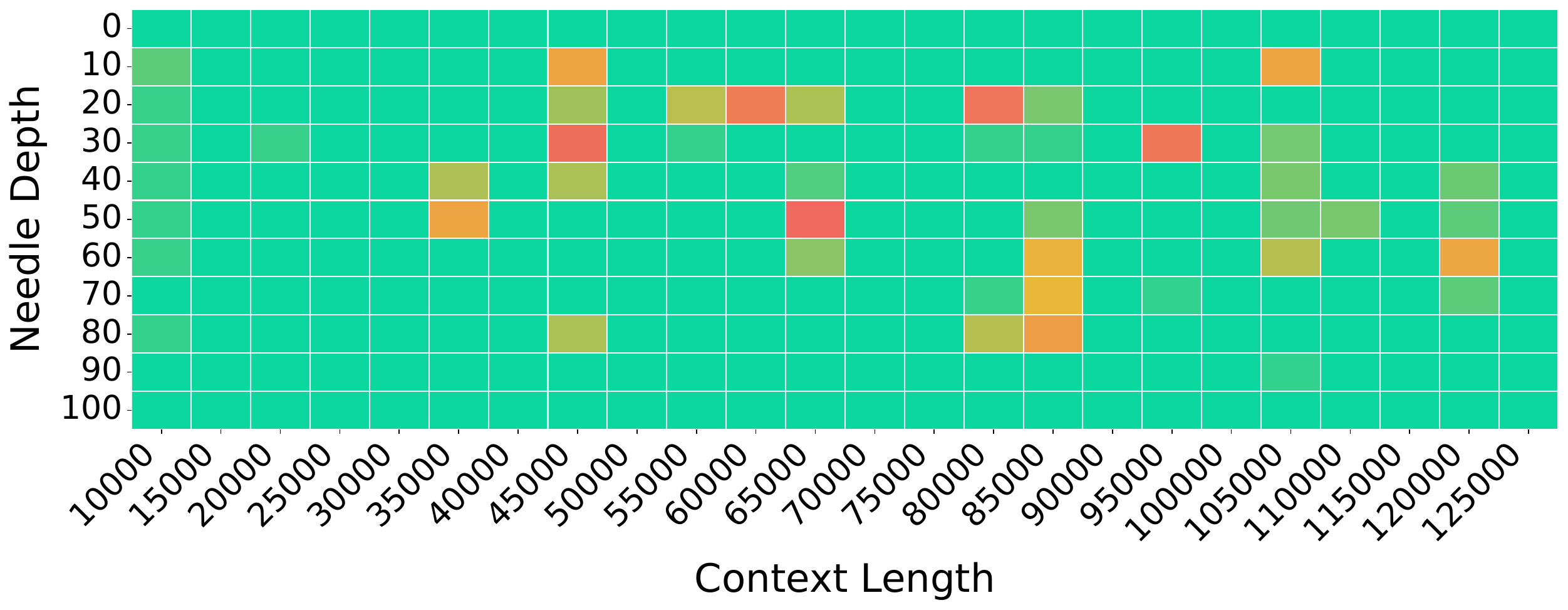}
    }
    \vspace{-0.5em}
    \subfigure[EA]{
    \includegraphics[width=0.3\linewidth]{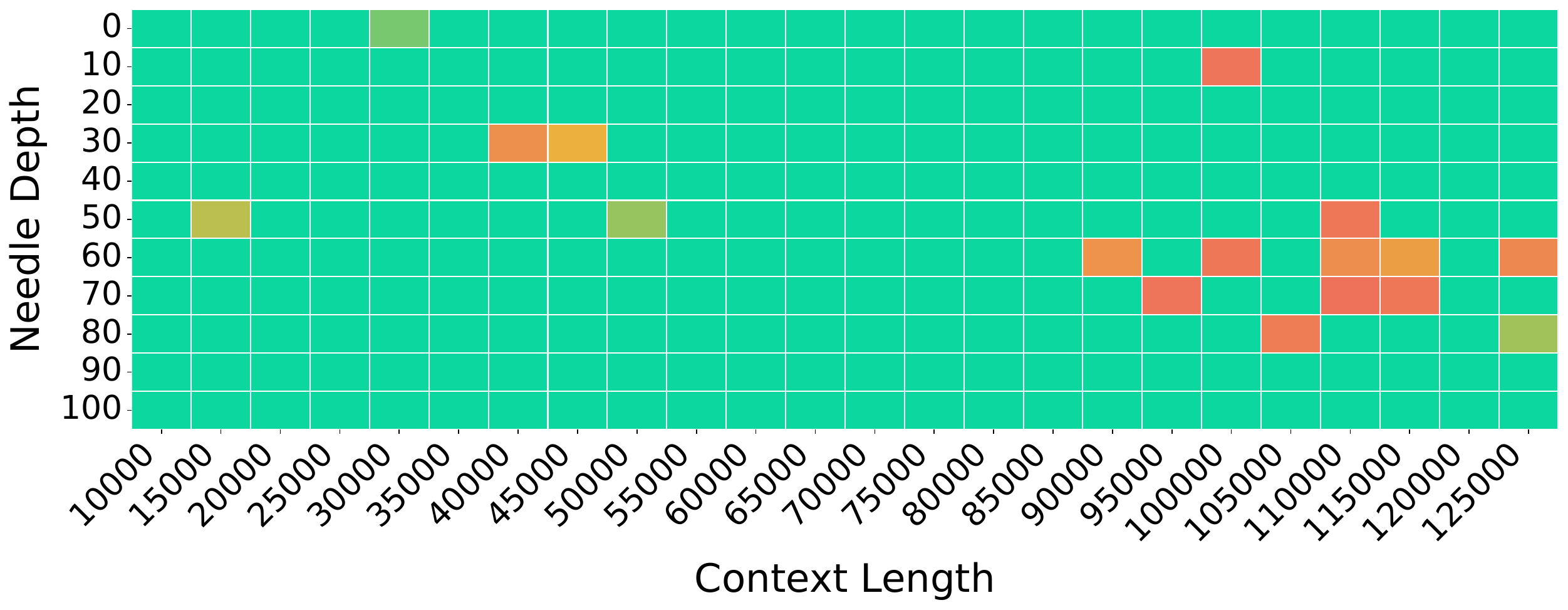}
    }
    \subfigure[PRKV-GLB]{
    \includegraphics[width=0.3\linewidth]{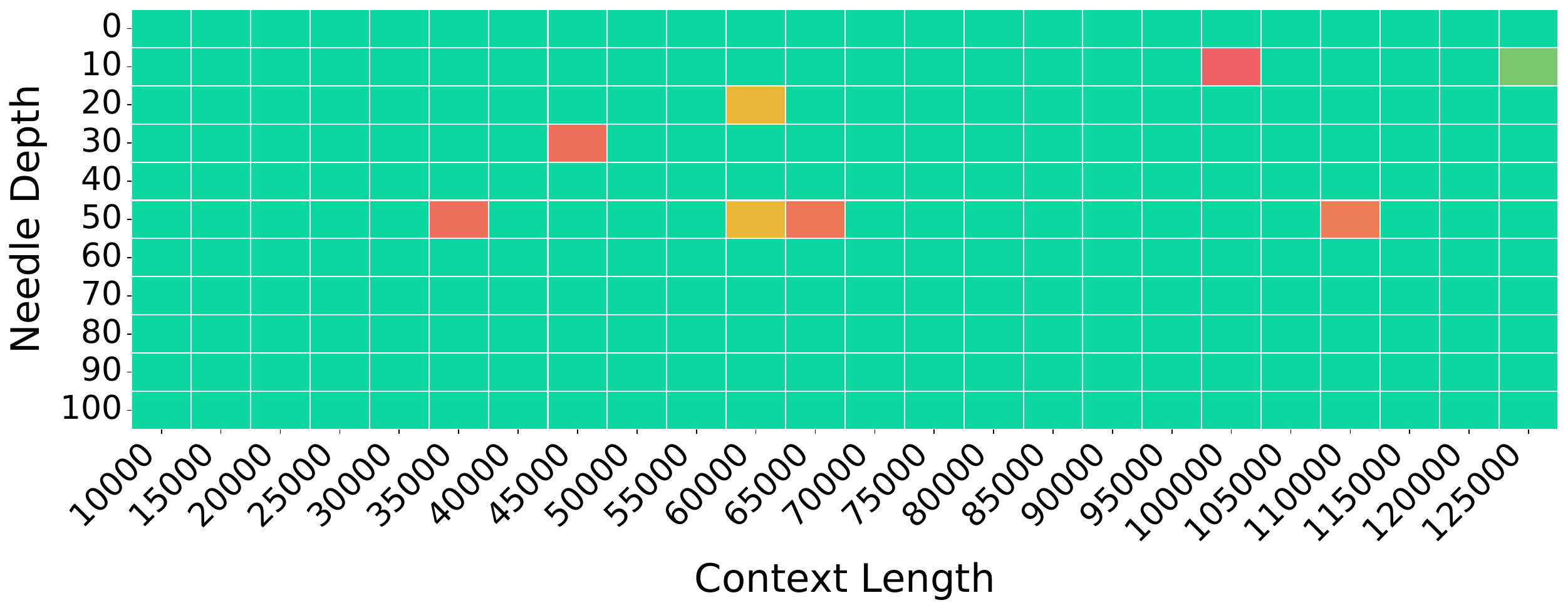}
    }
    \subfigure[PRKV]{
    \includegraphics[width=0.3\linewidth]{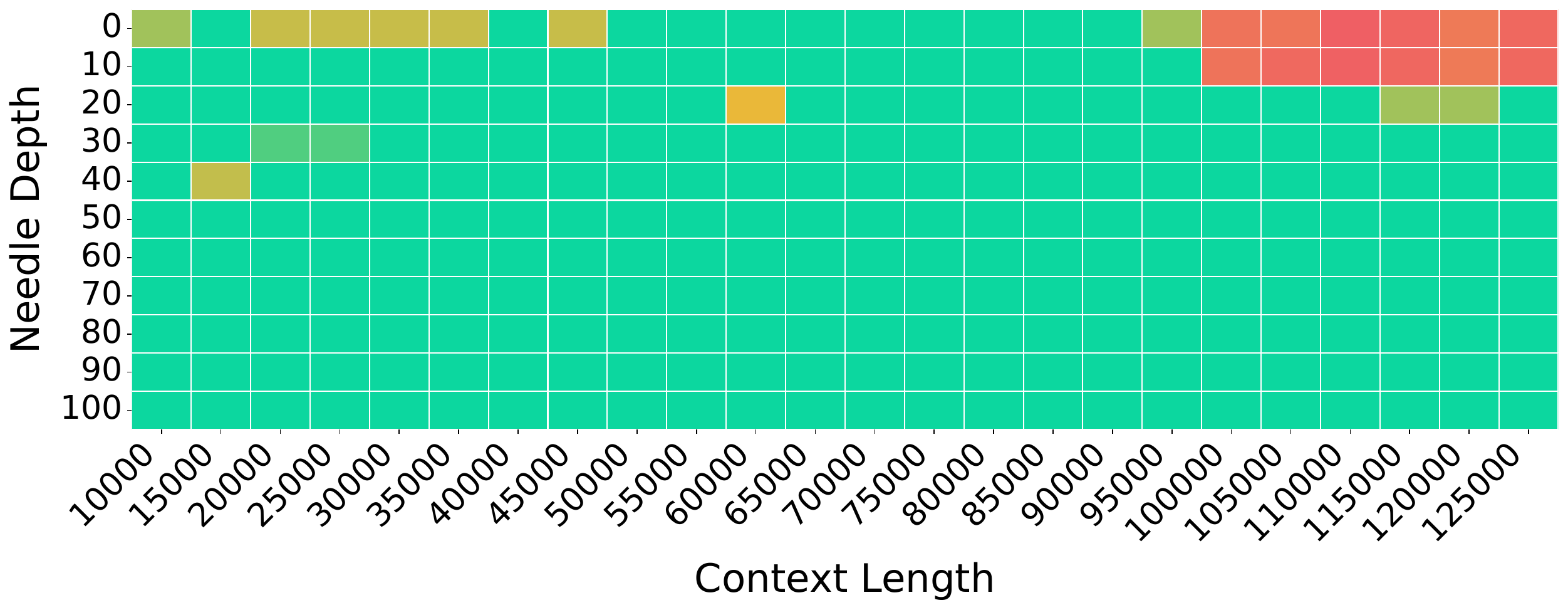}
    }
    \
\caption{\textbf{NIAH accuracy under varying sequence lengths.} We report NIAH results with 50\% compression ratio on Llama-3.1-8B. Full KV refers to an uncompressed KV cache. Color indicates retrieval accuracy on a continuous green–to–red scale (green = 1.0, red = 0.0).
}
\vspace{-1em}
\label{fig:needle}
\end{figure*}

\paragraph{NIAH}
Figure~\ref{fig:needle} reports NIAH accuracy for Llama-3.1-8B.
PRKV-GLB shows more stable retrieval across positions, while PRKV is more sensitive to needle location, particularly at 0--10\% depth.
At these early positions, isolated needles may be poorly represented by PRKV's prefix landmarks or weakly connected to recent-window nodes.
This is consistent with the role of global co-attention propagation in recovering isolated or weakly recurrent evidence.
By aggregating co-attended keys across the cache, PRKV-GLB produces smoother importance estimates and reduces reliance on the recent window alone.

\subsection{Decoding Eviction}

\paragraph{Setup}
We evaluate PRKV on DeepSeek-R1-Distill-Qwen-7B
~\citep{guo2025deepseek} using AIME25~\citep{opencompass_aime2025} and MATH-500~\citep{lightman2023let}, which require long chain-of-thought decoding and impose large KV memory demands.
$\mathbf{\times n}$ denotes a KV cache budget $n \times$ smaller than \textbf{Max Gen.}, which denotes the maximum generation length on each dataset.
We report exact accuracy (correct / total) and include StreamingLLM~\citep{xiao2024efficient} and KNorm~\citep{devoto2024simple} as strong decoding baselines. Detailed setup is deferred to Appendix~\ref{appendix:experimental_details}.

\begin{table}[t]
\small
\centering
\caption{\textbf{Decoding eviction accuracy under varying KV cache budgets on DeepSeek-R1-Distill-Qwen-7B.} Bold values indicate the best performance for each budget size.}
\vspace{.5em}
\label{tab:aime25_math500}
\resizebox{\linewidth}{!}{
\begin{tabular}{l|l|ccccc|l|cccc}
\toprule
\textbf{Method} & \textbf{\shortstack{Dataset\\(Max Gen.)}} & $\mathbf{\times 16}$ & $\mathbf{\times 8}$ & $\mathbf{\times 4}$ & $\mathbf{\times 2}$ & $\mathbf{\times 1}$ &
\textbf{\shortstack{Dataset\\(Max Gen.)}}  & $\mathbf{\times 8}$ & $\mathbf{\times 4}$ & $\mathbf{\times 2}$ & $\mathbf{\times 1}$
\\ \midrule
PRKV-GLB & \multirow{7}{*}{\shortstack{AIME25\\(32K)}} & \textbf{0.233} & 0.233 & 0.300 & 0.300 & 0.300 
& \multirow{7}{*}{\shortstack{MATH-500\\(4K)}} &  \textbf{0.524} & \textbf{0.550} & \textbf{0.560} & \textbf{0.524} \\ 
PRKV & & \textbf{0.233} & 0.200 & 0.300 & 0.300 & 0.300
& & 0.474 & 0.542 & 0.544 & 0.522\\ 
EA & & \textbf{0.233} & 0.233 & 0.300 & 0.300 & 0.300
& & 0.460 & 0.532 & 0.522 & \textbf{0.524} \\ 
KeyDiff & & 0.167 & 0.200 & 0.300 & 0.300 & 0.300
& & 0.288 & 0.458 & 0.532 & 0.522 \\
KNorm & & 0.100 & 0.133 & 0.267 & 0.300 & 0.300
& & 0.052 & 0.246 & 0.436 & \textbf{0.524} \\
StreamingLLM & & 0.133 & 0.267 & 0.267 & 0.300 & 0.300 
& & 0.352 & 0.490 & 0.528 & \textbf{0.524}\\
TOVA & & 0.167 & \textbf{0.367}& 0.300 & 0.300 & 0.300 
& & 0.348 & 0.498 & 0.534 & 0.522\\ 
\bottomrule
\end{tabular}
}
\vspace{-.5em}
\end{table}

\paragraph{AIME25 and MATH-500}
Table~\ref{tab:aime25_math500} summarizes accuracy under varying KV cache budgets.
Across both reasoning benchmarks, PRKV variants remain competitive with decoding-aware baselines under compression.
PRKV-GLB is especially strong on MATH-500, achieving the best accuracy at most budgets, while also matching the top AIME25 accuracy at every budget except $\times 8$, where TOVA performs best. 
Because AIME25 contains only 30 problems, accuracy is coarse-grained: each problem corresponds to 0.033, so several methods can share the same score (e.g., 0.300, or 9/30 correct).
These results suggest that PPR-based propagation, particularly with global co-attention, can help preserve useful long-range reasoning context during decoding.

\section{In-Depth Analysis}

\subsection{PRKV vs. PRKV-GLB Selection}
As shown in Figure~\ref{fig:convergence_score}, different PPR propagation of PRKV variants produces different score biases: PRKV yields sharper, more localized peaks, while PRKV-GLB produces smoother and more globally consistent scores. This suggests a simple selection rule. PRKV-GLB should be preferred as the default when the relevant evidence may be sparse, distributed, or weak in local window. Global co-attention improves recall by aggregating low-amplitude signals before PPR propagation, which explains its strong performance in NIAH, decoding, and many RULER/LongBench settings. PRKV is preferred when local or intermediate attention paths are expected to be clean. Its sharper score distribution avoids over-smoothing and can be more robust when the global graph becomes noisy, as observed for Gemma-3-12B on RULER-16K and Llama-3.1-8B on LongBench. 
In practice, PRKV-GLB is the safer default for sparse, distributed, or long-horizon dependencies, while PRKV is preferable when clean local relevance paths remain more reliable than global aggregation.

\subsection{Approximating Global Importance with Local PPR}

\begin{figure*}[t]
    \centering
    \subfigure[Token-level scores\label{fig:convergence_score}]{
    \includegraphics[width=0.3\linewidth]{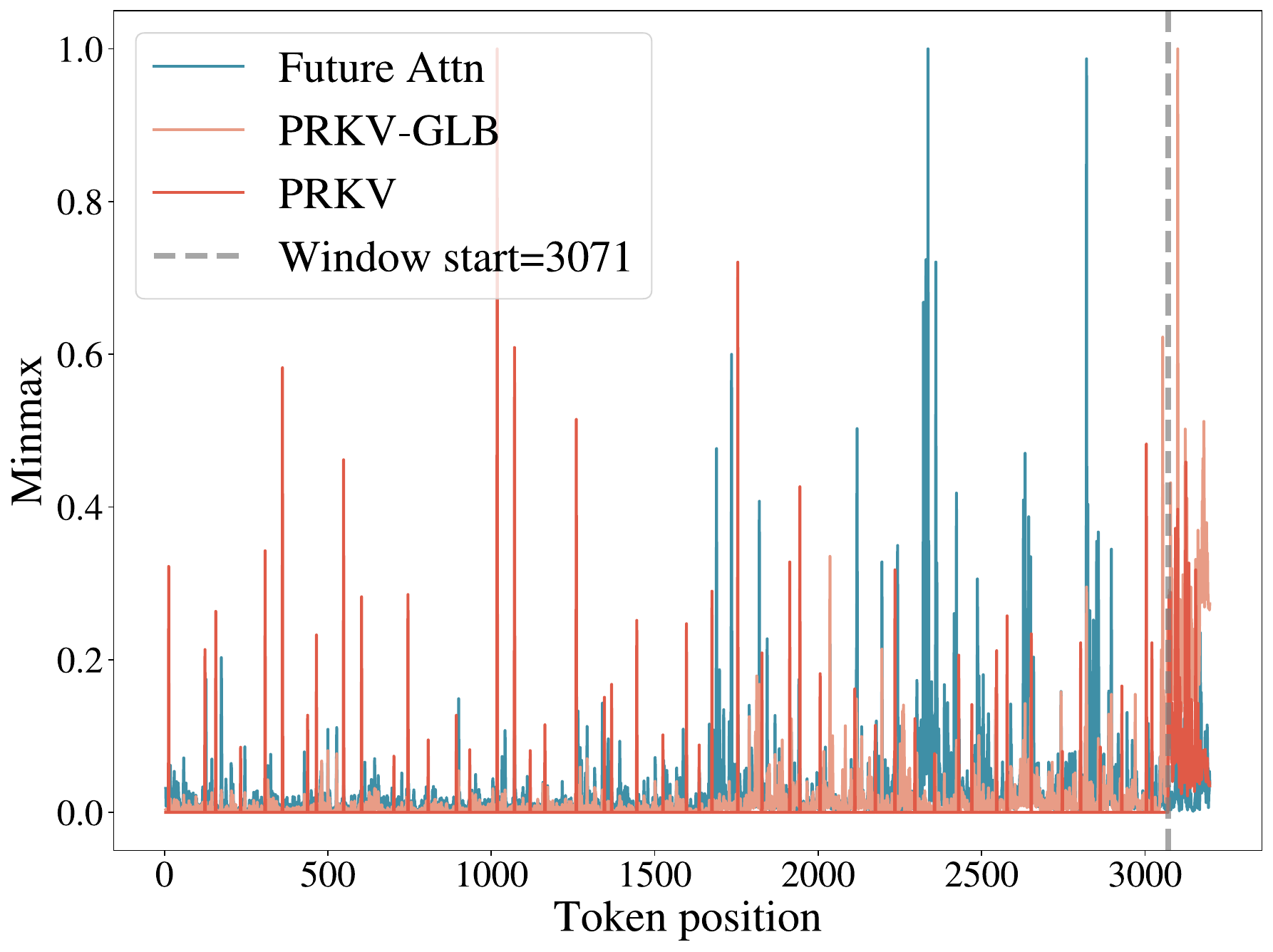}
    }
    \subfigure[Rank correlation\label{fig:convergence_rank}]{
    \includegraphics[width=0.3\linewidth]{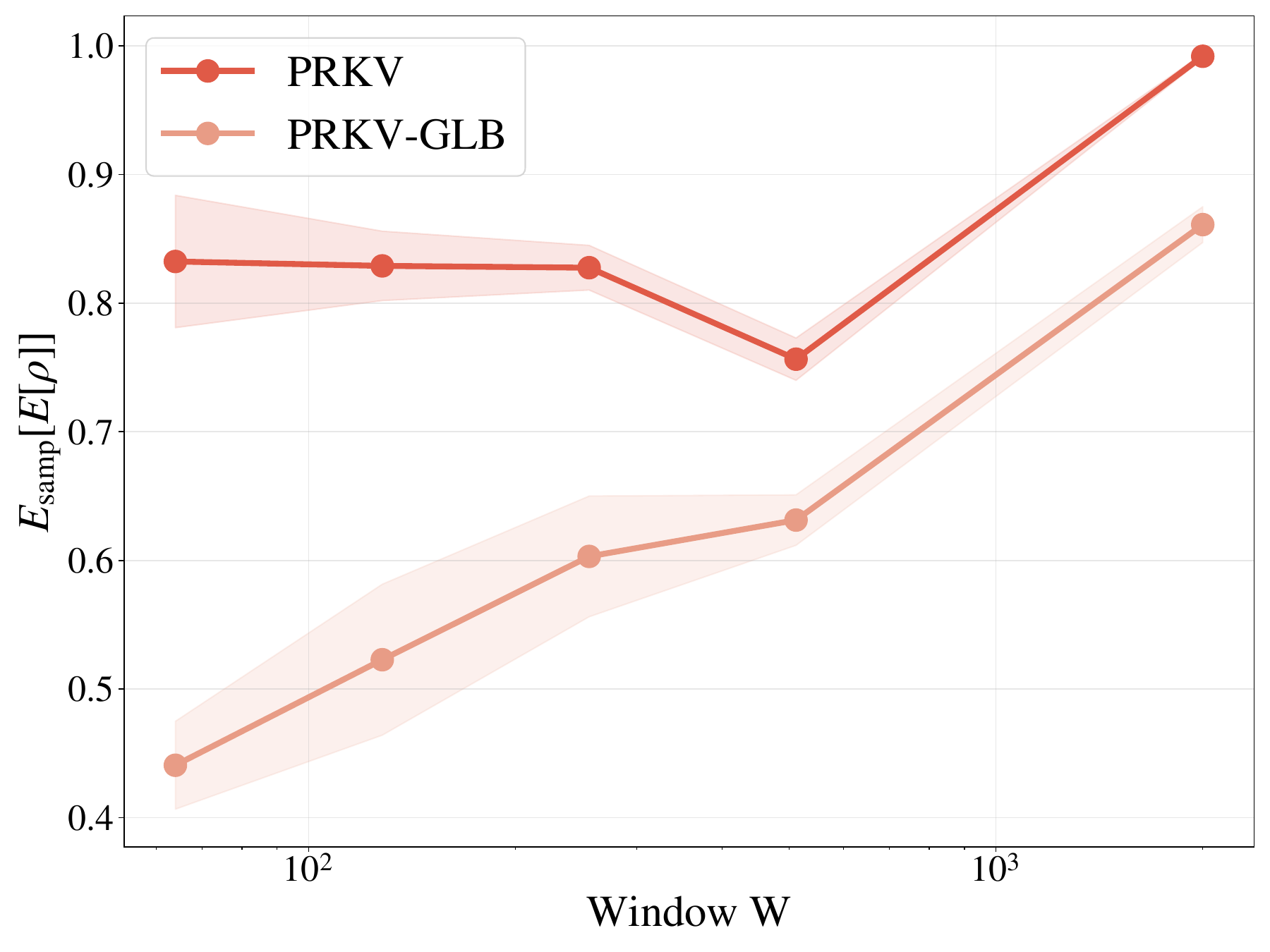}
    }
    \subfigure[Convergence of local PPR\label{fig:convergence_tau}]{
    \includegraphics[width=0.3\linewidth]{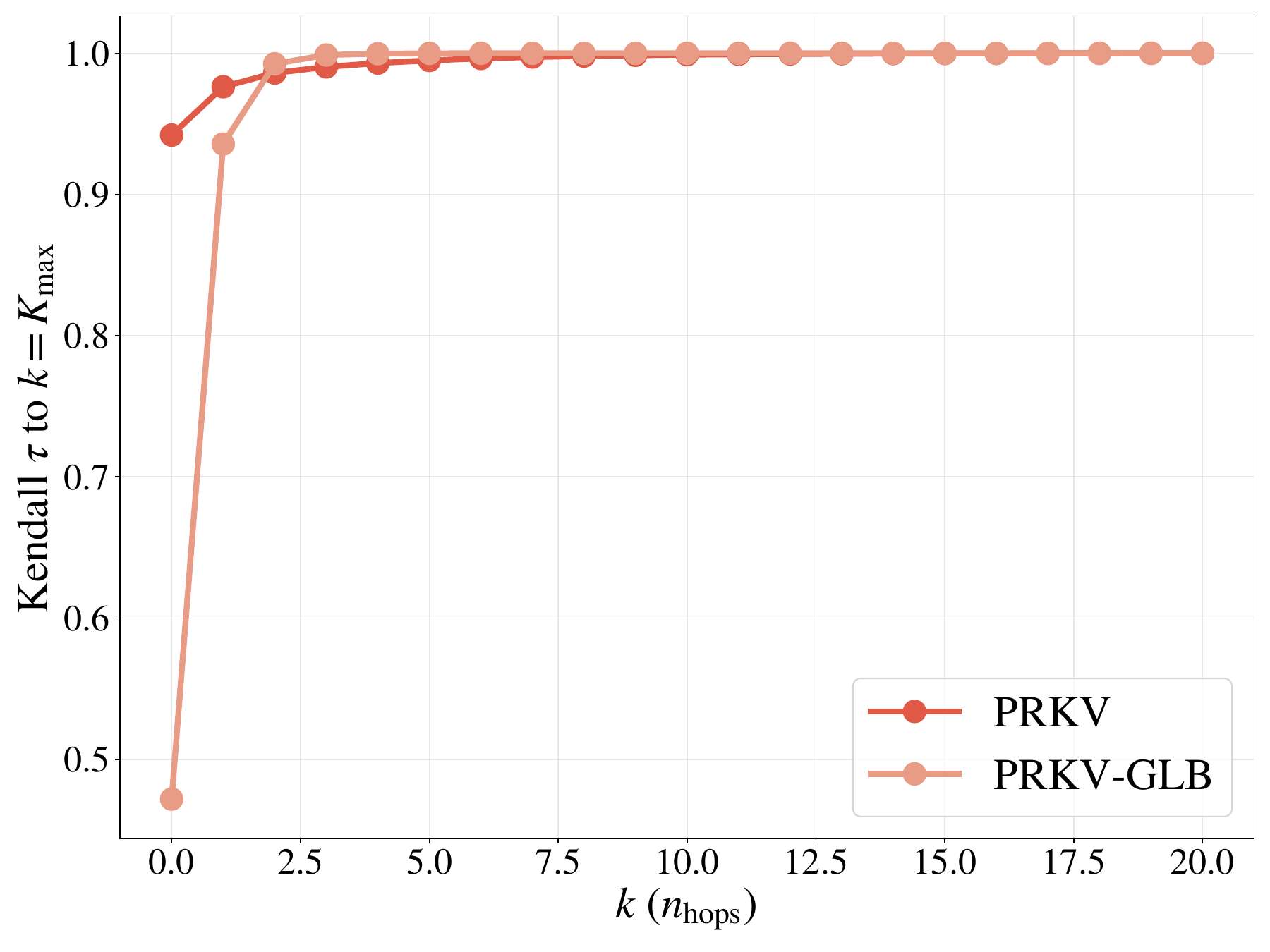}
    }
    \vspace{-0.5em}
    \caption{\textbf{Visualization and convergence of PRKV scores.}
    (a) Token-level scores produced by PRKV and PRKV-GLB compared to the future attention signal.
    (b) Rank correlation between PRKV and $\pi^\ast$ as a function of window size $W$.
    (c) Convergence of PPR scores to their converged values under finite iterations.
    All diagnosed on Llama-3.1-8B, 20\textsuperscript{th} layer.
    }
    \vspace{-1.5em}
    \label{fig:convergence}
\end{figure*}

\paragraph{Local-to-global consistency}
We first examine whether PPR computed from local attention windows  recovers the global importance induced by the full attention graph (see Appendix \ref{appendix:ppr_perturbation} for a perturbation bound). For a fixed input sequence, we compute the global PPR $\pi^\ast$ using the full attention matrix, and compare it with window-restricted PPR $\pi_W$ under varying window sizes $W$, averaged over five NarrativeQA samples. Figure~\ref{fig:convergence_rank} reports Spearman’s rank correlation $\rho$ between $\pi_W$ and $\pi^\ast$. As the window size $W$ increases, the correlation generally increases, indicating that local PPR progressively approaches the global ranking.
For small windows, PRKV shows stronger rank correlation with the global reference, suggesting that purely local propagation is effective when only limited context is available. As the window grows, PRKV-GLB benefits more from the additional context, indicating that incorporating a global component becomes useful under larger  windows.

\paragraph{Finite-iteration stability}
We next examine whether a small number of PageRank iterations suffices to obtain stable importance estimates. We measure stability using Kendall’s $\tau$ between rankings obtained after $n_\text{hops}$ iterations and those after convergence on five NarrativeQA samples. 
Figure~\ref{fig:convergence_tau} shows that both PRKV and PRKV-GLB rapidly stabilize within a small number of iterations, exceeding a correlation of $0.95$ after only a few steps and saturating well before $n_{\text{hops}}=20$. This rapid stabilization suggests that a small, fixed iteration budget suffices to recover nearly identical token rankings, making finite-iteration PPR a practical choice for eviction.

\subsection{Memory and Latency Analysis}

\begin{wrapfigure}[15]{r}{0.6\textwidth}
    \centering
    \vspace{-1em}
    \includegraphics[width=0.475\linewidth]{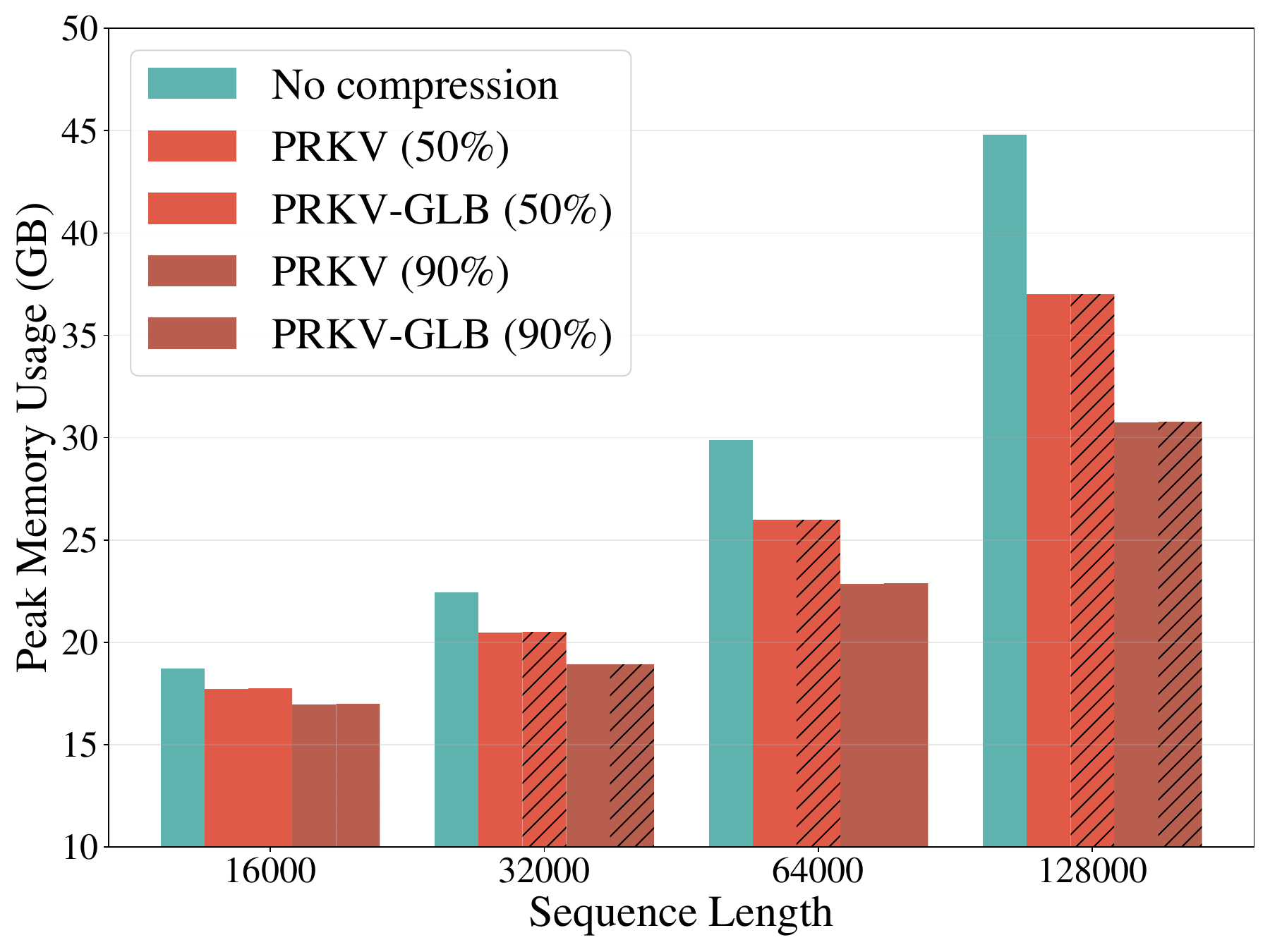}%
\includegraphics[width=0.475\linewidth]{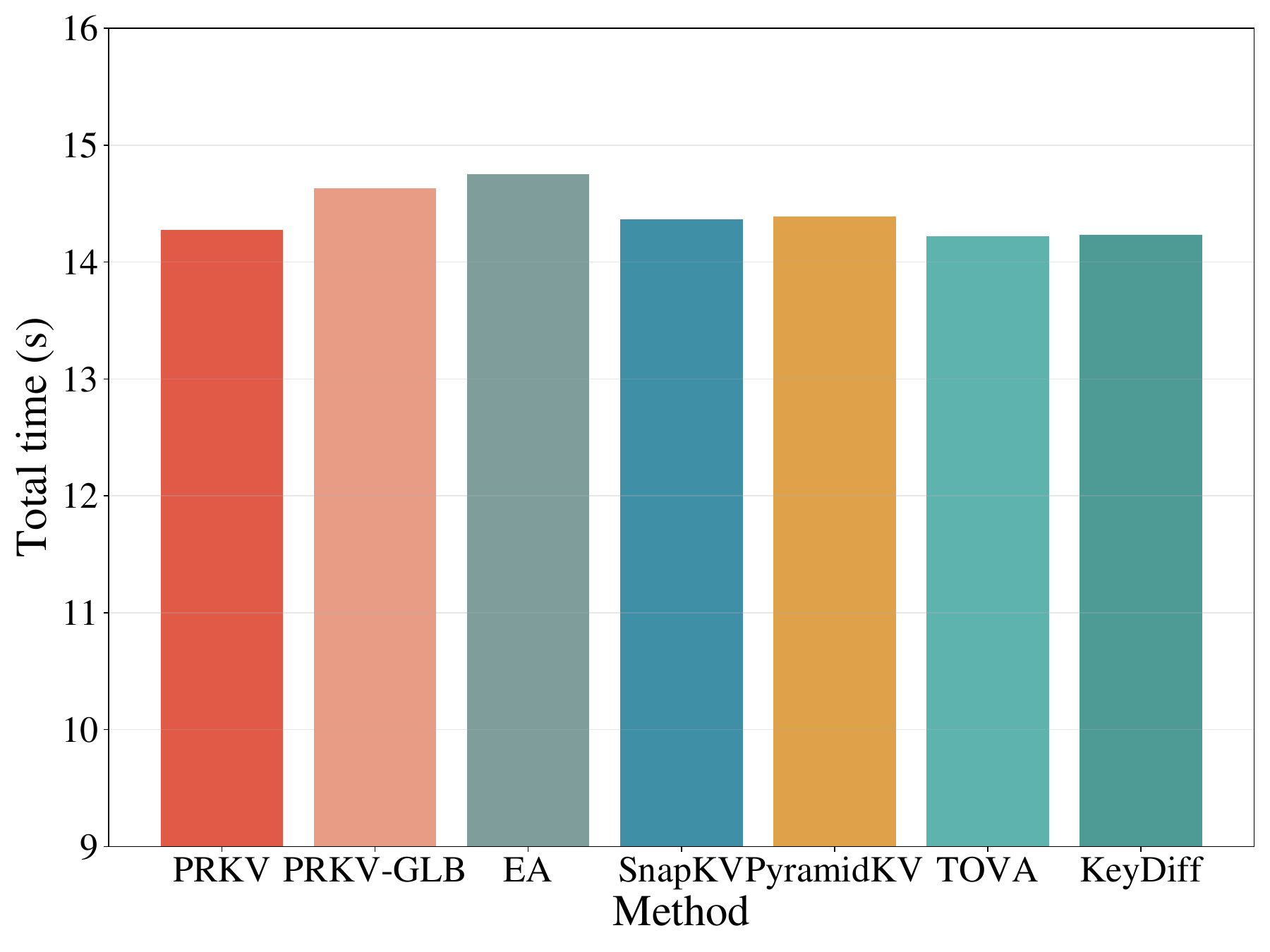}%
    \caption{\textbf{Latency and memory analysis of PRKV.}
    (Left) Peak memory usage across sequence lengths (16K--128K) at compression ratios of 0.5 and 0.9.
    (Right) End-to-end latency at 64K sequence length with compression ratio 0.5.
    Results are obtained on Llama-3.1-8B.}
    \vspace{-1.5em}
    \label{fig:runtime}
\end{wrapfigure}

As shown in Figure~\ref{fig:runtime}(Left), both PRKV and PRKV-GLB consistently reduce peak memory usage across all sequence lengths. The memory reduction becomes more pronounced as the sequence length increases, indicating that PRKV scales favorably to long-context settings. Figure~\ref{fig:runtime}(Right) reports total end-to-end generation time evaluated at a sequence length of 64K with a compression ratio of 0.5. All methods exhibit comparable total wall-clock time. PRKV-GLB incurs a modest additional cost from global co-attention scoring, but both variants maintain low end-to-end overhead. 
To distinguish the cost of compression from its effect on subsequent decoding, we additionally measure time to first token (TTFT) and decode throughput. TTFT includes prefill, importance scoring, and cache compression, and all methods generate the same number of tokens. On RULER-16K with Llama-3.1-8B-Instruct at 50\% compression ratio, PRKV and PRKV-GLB increase TTFT by 2.4\% and 6.3\%, respectively, relative to Full KV. Decode throughput increases from 46.49 tokens/s with Full KV to 49.12 and 49.14 tokens/s, respectively, and remains comparable to other baselines under the same cache budget. Under decoding-time compression on AIME25, they decode at 49.74 and 49.95 tokens/s, on par with baselines. Full TTFT, memory overhead, and theoretical complexity are detailed in Appendix~\ref{appendix:memory}.

\subsection{Ablation and Sensitivity Studies}

\begin{table*}[t]
  \centering
  \small
  \caption{\textbf{Ablation studies of PRKV and PRKV-GLB on Qwen3-8B.} We report the average of sixteen tasks from LongBench benchmarks over five compression ratios. 0\% indicates average results without compression. Bold values indicate the best performance within each block.}
  \label{tab:ablation_main}
  \setlength{\tabcolsep}{6pt}
  \renewcommand{\arraystretch}{1.15}
  \resizebox{0.95\textwidth}{!}{
  \begin{tabular}{@{} l l | *{6}{r} |*{6}{r} @{}}
    \toprule
    \multirow{2}{*}{\textbf{Component}} &
    \multirow{2}{*}{\textbf{Variant}} &
    \multicolumn{6}{c}{\textbf{PRKV}} &
    \multicolumn{6}{c}{\textbf{PRKV-GLB}} \\
    \cmidrule(lr){3-8} \cmidrule(lr){9-14} 
    & & 0\% & 10\% & 25\% & 50\% & 75\% & 90\% 
    & 0\% & 10\% & 25\% & 50\% & 75\% & 90\% \\
    \midrule
    \multirow{3}{*}{$\alpha$} & 0.75 & \textbf{47.27} & 47.68 & 48.51 & 48.96 & 46.02 & \textbf{37.60} &
    \textbf{47.27} & 47.34 & \textbf{47.99} & 48.63 & 45.74 & 38.80 \\
    & 0.85 & \textbf{47.27} & 47.74 & 48.56 & 48.92 & 46.26 & 37.59 &
    \textbf{47.27} & \textbf{47.52} & 47.86 & \textbf{48.81} & \textbf{45.84} & 39.09 \\
    & 0.95 & \textbf{47.27} & \textbf{47.80} & \textbf{48.82} & \textbf{49.23} & \textbf{46.32} & 37.27 &
    \textbf{47.27} & 47.47 & 47.74 & 48.77 & 45.78 & \textbf{39.17} \\
    \midrule
    
    \multirow{3}{*}{$W$}
    & 64 & \textbf{47.27} & \textbf{47.74} & \textbf{48.56} & \textbf{48.92} & \textbf{46.26} & \textbf{37.59} &
    \textbf{47.27} & \textbf{47.52} & 47.86 & \textbf{48.81} & \textbf{45.84} & \textbf{39.09} \\
    & 128 & \textbf{47.27} & 47.66 & 48.08 & 48.68 & 45.82 & 37.14 &
    \textbf{47.27} & 47.47 & \textbf{47.87} & 48.58 & 45.59 & 38.80 \\
    & 256 & \textbf{47.27} & 47.65 & 47.70 & 48.80 & 45.35 & 35.66 & 
    \textbf{47.27} & 47.37 & 47.79 & 48.59 & 45.47 & 38.00 \\ \midrule
    \multirow{3}{*}{$n_\text{hops}$} & 2 & \textbf{47.27} & \textbf{47.74} & 48.47 & \textbf{49.10} & 45.97 & 37.38 &
    \textbf{47.27} & \textbf{47.56} & \textbf{47.92} & 48.75 & 45.62 & \textbf{39.15} \\
    & 3 & \textbf{47.27} & 47.74 & 48.56 & 48.92 & \textbf{46.26} & \textbf{37.59} &
    \textbf{47.27} & 47.52 & 47.86 & \textbf{48.81} & \textbf{45.84} & 39.09 \\
    & 10 & \textbf{47.27} & 47.52 & \textbf{48.63} & 48.76 & 46.09 & 37.56 &
    \textbf{47.27} & \textbf{47.56} & 47.72 & 48.70 & 45.67 & 38.79 \\
    \midrule
    \multirow{3}{*}{\makecell{Value-norm\\scaling}}
    & PRKV (PPR+VNorm)
    & \textbf{47.27} & 47.74 & 48.56
    & \textbf{48.92} & \textbf{46.26} & \textbf{37.59}
    & \textbf{47.27} & 47.52 & 47.86
    & \textbf{48.81} & \textbf{45.84} & \textbf{39.09} \\
    & Only PPR & \textbf{47.27} & 46.05 & 44.30 & 41.24 & 37.42 & 32.86 & \textbf{47.27} & 47.10 & 46.70 & 45.02 & 41.86 & 35.93 \\
    & Only VNorm
    & \textbf{47.27} & \textbf{48.68} & \textbf{48.76}
    & 48.53 & 44.66 & 35.17
    & \textbf{47.27} & \textbf{48.68} & \textbf{48.76}
    & 48.53 & 44.66 & 35.17 \\
    \bottomrule
  \end{tabular}
  }
  \vspace{-.5em}
\end{table*}

Table~\ref{tab:ablation_main} examines the sensitivity of both PRKV and its GLB variant to key design choices, including damping factor $\alpha$, window size $W$, and the number of PPR iterations $n_{\text{hops}}$. Across both variants, performance is largely robust to reasonable hyperparameter choices. Results are stable over a broad range of damping factors, with a moderate value (e.g., $\alpha=0.85$) performing reliably. Increasing the window size does not consistently improve accuracy, suggesting diminishing returns from aggregating broader context. Also, while PPR scores converge as the number of hops increases, a small number of iterations already achieves strong performance, offering a favorable balance between accuracy and efficiency. 
We further examine the contribution of value-norm scaling by comparing the full method with Only PPR and Only VNorm. Removing value-norm scaling consistently reduces performance under compression, indicating that propagation-based relevance benefits from accounting for value magnitude. Conversely, Only VNorm performs strongly under mild compression, but the combined score achieves better performance as the cache budget tightens. These results support the complementary roles of PPR and value-norm scaling, particularly under aggressive compression. Additional ablations including comparisons across $\epsilon$ values are provided in Appendix~\ref{appendix:ablation}.

\section{Related Works}

\paragraph{KV eviction}
Recent work on KV cache compression has primarily focused on eviction-based methods that estimate token importance during autoregressive decoding.
Scissorhands~\citep{liu2023scissorhands} exploits short-term temporal persistence of influential tokens and evicts those with consistently low attention across decoding steps.
SnapKV~\citep{li2024snapkv} performs permanent eviction using attention patterns from recent queries within a fixed observation window, while PyramidKV~\citep{cai2025pyramidkvdynamickvcache} exploits pyramidal information flow across layers to dynamically adjust per-layer cache sizes.
Moreover, TOVA~\citep{oren2024transformers} removes low-attention tokens by viewing transformers as multi-state RNNs.
Beyond instantaneous attention, Expected Attention~\citep{devoto2025expectedattention} estimates importance via expected future attention, and KeyDiff~\citep{park2025keydiff} leverages key-space similarity to retain diverse tokens.

\paragraph{Graph-based KV caching}
In parallel, several works explore caching mechanisms that operate over explicit graph structures rather than individual tokens.
Graph-KV~\citep{wang2025graphkv} organizes KV states according to graph-structured dependencies, enabling sparse, structure-aware attention and demonstrating gains on graph-centric and RAG benchmarks.
SubGCache~\citep{zhu2025subgcache} further exploits structural similarity by caching and reusing representative subgraph-level KV states across queries in graph-based RAG systems.
Similarly, DepCache~\citep{yuan2025depcache} introduces a dependency-aware cache management framework that aligns KV reuse and eviction.
GraphKV~\citep{li2025graphkv} propagates decay signals over a key-similarity graph to diversify the tokens retained by an existing eviction policy, whereas PRKV estimates importance via PPR on attention graphs.

\paragraph{PageRank and graph diffusion}
PageRank has been studied as a mechanism for aggregating global importance from local graph connectivity, with early work establishing its theoretical foundations and convergence properties~\citep{page1999pagerank,langville2004deeperip}. Beyond web ranking, PageRank and its variants have been used to model long-range relevance in applications such as information retrieval and recommendation~\citep{gleich2015pagerank}. Diffusion-based propagation has also been explored in graph learning to capture multi-hop dependencies beyond one-hop GNN aggregation~\citep{atwood2016diffusion,xu2019graph}, including PPR-based node propagation and representation smoothing~\citep{gasteiger2019diffusion}. Diffuser~\citep{feng2023diffuser} extends this diffusion perspective to Transformers, showing that multi-hop kernels can approximate full attention.

\section{Conclusion}
We introduced PRKV, a training-free KV eviction method that exploits the graph structure of self-attention and estimates global influence via PPR. By capturing multi-hop propagation, PRKV preserves long-range dependencies under aggressive compression with minimal overhead. Our results show that this mechanism is effective across diverse benchmarks, and our analyses further demonstrate that PPR produces stable and interpretable importance estimates, supporting global importance estimation as a principled foundation for scalable memory-efficient inference.

\subsubsection*{Acknowledgments}
This research was supported by the MSIT (Ministry of Science and ICT), Korea, under the Development and Demonstration of Agentic AI Technology for Autonomous Physics Simulation Based on MCP Standard with Multi-Agent Collaboration (RS-2026-25587486), Global Research Support Program in the Digital Field program (RS-2024-00436680), AI Research Hub Project (RS-2024-00457882), High-Efficiency Neural Networks for Artificial General Intelligence (HERMES-Net) (RS-2026-25526850), and LG AI STAR Talent Development Program for Leading Large-Scale Generative AI Models in the Physical AI Domain (RS-2025-25442149), supervised by the IITP (Institute for Information \& Communications Technology Planning \& Evaluation). This work was supported by the National Research Foundation of Korea(NRF) grant funded by the Korea government(MSIT) (No. RS-2026-25607520, Center for Specialized Orchestration and Training of AI for Frontier Mathematics (SOTA-MATH)). This project is supported by Microsoft Research Asia.

\bibliographystyle{unsrtnat}
\bibliography{reference.bib}

%%%%%%%%%%%%%%%%%%%%%%%%%%%%%%%%%%%%%%%%%%%%%%%%%%%%%%%%%%%%
\clearpage
\appendix

\section{Broader Impact} \label{appendix:broader_impact}
This work focuses on improving the computational efficiency of long-context inference in LLMs through training-free KV eviction. By reducing memory usage and inference overhead, our method may lower the computational and hardware requirements needed to deploy long-context models, potentially improving accessibility to advanced language technologies in resource-constrained settings. PRKV is primarily technical in nature and does not introduce new model capabilities or training data. While cache eviction may alter model behavior due to compression, we empirically study these effects across multiple benchmarks and memory budgets. We do not foresee significant negative societal impacts specific to this work beyond those generally associated with the deployment of LLMs.

\section{Experiment Details} \label{appendix:experimental_details}
\subsection{Basic Settings}
Our experiments were conducted on a computing server equipped with dual \textrm{AMD EPYC 9654} CPUs (192 cores and 384 threads in total, 1.5--3.7\,GHz, 768\,MiB L3 cache). Most experiments were performed on an \textrm{NVIDIA A100 80GB PCIe} GPU with \textrm{CUDA} 12.4. 
Appendices~\ref{app:rb-experiments}, \ref{app:rb-quantization}, \ref{app:rb-latency}, and \ref{app:rb-tp} experiments were conducted on four \textrm{NVIDIA RTX PRO 6000 Blackwell Max-Q Workstation Edition} GPUs with \textrm{CUDA} 12.8. Within each reported comparison, all methods were evaluated under the same hardware and software configuration.
\subsection{PRKV and Baseline Settings}
For all evaluations, we fix damping factor $\alpha=0.85$, iteration $n_\text{hops}=3$, chunk size $C=64$, recency ratio $f=0.25$, recency boost $b=2$, the number of sink tokens $n_\text{sink}=4$. For RULER, we use window size $W=128$, while we use $W=64$ for LongBench, reflecting their different context regimes. We match the window size of window-based baselines (SnapKV, PyramidKV) to PRKV setting for fair comparison. 
For value-normalization, a small constant $\epsilon$ is used for numerical stabilization. 
For long-context modeling, we use $\epsilon=1e-3$ for LongBench and RULER 16K on Qwen3-8B, $\epsilon=1e-6$ for LongBench on Gemma-3-12B and NIAH, and $\epsilon=1e-2$ for all other experiments. 
For long-context benchmarks, we prune KV cache at prefilling phase. For reasoning benchmarks, eviction is done in decoding phase. 
Here, we use window size $W=128$, $\epsilon=1e-2$ and we cap the number of prefix landmarks as 5\% of target KV size for PRKV (about 1.6\% in practice with $C=64$). 
All other configurations are kept same to prefill eviction. 
Our code is available at \url{https://github.com/sjiinkim/PRKV}. All baseline implementations follow \texttt{kvpress}~\citep{devoto2025expectedattention} defaults. We use FlashAttention-2~\citep{dao2023flashattention} for default attention kernel.

\subsection{LongBench Evaluation Details} \label{appendix:longbench_details}

For the LongBench~\citep{bai2024longbench} evaluation, we assessed all English-language tasks, covering sixteen benchmarks grouped into six categories for comprehensive evaluation. Task descriptions, abbreviation, and associated metrics are provided below. During inference, we set the maximum sequence length to each model's trained length. 
\begin{itemize}
    \item Single-Document QA
    \begin{itemize}
        \item Qasper~\citep{dasigi2021dataset} (QA, F1 score)
        \item NarrativeQA~\citep{kovcisky2018narrativeqa} (NQA, F1 score)
        \item MultiFieldQA-en (MQA, F1 score)
    \end{itemize}
    \item Multi-Document QA
    \begin{itemize}
        \item HotpotQA~\citep{yang2018hotpotqa} (HQA, F1 score)
        \item 2WikiMultihopQA~\citep{ho-etal-2020-constructing} (2WM, F1 score)
        \item MuSiQue~\citep{trivedi2022musique} (MSQ, F1 score)
    \end{itemize}
    \item Summarization
    \begin{itemize}
        \item QMSum~\citep{zhong2021qmsum} (QM, ROUGE score)
        \item MultiNews~\citep{fabbri2019multi} (MN, ROUGE score)
        \item GovReport~\citep{huang2021efficient} (GR, ROUGE score)
    \end{itemize}
    \item Few-shot Learning
    \begin{itemize}
        \item TREC~\citep{li2002learning} (TR, classification score)
        \item TriviaQA~\citep{joshi2017triviaqa} (TQA, F1 score)
        \item SAMSum~\citep{gliwa2019samsum} (SS, ROUGE score)
    \end{itemize}
    \item Code Completion
    \begin{itemize}
        \item LCC~\citep{guo2023longcoder} (similarity score)
        \item RepoBench-p~\citep{liu2023repobench} (RB, similarity score)
    \end{itemize}
    \item Synthetic
    \begin{itemize}
        \item PassageCount (PC, Accuracy)
        \item PassageRetrieval (PR, Accuracy)
    \end{itemize}
\end{itemize}

\subsection{RULER Evaluation Details}

For the RULER~\citep{hsiehruler} benchmark, we evaluated across thirteen synthetic tasks spanning four categories: retrieval, multi-hop tracing, aggregation, and question answering. During inference, we set the maximum sequence length to each model's trained length. Below we list task categories, abbreviations, and associated evaluation metrics, along with brief descriptions.

\begin{itemize}
    \item Retrieval
    \begin{itemize}
        \item Single NIAH (S-NIAH, Accuracy): Retrieve one “needle” (key-value pair) hidden in a long “haystack.” Three needle types are used (Word–Number, Word–UUID, UUID–UUID).
        \item Multi-keys NIAH (MK-NIAH, Accuracy): Retrieve the correct value when multiple distractor keys are inserted.
        As with S-NIAH, three needle types are tested.
        \item Multi-values NIAH (MV-NIAH, Accuracy): Retrieve all values associated with a single key.
        \item Multi-queries NIAH (MQ-NIAH, Accuracy): Retrieve values for multiple distinct keys simultaneously.
    \end{itemize}
    \item Multi-hop Tracing
    \begin{itemize}
        \item Variable Tracking (VT, Accuracy): Trace variable assignments through chains of references to find all variables pointing to the same value.
    \end{itemize}
    \item Aggregation
    \begin{itemize}
        \item Common Words Extraction (CWE, Accuracy): Identify a fixed set of common words among distractors in long sequences.
        \item Frequent Words Extraction (FWE, Accuracy): Return the top-$K$ most frequent words sampled from a Zipfian distribution.
    \end{itemize}
    \item Question Answering
    \begin{itemize}
        \item Long-context QA (QA, String match accuracy): Answer questions when golden paragraphs are mixed with many distractors, extending short-context QA datasets (e.g., SQuAD, HotpotQA).
    \end{itemize}
\end{itemize}

\subsection{Decoding Evaluation Details}
For decoding experiments, we evaluate our method on DeepSeek-R1-Distill-Qwen-7B~\citep{guo2025deepseek}, as reasoning-oriented models produce long chain-of-thought outputs that heavily stress the KV cache. We use two math reasoning benchmarks: AIME25~\citep{opencompass_aime2025} and MATH-500~\citep{lightman2023let}. AIME25 consists of thirty competition-style problems requiring multi-step reasoning and precise computation, while MATH-500 covers 500 questions from a broad range of mathematical domains with varying difficulty.

During generation, the KV cache is first allowed to grow to a fixed capacity, after which token eviction is applied. We apply compression every 128 steps. For all baselines, we set buffer size to 128, following implementation of \texttt{kvpress}~\citep{devoto2025expectedattention}. We use a maximum generation length of 4K for MATH-500 and 32K for AIME25 and report exact accuracy (correct / total).

As baselines, we omit SnapKV and PyramidKV, since they both target eviction at the prefill stage. We therefore compare against decoding-aware baselines (StreamingLLM, KNorm, EA, KeyDiff, TOVA) that operate online eviction during autoregressive generation. 

\subsection{Motivational Analyses Details}
\label{appendix:motiv_details}

All motivational analyses are conducted on long-context inputs from the PaulGrahamEssays corpus using Llama-3.1-8B as the backbone model. Each document is loaded as a single string, truncated to at most 20{,}000 characters, and tokenized with the model's tokenizer. We run inference in teacher-forcing mode when needed. These analyses are retrospective diagnostics: we evaluate whether importance scores computed at an anchor position $t$ anticipate the subsequent usage of tokens already present in the cache. Attention from future query positions is used only as a held-out evaluation reference and is never used to compute the importance scores.

\paragraph{Evaluation protocol}
To reduce sensitivity to the choice of excerpt and anchor position, we use multiple slices and anchors. From each tokenized document, we uniformly sample $N_{\mathrm{seq}}=5$ offsets and extract slices of $S_{\mathrm{slice}}=4096$ tokens. Token positions are indexed within each slice. Within each slice, we uniformly sample $N_{\mathrm{anchor}}=25$ anchor positions without replacement, subject to the availability of both a past key window of size $W$ ending at $t$ and a later query window. At each anchor, all methods score the same candidate window $\mathcal{W}_t=\{t-W+1,\ldots,t\}$ using only information available up to $t$. Reported metrics are averaged over anchors and slices.

\paragraph{Attention notation and future-usage reference}
Let $\bar{A}_{u,i}$ denote the attention from query position $u$ to key position $i$, obtained by averaging the attention weights in Eq.~\ref{eq:attn} over heads. For anchor $t$, let $\mathcal{Q}_t$ denote the later query positions used for evaluation, with $u>t$ for every $u\in\mathcal{Q}_t$. We define the normalized future-attention reference as
\[
a_i=\frac{\sum_{u\in\mathcal{Q}_t}\bar{A}_{u,i}}{\sum_{j\in\mathcal{W}_t}\sum_{u\in\mathcal{Q}_t}\bar{A}_{u,j}}, \qquad i\in\mathcal{W}_t.
\]
The vector $\mathbf{a}=(a_i)_{i\in\mathcal{W}_t}$ represents the relative later-window attention mass received by the candidate tokens. It serves as an empirical proxy for subsequent usage rather than as semantic ground-truth importance. For quantities associated with a fixed anchor, we omit their dependence on $t$ when unambiguous.

\begin{table*}[t]
\centering
\vspace{-1em}
\caption{\textbf{Definitions of the diagnostic scoring rules in Figure~\ref{fig:motivational_analysis}.}}
\label{tab:motivational_scores}
\small
\begin{tabular}{@{}p{0.12\textwidth}p{0.3\textwidth}p{0.47\textwidth}@{}}
\toprule
\textbf{Method} & \textbf{Definition of $s_i$} & \textbf{Signal encoded} \\
\midrule
Random & $s_i\overset{\mathrm{i.i.d.}}{\sim}\mathcal{U}(0,1)$ & No information; chance ranking. \\[5pt]
Recency & $\displaystyle s_i=\frac{1}{t-i+1}$ & Positional recency relative to the anchor. \\
Frequency & $\displaystyle s_i=\sum_{j=1}^{t}\mathbf{1}\{\operatorname{tok}(j)=\operatorname{tok}(i)\}$ & Number of occurrences of the candidate token ID in the observed prefix. \\
Attention & $s_i=\bar{A}_{t,i}$ & Direct head-averaged attention from the anchor position. \\
PPR & $s_i=r_i^{(n_{\mathrm{hops}})}$ & Propagation of the same anchor-attention signal over the causal attention-induced token graph. \\
\bottomrule
\end{tabular}
\vspace{-1em}
\end{table*}

\paragraph{Diagnostic scores}
Table~\ref{tab:motivational_scores} summarizes the scores
$\mathbf{s}=(s_i)_{i\in\mathcal{W}_t}$ assigned by each method.
The Attention baseline directly uses the head-averaged anchor
attention, $s_i=\bar{A}_{t,i}$.
For PPR, we normalize this signal over the candidate window
to obtain the personalization vector:
\[
u_i=
\frac{\bar{A}_{t,i}}
{\sum_{j\in\mathcal{W}_t}\bar{A}_{t,j}},
\qquad i\in\mathcal{W}_t.
\]
Let $P$ denote the transition matrix obtained by restricting
head-averaged causal attention to $\mathcal{W}_t$ and
normalizing each row to sum to one.
Its entry $P_{ij}$ represents a transition from query token
$i$ to key token $j$, with $P_{ij}=0$ for $j>i$.
Using row-vector distributions, we initialize and update PPR as
\[
\mathbf{r}^{(0)}=\mathbf{u},
\qquad
\mathbf{r}^{(k)}
=
\alpha\mathbf{r}^{(k-1)}P
+
(1-\alpha)\mathbf{u},
\qquad k=1,\ldots,n_{\mathrm{hops}},
\]
and set $s_i=r_i^{(n_{\mathrm{hops}})}$.
Comparing Attention and PPR therefore tests the effect of
propagating a common anchor-attention signal.
This diagnostic uses attention-based personalization,
whereas deployed PRKV uses the recency-biased personalization
described in Section~\ref{sec:method_unified}.
For the Frequency baseline, $\operatorname{tok}(j)$ denotes
the token ID at position $j$, and $\mathbf{1}\{\cdot\}$ is
the indicator function. Counts are computed over the observed
slice prefix $\{1,\ldots,t\}$.

\paragraph{Metrics}
Figure~\ref{fig:motivational_analysis}(a--c) evaluates agreement with the future-attention reference, while Figure~\ref{fig:motivational_analysis}(d) measures output preservation under eviction. Let $\operatorname{TopK}(\mathbf{s},K)$ denote the indices of the $K$ highest-scoring candidates under the ordering used for retention. For a retention fraction $f\in(0,1]$, we set $K(f)=\max(1,\lfloor fW\rfloor)$.

\begin{itemize}

\item \textbf{(a) Top-$K$ overlap.} We measure agreement between the retained set selected using $\mathbf{s}$ and the reference set selected using $\mathbf{a}$:
\[
\operatorname{Overlap}(\mathbf{s},\mathbf{a};f)=\frac{\left|\operatorname{TopK}(\mathbf{s},K(f))\cap\operatorname{TopK}(\mathbf{a},K(f))\right|}{K(f)}.
\]
This is the fraction of selected tokens that also belong to the reference top-$K$ set. Higher values indicate better identification of tokens that subsequently receive high attention under the same retention budget. Panel (a) reports this metric as a function of $f$.

\item \textbf{(b) Spearman correlation with later-window attention.} We compute the Spearman rank correlation $\rho_{\mathrm{S}}(\mathbf{r}^{(n_{\mathrm{hops}})},\mathbf{a})$ between the PPR scores and the future-attention reference. Panel (b) reports this correlation as a function of $n_{\mathrm{hops}}$, measuring how full-ranking agreement with subsequent token usage changes with propagation depth. Higher values indicate stronger rank alignment.

\item \textbf{(c) Attention-retention AUC.} When retaining the $K$ highest-scoring candidates, the preserved future-attention mass is
\[
R(\mathbf{s},\mathbf{a};K)=\sum_{i\in\operatorname{TopK}(\mathbf{s},K)}a_i.
\]
We summarize this quantity across retention budgets using the discrete attention-retention AUC:
\[
\mathrm{AUC}_{\mathrm{retained}}(\mathbf{s},\mathbf{a})=\frac{1}{W}\sum_{K=1}^{W}R(\mathbf{s},\mathbf{a};K).
\]
This is equivalent to averaging the remaining future-attention mass as candidates are progressively evicted in increasing score order. Higher values indicate that more future-attention mass within the candidate window is preserved across retention budgets. Panel (c) compares this metric across scoring methods.

\item \textbf{(d) Eviction-induced KL divergence.} Let $p_{\ell}^{\mathrm{full}}$ and $p_{\ell}^{\mathrm{evict}}$ denote the next-token distributions at continuation index $\ell$ under the full-KV model and the model using score-based KV eviction, respectively. Both distributions are obtained from the model logits over a vocabulary of size $V$. We compute the forward KL divergence
\[
D_{\mathrm{KL}}\!\left(p_{\ell}^{\mathrm{full}}\middle\|p_{\ell}^{\mathrm{evict}}\right)=\sum_{v=1}^{V}p_{\ell}^{\mathrm{full}}(v)\log\frac{p_{\ell}^{\mathrm{full}}(v)}{p_{\ell}^{\mathrm{evict}}(v)},
\]
and average it over the $N_{\mathrm{cont}}$ evaluated continuation positions:
\[
\mathrm{MeanKL}=\frac{1}{N_{\mathrm{cont}}}\sum_{\ell=1}^{N_{\mathrm{cont}}}D_{\mathrm{KL}}\!\left(p_{\ell}^{\mathrm{full}}\middle\|p_{\ell}^{\mathrm{evict}}\right).
\]
Lower values indicate less eviction-induced output distortion and better preservation of the full-KV output distribution. Panel (d) compares $\mathrm{MeanKL}$ across scoring methods.

\end{itemize}

\section{Algorithm}
\label{appendix:algorithm}
We provide pseudocode for the inference-time cache-management procedure of PRKV. Algorithm~\ref{alg:prkv-overview} shows when compression is invoked in the prefill and decoding eviction modes, and Algorithm~\ref{alg:prkv} details \textsc{Compress}, the scoring and eviction procedure performed at each compression step. Within \textsc{Compress}, PRKV selects prefix landmarks with Algorithm~\ref{alg:landmark}, and the transition operator $\textsc{ApplyTransition}(\cdot)$ is given by Algorithm~\ref{alg:apply-nodes} for PRKV and by Algorithm~\ref{alg:apply-global} for PRKV-GLB. Prefill and decoding share this mechanism; they differ in when compression is invoked and in the scoring states available at each compression step, which for PRKV determine the node set and whether landmarks are pinned (Section~\ref{sec:method_landmark}). The algorithms are written for a single KV head and are applied independently per KV head, except that landmarks are selected from head-averaged keys and shared across heads.

\begin{algorithm}[H]
\caption{PRKV Inference-Time Cache Management}
\label{alg:prkv-overview}
\begin{algorithmic}[1]
\STATE{\textbf{Input:} Cache budget $K$, compression interval $\Delta$, variant $\mathcal{V} \in \{\text{PRKV}, \text{PRKV-GLB}\}$}
\STATE{/* Prefill eviction mode */}
\STATE Process the prompt; construct the KV cache $\mathcal{C}$ and collect hidden states $\mathcal{H}_{\mathrm{pre}}$
\STATE $\mathcal{C} \leftarrow \textsc{Compress}(\mathcal{C}, \mathcal{H}_{\mathrm{pre}}, K, \mathcal{V}, \text{prefill})$
\STATE{/* Decoding eviction mode */}
\WHILE{generation has not terminated}
  \STATE Generate the next token; append its KV state to $\mathcal{C}$ and its hidden state to $\mathcal{H}_{\mathrm{dec}}$
  \IF{$\Delta$ tokens have been generated since the last compression}
    \STATE $\mathcal{C} \leftarrow \textsc{Compress}(\mathcal{C}, \mathcal{H}_{\mathrm{dec}}, K, \mathcal{V}, \text{decoding})$; \ $\mathcal{H}_{\mathrm{dec}} \leftarrow \emptyset$
  \ENDIF
\ENDWHILE
\end{algorithmic}
\end{algorithm}
\begin{algorithm}[h]
\caption{Compress: PRKV Eviction at One Compression Step}
\label{alg:prkv}
\begin{algorithmic}[1]
\STATE{\textbf{Input:} Cache $\mathcal{C} = \{(k_i, v_i)\}$, scoring states $\mathcal{H}$, budget $K$, variant $\mathcal{V}$, mode}
\STATE{\textbf{Input:} Window size $W$, chunk size $C$, recency fraction $f$, boost $b$}
\STATE{\textbf{Input:} Damping $\alpha$, steps $n_{\mathrm{hops}}$, $\epsilon > 0$}
\STATE{\textbf{Output:} Retained token set $S \subseteq \mathcal{C}$}
\STATE{/* Node set and protected tokens */}
\STATE $\mathcal{R} \leftarrow$ the $W$ most recent tokens
\IF{$\mathcal{V} = \text{PRKV}$}
  \STATE $\mathcal{L} \leftarrow \textsc{SelectLandmarks}(\{k_i\}, W, C)$ \quad (Alg.~\ref{alg:landmark})
  \STATE $\mathcal{N} \leftarrow \mathcal{L} \cup \mathcal{R}$ in prefill (Eq.~\ref{eq:eq9}), and $\mathcal{N} \leftarrow \mathcal{R}$ in decoding (Eq.~\ref{eq:eq11})
\ELSE
  \STATE $\mathcal{N} \leftarrow \mathcal{C}$
\ENDIF
\STATE $\mathcal{P} \leftarrow$ sink tokens (and up to $5\%$ of $K$ landmarks from $\mathcal{L}$, for PRKV in decoding)
\STATE{/* Recency-biased personalization */}
\STATE $\mathcal{U} \leftarrow \mathcal{N}$ for PRKV and $\mathcal{R}$ for PRKV-GLB; \ $n \leftarrow \max\{1, \mathrm{round}(f\,|\mathcal{U}|)\}$
\STATE $u_i \leftarrow \mathbf{1}\{i \in \mathcal{U}\}$ for $i \in \mathcal{N}$ \quad ($u \in \mathbb{R}^{|\mathcal{N}|}$)
\STATE $u_i \leftarrow b \cdot u_i$ for the $n$ most recent tokens $i \in \mathcal{U}$
\STATE $u \leftarrow \textsc{Normalize}(u)$ \quad (rescale to sum to one)
\STATE{/* PPR diffusion */}
\STATE $r \leftarrow u$
\FOR{$k = 1$ \textbf{to} $n_{\mathrm{hops}}$}
    \STATE $r \leftarrow \alpha\, \textsc{ApplyTransition}(r) + (1-\alpha)\, u$ \quad (Alg.~\ref{alg:apply-nodes} or~\ref{alg:apply-global})
  \STATE $r \leftarrow \textsc{Normalize}(r)$
\ENDFOR
\STATE{/* Value-aware scoring and selection */}
\STATE $\hat\pi_i \leftarrow r_i$ for $i \in \mathcal{N}$, \ $\hat\pi_i \leftarrow 0$ for $i \in \mathcal{C} \setminus \mathcal{N}$
\STATE $s_i \leftarrow (\hat\pi_i + \epsilon)\, \|v_i\|_2$ for each $i \in \mathcal{C}$
\STATE $S \leftarrow \mathcal{P} \cup \operatorname{TopK}\bigl(\{s_i\}_{i \in \mathcal{C} \setminus \mathcal{P}},\, K - |\mathcal{P}|\bigr)$
\STATE \textbf{Return} $S$
\end{algorithmic}
\end{algorithm}
\vspace{-1em}
\begin{algorithm}[H]
\caption{SelectLandmarks: Prefix Landmark Selection}
\label{alg:landmark}
\begin{algorithmic}[1]
\STATE{\textbf{Input:} Cached keys $\{k_i\}_{i=1}^{T}$, window size $W$, chunk size $C$}
\STATE{\textbf{Output:} Landmark set $\mathcal{L}$}
\STATE $\mathcal{Z} \leftarrow \{n_{\mathrm{sink}}+1, \dots, T-W\}$ \quad (non-sink prefix)
\STATE $\tilde{k}_i \leftarrow \bar{k}_i / \lVert \bar{k}_i \rVert_2$ for $i \in \mathcal{Z}$ \quad ($\bar{k}_i$: key averaged over KV heads)
\STATE $\bar{a} \leftarrow \sum_{i \in \mathcal{Z}} \tilde{k}_i \,/\, \lVert \sum_{i \in \mathcal{Z}} \tilde{k}_i \rVert_2$
\STATE Partition $\mathcal{Z}$ into consecutive chunks $\{\mathcal{Z}_c\}$ of size $C$
\STATE $\ell_c \leftarrow \arg\min_{i \in \mathcal{Z}_c} \langle \tilde{k}_i, \bar{a} \rangle$ for each chunk $c$
\STATE \textbf{Return} $\mathcal{L} = \{\ell_c\}_c$
\end{algorithmic}
\end{algorithm}
\vspace{-1em}
\begin{algorithm}[H]
\caption{ApplyTransition: Node Attention}
\label{alg:apply-nodes}
\begin{algorithmic}[1]
\STATE{\textbf{Input:} Node set $\mathcal{N}$ with keys $\{k_p\}_{p \in \mathcal{N}}$ and queries $\{q_p\}_{p \in \mathcal{N}}$ computed from $\mathcal{H}$}
\STATE{\textbf{Input:} Current vector $r \in \mathbb{R}^{|\mathcal{N}|}$}
\STATE{\textbf{Output:} $rP \in \mathbb{R}^{|\mathcal{N}|}$}
\STATE $P_{p,:} \leftarrow \operatorname{softmax}_{j}\bigl(q_p k_j^{\top}/\sqrt{d_k}\bigr)$ over $j \in \mathcal{N}$, $j \le p$ \quad (causal mask on token positions)
\STATE $rP_j \leftarrow \sum_{p \in \mathcal{N}} r_p P_{p,j}$ for $j \in \mathcal{N}$
\STATE \textbf{Return} $rP$
\end{algorithmic}
\end{algorithm}
\vspace{-1em}
\begin{algorithm}[H]
\caption{ApplyTransition: Global Co-Attention}
\label{alg:apply-global}
\begin{algorithmic}[1]
\STATE{\textbf{Input:} Window attention $A \in \mathbb{R}^{W \times T}$ of the window queries computed from $\mathcal{H}$}
\STATE{\textbf{Input:} Current vector $r \in \mathbb{R}^{T}$}
\STATE{\textbf{Output:} $rP \in \mathbb{R}^{T}$}
\STATE $c_j \leftarrow \sum_{u=1}^W A_{u,j}$ \quad for $j=1,\dots,T$
\STATE $c \leftarrow \max(c,\epsilon)$
\STATE $w_j \leftarrow r_j / c_j$
\STATE $z_u \leftarrow \sum_{j=1}^T w_j A_{u,j}$ \quad for $u=1,\dots,W$
\STATE $rP_j \leftarrow \sum_{u=1}^W z_u A_{u,j}$
\STATE \textbf{Return} $rP$
\end{algorithmic}
\end{algorithm}
\vspace{-1em}

\section{Perturbation bound of local PPR} \label{appendix:ppr_perturbation}
We provide a formal perturbation bound for PPR that quantifies the gap between the global attention transition $P$ and the local/approximated transition $Q$. For row-stochastic $P$, $Q$ and distributions $u$, $w$, the PPR fixed point satisfies the Lipschitz bound
\begin{equation}
\lVert{\pi(P,u)-\pi(Q,w)}\rVert_1 \le \lVert{u-w}\rVert_1 +
\frac{\alpha}{1-\alpha}\lVert{P-Q}\rVert_{1, 1}.
\end{equation}

We provide detailed proof structure below. 
Recall the fixed point equation of PPR (Eq. ~\ref{eq:ppr}). 
We use the operator norm induced by the $\ell_1$ norm on row vectors,
\(
  \|A\|_{1,1} := \sup_{x \neq 0} \frac{\|xA\|_1}{\|x\|_1}
             = \max_i \sum_j |A_{ij}|,
\)
i.e., the maximum absolute row sum. Hence $\|P\|_{1,1}=1$ for any row-stochastic $P$.

\begin{lemma}[Solution of PPR]\label{lemma1}
For any row-stochastic $P$ and any distribution $u$, the unique solution to Eq. ~\ref{eq:ppr} is
\begin{equation}
\pi(P,u)=(1-\alpha)u(I-\alpha P)^{-1}. 
\label{eq:eq17}
\end{equation}

Moreover,
\begin{equation}    
\|{(I-\alpha P)^{-1}}\|_{1, 1}\le\frac{1}{1-\alpha}. 
\label{eq:eq18}
\end{equation}
\end{lemma}

\begin{proof}
Since $\lVert{\alpha P}\rVert_{1, 1}=\alpha<1$, the Neumann series converges:
$(I-\alpha P)^{-1}=\sum_{t=0}^{\infty}(\alpha P)^t,$
hence $I-\alpha P$ is invertible and Eq.~\ref{eq:eq17} follows.

For Eq.~\ref{eq:eq18}, use submultiplicativity and $\lVert{P^t}\rVert_{1, 1}\le \lVert{P}\rVert_{1, 1}^t = 1$:
\begin{equation}
\lVert{(I-\alpha P)^{-1}}\rVert_{1, 1}
\le \sum_{t=0}^{\infty}\lVert{(\alpha P)^t}\rVert_{1, 1}
\le \sum_{t=0}^{\infty}\alpha^t
= \frac{1}{1-\alpha}. 
\end{equation}
In this regard, we can get a perturbation bound of PPR for row-stochastic matrices:
\end{proof}

\begin{theorem}
    Let $P,Q \in \mathbb{R}^{n\times n}$ be row-stochastic, and let $u,w \in \mathbb{R}^n$ be distributions. Then
    \begin{equation}
    \lVert{\pi(P,u)-\pi(Q,w)}\rVert_1 \le \lVert{u-w}\rVert_1 +
    \frac{\alpha}{1-\alpha}\lVert{P-Q}\rVert_{1, 1}.
    \label{eq:eq21}
    \end{equation}
\end{theorem} 

\begin{proof}
    
Using Lemma~\ref{lemma1}
\begin{equation}    
\pi(P,u)-\pi(Q,w) = (1-\alpha)\bigl(u(I-\alpha P)^{-1}-w(I-\alpha Q)^{-1}\bigr).
\end{equation}

Add and subtract $(1-\alpha)w(I-\alpha P)^{-1}$:
\begin{align*}
\pi(P,u)-\pi(Q,w)&=(1-\alpha)(u-w)(I-\alpha P)^{-1} \\ &+(1-\alpha)w\bigl((I-\alpha P)^{-1}-(I-\alpha Q)^{-1}\bigr). 
\end{align*}

For the second term use the resolvent identity
\begin{equation}
(I-\alpha P)^{-1}-(I-\alpha Q)^{-1} = (I-\alpha P)^{-1}\,\alpha(P-Q)\,(I-\alpha Q)^{-1}. 
\end{equation}
Taking $\ell_1$ norms and applying submultiplicativity yields

\begin{equation}
\begin{split}
\lVert{\pi(P,u)-\pi(Q,w)}\rVert_1 &\le (1-\alpha)\lVert{u-w}\rVert_1 \lVert{(I-\alpha P)^{-1}}\rVert_{1, 1} \\
&+(1-\alpha)\lVert{w}\rVert_1 \lVert{(I-\alpha P)^{-1}}\rVert_{1, 1}\,\alpha\lVert{P-Q}\rVert_{1, 1}\lVert{(I-\alpha Q)^{-1}}\rVert_{1, 1}.
\end{split}
\end{equation}

Since $w$ is a distribution, $\lVert{w}\rVert_1=1$. Applying Lemma~\ref{lemma1} to both resolvents gives
\begin{equation}
\begin{split}
\lVert{\pi(P,u)-\pi(Q,w)}\rVert_1 &\le (1-\alpha)\lVert{u-w}\rVert_1\cdot \frac{1}{1-\alpha} \\
&\quad + (1-\alpha)\cdot 1\cdot \frac{1}{1-\alpha}\cdot \alpha\lVert{P-Q}\rVert_{1, 1}\cdot \frac{1}{1-\alpha}.
\end{split}
\label{eq:ppr_bound_expand}
\end{equation}

The first term equals $\lVert{u-w}\rVert_1$ and the second simplifies to $\frac{\alpha}{1-\alpha}\lVert{P-Q}\rVert_{1, 1}$, proving Eq.~\ref{eq:eq21}. 
\end{proof}

For window truncation, with $Q$ lifted to the full state space by restricting each retained row of $P$ to the retained window and renormalizing it, while copying the remaining rows of $P$, $\lVert P-Q \rVert_{1,1}$ is controlled by the omitted attention mass $m_i$ (outside the retained window), yielding error $$\lVert \pi(P,u)-\pi(Q,w)\rVert_1 \le \frac{2\alpha}{1-\alpha}\max_i m_i, $$ (for the same personalization). 
PRKV is designed to further reduce this error via landmark nodes: intuitively, the error can be attributed to (i) uncovered mass outside the node set and (ii) the error of representing each chunk by its landmark, suggesting the approximation is tight when attention is locally concentrated or well
summarized by landmarks.

\paragraph{Empirical locality in attention logits}

\begin{figure}[h]
    \centering
    \subfigure[Outside mass\label{fig:locality_outmass}]{
    \includegraphics[width=0.3\linewidth]{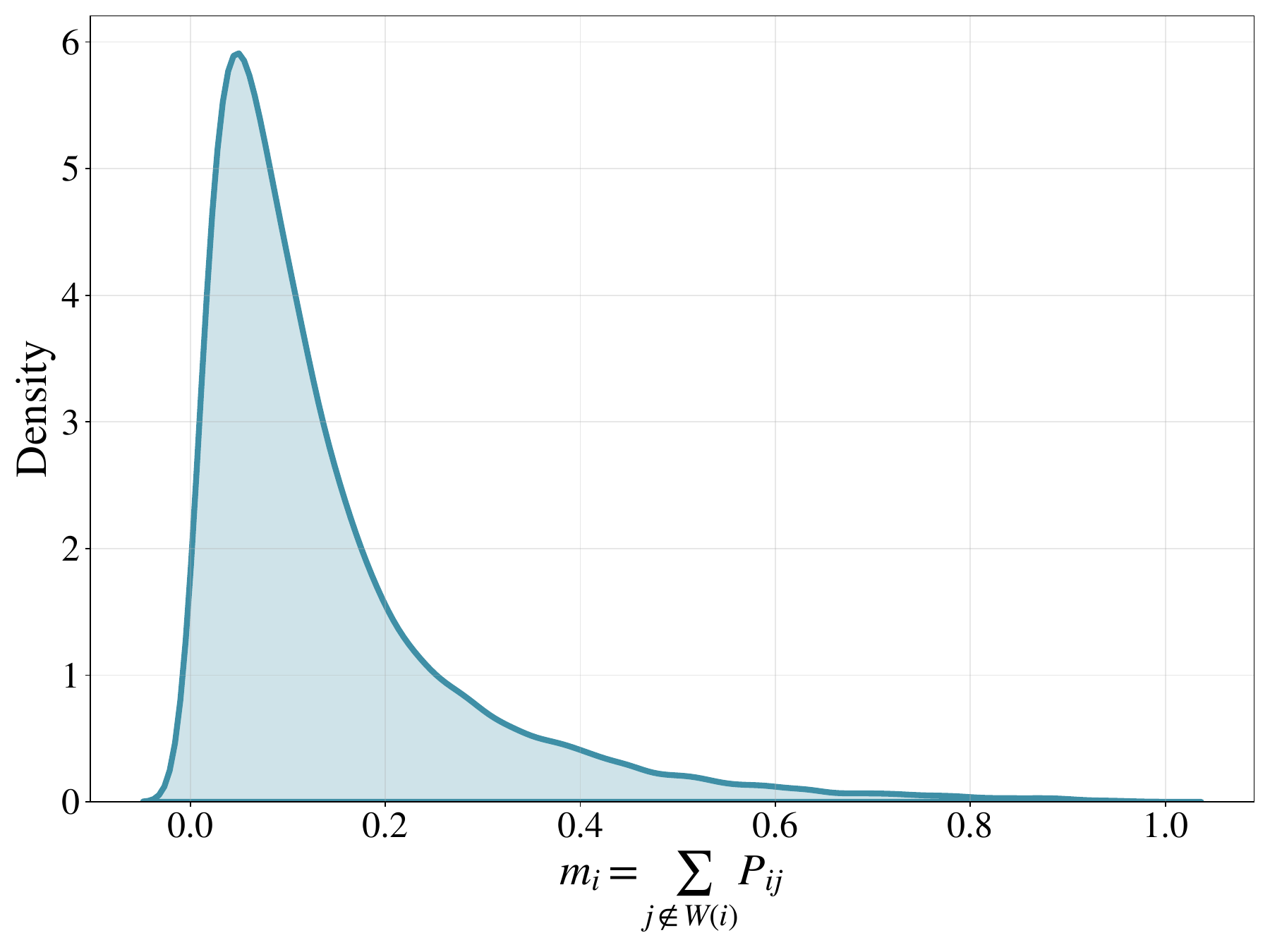}
    }
    \subfigure[Logit separation margin\label{fig:locality_logit}]{
    \includegraphics[width=0.3\linewidth]{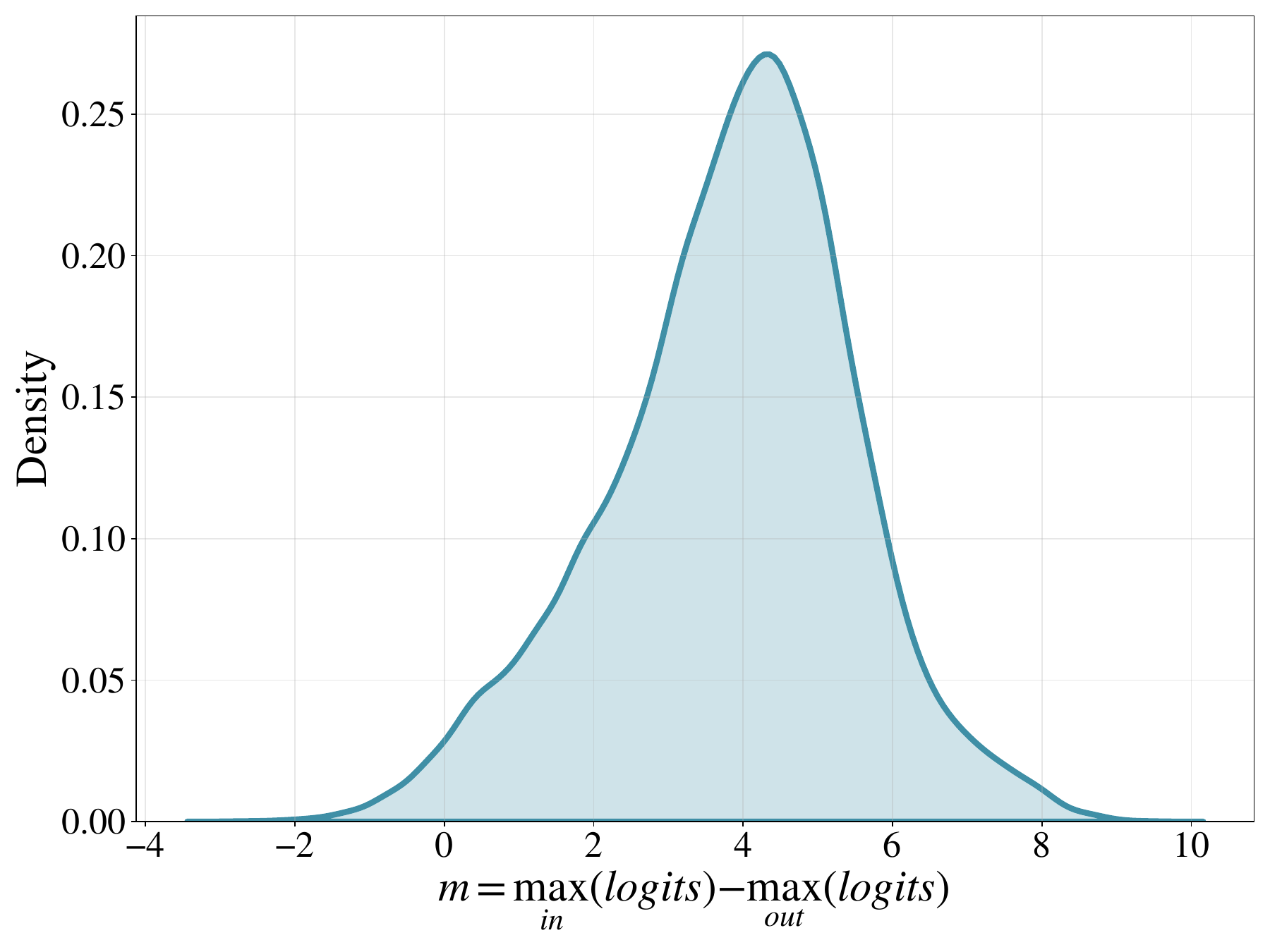}
    }
    \subfigure[Exponential transform\label{fig:locality_exp}]{
    \includegraphics[width=0.3\linewidth]{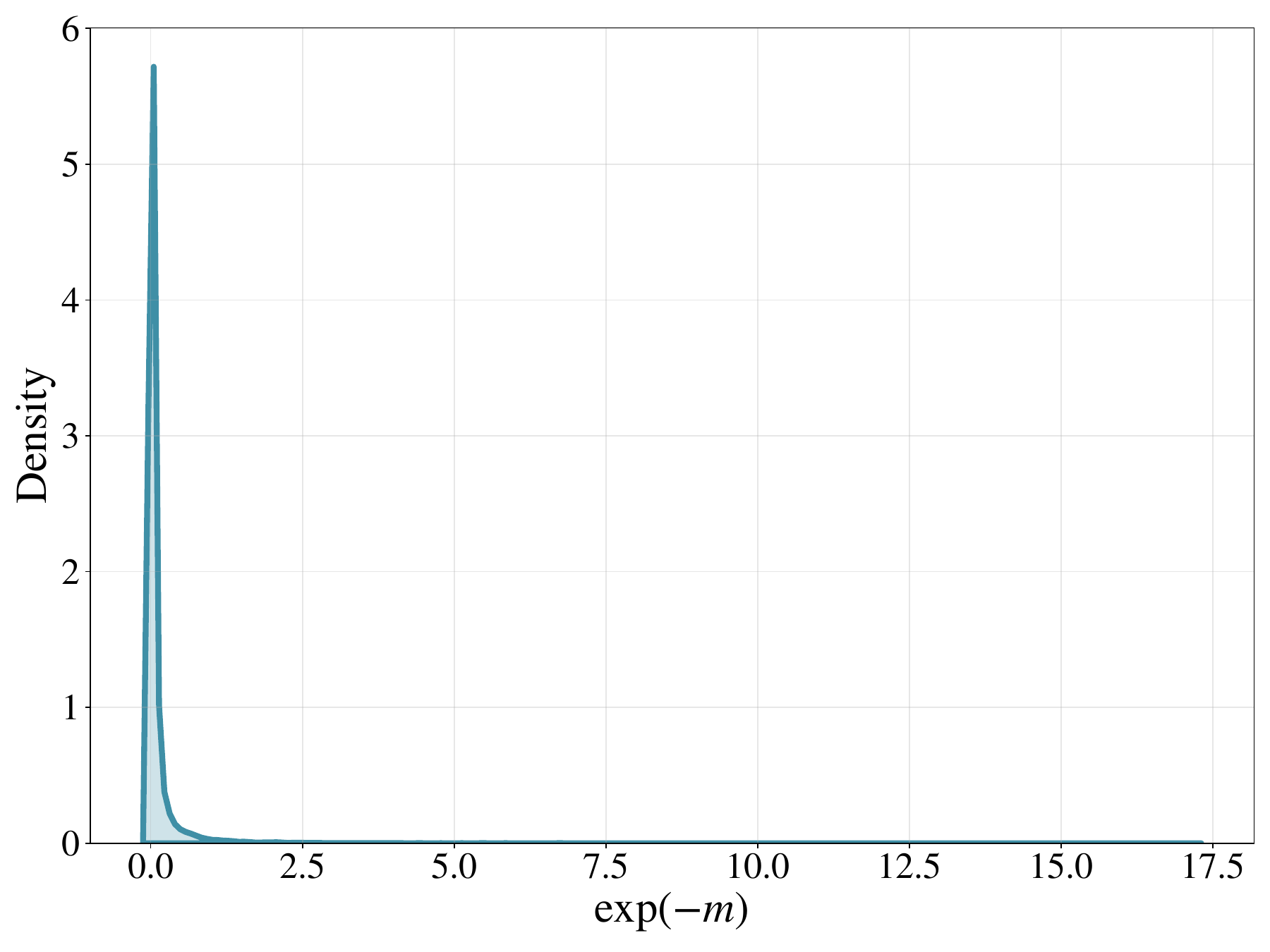}
    }
    \caption{\textbf{Empirical analyses of locality.} (a) Empirical distributions of outside-window attention mass, (b) logit separation margins, and (c) their exponential transform, measured on 13\textsuperscript{th} layer of Llama-3.1-8B using ten samples from LongBench NarrativeQA. The distributions indicate strong preference for in-window tokens, supporting the locality assumptions used in our analysis.}
    \label{fig:locality}
\end{figure}

To empirically validate the locality assumption underlying our analysis, we examine the distribution of attention separation margins in a setting where locality is expected to be pronounced. Let \(P \in \mathbb{R}^{T \times T}\) denote the row-stochastic attention matrix, where \(P_{ij}\) is the attention probability from query \(i\) to key \(j\). 
Specifically, we sample ten instances from the NarrativeQA split of LongBench and focus on a representative layer that exhibits strong locality behavior. 
Figure~\ref{fig:locality} visualizes three complementary statistics. Figure~\ref{fig:locality_outmass} shows the distribution of outside-window attention mass $m_i = \sum_{j \notin W(i)} P_{ij}$, which concentrates near zero, indicating that only a small fraction of attention mass flows outside the local window for most queries.
Figure~\ref{fig:locality_logit} reports the logit separation margin between the strongest in-window and out-of-window tokens, revealing consistently positive margins with a peak well above zero. This implies that, at the logit level, local tokens typically dominate their non-local competitors.
Finally, Figure~\ref{fig:locality_exp} shows the distribution of $\exp(-m)$, which is sharply concentrated near zero, further confirming strong separation in favor of local context. Together, these observations demonstrate that, in practice, attention in this regime is strongly biased toward local context: both probability mass and logit-level competition are dominated by tokens within the window. This empirical behavior explains why window-restricted importance estimation remains stable in typical cases.

\section{Extended Empirical Analyses}\label{appendix:score_behavior}

\paragraph{Case study: token selection under eviction}

\begin{figure*}[h]
    \centering
    \subfigure[Observed Attention]{
    \includegraphics[width=.3\linewidth]{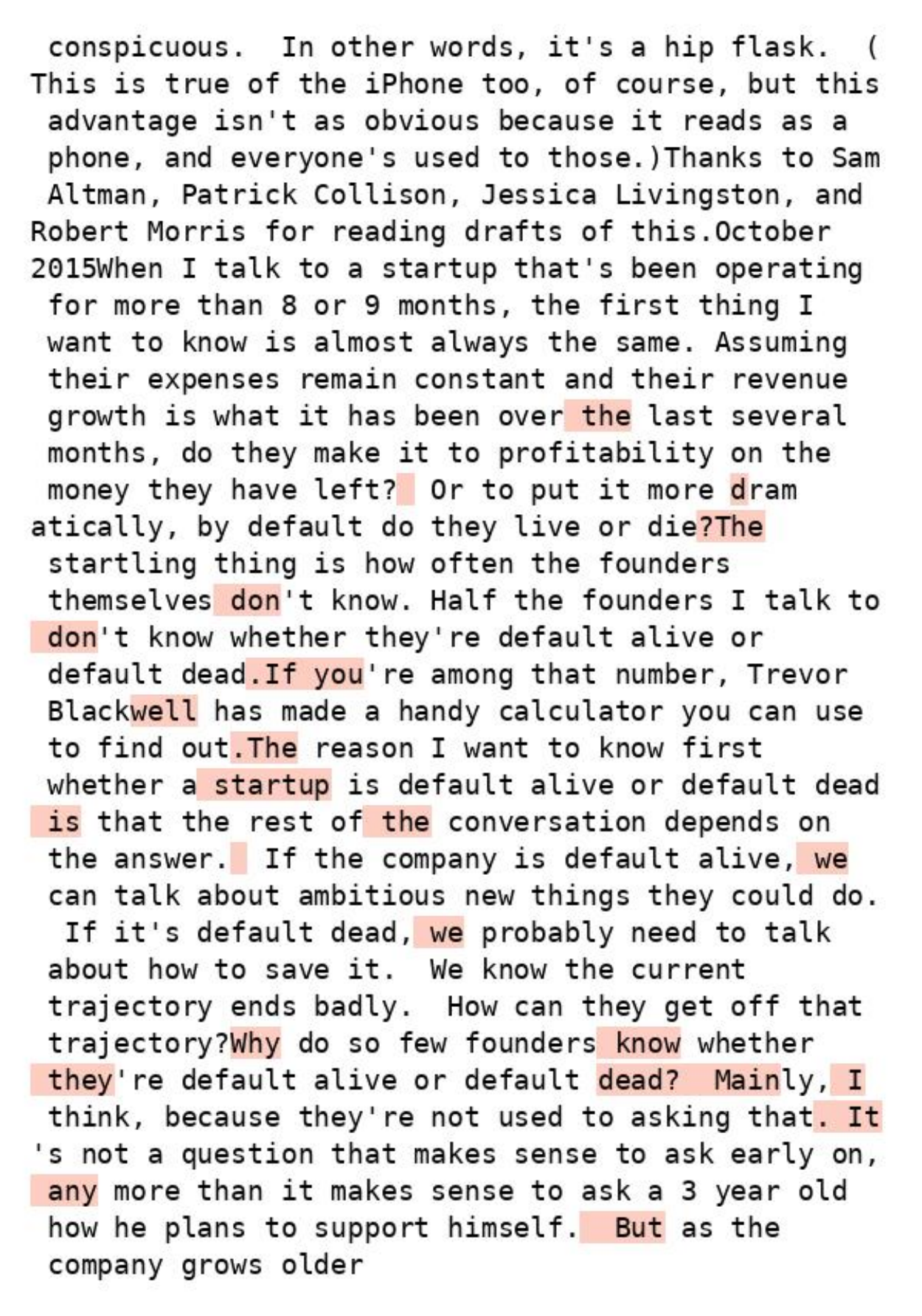}
    }
    \subfigure[PRKV]{
    \includegraphics[width=.3\linewidth]{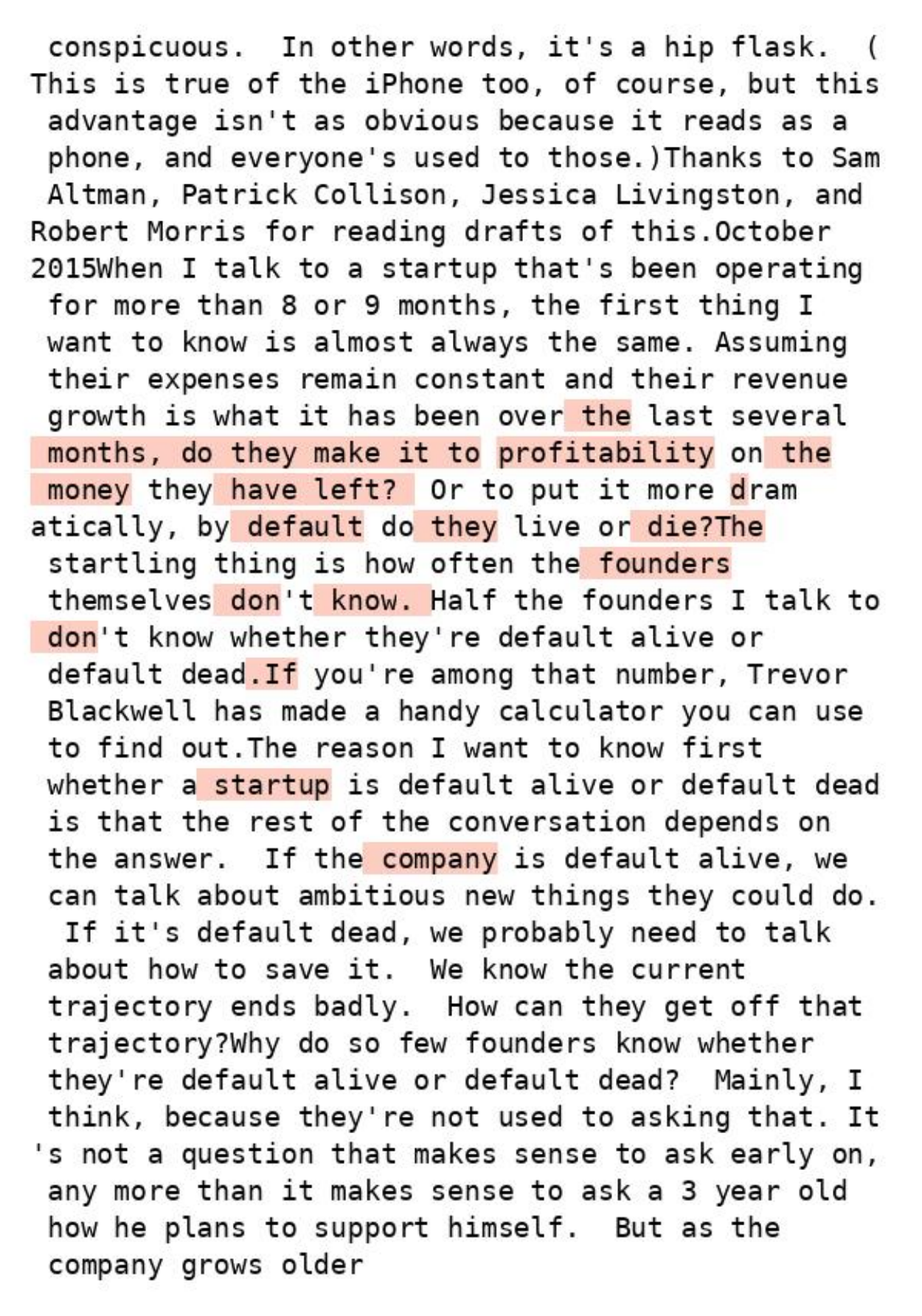}
    }
    \subfigure[PRKV-GLB]{
    \includegraphics[width=.3\linewidth]{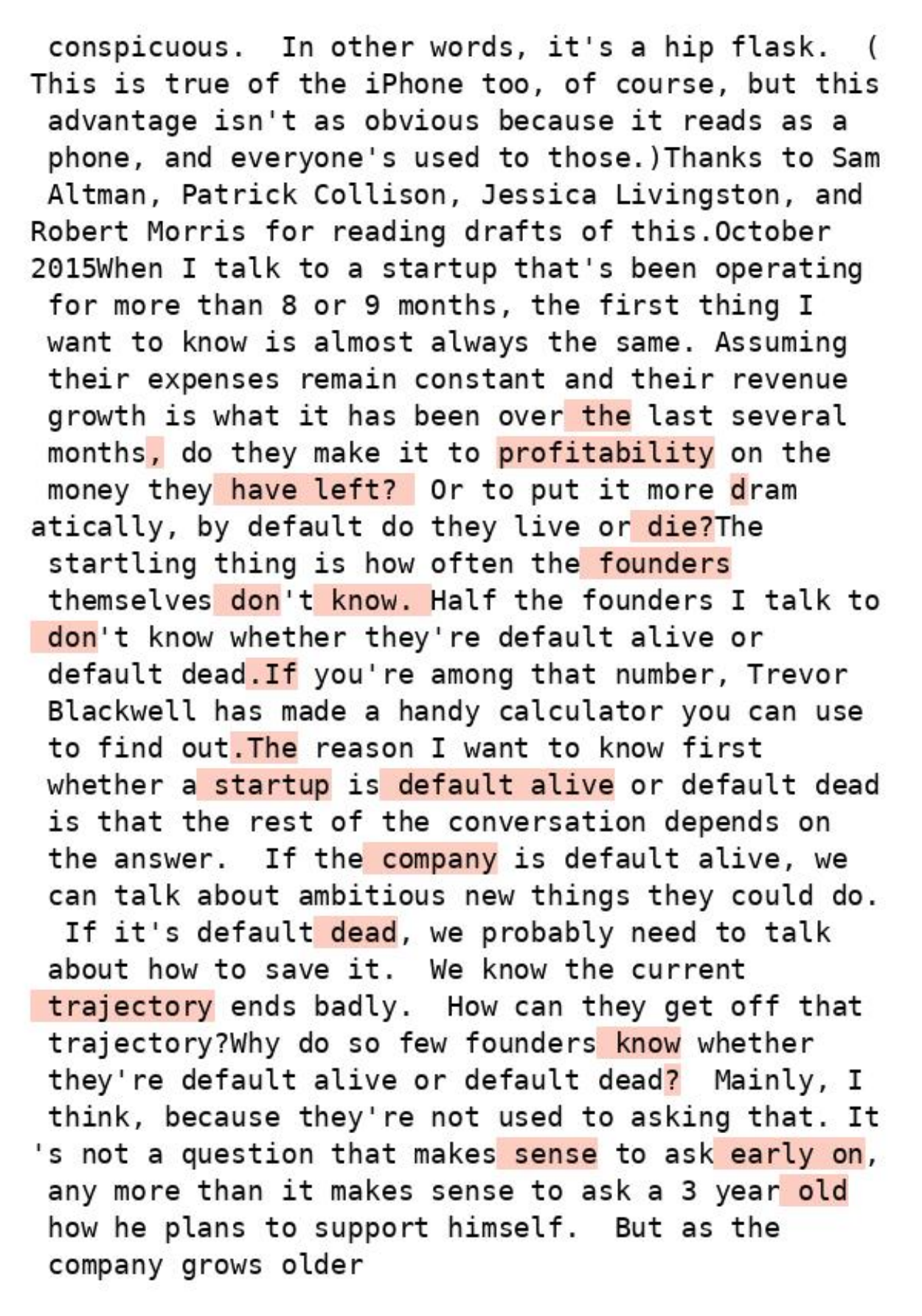}
    }
    \caption{\textbf{Case study of token selection.}
    We report token selection result of (a) observed attention, (b) PRKV, and (c) PRKV-GLB. All measurements are conducted using the \texttt{bfloat16} data type on 11\textsuperscript{th} layer, 27\textsuperscript{th} head of Llama-3.1-8B with PaulGrahamEssays dataset.}
    \label{fig:topk_selection}
\end{figure*}

Figure~\ref{fig:topk_selection} presents a qualitative comparison of token selection under top-$K$ eviction. We visualize the tokens retained by (a) observed attention, (b) PRKV, and (c) PRKV-GLB on a long input from the PaulGrahamEssays dataset. Compared to observed attention, PRKV consistently preserves tokens that
correspond to semantically salient words and phrases, such as content-bearing nouns and verbs that directly convey the meaning
of the surrounding context. This behavior reflects the locality-aware nature of PRKV: importance is primarily determined by how strongly tokens participate in local attention interactions, leading to a more focused selection of window-relevant content. PRKV-GLB, on the other hand, tends to retain a broader set of tokens, including words that are contextually related but not always locally dominant. By incorporating global propagation, PRKV-GLB captures a wider neighborhood of influence, which can be beneficial for long-range coherence, but may also result in preserving surrounding or structurally adjacent tokens in addition to the most locally salient ones.

\paragraph{Behavior analysis of PRKV and PRKV-GLB}

\begin{figure*}[h]
    \centering
    \subfigure[PRKV Score\label{fig:behavior_analysis_prkv_score}]{
    \includegraphics[width=0.23\linewidth]{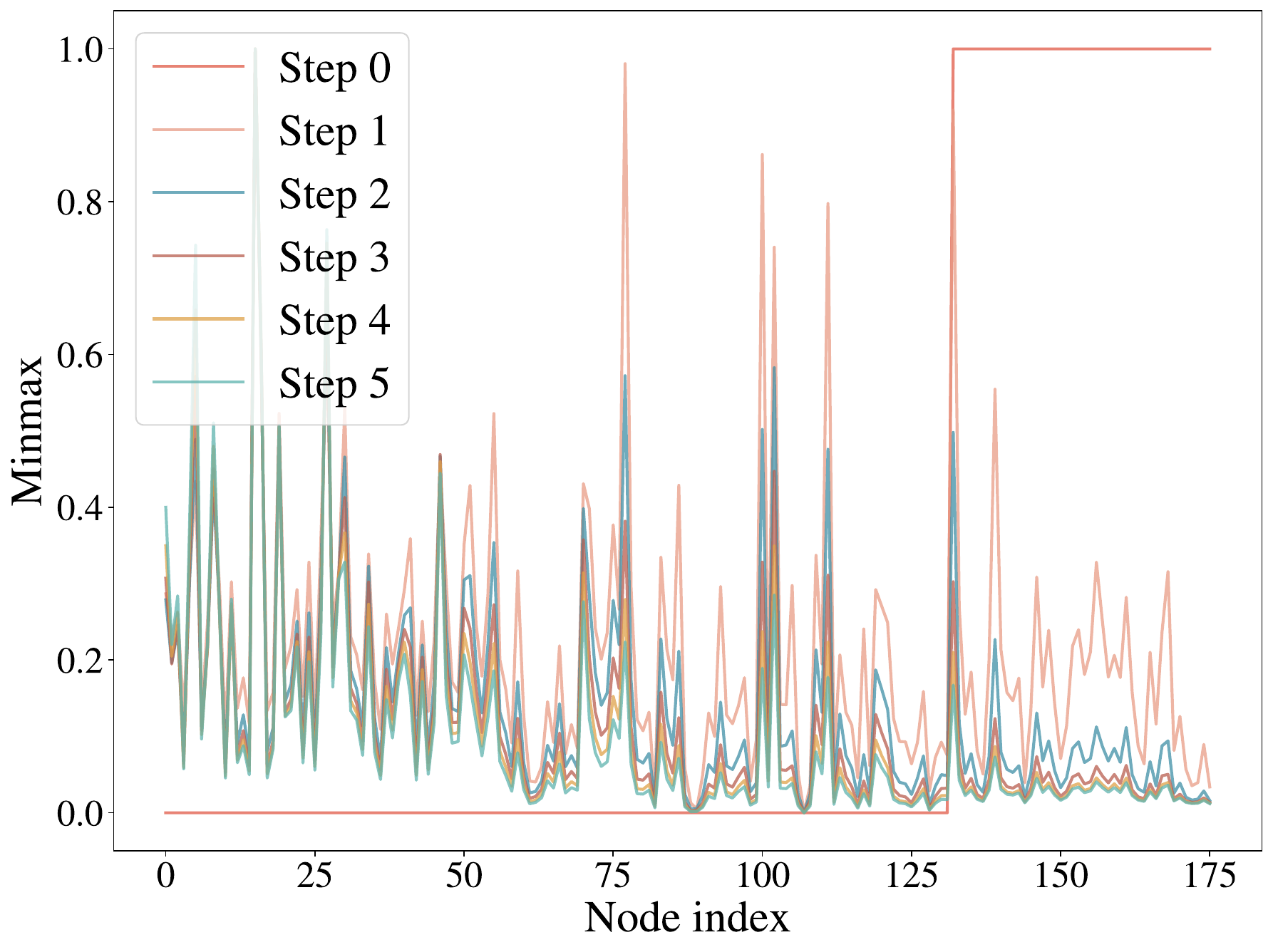}
    }
    \subfigure[PRKV Transition\label{fig:behavior_analysis_prkv_transition}]{
    \includegraphics[width=0.23\linewidth]{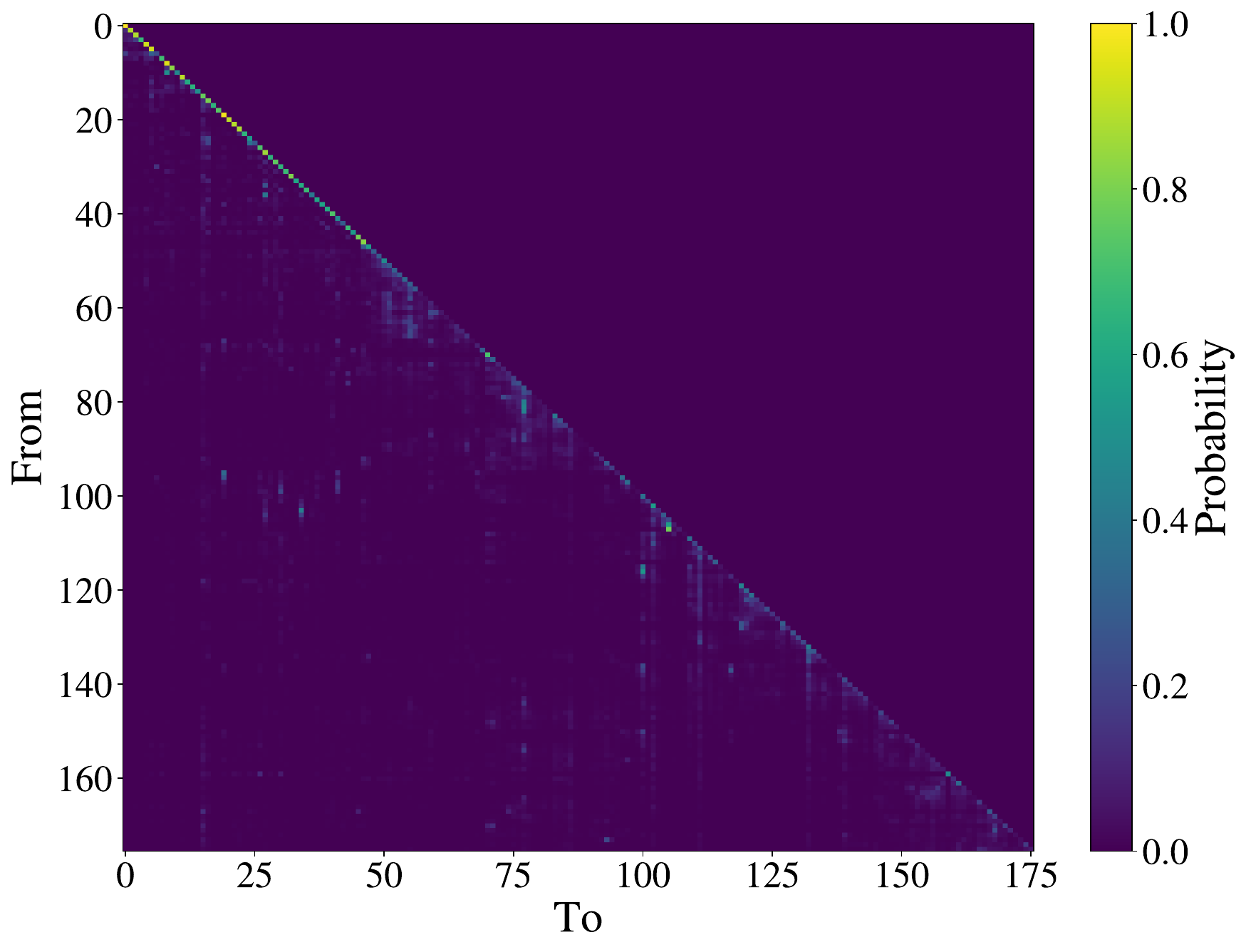}
    }
    \subfigure[PRKV-GLB Score\label{fig:behavior_analysis_prkv_glb_score}]{
    \includegraphics[width=0.23\linewidth]{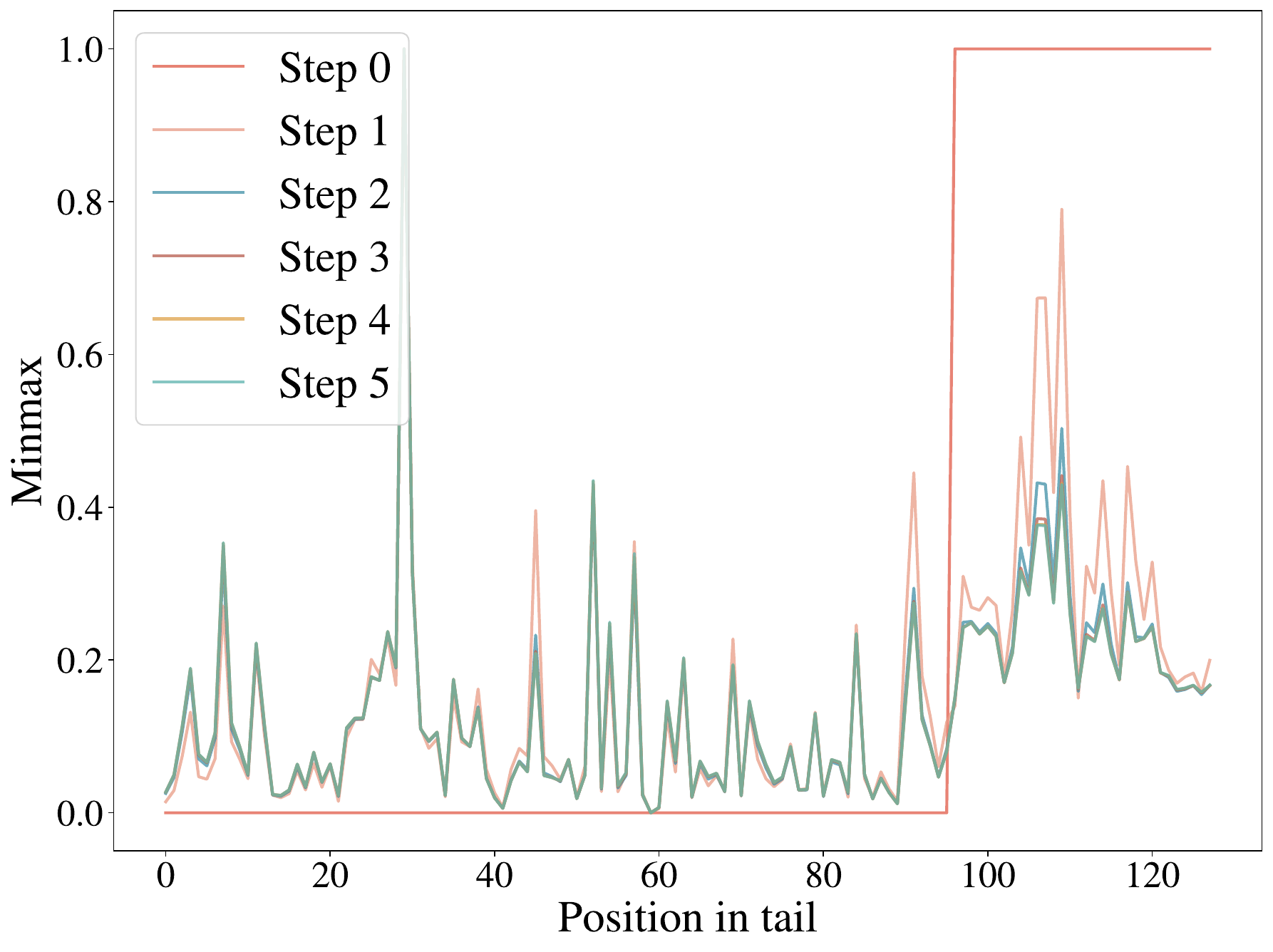}
    }
    \subfigure[PRKV-GLB Transition\label{fig:behavior_analysis_prkv_glb_transition}]{
    \includegraphics[width=0.23\linewidth]{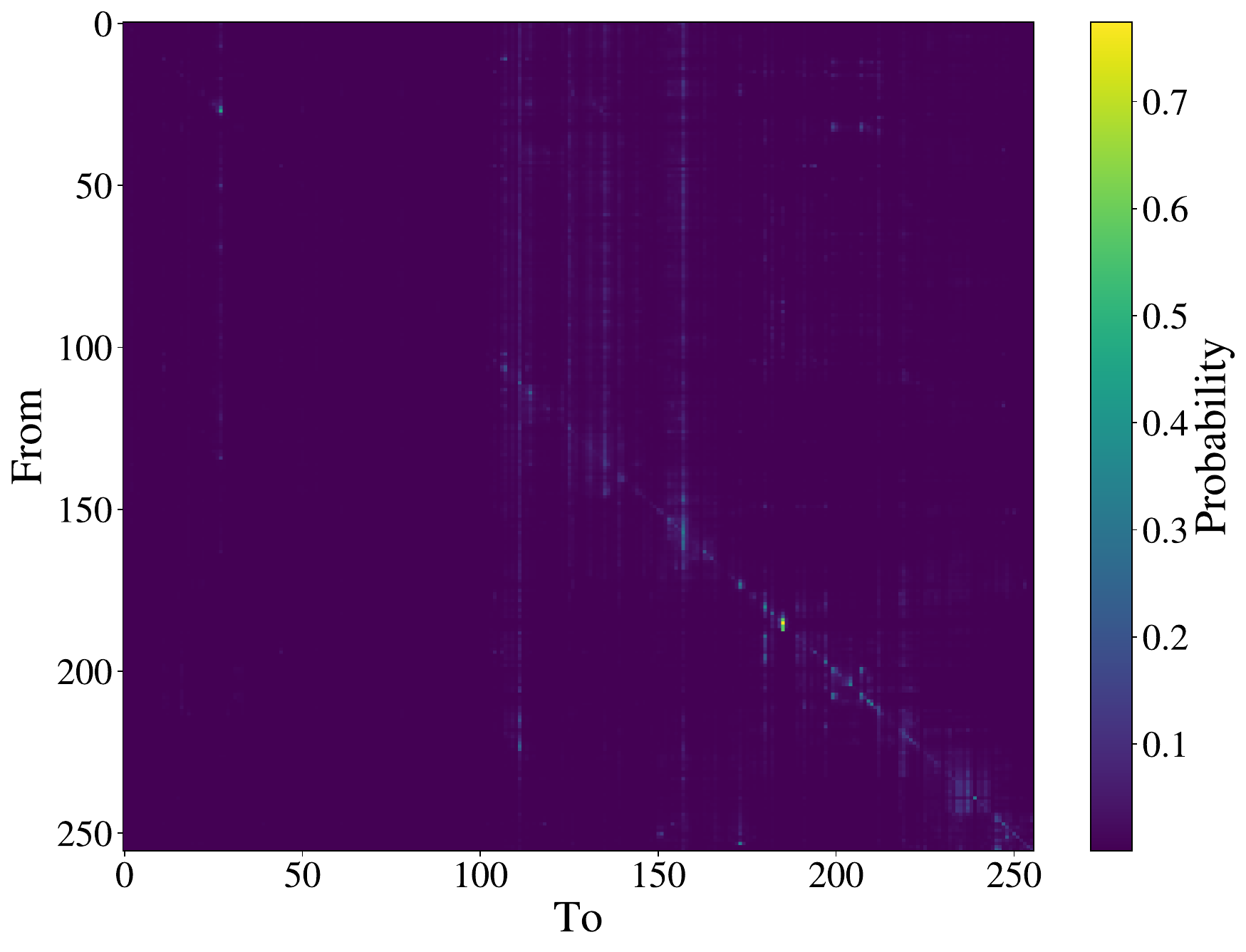}
    }
    \caption{\textbf{Behavior analysis of PRKV.} We report how transition matrix is formed and how actual PPR score evolves from personalization vector of both PRKV and PRKV-GLB. Results are obtained from 20\textsuperscript{th} layer of Llama-3.1-8B with context from PaulGrahamEssays dataset.}
    \label{fig:behavior_analysis}
\end{figure*}

Figure~\ref{fig:behavior_analysis} illustrates how the transition matrix is constructed and how PPR scores evolve from the initial personalization vector for both PRKV and PRKV-GLB. For PRKV, the transition matrix is derived from a locality-restricted
attention graph that includes both window tokens and a small set of landmark nodes. As shown in Figure~\ref{fig:behavior_analysis_prkv_transition}, this construction results in a sparse, causal transition structure, where probability mass primarily flows from later tokens to earlier, locally relevant tokens, with occasional connections mediated through landmarks.
In contrast, PRKV-GLB is based on co-attention $A^\top A$ between keys aggregated over the sequence. As a result, its transition matrix (Figure~\ref{fig:behavior_analysis_prkv_glb_transition}) exhibits broader connectivity that is not confined to a lower-triangular structure. We emphasize that this difference is a consequence of the construction: PRKV-GLB is not intended to represent a causal transition graph, but rather a globally smoothed influence matrix. Figures~\ref{fig:behavior_analysis_prkv_score} and \ref{fig:behavior_analysis_prkv_glb_score} track the evolution of PPR scores over iterations, starting from a uniform recent personalization vector. Across iterations, scores gradually deviate from the initialization and increasingly reflect the structure of the underlying attention graph. In PRKV, this process amplifies multi-step influence along local attention paths, leading to the emergence of tokens with sustained long-range importance. A similar evolution is observed for PRKV-GLB, albeit with smoother and more diffuse score distributions due to global propagation.

\section{Extended Ablation and Sensitivity Studies} \label{appendix:ablation}

Table~\ref{tab:ablation} reports ablations of PRKV and PRKV-GLB on Qwen3-8B. Both variants are relatively insensitive to the personalization strategy: recent-window, query-based, and query-EMA personalization yield comparable results, with query-based variants providing occasional gains at the cost of additional computation. We therefore use the recent-window strategy as a lightweight default. PRKV-specific ablations in Table~\ref{tab:ablation_lm_more} show that chunk size and landmark selection have limited impact, with cosine diversity giving a balanced default but other landmark criteria remaining competitive.

\begin{table*}[h]
  \centering
  \small
  \caption{\textbf{Extended ablation studies of PRKV and PRKV-GLB on Qwen3-8B.} We report the average of sixteen tasks from LongBench benchmarks over five compression ratios. 0\% indicates average results without compression. Bold values indicate the best performance within each block.}
  \label{tab:ablation}
  \setlength{\tabcolsep}{6pt}
  \renewcommand{\arraystretch}{1.15}
  \resizebox{\textwidth}{!}{
  \begin{tabular}{@{} l l | *{6}{r} |*{6}{r} @{}}
    \toprule
    \multirow{2}{*}{\textbf{Component}} &
    \multirow{2}{*}{\textbf{Variant}} &
    \multicolumn{6}{c}{\textbf{PRKV}} &
    \multicolumn{6}{c}{\textbf{PRKV-GLB}} \\
    \cmidrule(lr){3-8} \cmidrule(lr){9-14} 
    & & 0\% & 10\% & 25\% & 50\% & 75\% & 90\% 
    & 0\% & 10\% & 25\% & 50\% & 75\% & 90\% \\
    \midrule
    
    \multirow{3}{*}{Personalization}

    & Recent & \textbf{47.27 }& 47.74 & 48.56 & 48.92 & 46.26 & 37.59 &
    \textbf{47.27} & \textbf{47.52} & \textbf{47.86} & 48.81 & 45.84 & 39.09 \\
    & Query & \textbf{47.27} & 47.63 & \textbf{48.92} & \textbf{49.00} & \textbf{46.35} & \textbf{38.11} &
    \textbf{47.27} & 47.34 & 47.74 & 48.90 & 45.59 & \textbf{39.78} \\
    & Query EMA & \textbf{47.27} & \textbf{47.82} & 48.59 & 48.86 & 46.01 & 37.47 &
    \textbf{47.27} & 47.35 & 47.66 & \textbf{48.93} & \textbf{46.16} & 39.49 \\ \midrule
    \multirow{5}{*}{\makecell{Value-norm \\scaling}}
    & Only VNorm & \textbf{47.27} & \textbf{48.68} & 48.76 & 48.53 & 44.66 & 35.17 
    & \textbf{47.27} & \textbf{48.68} & \textbf{48.76} & 48.53 & 44.66 & 35.17 \\
    & Only PPR & \textbf{47.27} & 46.05 & 44.30 & 41.24 & 37.42 & 32.86 & \textbf{47.27} & 47.10 & 46.70 & 45.02 & 41.86 & 35.93 \\

    & PRKV ($\epsilon=1e-2$) & \textbf{47.27} & 47.92 & \textbf{48.87} & \textbf{49.01} & 45.79 & 36.55 
    & \textbf{47.27} & 47.16 & 47.96 & \textbf{48.82} & \textbf{46.16} & 36.98 \\
    & PRKV ($\epsilon=1e-3$) & \textbf{47.27} & 47.74 & 48.56 & 48.92 & \textbf{46.26} & 37.59
    & \textbf{47.27} & 47.52 & 47.86 & 48.81 & 45.84 & \textbf{39.09} \\
    & PRKV ($\epsilon=1e-6$) & \textbf{47.27} & 47.70 & 48.56 & 48.95 & 46.12 & \textbf{37.76}
    & \textbf{47.27} & 47.22 & 47.01 & 45.49 & 42.53 & 37.03 \\
    \bottomrule
  \end{tabular}
  }
\end{table*}

\begin{table*}[h]
  \centering
  \small
  \caption{\textbf{Ablation studies of PRKV on Qwen3-8B.} We report the average of sixteen tasks from LongBench benchmarks over five compression ratios. 0\% indicates average results without compression.
  Bold values indicate the best performance within each block.
  }
  \vspace{0.5em}
  \label{tab:ablation_lm_more}
  \setlength{\tabcolsep}{6pt}
  \renewcommand{\arraystretch}{1.15}
  \resizebox{\textwidth}{!}{
  \begin{tabular}{@{} l l *{6}{r} l l *{6}{r} @{}}
    \toprule
    \multirow{2}{*}{\textbf{Component}} &
    \multirow{2}{*}{\textbf{Variant}} &
    \multicolumn{6}{c}{\textbf{LongBench}} & \multirow{2}{*}{\textbf{Component}} &
    \multirow{2}{*}{\textbf{Variant}} &
    \multicolumn{6}{c}{\textbf{LongBench}} \\
    \cmidrule(lr){3-8} \cmidrule(lr){11-16} 
    & & 0\% & 10\% & 25\% & 50\% & 75\% & 90\% &
    & & 0\% & 10\% & 25\% & 50\% & 75\% & 90\% \\
    \midrule
    
    \multirow{3}{*}{$C$}
    & 64 & \textbf{47.27} & 47.74 & \textbf{48.56} & \textbf{48.92} & \textbf{46.26} & \textbf{37.59} &
    \multirow{3}{*}{\makecell{Landmark\\ selection}} & Cosine diversity & \textbf{47.27} & \textbf{47.74} & \textbf{48.56} & \textbf{48.92} & 46.26 & 37.59 \\
    & 128 & \textbf{47.27} & 47.81 & 48.55 & 48.72 & 45.84 & 36.90 &
    & Cosine similarity & \textbf{47.27} & 47.23 & 48.33 & 48.75 & \textbf{46.46} & \textbf{37.89} \\
    & 256 & \textbf{47.27} & \textbf{47.90} & 48.38 & 48.63 & 46.03 & 37.31 &
    & L2 Norm & \textbf{47.27} & 47.23 & 47.83 & 48.57 & 45.74 & 37.06 \\
    \bottomrule
  \end{tabular}
  }

\end{table*}

Value-norm scaling is another important design choice. Using PPR alone (Only PPR) leads to clear degradation, particularly under high compression, because graph propagation estimates relational importance but ignores the magnitude of the cached value vectors. Using only value norms (Only VNorm) is also insufficient: it can be competitive under mild compression, but degrades as the budget tightens because it does not capture whether a token lies on relevant attention paths.
The combined PPR + VNorm formulation (PRKV) is consistently more stable, supporting our use of both propagation-based relevance and value magnitude.
We also observe that the scaling constant $\epsilon$ affects numerical stability. This is especially pronounced for PRKV-GLB, where global co-attention smoothing can produce smaller, more diffuse scores; choosing $\epsilon$ too small can over-amplify near-zero normalized values and hurt high-compression performance.

\paragraph{Hyperparameter Sensitivity}
\label{appendix:hp_robustness}
\begin{wraptable}{r}{0.55\textwidth}
\vspace{-1em}
\centering
\caption{\textbf{Hyperparameter Sensitivity}. We provide performance spread across 25 PRKV hyperparameter variants on LongBench (Qwen3-8B). CV denotes coefficient of variation (\%). Low CV values indicate robustness to hyperparameter choices.}
\label{tab:hp_spread}
\small
\resizebox{\linewidth}{!}{
\begin{tabular}{l ccc ccc}
\toprule
 & \multicolumn{3}{c}{\textbf{PRKV}} & \multicolumn{3}{c}{\textbf{PRKV-GLB}} \\
Ratio & Mean & Std & CV(\%) & Mean & Std & CV(\%) \\
\midrule
10\% &  47.8 & 0.16 & 0.3 &  47.3 & 0.12 & 0.3 \\
25\% &  48.3 & 0.35 & 0.7 &  47.4 & 0.51 & 1.1 \\
50\% &  48.9 & 0.11 & 0.2 &  47.1 & 1.52 & 3.2 \\
75\% &  46.0 & 0.26 & 0.6 &  44.0 & 1.59 & 3.6 \\
90\% &  37.4 & 0.52 & 1.4 &  37.5 & 1.33 & 3.6 \\
\bottomrule
\end{tabular}
}
\vspace{-1em}
\end{wraptable}

Since our method is deterministic given a fixed model and input, varying the random seed does not introduce any performance variance. Instead, we analyze robustness by examining the sensitivity of performance to hyperparameter choices. Specifically, we select the top 25 variants of PRKV ranked by average LongBench score at 50\% compression (Qwen3-8B), each differing in hyperparameters. Table~\ref{tab:hp_spread} reports the performance spread across all variants at each compression ratio. For PRKV, the coefficient of variation (CV) remains below 1.4\% across all compression ratios, indicating that performance is largely stable across a wide range of hyperparameter configurations. PRKV-GLB shows slightly higher variation at aggressive compression ratios (CV up to 3.6\%), yet remains well within a tight range. These low CV values demonstrate that PRKV is not sensitive to specific hyperparameter configurations, and practitioners can expect robust performance without extensive tuning.

\section{Extended Experiments on Large Models}
\label{app:rb-experiments}
\begin{table}[htbp]
\centering\small
\vspace{-1em}
\caption{\textbf{Performance comparison on RULER-4K with a larger dense model and an MoE model.} We report the categorical average score of thirteen tasks from RULER benchmarks at 50 \% compression ratio. Full KV refers to an uncompressed KV cache. Higher scores are better.}
\vspace{.5em}
\label{tab:rb-large-models}
\begin{tabular}{llrrrrr}
\toprule
\textbf{Model} & \textbf{Method} & \textbf{Retrieval} & \textbf{Tracing} & \textbf{Aggregation} & \textbf{QA} & \textbf{AVG.} \\
\midrule
\multirow{4}{*}{Qwen3-32B} & Full KV & 100.0 & 100.0 & 96.3 & 75.2 & 95.6 \\
 & SnapKV & 49.4 & 93.5 & 96.4 & 68.9 & 63.0 \\
 & PRKV & 83.8 & 100.0 & 93.6 & 71.1 & 84.6 \\
 & PRKV-GLB & 83.9 & 100.0 & 94.3 & 71.9 & 84.9 \\
\midrule
\multirow{4}{*}{Qwen3-30B-A3B (MoE)} & Full KV & 100.0 & 100.0 & 96.0 & 70.2 & 94.8 \\
 & SnapKV & 27.6 & 27.3 & 68.0 & 54.9 & 38.0 \\
 & PRKV & 84.4 & 100.0 & 94.9 & 64.9 & 84.2 \\
 & PRKV-GLB & 84.7 & 100.0 & 95.0 & 64.2 & 84.3 \\
\bottomrule
\end{tabular}
\end{table}

We extend the evaluation to larger dense (Qwen3-32B) and mixture-of-experts (MoE) models (Qwen3-30B-A3B) in Table~\ref{tab:rb-large-models}.
We evaluate on RULER-4K at 50\% compression. All compressed methods use the same retained-token budget. We group the 13 tasks into retrieval (eight needle-in-a-haystack tasks), multi-hop tracing (variable tracking), aggregation (CWE and FWE), and question answering (two tasks). Each category is an unweighted task average; the overall average is over all 13 tasks.
On Qwen3-32B, PRKV and PRKV-GLB achieve 84.6 and 84.9, compared with 63.0 for SnapKV and 95.6 for Full KV. On Qwen3-30B-A3B, the corresponding scores are 84.2 and 84.3, compared with 38.0 and 94.8. These results extend the evidence beyond the smaller dense backbones, while the remaining gaps to Full KV show that compression is not lossless in this setting.

\section{Extended Memory and Time Analyses} \label{appendix:memory}
\begin{wrapfigure}[14]{r}{0.35\textwidth}
    \centering
    \vspace{-2.5em}
    \includegraphics[width=\linewidth]{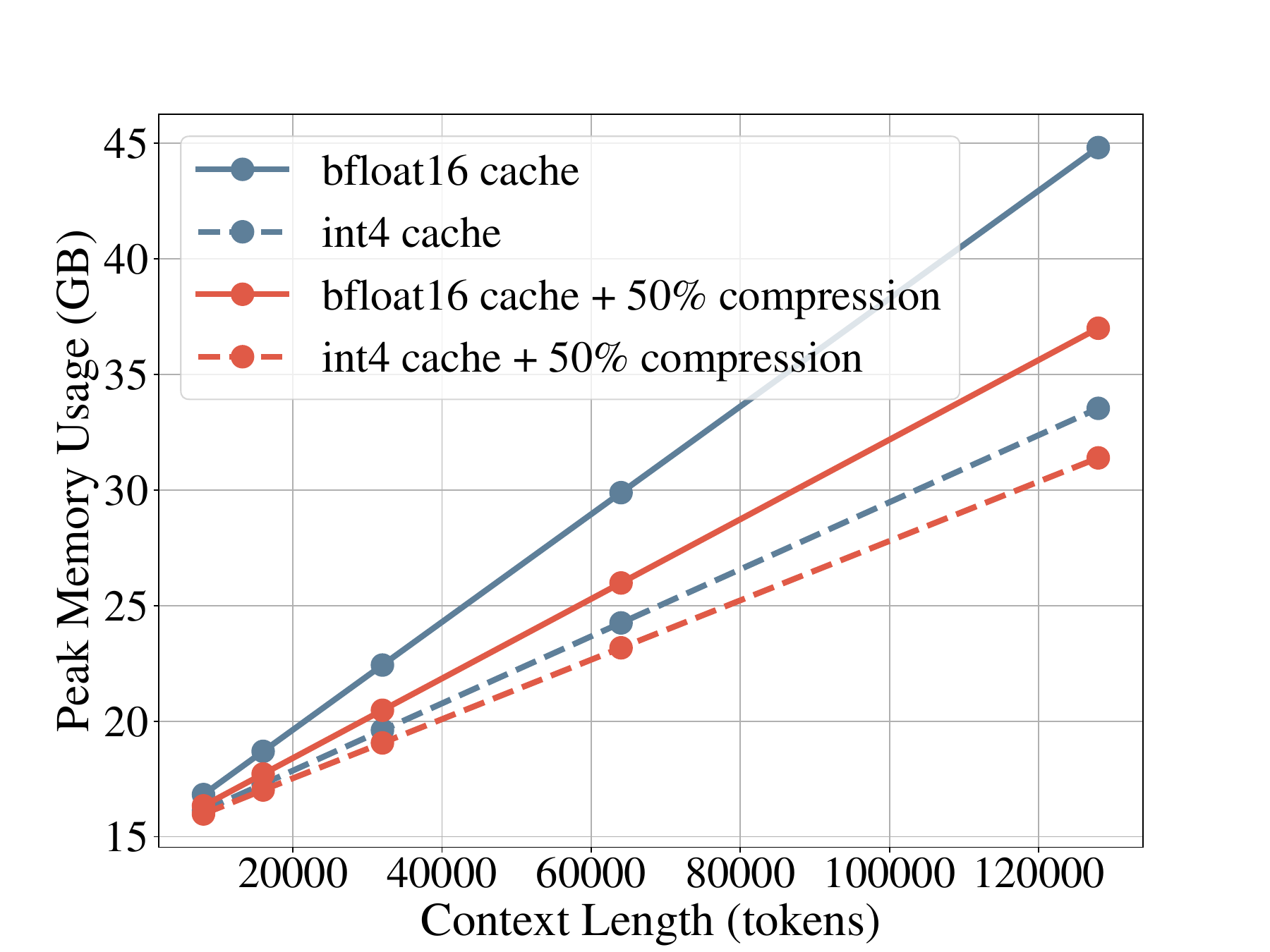}
    \vspace{-1em}
    \caption{\textbf{Effect of cache quantization on peak memory usage.} All measurements are conducted on {Llama-3.1-8B}.}
    \label{fig:quantization}
\end{wrapfigure}
\subsection{Quantization}
\label{app:rb-quantization}
We analyze the effect of cache quantization on memory usage. Since our method operates independently of the numerical precision of the KV cache, it is orthogonal to cache quantization and can be seamlessly combined with the quantized KV cache implementations provided by the \texttt{transformers} library. Applying our method to quantized caches results in substantially lower memory consumption compared to using quantization alone, as illustrated in Figure~\ref{fig:quantization}. We evaluate peak memory usage across varying context lengths for both \texttt{bfloat16} and \texttt{int4} KV caches, under settings with and without 50\% compression. The results indicate that cache quantization and compression act in a complementary manner, leading to greater memory savings, particularly for longer contexts.

\subsection{Memory and Latency Behavior under Compression}
\begin{figure*}[h]
    \centering
    \subfigure[Peak memory usage]{
    \includegraphics[width=0.3\linewidth]{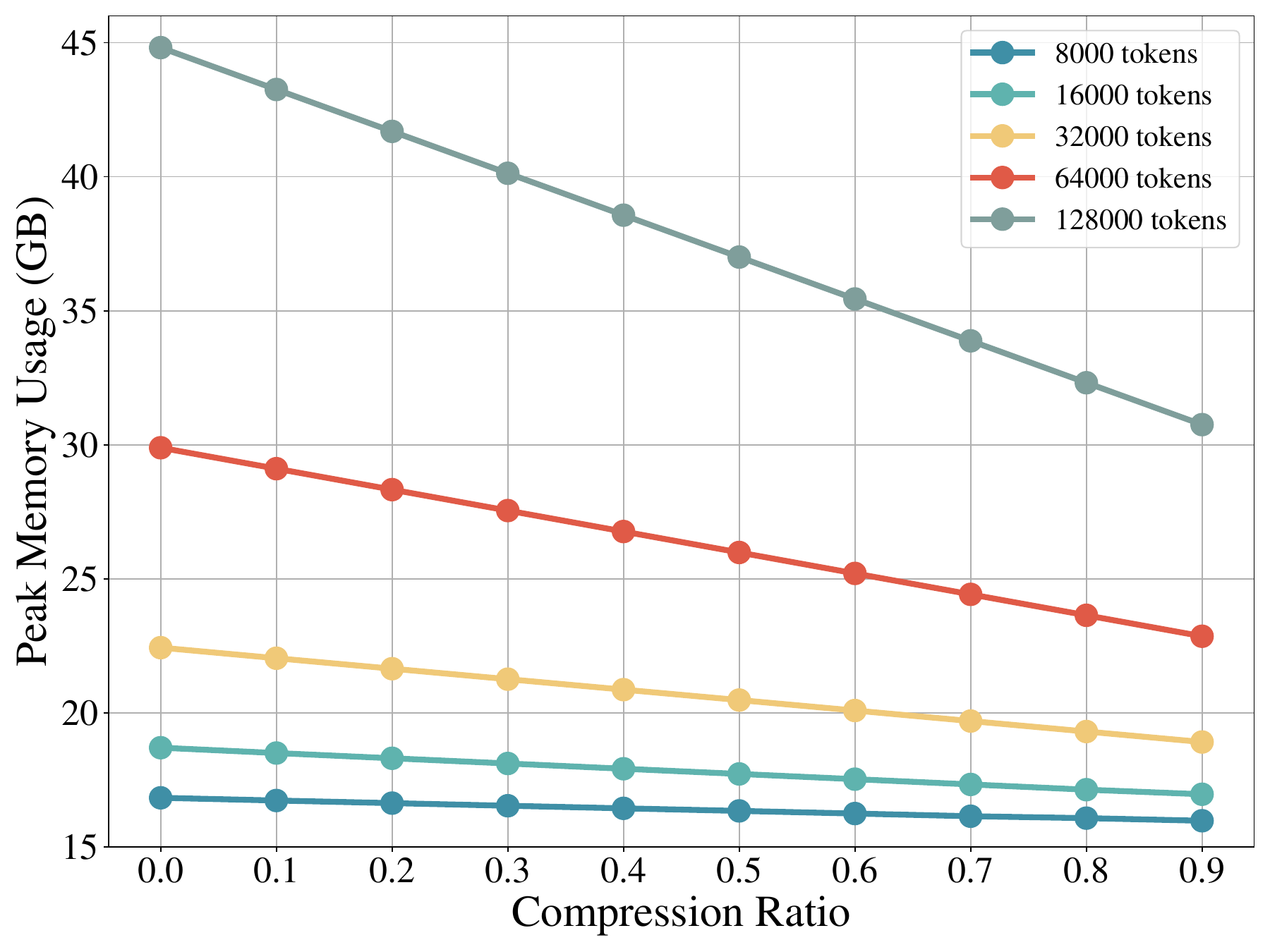}
    }
    \subfigure[Cache size]{
    \includegraphics[width=0.3\linewidth]{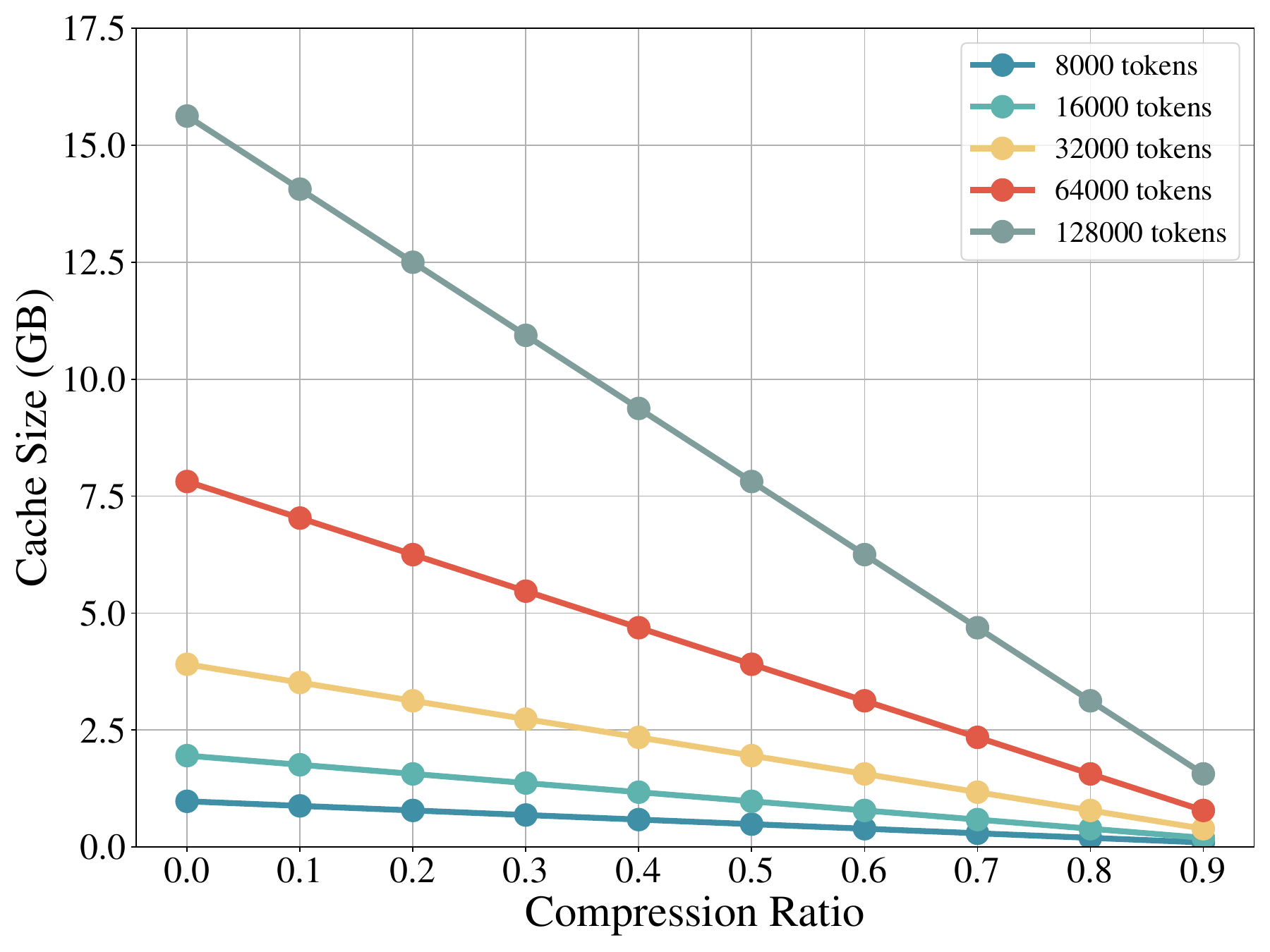}
    }
    \subfigure[Generation time]{
    \includegraphics[width=0.3\linewidth]{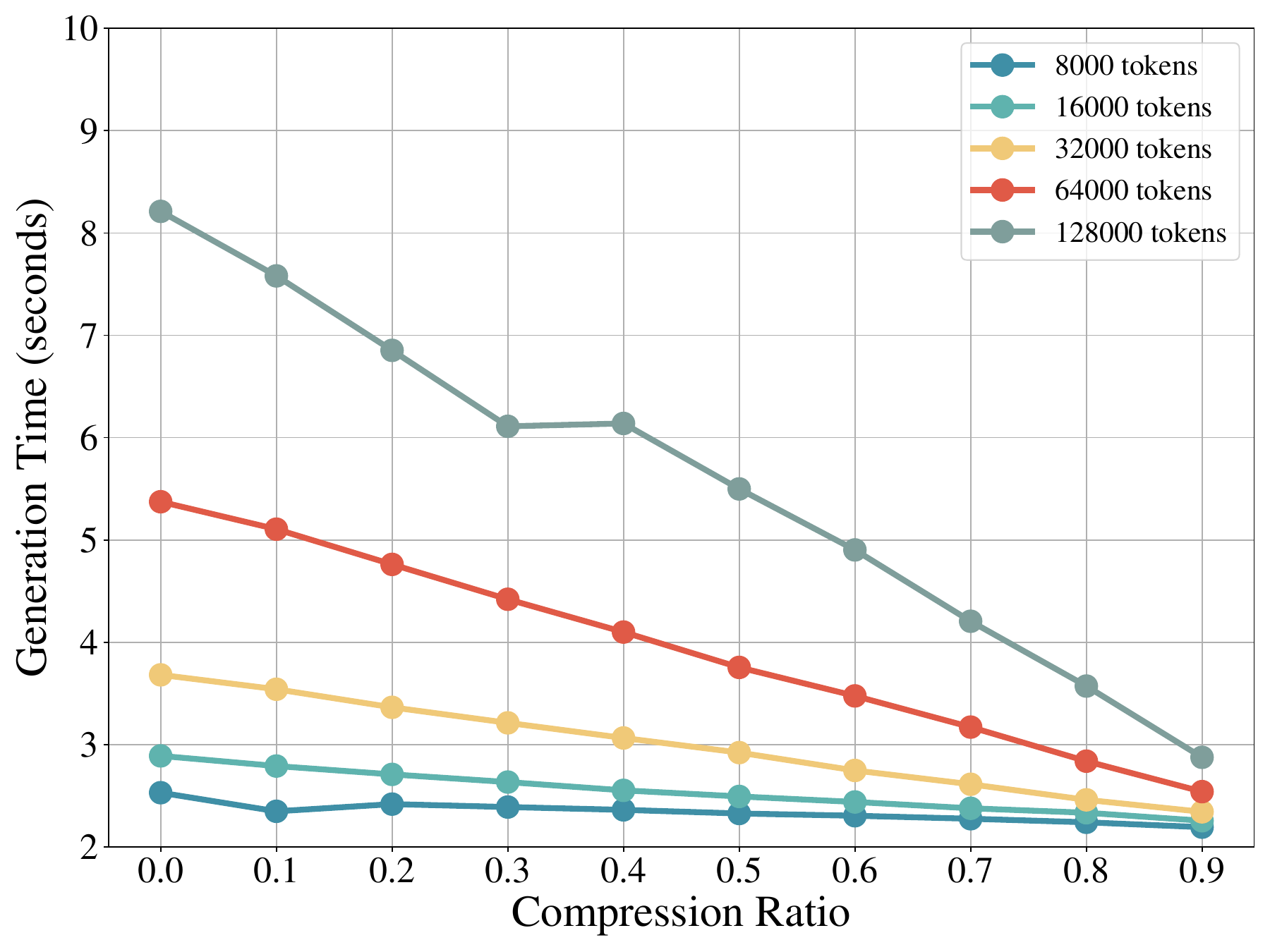}
    }
    \caption{\textbf{Memory consumption and generation latency under varying compression ratios.}
    We report (a) peak memory usage, (b) KV cache size, and (c) generation time as functions of the compression ratio. Results are shown for sequence lengths ranging from 8K to 128K tokens. All measurements are conducted using the \texttt{bfloat16} data type on {Llama-3.1-8B}.}
    \label{fig:memory_bf16}
\end{figure*}

Figure~\ref{fig:memory_bf16} evaluates the memory footprint and generation latency as functions of the compression ratio. We report peak GPU memory usage, KV cache size, and generation time for sequence lengths ranging from 8K to 128K tokens. Across all sequence lengths, increasing the compression ratio consistently reduces both the KV cache size and peak memory usage. This trend closely follows the intended design of the compression mechanism, indicating that retained tokens scale proportionally with the specified budget. Notably, the reduction is more pronounced for longer contexts, where KV caching dominates the memory footprint. In addition to memory savings, generation latency also decreases with higher compression ratios. This reflects that compression yields practical speedups rather than merely lowering peak memory usage.

\paragraph{Scoring overhead}
\begin{table}[h]
\centering
\vspace{-1.5em}
\caption{
\textbf{Scoring overhead of PRKV variants.} Measured at 64K context with 50\% compression on Llama-3.1-8B.
Score time measures only importance-score computation and is reported together with its fraction of total end-to-end runtime.
}
\label{tab:latency_breakdown}
\resizebox{0.95\textwidth}{!}{%
\begin{tabular}{cccccc}
\toprule
 \textbf{Method} &
   \textbf{Total Time (sec)} &
   \textbf{Prefill Time (sec)} &
   \textbf{Score Time (sec)} &
   \shortstack{\textbf{Score Time /}\\ \textbf{Prefill Time (\%)}} &
   \shortstack{\textbf{Score Time  /} \\ \textbf{Total Time (\%)}} \\
\midrule
PRKV &
  14.3757 &
  10.4953 &
  0.1379 &
  1.3138 &
  0.9591 \\
PRKV-GLB &
  14.7180 &
  10.7578 &
  0.4602 &
  4.2782 &
  3.1271 \\
\bottomrule
\end{tabular}%
}
\end{table}
To separate scoring overhead from end-to-end latency, we additionally measure the time spent only on importance-score computation at 64K context with 50\% compression in Table~\ref{tab:latency_breakdown}. PRKV spends 0.1379$s$ on scoring, corresponding to 0.96\% of total generation time, while PRKV-GLB spends 0.4602$s$, corresponding to 3.13\% of total time. This confirms that the global co-attention variant incurs higher scoring cost than PRKV, but its overhead remains a small fraction of the overall runtime.

\subsection{Latency and efficiency analyses}
\label{app:rb-latency}

We measure all compared policies with the same single-GPU, bfloat16, FlashAttention-2 configuration within each benchmark. Generation is greedy and EOS termination is disabled so that methods generate the same number of tokens. Time to first token (TTFT) includes context prefill, scoring and compression, question processing, and production of the first token. Decode throughput counts subsequent generated tokens per second and includes any periodic scoring and eviction. Peak memory is allocated GPU memory over the measured prefill and generation pass, including model weights and temporary workspace, reported in GB ($2^{30}$ bytes).

\paragraph{Prefill eviction}
We use Llama-3.1-8B-Instruct and retain 50\% of the context KV cache. LongBench measurements cover 192 examples across 16 English tasks, with a mean context length of approximately 10.1K tokens. RULER measurements use 15 examples per task across all 13 tasks (195 examples per context-length setting); the 4K and 16K configurations have mean tokenized context lengths of approximately 3.6K and 15.6K, respectively.

\begin{table}[htbp]
\centering\small
\caption{\textbf{Prefill eviction efficiency comparison.} We report  TTFT in seconds and decode throughput in tokens/s on each benchmark on Llama-3.1-8B-Instruct at 50\% compression ratio. Full KV refers to an uncompressed KV cache.}
\label{tab:rb-prefill-latency}
\vspace{.5em}
\begin{tabular}{lrrrrrr}
\toprule
\multirow{2}{*}{\textbf{Method}}& \multicolumn{2}{c}{\textbf{LongBench}} & \multicolumn{2}{c}{\textbf{RULER-4K}} & \multicolumn{2}{c}{\textbf{RULER-16K}} \\
\cmidrule(lr){2-3}\cmidrule(lr){4-5}\cmidrule(lr){6-7}
 & TTFT & tok/s & TTFT & tok/s & TTFT & tok/s \\

\midrule
Full KV & 0.879 & 48.09 & 0.259 & 49.78 & 1.327 & 46.49 \\
SnapKV & 0.912 & 49.73 & 0.272 & 50.39 & 1.366 & 49.21 \\
PyramidKV & 0.914 & 49.65 & 0.273 & 50.41 & 1.366 & 48.93 \\
TOVA & 0.899 & 49.74 & 0.268 & 50.35 & 1.341 & 49.08 \\
KeyDiff & 0.895 & 49.80 & 0.264 & 50.36 & 1.337 & 49.12 \\
EA & 0.980 & 49.76 & 0.300 & 50.36 & 1.463 & 49.08 \\
PRKV & 0.915 & 49.82 & 0.287 & 50.37 & 1.359 & 49.12 \\
PRKV-GLB & 0.944 & 49.79 & 0.283 & 50.38 & 1.410 & 49.14 \\
\bottomrule
\end{tabular}
\end{table}

At RULER-16K, PRKV and PRKV-GLB add approximately 2.4\% and 6.3\% TTFT relative to Full KV, while reaching approximately 49.1 tokens/s versus 46.49 for Full KV (Table~\ref{tab:rb-prefill-latency}). Their throughput is close to the other eviction policies under the matched cache budget. 

\paragraph{Decoding eviction}
We measure decoding eviction efficiency in Table~\ref{tab:rb-decoding-latency}. We use DeepSeek-R1-Distill-Qwen-7B with a target cache size of 2,048 tokens and compression every 128 decoding steps. MATH-500 timing uses three examples with 4,096 generated tokens each while AIME25 timing uses one example with 32,768 generated tokens.

\begin{table}[htbp]
\centering\small
\caption{\textbf{Decoding eviction efficiency comparison.} We report peak memory allocation (GB) and throughput (token/s) on each benchmark on DeepSeek-R1-Distill-Qwen-7B at 2,048-token target cache and an eviction interval of 128 steps. Full KV refers to an uncompressed KV cache.}
\vspace{.5em}
\label{tab:rb-decoding-latency}
\begin{tabular}{lrrrr}
\toprule
\multirow{2}{*}{\textbf{Method}} & \multicolumn{2}{c}{\textbf{MATH-500}} & \multicolumn{2}{c}{\textbf{AIME25}}  \\
\cmidrule(lr){2-3}\cmidrule(lr){4-5}
 & tok/s & GB & tok/s & GB \\
\midrule
Full KV & 46.20 & 14.46 & 52.19 & 16.04 \\
KNorm & 51.05 & 14.39 & 50.23 & 14.40 \\
StreamingLLM & 51.04 & 14.39 & 50.29 & 14.40 \\
KeyDiff & 51.04 & 14.40 & 50.30 & 14.41 \\
TOVA & 50.94 & 14.40 & 50.22 & 14.42 \\
EA & 50.80 & 14.44 & 49.91 & 14.45 \\
PRKV & 50.33 & 14.39 & 49.74 & 14.41 \\
PRKV-GLB & 50.35 & 14.49 & 49.95 & 14.50 \\
\bottomrule
\end{tabular}
\vspace{-.5em}
\end{table}

On MATH-500, PRKV increases throughput from 46.20 to 50.33 tokens/s. On AIME25, it reduces peak allocated memory from 16.04 to 14.41 GB, while throughput decreases from 52.19 to 49.74 tokens/s. Thus the observed benefit is a memory--accuracy trade-off with throughput comparable to other compressed policies, rather than a universal throughput improvement over Full KV. PRKV-GLB uses approximately 0.10 GB more peak memory than PRKV in these online measurements.
\subsection{Tensor-Parallel Efficiency}
\label{app:rb-tp}

We examine whether PRKV's memory savings and latency persist under tensor parallelism (TP). We evaluate Qwen3-32B in bfloat16 using HuggingFace native TP (\texttt{tp\_plan="auto"}) with TP degrees of 1, 2, and 4, disabling NCCL peer-to-peer transport (\texttt{NCCL\_P2P\_DISABLE=1}). At 32,000 context tokens, each result is averaged over two inputs with 32 generated tokens, and all compressed methods retain 16,000 context KV entries. TTFT includes scoring and compression, and memory denotes the rank-local peak allocation per GPU.

\begin{table}[htbp]
\centering\small
\caption{\textbf{TP scaling on Qwen3-32B at 32,000 context tokens.} Compressed methods retain 16,000 context KV entries. Mem denotes peak allocated memory per GPU. All policies use the same measurement configuration within each TP degree.}
\vspace{.5em}
\label{tab:rb-tp}
\begin{tabular}{lrrrrrr}
\toprule
 & \multicolumn{2}{c}{\textbf{TP=1}} & \multicolumn{2}{c}{\textbf{TP=2}} & \multicolumn{2}{c}{\textbf{TP=4}} \\
\cmidrule(lr){2-3} \cmidrule(lr){4-5} \cmidrule(lr){6-7}
Method & TTFT (s) & Mem (GB) & TTFT (s) & Mem (GB) & TTFT (s) & Mem (GB) \\
\midrule
Full KV  & 14.66 & 74.66 & 9.02 & 39.41 & 6.98 & 21.78 \\
KNorm    & 14.51 & 70.76 & 8.74 & 37.46 & 6.48 & 20.80 \\
SnapKV   & 14.85 & 70.76 & 8.77 & 37.46 & 6.42 & 20.80 \\
PRKV     & 14.65 & 70.76 & 8.71 & 37.46 & 6.40 & 20.80 \\
PRKV-GLB & 15.19 & 70.75 & 8.87 & 37.46 & 6.46 & 20.80 \\
\bottomrule
\end{tabular}
\end{table}

As shown in Table~\ref{tab:rb-tp}, KV compression reduces per-GPU peak memory at every TP degree, by approximately 3.90, 1.95, and 0.98\,GB at TP=1, 2, and 4, respectively. This roughly inverse scaling is consistent with the removable KV state being distributed across more ranks. Both PRKV variants match the peak memory of KNorm and SnapKV, indicating that importance estimation adds no visible memory footprint. PRKV's TTFT decreases from 14.65\,s at TP=1 to 6.40\,s at TP=4 and remains close to KNorm and SnapKV at each degree, while PRKV-GLB incurs a modest additional latency at TP=1 that narrows at higher degrees. We do not interpret the small differences between methods as a systematic speed advantage.

\begin{table}[htbp]
\centering\small
\caption{\textbf{Peak allocated memory per GPU (GB) at TP=4 on Qwen3-32B.} Both PRKV variants retain 50\% of the context KV entries. Saving is Full KV minus PRKV, calculated before rounding.}
\vspace{.5em}
\label{tab:rb-tp-context}
\begin{tabular}{lrrrr}
\toprule
\textbf{Context tokens} & \textbf{Full KV} & \textbf{PRKV} & \textbf{PRKV-GLB} & Saving \\
\midrule
32,000  & 21.78 & 20.80 & 20.80 & 0.98 \\
64,000  & 26.11 & 24.16 & 24.16 & 1.95 \\
128,000 & 34.78 & 30.87 & 30.87 & 3.91 \\
\bottomrule
\end{tabular}
\end{table}
To examine context dependence, we fix TP=4 and extend to 64,000 and 128,000 context tokens with 50\% KV retention, using one input and eight generated tokens; these runs characterize memory rather than latency. As shown in Table~\ref{tab:rb-tp-context}, the per-GPU saving grows roughly in proportion to context length, from 0.98\,GB at 32,000 tokens to 3.91\,GB at 128,000 tokens. Since peak memory also includes non-KV components such as model weights, 50\% KV retention does not imply a 50\% reduction in total peak memory; the benefit is additional memory headroom at a given TP degree, which grows with context length.

\subsection{Theoretical Complexity Analysis}
We report complexities per batch element and per attention head. The total cost scales linearly with batch size and the number of heads. We assume a FlashAttention-style kernel, which does not expose intermediate attention matrices.

\paragraph{Notation}
We use the following notation:
\begin{itemize}
    \item $d$: head dimension
    \item $T$: sequence length
    \item $W$: window size ($W \le T$)
    \item $C$: prefix chunk size
    \item $\mathcal{L}$: set of prefix landmarks, selected over chunks of size $C$
    \item $L = |\mathcal{L}| \le \lceil (T - W)/C \rceil$: number of landmarks
    \item $N = W + L$: number of nodes in the node-attention graph
    \item $n_{\text{hops}}$: number of PageRank iterations
\end{itemize}

\paragraph{PRKV (Node Attention)}

In prefill eviction, PRKV constructs a node graph over the recent window and a small set
of prefix landmarks. For a single head and batch element, the required
computation is:
\begin{enumerate}
    \item \textbf{Landmark selection.}
    We compute a global anchor over prefix keys and select one landmark per
    prefix chunk using cosine dissimilarity, incurring
    \[
        \mathcal{O}(T d)
    \]
    for normalized key computation and similarity scores.
    \item \textbf{Node attention.}
    We form an explicit node-level attention matrix over
    $\mathcal{N} = \mathcal{L} \cup \{\text{recent window}\}$ of size
    $N = W + L$.
    Scaled dot-product attention over this node set costs
    \[
        \mathcal{O}(N^2 d).
    \]
    \item \textbf{PPR propagation.}
    PPR is applied to the resulting $N \times N$ transition matrix. Each iteration 
    costs $\mathcal{O}(N^2)$, yielding
    \[
        \mathcal{O}(n_{\text{hops}} N^2)
    \]
    over $n_{\text{hops}}$ iterations.
\end{enumerate}

The overall prefill complexity of PRKV (per head, per batch) is therefore
\[
\boxed{
\text{Time}_{\mathrm{PRKV}}^{\mathrm{prefill}}
= \mathcal{O}(T d + N^2 (d + n_{\text{hops}})),
\qquad
\text{Memory}_{\mathrm{PRKV}}^{\mathrm{prefill}} = \mathcal{O}(N^2).
}
\]

During decoding eviction, prefix landmarks are selected and simply retained in the cache, while importance propagation is restricted to the most recent window tokens. For a single head and batch element, we require:
\begin{enumerate}
    \item \textbf{Landmark update.}
    Prefix landmarks do not participate as nodes in the decoding transition, but are updated using cosine dissimilarity against a global anchor as new tokens arrive. This incurs
    $\mathcal{O}(T d)$ but is performed infrequently (e.g., every $n_\text{dec}$ steps), so the amortized per-step cost is
    \[
    \mathcal{O}\!\left(\tfrac{T d}{n_\text{dec}}\right).
    \]
    \item \textbf{Window attention.}
    In decoding, the node set reduces to the window, $N = W$, and landmarks do not participate in node attention. PRKV computes attention only among the $W$ most recent tokens, using both queries and keys from the tail window. This window-local attention costs
    \[
        \mathcal{O}(W^2 d).
    \]
    \item \textbf{PPR propagation.}
    PPR on the $W \times W$ transition thus costs
    \[
        \mathcal{O}(n_{\text{hops}} W^2).
    \]
\end{enumerate}

This yields the decoding complexity
\[
\boxed{
\mathrm{Time}^{\mathrm{dec}}_{\mathrm{PRKV}}
=
\mathcal{O}\!\left(\tfrac{T d}{n_\text{dec}} + W^2(d + n_{\mathrm{hops}})\right),
\qquad
\mathrm{Memory}^{\mathrm{dec}}_{\mathrm{PRKV}}
=
\mathcal{O}(W^2)
}
\]

\paragraph{PRKV-GLB (Global Co-Attention)}

PRKV-GLB defines a global transition over all $T$ tokens using key-key
co-attention derived from the last $W$ query positions. For a single head and
batch element, the construction and propagation proceed as follows.
\begin{enumerate}
    \item \textbf{Window attention.}
    Computing a window attention matrix $A$ costs
    \[
        \mathcal{O}(W T d).
    \]
    \item \textbf{PPR propagation.}
    PRKV-GLB uses the implicit low-rank affinity
    $S = A^\top A$ and applies the global transition via
    $r \mapsto r D^{-1} A^\top A$ without materializing a full $T \times T$
    matrix. Each PPR iteration can be implemented as
    \[
        r \;\mapsto\; (r D^{-1}) A^\top \;\mapsto\; ((r D^{-1}) A^\top) A,
    \]
    costing $\mathcal{O}(W T)$ per iteration. Over $n_{\text{hops}}$ iterations,
    this yields
    \[
        \mathcal{O}(n_{\text{hops}} W T).
    \]
\end{enumerate}

Thus, the per-head, per-batch complexity of PRKV-GLB is
\[
\boxed{
\text{Time}_{\mathrm{GLB}}
= \mathcal{O}(W T (d+n_{\text{hops}})),
\qquad
\text{Memory}_{\mathrm{GLB}} = \mathcal{O}(W T).
}
\]

\paragraph{Comparison}
Under FlashAttention, PRKV incurs an $N^2$ node-attention cost during prefill and a $W^2$ cost during decoding, while PRKV-GLB incurs a $W T$ cost in both modes. Consequently, PRKV is more favorable when $N$ remains small (few landmarks and moderate
windows).

\paragraph{Remark (Eager Attention)} If the attention implementation returns attention weights (e.g., eager attention in HuggingFace \texttt{transformers}), the $W \times T$ window attention of the prefill pass could be reused instead of recomputed, leaving only the $O(n_{\mathrm{hops}}WT)$ co-attention propagation cost for PRKV-GLB. During decoding, each step exposes only the attention row of the current token, so reuse would additionally require buffering the rows of the last $W$ steps; with such a buffer, the same saving applies to PRKV-GLB, and PRKV reduces to $O(Td/n_{\mathrm{dec}} + n_{\mathrm{hops}}W^2)$. FlashAttention and PyTorch SDPA return only the attention output, and our implementation always recomputes the window attention, as assumed above.

\section{Full Experiment Results} \label{appendix:full_exp}
In this section, we provide full result for LongBench and RULER on 4K and 16K sequence length by model types (from Table~\ref{tab:ruler_full_4k_qwen} to Table~\ref{tab:longbench-big_gemma}).

\begin{table*}[h]
\small
\centering
\caption{RULER 4K Full Result on Qwen3-8B}
\label{tab:ruler_full_4k_qwen}
\vspace{0.5em}
\resizebox{\textwidth}{!}{
\begin{tabular}{c|l|ccccccccccccc|c}
\toprule
\textbf{Comp.Ratio} & \textbf{Method} & \textbf{S-NIAH-1} & \textbf{S-NIAH-2} & \textbf{S-NIAH-3} & \textbf{MK-NIAH-1} & \textbf{MK-NIAH-2} & \textbf{MK-NIAH-3} & \textbf{MV-NIAH} & \textbf{MQ-NIAH} & \textbf{VT} & \textbf{CWE} & \textbf{FWE} & \textbf{QA-1} & \textbf{QA-2} & \textbf{AVG} \\ \midrule
\multicolumn{2}{c|}{\textbf{Full KV}} & 100.00 & 100.00 & 100.00 & 100.00 & 100.00 & 100.00 & 100.00 & 99.90 & 100.00 & 98.96 & 95.33 & 81.80 & 63.80 & 95.37 \\ \midrule
 \multirow{7}{*}{\textbf{10\%}}  
 & SnapKV & 93.80 & 100.00 & 99.60 & 98.40 & 95.20 & 80.00 & 99.00 & 99.45 & 98.56 & 98.74 & 95.80 & 80.80 & 61.60 & 92.38 \\ 
 & PyramidKV & 92.40 & 100.00 & 98.40 & 96.60 & 93.80 & 77.20 & 98.10 & 98.20 & 97.84 & 98.80 & 96.00 & 81.00 & 61.80 & 91.55 \\ 
 & EA & 100.00 & 100.00 & 100.00 & 99.80 & 100.00 & 99.80 & 99.95 & 99.90 & 100.00 & 99.04 & 95.33 & 80.60 & 63.20 & 95.20 \\
 & KeyDiff & 100.00 & 100.00 & 99.80 & 98.80 & 99.20 & 93.20 & 100.00 & 100.00 & 100.00 & 98.10 & 93.60 & 80.80 & 55.40 & 93.76 \\
 & TOVA & 100.00 & 100.00 & 100.00 & 100.00 & 69.80 & 51.60 & 100.00 & 99.90 & 100.00 & 98.84 & 94.67 & 81.60 & 62.60 & 89.15 \\
 & PRKV & 100.00 & 100.00 & 100.00 & 99.80 & 100.00 & 99.80 & 99.95 & 99.95 & 100.00 & 99.02 & 95.33 & 81.40 & 63.00 & 95.25 \\
 & PRKV-GLB & 100.00 & 100.00 & 100.00 & 99.80 & 100.00 & 99.80 & 99.95 & 99.95 & 100.00 & 98.98 & 95.27 & 81.20 & 63.20& 95.24 \\

\midrule

\multirow{7}{*}{\textbf{25 \%}}
 & SnapKV & 88.20 & 100.00 & 55.80 & 78.20 & 73.20 & 52.00 & 82.95 & 83.45 & 95.56 & 98.88 & 96.27 & 80.00 & 58.60 & 80.24 \\
 & PyramidKV & 84.40 & 97.60 & 27.20 & 60.80 & 71.40 & 47.80 & 69.65 & 69.10 & 91.20 & 98.94 & 96.27 & 78.80 & 58.60 & 73.21 \\
 & EA & 100.00 & 99.80 & 100.00 & 99.20 & 100.00 & 83.80 & 99.90 & 99.95 & 100.00 & 99.26 & 95.40 & 80.00 & 62.60 & 93.84 \\ 
 & KeyDiff & 100.00 & 100.00 & 99.80 & 96.20 & 89.60 & 68.20 & 99.90 & 99.85 & 100.00 & 97.06 & 91.47 & 72.80 & 46.60 & 89.34 \\
 & TOVA & 100.00 & 100.00 & 100.00 & 100.00 & 30.00 & 12.00 & 100.00 & 99.90 & 100.00 & 96.34 & 93.13 & 80.40 & 60.00 & 82.44 \\
 & PRKV & 100.00 & 100.00 & 100.00 & 99.00 & 100.00 & 83.40 & 99.95 & 99.95 & 100.00 & 99.02 & 95.20 & 79.40 & 61.80 & 93.67 \\
 & PRKV-GLB & 100.00 & 100.00 & 100.00 & 99.40 & 100.00 & 85.00 & 99.95 & 99.95 & 100.00 & 99.14 & 95.47 & 80.40 & 63.00 & 94.02 \\

\midrule

\multirow{7}{*}{\textbf{50\%}} 
 & SnapKV & 74.20 & 73.20 & 9.80 & 36.60 & 30.20 & 15.20 & 36.10 & 39.15 & 82.20 & 98.52 & 95.67 & 72.40 & 54.80 & 55.23 \\
 & PyramidKV & 67.60 & 30.40 & 6.00 & 23.20 & 29.60 & 11.00 & 22.15 & 24.75 & 65.80 & 81.16 & 94.07 & 67.80 & 44.80 & 43.72 \\ 
 & EA & 100.00 & 100.00 & 99.20 & 97.80 & 100.00 & 6.40 & 99.95 & 100.00 & 100.00 & 97.28 & 95.80 & 72.20 & 58.20 & 86.68 \\ 
 & KeyDiff & 100.00 & 99.80 & 99.00 & 95.20 & 53.80 & 19.40 & 99.90 & 99.35 & 100.00 & 90.70 & 85.27 & 52.40 & 28.60 & 78.72 \\
 & TOVA & 100.00 & 100.00 & 97.60 & 99.80 & 6.80 & 0.40 & 99.90 & 99.90 & 100.00 & 84.36 & 90.40 & 75.60 & 54.80 & 77.66 \\
 & PRKV  & 100.00 & 100.00 & 98.80 & 98.40 & 100.00 & 5.00 & 99.90 & 100.00 & 100.00 & 97.28 & 95.93 & 73.40 & 56.00 & 86.52 \\
 & PRKV-GLB & 100.00 & 100.00 & 98.20 & 98.20 & 99.80 & 6.00 & 99.95 & 99.95 & 100.00 & 97.84 & 95.67 & 73.20 & 56.80 & 86.59 \\

\midrule

\multirow{7}{*}{\textbf{75\%}}
 & SnapKV & 37.00 & 14.60 & 5.60 & 24.00 & 10.40 & 2.80 & 19.45 & 23.75 & 41.08 & 93.08 & 91.87 & 65.80 & 41.00 & 36.19 \\ 
 &PyramidKV & 37.20 & 14.80 & 5.60 & 24.00 & 10.40 & 3.00 & 19.45 & 23.70 & 41.08 & 92.98 & 91.93 & 66.00 & 41.20 & 36.26 \\ 
 & EA & 100.00 & 97.40 & 52.00 & 93.80 & 13.20 & 0.00 & 98.10 & 98.55 & 100.00 & 73.32 & 94.47 & 54.40 & 47.80 & 71.00 \\ 
 & KeyDiff & 100.00 & 93.60 & 93.00 & 78.40 & 12.40 & 2.00 & 94.10 & 93.05 & 99.04 & 65.46 & 67.73 & 24.80 & 12.20 & 64.29 \\
 & TOVA & 82.20 & 100.00 & 15.40 & 95.00 & 1.20 & 0.00 & 96.80 & 96.80 & 96.44 & 52.32 & 81.47 & 55.60 & 40.60 & 62.60 \\
 & PRKV & 100.00 & 98.80 & 49.80 & 95.00 & 15.40 & 0.00 & 98.65 & 98.90 & 100.00 & 82.92 & 94.27 & 59.20 & 45.20 & 72.16 \\
 & PRKV-GLB & 100.00 & 98.20 & 53.00 & 95.00 & 18.60 & 0.00 & 98.20 & 98.70 & 100.00 & 79.66 & 94.80 & 58.80 & 46.20 & 72.40 \\

\midrule

\multirow{7}{*}{\textbf{90\%}} 
 & SnapKV & 5.80 & 5.80 & 5.60 & 19.60 & 4.20 & 0.40 & 17.95 & 20.55 & 16.48 & 45.08 & 80.93 & 53.20 & 28.00 & 23.35 \\
 & PyramidKV & 5.80 & 5.80 & 5.60 & 19.60 & 4.20 & 0.40 & 17.95 & 20.55 & 16.56 & 45.36 & 80.87 & 53.80 & 28.60 & 23.47 \\
 & EA & 99.40 & 93.60 & 5.00 & 89.20 & 0.20 & 0.00 & 94.05 & 94.75 & 99.56 & 29.82 & 83.73 & 41.00 & 34.80 & 58.85 \\
 & KeyDiff & 100.00 & 56.20 & 66.20 & 53.20 & 1.00 & 0.00 & 57.40 & 65.95 & 32.28 & 7.04 & 36.80 & 8.60 & 10.40 & 38.08 \\
 & TOVA & 17.80 & 82.60 & 0.00 & 28.00 & 0.20 & 0.00 & 22.15 & 22.10 & 15.12 & 15.60 & 58.93 & 33.20 & 26.40 & 24.78 \\
 & PRKV & 99.00 & 91.20 & 8.60 & 88.00 & 2.60 & 0.00 & 92.30 & 93.30 & 99.36 & 36.02 & 69.93 & 45.60 & 34.20 & 58.47 \\
 & PRKV-GLB & 99.60 & 91.40 & 6.80 & 86.80 & 2.20 & 0.00 & 92.35 & 93.10 & 99.48 & 34.08 & 81.33 & 47.00 & 35.20 & 59.18 \\

\bottomrule
\end{tabular}}

\end{table*}

\begin{table*}[h]
\small
\centering
\caption{RULER 16K Full Result on Qwen3-8B}
\label{tab:ruler_full_16k_qwen}
\vspace{0.5em}
\resizebox{\textwidth}{!}{
\begin{tabular}{c|l|ccccccccccccc|c}
\toprule
\textbf{Comp.Ratio} & \textbf{Method} & \textbf{S-NIAH-1} & \textbf{S-NIAH-2} & \textbf{S-NIAH-3} & \textbf{MK-NIAH-1} & \textbf{MK-NIAH-2} & \textbf{MK-NIAH-3} & \textbf{MV-NIAH} & \textbf{MQ-NIAH} & \textbf{VT} & \textbf{CWE} & \textbf{FWE} & \textbf{QA-1} & \textbf{QA-2} & \textbf{AVG} \\ \midrule
\multicolumn{2}{c|}{\textbf{Full KV}} & 100.00 & 100.00 & 99.80 & 99.60 & 100.00 & 99.60 & 99.60 & 100.00 & 100.00 & 83.06 & 94.00 & 75.40 & 58.20 & 93.02 \\ \midrule
 \multirow{7}{*}{\textbf{10\%}}  
 & SnapKV & 100.00 & 100.00 & 75.20 & 99.60 & 95.00 & 80.40 & 99.60 & 99.85 & 99.96 & 83.46 & 94.27 & 69.80 & 55.60 & 88.67 \\ 
 & PyramidKV & 100.00 & 99.80 & 61.00 & 99.40 & 94.20 & 74.20 & 98.60 & 99.60 & 99.92 & 83.04 & 94.13 & 69.00 & 56.60 & 86.88 \\ 
 & EA & 100.00 & 100.00 & 100.00 & 99.40 & 99.80 & 99.80 & 99.55 & 100.00 & 100.00 & 87.24 & 93.60 & 74.00 & 58.20 & 93.20 \\
 & KeyDiff & 100.00 & 100.00 & 99.80 & 96.60 & 87.20 & 83.20 & 99.35 & 99.75 & 100.00 & 84.44 & 95.67 & 60.20 & 51.00 & 89.02 \\
 & TOVA & 100.00 & 100.00 & 99.80 & 99.60 & 75.40 & 63.20 & 99.60 & 100.00 & 100.00 & 83.78 & 94.67 & 74.00 & 57.80 & 88.30 \\
 & PRKV & 100.00 & 100.00 & 99.80 & 99.40 & 99.80 & 99.80 & 99.50 & 100.00 & 100.00 & 83.92 & 94.20 & 73.60 & 58.00 &92.92 \\
 & PRKV-GLB & 100.00 & 100.00 & 99.80 & 99.60 & 99.80 & 99.80 & 99.55 & 100.00 & 100.00 & 83.36 & 94.27 & 74.40 & 58.00 & 92.97 \\

\midrule

\multirow{7}{*}{\textbf{25 \%}}
 & SnapKV & 99.80 & 99.00 & 38.80 & 96.00 & 76.40 & 54.00 & 95.15 & 96.80 & 99.84 & 83.58 & 94.07 & 63.00 & 50.60 & 80.54 \\ 
 & PyramidKV & 99.20 & 96.80 & 14.20 & 89.80 & 72.40 & 49.40 & 90.45 & 91.20 & 99.72 & 82.18 & 94.07 & 60.40 & 51.00 & 76.22 \\ 
 & EA & 100.00 & 100.00 & 100.00 & 98.60 & 99.60 & 74.40 & 99.45 & 99.55 & 100.00 & 86.78 & 93.07 & 74.00 & 57.20 & 90.97 \\ 
 & KeyDiff & 100.00 & 100.00 & 99.80 & 87.40 & 74.80 & 51.80 & 98.70 & 98.45 & 100.00 & 85.56 & 95.20 & 47.80 & 43.20 & 83.29 \\
 & TOVA & 100.00 & 100.00 & 99.80 & 99.60 & 36.00 & 23.80 & 99.70 & 100.00 & 100.00 & 83.62 & 95.67 & 68.20 & 55.40 & 81.68 \\
 & PRKV & 100.00 & 100.00 & 99.80 & 99.20 & 99.80 & 73.80 & 99.55 & 99.95 & 100.00 & 84.30 & 93.67 & 74.20 & 58.00 & 90.94 \\
 & PRKV-GLB & 100.00 & 100.00 & 99.80 & 99.20 & 99.80 & 76.20 & 99.55 & 99.90 & 100.00 & 83.80 & 93.80 & 75.40 & 58.00 & 91.19 \\

\midrule

\multirow{7}{*}{\textbf{50\%}} 
 & SnapKV & 98.20 & 91.60 & 4.40 & 66.00 & 35.40 & 20.00 & 57.50 & 63.85 & 98.68 & 79.36 & 94.47 & 46.80 & 44.20 & 61.57 \\ 
 & PyramidKV & 96.80 & 63.60 & 2.40 & 27.00 & 31.00 & 12.20 & 22.65 & 26.25 & 93.32 & 41.40 & 93.80 & 43.00 & 38.20 & 45.51 \\ 
 & EA & 100.00 & 99.80 & 97.20 & 96.40 & 98.20 & 5.40 & 99.30 & 99.30 & 100.00 & 82.92 & 92.93 & 66.40 & 52.20 & 83.85 \\
 & KeyDiff & 100.00 & 100.00 & 99.00 & 82.40 & 32.00 & 14.00 & 98.70 & 97.85 & 100.00 & 86.40 & 94.53 & 32.60 & 30.20 & 74.44 \\
 & TOVA & 100.00 & 100.00 & 99.80 & 99.60 & 9.60 & 4.00 & 99.65 & 99.90 & 100.00 & 77.82 & 96.73 & 54.20 & 50.00 & 76.25 \\
 & PRKV & 100.00 & 100.00 & 98.40 & 97.60 & 98.40 & 6.40 & 99.50 & 99.70 & 100.00 & 81.44 & 93.20 & 64.40 & 51.60 & 83.90 \\
 & PRKV-GLB & 100.00 & 100.00 & 97.60 & 97.80 & 98.40 & 6.40 & 99.60 & 99.70 & 100.00 & 82.86 & 93.60 & 66.20 & 52.80 & 84.23 \\
 
\midrule

\multirow{7}{*}{\textbf{75\%}}
 & SnapKV & 92.40 & 38.40 & 2.60 & 18.80 & 8.60 & 3.40 & 13.85 & 17.55 & 94.32 & 61.92 & 94.40 & 34.80 & 36.40 & 39.80 \\ 
 & PyramidKV & 88.60 & 19.20 & 2.40 & 13.00 & 8.60 & 2.60 & 10.85 & 12.65 & 82.04 & 20.24 & 94.93 & 30.80 & 34.40 & 32.33 \\
 & EA & 100.00 & 96.60 & 58.60 & 92.40 & 11.60 & 0.00 & 96.15 & 97.65 & 100.00 & 43.88 & 91.47 & 50.60 & 41.40 & 67.72 \\ 
 & KeyDiff & 100.00 & 97.80 & 95.20 & 70.40 & 3.60 & 0.80 & 96.95 & 94.40 & 100.00 & 82.98 & 89.20 & 19.80 & 18.40 & 66.89 \\
 & TOVA & 98.40 & 100.00 & 63.80 & 99.40 & 1.20 & 0.40 & 99.20 & 99.90 & 100.00 & 56.94 & 97.60 & 36.60 & 39.20 & 68.66 \\
 & PRKV & 100.00 & 98.20 & 53.20 & 95.60 & 7.60 & 0.60 & 97.35 & 98.30 & 100.00 & 62.28 & 91.60 & 48.40 & 41.00 & 68.78 \\
 & PRKV-GLB & 100.00 & 99.00 & 54.20 & 94.40 & 15.40 & 0.80 & 96.85 & 98.30 & 100.00 & 49.94 & 92.07 & 50.40 & 42.20 & 68.74 \\

\midrule

\multirow{7}{*}{\textbf{90\%}} 
 & SnapKV & 73.80 & 6.60 & 2.40 & 12.20 & 2.60 & 1.00 & 9.80 & 10.55 & 66.44 & 22.52 & 92.60 & 27.80 & 28.80 & 27.47 \\ 
 & PyramidKV & 73.80 & 6.60 & 2.40 & 12.20 & 2.60 & 1.00 & 9.80 & 10.55 & 66.52 & 22.46 & 92.67 & 27.80 & 28.80 & 27.48 \\ 
 & EA & 100.00 & 94.00 & 15.80 & 81.20 & 0.00 & 0.00 & 92.10 & 94.25 & 99.80 & 9.20 & 78.40 & 36.00 & 34.20 & 56.53 \\ 
 & KeyDiff & 100.00 & 70.00 & 80.80 & 52.00 & 0.20 & 0.00 & 79.10 & 79.65 & 79.68 & 64.58 & 66.47 & 8.40 & 12.80 & 53.36 \\
 & TOVA & 65.60 & 99.20 & 2.80 & 86.40 & 0.40 & 0.00 & 83.05 & 85.35 & 88.24 & 20.06 & 96.73 & 26.40 & 26.40 & 52.36 \\
 & PRKV & 99.00 & 92.20 & 13.40 & 84.00 & 0.40 & 0.80 & 92.40 & 93.95 & 99.56 & 20.64 & 81.73 & 34.60 & 30.60 & 57.18 \\
 & PRKV-GLB & 100.00 & 94.40 & 10.00 & 84.80 & 0.40 & 0.60 & 92.25 & 94.45 & 99.84 & 16.58 & 88.27 & 34.60 & 34.20 & 57.72 \\

\bottomrule
\end{tabular}}
\end{table*}

\begin{table*}[h]
\small
\centering
\caption{RULER 4K Full Result on Gemma-3-12B}
\label{tab:ruler_full_4k_gemma}
\vspace{0.5em}
\resizebox{\textwidth}{!}{
\begin{tabular}{c|l|ccccccccccccc|c}
\toprule
\textbf{Comp.Ratio} & \textbf{Method} & \textbf{S-NIAH-1} & \textbf{S-NIAH-2} & \textbf{S-NIAH-3} & \textbf{MK-NIAH-1} & \textbf{MK-NIAH-2} & \textbf{MK-NIAH-3} & \textbf{MV-NIAH} & \textbf{MQ-NIAH} & \textbf{VT} & \textbf{CWE} & \textbf{FWE} & \textbf{QA-1} & \textbf{QA-2} & \textbf{AVG} \\ \midrule
\multicolumn{2}{c|}{\textbf{Full KV}} & 100.00 & 100.00 & 100.00 & 99.80 & 98.80 & 99.80 & 99.75 & 99.95 & 99.68 & 95.10 & 97.33 & 87.60 & 60.80 & 95.28 \\ \midrule
 \multirow{7}{*}{\textbf{10\%}}  
 & SnapKV & 99.60 & 98.60 & 6.80 & 84.80 & 97.40 & 92.20 & 85.25 & 89.40 & 99.16 & 95.98 & 97.67 & 87.60 & 59.80 & 84.17 \\
 & PyramidKV & 99.00 & 98.60 & 6.40 & 79.00 & 98.20 & 90.00 & 80.05 & 84.30 & 99.44 & 95.66 & 97.53 & 85.40 & 59.80 & 82.57 \\
 & EA & 100.00 & 100.00 & 99.80 & 99.60 & 98.60 & 99.40 & 99.30 & 99.90 & 99.68 & 94.92 & 97.27 & 87.60 & 60.80 & 95.14 \\ 
 & KeyDiff & 100.00 & 100.00 & 100.00 & 99.40 & 96.80 & 92.60 & 99.85 & 100.00 & 99.64 & 95.38 & 97.07 & 86.60 & 60.00 & 94.41 \\
 & TOVA & 100.00 & 100.00 & 100.00 & 99.80 & 59.00 & 67.40 & 99.70 & 99.95 & 99.60 & 94.88 & 97.33 & 87.40 & 60.40 & 89.65 \\
 & PRKV & 100.00 & 100.00 & 99.80 & 99.60 & 98.60 & 99.40 & 99.40 & 99.95 & 99.68 & 94.96 & 97.27 & 87.40 & 60.60 & 95.13 \\
 & PRKV-GLB & 100.00 & 100.00 & 99.80 & 99.60 & 98.60 & 99.20 & 99.50 & 99.95 & 99.68 & 94.82 & 97.27 & 87.60 & 60.80 & 95.14 \\

\midrule

\multirow{7}{*}{\textbf{25 \%}}
 & SnapKV & 99.40 & 95.00 & 5.60 & 58.80 & 88.40 & 70.80 & 50.25 & 58.65 & 97.68 & 96.30 & 97.93 & 85.20 & 57.20 & 73.94 \\ 
 & PyramidKV & 99.00 & 93.60 & 5.60 & 45.60 & 89.60 & 56.00 & 39.35 & 45.35 & 98.68 & 96.28 & 97.60 & 83.40 & 57.20 & 69.79 \\ 
 & EA & 100.00 & 100.00 & 99.80 & 99.60 & 99.20 & 98.20 & 97.70 & 99.75 & 99.60 & 96.18 & 97.33 & 85.40 & 59.80 & 94.81 \\ 
 & KeyDiff & 100.00 & 100.00 & 99.60 & 98.80 & 93.20 & 69.00 & 99.80 & 99.95 & 99.60 & 95.38 & 96.13 & 75.00 & 53.20 & 90.74 \\
 & TOVA & 100.00 & 100.00 & 100.00 & 99.60 & 10.40 & 9.80 & 99.65 & 99.95 & 99.64 & 94.86 & 96.93 & 84.60 & 59.00 & 81.11 \\
 & PRKV & 100.00 & 100.00 & 99.80 & 99.40 & 99.20 & 98.20 & 97.90 & 99.85 & 99.64 & 95.86 & 97.33 & 85.40 & 59.80 & 94.80 \\
 & PRKV-GLB & 100.00 & 100.00 & 99.80 & 99.40 & 99.00 & 98.40 & 97.45 & 99.95 & 99.68 & 95.98 & 97.40 & 86.40 & 60.40 & 94.91 \\

\midrule

\multirow{7}{*}{\textbf{50\%}} 
 & SnapKV & 98.40 & 81.80 & 5.60 & 31.20 & 46.40 & 27.60 & 24.25 & 28.40 & 90.60 & 96.40 & 98.20 & 78.00 & 55.60 & 58.65 \\ 
 & PyramidKV & 98.40 & 66.20 & 5.60 & 25.20 & 43.60 & 10.80 & 19.15 & 22.15 & 93.32 & 94.02 & 97.40 & 73.60 & 49.80 & 53.79 \\ 
 & EA & 100.00 & 100.00 & 97.60 & 98.40 & 97.40 & 37.60 & 90.80 & 99.20 & 99.48 & 95.82 & 97.07 & 77.20 & 55.80 & 88.18 \\ 
 & KeyDiff & 100.00 & 99.00 & 100.00 & 97.20 & 65.00 & 5.80 & 98.70 & 99.05 & 99.60 & 85.06 & 90.60 & 58.20 & 39.60 & 79.83 \\
 & TOVA & 99.80 & 100.00 & 86.00 & 97.80 & 1.00 & 0.20 & 98.75 & 98.30 & 99.60 & 94.66 & 94.53 & 71.00 & 54.20 & 76.60 \\
 & PRKV & 100.00 & 100.00 & 99.00 & 98.80 & 97.20 & 34.40 & 93.85 & 99.50 & 99.64 & 96.32 & 97.27 & 76.40 & 55.40 & 88.29 \\
 & PRKV-GLB & 100.00 & 100.00 & 98.60 & 99.00 & 97.60 & 36.20 & 92.55 & 99.45 & 99.56 & 96.08 & 97.13 & 77.80 & 57.20 & 88.55 \\

\midrule

\multirow{7}{*}{\textbf{75\%}}
 & SnapKV & 96.40 & 38.00 & 5.60 & 22.00 & 12.00 & 1.20 & 18.35 & 21.05 & 72.72 & 94.68 & 97.33 & 66.20 & 45.00 & 45.43 \\ 
 & PyramidKV & 96.40 & 37.80 & 5.60 & 21.80 & 12.00 & 1.20 & 18.40 & 21.10 & 72.48 & 94.66 & 97.40 & 66.20 & 45.00 & 45.39 \\ 
 & EA & 100.00 & 99.80 & 90.60 & 96.20 & 60.80 & 0.20 & 86.70 & 97.80 & 99.32 & 92.36 & 94.87 & 63.20 & 48.00 & 79.22 \\ 
 & KeyDiff & 99.80 & 81.60 & 91.60 & 82.00 & 13.40 & 0.00 & 87.15 & 88.25 & 97.40 & 34.22 & 78.47 & 30.00 & 21.00 & 61.91 \\
 & TOVA & 96.80 & 98.40 & 11.80 & 65.40 & 0.00 & 0.00 & 57.55 & 59.55 & 93.32 & 90.80 & 85.73 & 51.60 & 45.60 & 58.20 \\
 & PRKV & 100.00 & 100.00 & 92.80 & 95.40 & 59.40 & 0.20 & 91.60 & 97.90 & 99.28 & 89.66 & 94.60 & 65.80 & 45.80 & 79.42 \\
 & PRKV-GLB & 100.00 & 100.00 & 91.40 & 96.00 & 60.20 & 0.20 & 89.40 & 97.75 & 99.40 & 93.42 & 95.40 & 66.80 & 46.20 & 79.71 \\

\midrule

\multirow{7}{*}{\textbf{90\%}} 
 & SnapKV & 61.20 & 6.60 & 4.80 & 17.40 & 4.20 & 0.40 & 16.80 & 19.80 & 46.92 & 83.94 & 91.53 & 51.60 & 30.80 & 33.54 \\ 
 & PyramidKV & 61.20 & 6.60 & 4.80 & 17.60 & 4.20 & 0.40 & 16.95 & 19.85 & 47.16 & 83.98 & 91.60 & 51.60 & 30.80 & 33.60 \\ 
 & EA & 100.00 & 97.00 & 68.60 & 84.00 & 4.20 & 0.00 & 83.15 & 91.55 & 98.76 & 61.94 & 89.27 & 44.20 & 37.00 & 66.13 \\ 
 & KeyDiff & 82.60 & 24.00 & 47.60 & 38.60 & 1.00 & 0.00 & 35.80 & 42.45 & 79.68 & 9.16 & 57.47 & 13.60 & 14.00 & 34.30 \\
 & TOVA & 37.60 & 63.40 & 0.00 & 8.40 & 0.00 & 0.00 & 1.60 & 1.70 & 26.52 & 53.96 & 65.60 & 33.80 & 34.80 & 25.18 \\
 & PRKV & 100.00 & 98.00 & 69.20 & 80.80 & 4.80 & 0.00 & 86.10 & 92.20 & 97.80 & 55.44 & 88.53 & 50.00 & 35.40 & 66.02 \\
 & PRKV-GLB & 100.00 & 98.60 & 65.80 & 82.00 & 6.00 & 0.00 & 85.10 & 92.80 & 99.00 & 71.42 & 89.40 & 50.80 & 36.00 & 67.46 \\

\bottomrule
\end{tabular}}
\end{table*}

\begin{table*}[h]
\small
\centering
\caption{RULER 16K Full Result on Gemma-3-12B}
\label{tab:ruler_full_16k_gemma}
\vspace{0.5em}
\resizebox{\textwidth}{!}{
\begin{tabular}{c|l|ccccccccccccc|c}
\toprule
\textbf{Comp.Ratio} & \textbf{Method} & \textbf{S-NIAH-1} & \textbf{S-NIAH-2} & \textbf{S-NIAH-3} & \textbf{MK-NIAH-1} & \textbf{MK-NIAH-2} & \textbf{MK-NIAH-3} & \textbf{MV-NIAH} & \textbf{MQ-NIAH} & \textbf{VT} & \textbf{CWE} & \textbf{FWE} & \textbf{QA-1} & \textbf{QA-2} & \textbf{AVG} \\ \midrule
\multicolumn{2}{c|}{\textbf{Full KV}} & 100.00 & 100.00 & 100.00 & 89.60 & 56.00 & 60.80 & 95.65 & 99.05 & 96.48 & 89.84 & 98.60 & 76.80 & 55.00 & 85.99 \\ \midrule
 \multirow{7}{*}{\textbf{10\%}}  
 & SnapKV & 100.00 & 88.60 & 5.80 & 77.20 & 54.80 & 55.40 & 78.75 & 87.00 & 93.72 & 89.76 & 99.07 & 72.80 & 53.40 & 73.56 \\ 
 & PyramidKV & 99.80 & 87.00 & 4.00 & 72.40 & 54.80 & 53.80 & 72.65 & 82.85 & 93.96 & 90.20 & 98.87 & 73.20 & 54.00 & 72.12 \\ 
 & EA & 100.00 & 100.00 & 100.00 & 85.80 & 53.80 & 42.60 & 88.65 & 99.20 & 93.72 & 85.40 & 97.87 & 72.80 & 53.00 & 82.53 \\ 
 & KeyDiff & 100.00 & 100.00 & 100.00 & 79.80 & 46.40 & 31.80 & 95.40 & 99.20 & 96.48 & 89.10 & 98.47 & 72.20 & 53.40 & 81.71 \\
 & TOVA & 100.00 & 100.00 & 80.00 & 89.40 & 31.20 & 27.80 & 95.05 & 98.75 & 96.48 & 89.64 & 98.73 & 75.40 & 54.00 & 79.73 \\
 & PRKV & 100.00 & 100.00 & 100.00 & 86.00 & 54.00 & 44.80 & 90.65 & 99.30 & 94.20 & 85.34 & 98.07 & 72.80 & 53.40 & 82.97 \\
 & PRKV-GLB & 100.00 & 100.00 & 100.00 & 85.40 & 53.60 & 42.60 & 89.00 & 99.10 & 94.04 & 85.06 & 97.80 & 72.60 & 53.00 & 82.48 \\
\midrule

\multirow{7}{*}{\textbf{25 \%}}
 & SnapKV & 99.80 & 73.00 & 2.60 & 51.00 & 45.20 & 40.20 & 46.40 & 57.50 & 92.16 & 89.94 & 99.13 & 63.60 & 50.20 & 62.36 \\ 
 & PyramidKV & 99.80 & 61.40 & 2.40 & 38.20 & 44.80 & 33.60 & 37.90 & 42.40 & 93.76 & 89.64 & 98.93 & 61.80 & 49.20 & 57.99 \\ 
 & EA & 100.00 & 100.00 & 99.60 & 83.80 & 55.40 & 35.60 & 85.40 & 98.75 & 93.56 & 87.34 & 98.00 & 71.00 & 52.40 & 81.60 \\
 & KeyDiff & 100.00 & 99.80 & 99.80 & 76.80 & 41.40 & 20.40 & 95.45 & 99.10 & 96.44 & 87.82 & 97.07 & 61.80 & 47.00 & 78.68 \\
 & TOVA & 100.00 & 98.20 & 43.60 & 87.60 & 8.80 & 8.20 & 90.95 & 95.65 & 96.60 & 90.14 & 98.87 & 74.00 & 52.60 & 72.71 \\
 & PRKV & 100.00 & 100.00 & 99.80 & 85.40 & 54.80 & 35.60 & 87.55 & 98.80 & 94.08 & 87.44 & 98.07 & 71.60 & 52.60 & 81.98 \\
 & PRKV-GLB & 100.00 & 100.00 & 99.60 & 85.00 & 55.20 & 36.20 & 86.00 & 98.85 & 93.80 & 86.92 & 98.00 & 71.00 & 52.00 & 81.74 \\

\midrule

\multirow{7}{*}{\textbf{50\%}} 
 & SnapKV & 99.80 & 42.20 & 2.40 & 22.60 & 21.20 & 8.80 & 15.00 & 21.95 & 91.04 & 90.02 & 99.07 & 46.40 & 44.60 & 46.54 \\ 
 & PyramidKV & 99.40 & 14.00 & 2.40 & 12.00 & 18.20 & 5.60 & 10.65 & 11.80 & 92.00 & 85.22 & 98.33 & 37.40 & 38.60 & 40.43 \\
 & EA & 100.00 & 99.80 & 79.80 & 77.80 & 52.40 & 12.40 & 77.20 & 97.25 & 92.80 & 86.16 & 97.87 & 62.00 & 48.00 & 75.65 \\
 & KeyDiff & 100.00 & 97.40 & 99.80 & 76.60 & 24.60 & 10.40 & 94.75 & 98.85 & 95.04 & 84.52 & 93.60 & 33.60 & 33.60 & 72.52 \\
 & TOVA & 100.00 & 89.60 & 5.60 & 72.00 & 1.40 & 0.80 & 71.80 & 81.45 & 96.56 & 90.34 & 98.40 & 58.20 & 47.60 & 62.60 \\
 & PRKV & 100.00 & 100.00 & 80.80 & 79.60 & 53.20 & 12.00 & 80.05 & 97.90 & 92.80 & 87.46 & 98.07 & 61.80 & 47.80 & 76.27 \\
 & PRKV-GLB & 100.00 & 100.00 & 79.00 & 78.20 & 52.60 & 12.20 & 77.65 & 97.50 & 91.84 & 87.20 & 98.00 & 61.00 & 47.40 & 75.58 \\

\midrule

\multirow{7}{*}{\textbf{75\%}}
 & SnapKV & 97.60 & 11.00 & 2.40 & 10.80 & 4.80 & 2.00 & 10.10 & 10.95 & 87.44 & 88.62 & 98.60 & 26.60 & 33.20 & 37.24 \\ 
 & PyramidKV & 95.20 & 6.00 & 2.40 & 11.00 & 5.00 & 1.80 & 9.80 & 10.45 & 87.36 & 77.02 & 97.13 & 24.40 & 30.20 & 35.21 \\
 & EA & 100.00 & 99.80 & 43.80 & 69.60 & 23.00 & 2.60 & 64.90 & 89.55 & 91.52 & 78.94 & 95.53 & 42.00 & 40.00 & 64.71 \\
 & KeyDiff & 100.00 & 65.40 & 95.60 & 59.20 & 6.00 & 0.00 & 88.85 & 91.30 & 93.56 & 42.08 & 87.73 & 11.40 & 19.60 & 58.52 \\
 & TOVA & 100.00 & 54.60 & 2.40 & 34.00 & 0.00 & 0.00 & 27.45 & 36.85 & 94.32 & 89.50 & 98.07 & 36.80 & 37.60 & 47.05 \\
 & PRKV & 100.00 & 99.40 & 50.40 & 71.80 & 22.00 & 3.20 & 71.10 & 91.20 & 91.52 & 80.44 & 95.13 & 39.40 & 40.20 & 65.83 \\
 & PRKV-GLB & 100.00 & 99.60 & 45.00 & 70.00 & 23.60 & 2.40 & 66.35 & 89.60 & 90.76 & 80.62 & 95.40 & 40.40 & 40.20 & 64.92 \\
\midrule

\multirow{7}{*}{\textbf{90\%}} 
 & SnapKV & 86.20 & 2.20 & 2.40 & 10.80 & 1.60 & 1.20 & 9.75 & 10.20 & 75.64 & 74.60 & 97.33 & 17.60 & 27.00 & 32.04 \\ 
 & PyramidKV & 86.20 & 2.20 & 2.40 & 10.80 & 1.60 & 1.20 & 9.75 & 10.25 & 75.56 & 74.68 & 97.27 & 17.80 & 27.20 & 32.07 \\ 
 & EA & 100.00 & 74.40 & 23.40 & 48.80 & 2.40 & 0.00 & 57.85 & 67.30 & 85.40 & 36.08 & 84.67 & 26.60 & 29.20 & 48.93 \\
 & KeyDiff & 100.00 & 16.20 & 46.60 & 33.80 & 0.60 & 0.00 & 52.65 & 58.15 & 83.16 & 11.42 & 62.53 & 5.20 & 12.00 & 37.10 \\
 & TOVA & 86.00 & 10.00 & 1.80 & 8.60 & 0.00 & 0.00 & 4.50 & 6.10 & 85.32 & 77.98 & 93.13 & 22.00 & 30.40 & 32.76 \\
 & PRKV & 100.00 & 82.80 & 25.80 & 52.00 & 3.20 & 0.00 & 62.65 & 73.95 & 87.76 & 36.54 & 83.20 & 25.60 & 30.20 & 51.05 \\
 & PRKV-GLB & 100.00 & 75.20 & 23.60 & 53.20 & 2.80 & 0.00 & 58.85 & 69.30 & 87.80 & 39.80 & 84.60 & 25.40 & 28.40 & 49.92 \\

\bottomrule
 \end{tabular}}

\end{table*}

\begin{table*}[h]
\centering
\small
\caption{LongBench Full Result on Llama-3.1-8B-Instruct}
\label{tab:longbench-big_llama}
\vspace{0.5em}
\setlength{\tabcolsep}{4.5pt}
\renewcommand{\arraystretch}{1.18}
\resizebox{\textwidth}{!}{
\begin{tabular}{@{} l |l|
*{3}{c} | % Single-Doc QA
*{3}{c} | % Multi-Doc QA
*{3}{c} | % Summarization
*{3}{c} | % Few-shot
*{2}{c} | % Synthetic
*{2}{c} | % Synthetic
c        % Avg
@{}}
\toprule
\multirow{2}{*}{\textbf{Ratio}} &
\multirow{2}{*}{\textbf{Method}} &
\multicolumn{3}{c|}{\textbf{Single-Document QA}} &
\multicolumn{3}{c|}{\textbf{Multi-Document QA}} &
\multicolumn{3}{c|}{\textbf{Summarization}} &
\multicolumn{3}{c|}{\textbf{Few-shot Learning}} &
\multicolumn{2}{c|}{\textbf{Synthetic}} &
\multicolumn{2}{c|}{\textbf{Code}} &
\multirow{2}{*}{\textbf{Avg}} \\
& &
\makecell{\texttt{NQA}} & \makecell{\texttt{QA}} & \makecell{\texttt{MQA}} & \makecell{\texttt{HQA}} &
\makecell{\texttt{2WM}} & \makecell{\texttt{MSQ}} & \makecell{\texttt{GR}} &
\makecell{\texttt{QM}} & \makecell{\texttt{MN}} & \makecell{\texttt{TR}} &
\makecell{\texttt{TQA}} & \makecell{\texttt{SS}} & \makecell{\texttt{PC}} &
\makecell{\texttt{PR}} & \makecell{\texttt{LCC}} & \makecell{\texttt{RB}} \\
\midrule

% ===== 100% =====
\textbf{0\%} & Full KV &
30.99 & 47.39 & 56.31 & 59.80 & 51.33 & 33.51 & 35.19 & 25.34 & 26.99 & 29.00 & 84.81 & 38.76 & 10.70 & 100.00 & 51.21 & 44.21 & 45.35 \\
\midrule

% ===== 10%  =====
\multirow{7}{*}{\textbf{10\%}} & SnapKV &
30.46 & 46.98 & 56.15 & 58.44 & 52.03 & 32.28 & 34.70 & 24.85 & 26.99 & 28.50 & 84.98 & 39.98 & 11.15 & 99.50 & 51.39 & 44.84 & 45.20 \\
& PyramidKV &
30.86 & 47.07 & 56.39 & 58.63 & 52.56 & 32.32 & 34.92 & 24.96 & 26.99 & 30.00 & 84.81 & 39.62 & 11.20 & 100.00 & 51.06 & 44.86 & 45.39 \\
& EA & 
31.00 & 48.61 & 55.76 & 59.87 & 51.92 & 33.80 & 34.98 & 24.91 & 27.38 & 30.50 & 86.96 & 38.90 & 13.64 & 100.00 & 51.15 & 45.22 & 45.91 \\ 
& KeyDiff & 30.75 & 47.67 & 55.19 & 58.29 & 51.24 & 33.90 & 34.92 & 24.97 & 26.98 & 31.00 & 85.56 & 39.61 & 10.15 & 99.50 & 50.29 & 43.81 & 45.24 \\
& TOVA & 31.46 & 46.88 & 55.59 & 59.45 & 52.25 & 33.27 & 35.06 & 24.95 & 26.92 & 28.00 & 84.01 & 38.84 & 11.15 & 100.00 & 51.12 & 44.19 & 45.20 \\
& PRKV & 30.61 & 48.49 & 56.44 & 59.42 & 50.93 & 33.00 & 35.12 & 25.01 & 27.17 & 31.50 & 86.71 & 38.74 & 12.54 & 100.00 & 50.85 & 44.93 & 45.72 \\
& PRKV-GLB & 30.60 & 47.86 & 55.41 & 59.29 & 50.63 & 32.47 & 34.93 & 24.86 & 26.95 & 30.50 & 86.88 & 38.99 & 13.54 & 100.00 & 51.44 & 45.26 & 45.60 \\

\midrule

% ===== 25% =====
\multirow{7}{*}{\textbf{25\%}} & SnapKV &
30.62 & 45.80 & 54.91 & 58.06 & 52.02 & 32.06 & 34.03 & 24.60 & 26.44 & 29.50 & 84.83 & 40.18 & 11.20 & 99.50 & 51.62 & 45.06 & 45.03 \\
& PyramidKV & 30.03 & 47.42 & 54.61 & 59.11 & 50.69 & 32.17 & 34.02 & 24.15 & 26.58 & 29.50 & 85.31 & 39.97 & 10.70 & 100.00 & 51.45 & 44.33 & 45.00 \\
& EA & 30.27 & 46.76 & 55.88 & 58.75 & 49.12 & 32.94 & 34.57 & 25.02 & 27.09 & 60.50 & 91.21 & 39.03 & 12.64 & 99.50 & 49.80 & 47.87 & 47.56 \\ 
& KeyDiff & 31.58 & 47.64 & 54.50 & 58.40 & 50.94 & 35.68 & 34.68 & 24.88 & 26.92 & 62.00 & 86.56 & 40.05 & 11.15 & 99.50 & 50.10 & 43.84 & 47.40 \\
& TOVA & 31.38 & 47.46 & 55.26 & 58.61 & 51.66 & 32.04 & 34.14 & 24.45 & 26.52 & 21.50 & 84.59 & 39.65 & 11.20 & 99.50 & 50.97 & 44.52 & 44.59 \\
& PRKV & 30.13 & 47.89 & 56.19 & 58.88 & 50.57 & 32.70 & 34.87 & 24.86 & 26.70 & 57.00 & 90.41 & 39.60 & 13.14 & 100.00 & 50.20 & 46.89 & 47.50 \\
& PRKV-GLB & 30.67 & 47.71 & 55.17 & 58.65 & 50.83 & 33.67 & 34.64 & 24.87 & 26.98 & 40.50 & 89.41 & 39.26 & 12.81 & 100.00 & 51.41 & 46.57 & 46.45 \\

\midrule

% ===== 50% =====
\multirow{7}{*}{\textbf{50\%}} & SnapKV &
29.08 & 41.53 & 45.59 & 56.43 & 49.33 & 29.84 & 32.16 & 23.32 & 25.40 & 35.00 & 85.13 & 40.29 & 11.05 & 99.50 & 50.63 & 44.75 & 43.69 \\
& PyramidKV &
28.47 & 41.28 & 49.15 & 58.05 & 48.13 & 29.96 & 31.61 & 24.23 & 25.70 & 29.50 & 85.83 & 41.31 & 10.00 & 99.50 & 52.06 & 45.20 & 43.75 \\
& EA & 30.06 & 45.32 & 54.11 & 57.59 & 50.28 & 30.38 & 33.89 & 24.43 & 26.83 & 65.50 & 90.98 & 39.50 & 11.04 & 98.50 & 47.13 & 48.27 & 47.11 \\
& KeyDiff & 32.35 & 44.96 & 52.89 & 55.53 & 46.52 & 30.59 & 32.50 & 24.04 & 25.81 & 63.00 & 86.29 & 40.14 & 10.65 & 99.50 & 42.61 & 43.45 & 45.68 \\
& TOVA & 30.08 & 44.27 & 51.02 & 57.56 & 49.06 & 30.93 & 32.70 & 23.25 & 25.54 & 13.00 & 85.43 & 39.64 & 11.65 & 99.50 & 51.75 & 43.79 & 43.07 \\
& PRKV & 30.76 & 45.82 & 53.22 & 56.59 & 49.78 & 31.30 & 33.26 & 24.51 & 26.29 & 65.00 & 92.26 & 40.14 & 12.70 & 99.50 & 51.27 & 47.15 & 47.47 \\
& PRKV-GLB & 31.04 & 46.17 & 52.41 & 58.26 & 50.29 & 30.43 & 33.81 & 24.52 & 26.44 & 49.50 & 91.04 & 41.01 & 13.00 & 98.50 & 52.04 & 47.21 & 46.60 \\

\midrule

% ===== 75% =====
\multirow{7}{*}{\textbf{75\%}} & SnapKV &
28.43 & 31.36 & 35.01 & 55.18 & 42.44 & 28.18 & 29.24 & 22.41 & 23.35 & 37.00 & 84.73 & 40.85 & 9.10 & 90.00 & 51.53 & 44.62 & 40.84 \\
& PyramidKV &
26.92 & 30.08 & 34.79 & 55.13 & 39.83 & 25.19 & 28.23 & 22.20 & 23.41 & 34.00 & 85.58 & 41.73 & 9.55 & 93.50 & 51.29 & 44.15 & 40.35 \\
& EA & 29.57 & 41.50 & 46.13 & 52.51 & 42.04 & 28.27 & 31.06 & 23.68 & 25.92 & 56.50 & 90.99 & 39.88 & 8.54 & 96.00 & 43.01 & 48.84 & 44.03 \\ 
& KeyDiff & 29.57 & 32.81 & 42.16 & 53.29 & 38.22 & 25.01 & 29.36 & 23.60 & 22.95 & 50.50 & 84.78 & 40.13 & 11.05 & 99.50 & 34.44 & 44.96 & 41.40 \\
& TOVA & 27.72 & 32.94 & 39.06 & 55.18 & 43.79 & 27.71 & 29.33 & 22.06 & 23.00 & 4.50 & 84.42 & 40.15 & 10.55 & 98.00 & 50.42 & 44.42 & 39.58 \\
& PRKV & 29.86 & 43.24 & 42.69 & 54.33 & 45.83 & 28.23 & 30.80 & 23.66 & 25.56 & 49.50 & 91.49 & 40.82 & 10.50 & 95.50 & 49.89 & 48.55 & 44.40 \\
& PRKV-GLB & 30.96 & 40.88 & 44.96 & 52.70 & 43.12 & 27.43 & 30.99 & 23.34 & 25.71 & 51.50 & 91.21 & 39.95 & 7.60 & 97.00 & 52.08 & 47.71 & 44.20 \\

\midrule

% ===== 90% =====
\multirow{7}{*}{\textbf{90\%}} & SnapKV &
24.57 & 20.58 & 23.21 & 44.32 & 23.71 & 20.52 & 25.15 & 20.04 & 20.19 & 34.50 & 81.38 & 39.05 & 6.00 & 54.00 & 48.55 & 45.83 & 33.23 \\
& PyramidKV &
22.82 & 20.66 & 23.49 & 45.98 & 24.69 & 19.20 & 24.98 & 20.62 & 20.14 & 34.50 & 81.88 & 38.19 & 6.00 & 54.50 & 48.33 & 46.48 & 33.28 \\
& EA & 29.04 & 30.55 & 32.68 & 45.52 & 32.42 & 21.92 & 27.46 & 22.44 & 23.70 & 29.00 & 90.52 & 36.21 & 6.00 & 64.50 & 33.60 & 47.06 & 35.79 \\ 
& KeyDiff & 28.03 & 16.77 & 35.38 & 40.83 & 24.56 & 15.49 & 26.15 & 21.19 & 18.80 & 40.00 & 78.88 & 39.18 & 10.06 & 89.00 & 22.55 & 45.07 & 34.50 \\
& TOVA & 24.04 & 14.74 & 26.84 & 46.39 & 30.30 & 20.02 & 25.29 & 20.12 & 19.58 & 2.50 & 80.56 & 40.01 & 9.00 & 90.50 & 48.82 & 44.64 & 33.96 \\
& PRKV & 26.94 & 32.02 & 30.92 & 44.45 & 31.87 & 22.11 & 26.74 & 22.81 & 22.69 & 23.25 & 91.04 & 38.17 & 10.00 & 72.50 & 47.28 & 47.27 & 36.88 \\
& PRKV-GLB & 28.29 & 27.34 & 30.77 & 46.21 & 29.54 & 21.30 & 27.22 & 22.22 & 22.98 & 29.00 & 91.46 & 37.12 & 7.25 & 64.50 & 49.92 & 45.97 & 36.32 \\

\bottomrule
\end{tabular}}

\end{table*}

\begin{table*}[h]
\centering
\small
\caption{LongBench Full Result on Qwen3-8B}
\label{tab:longbench-big_qwen}
\vspace{0.5em}
\setlength{\tabcolsep}{4.5pt}
\renewcommand{\arraystretch}{1.18}
\resizebox{\textwidth}{!}{
\begin{tabular}{@{} l |l|
*{3}{c} | % Single-Doc QA
*{3}{c} | % Multi-Doc QA
*{3}{c} | % Summarization
*{3}{c} | % Few-shot
*{2}{c} | % Synthetic
*{2}{c} | % Synthetic
c        % Avg
@{}}
\toprule
\multirow{2}{*}{\textbf{Ratio}} &
\multirow{2}{*}{\textbf{Method}} &
\multicolumn{3}{c|}{\textbf{Single-Document QA}} &
\multicolumn{3}{c|}{\textbf{Multi-Document QA}} &
\multicolumn{3}{c|}{\textbf{Summarization}} &
\multicolumn{3}{c|}{\textbf{Few-shot Learning}} &
\multicolumn{2}{c|}{\textbf{Synthetic}} &
\multicolumn{2}{c|}{\textbf{Code}} &
\multirow{2}{*}{\textbf{Avg}} \\
& &
\makecell{\texttt{NQA}} & \makecell{\texttt{QA}} & \makecell{\texttt{MQA}} & \makecell{\texttt{HQA}} &
\makecell{\texttt{2WM}} & \makecell{\texttt{MSQ}} & \makecell{\texttt{GR}} &
\makecell{\texttt{QM}} & \makecell{\texttt{MN}} & \makecell{\texttt{TR}} &
\makecell{\texttt{TQA}} & \makecell{\texttt{SS}} & \makecell{\texttt{PC}} &
\makecell{\texttt{PR}} & \makecell{\texttt{LCC}} & \makecell{\texttt{RB}} \\
\midrule

% ===== 100% =====
\textbf{0\%} & Full KV
& 28.64 & 43.54 & 55.65 & 62.80 & 48.88 & 35.92 & 33.34 & 24.43 & 24.68 & 41.00 & 90.31 & 39.98 & 10.50 & 92.02 & 65.03 & 59.61 & 47.27 \\
\midrule

% ===== 10% =====
\multirow{7}{*}{\textbf{10\%}}
 & SnapKV & 29.20 & 43.91 & 54.73 & 62.18 & 49.16 & 34.73 & 33.08 & 24.54 & 24.79 & 39.50 & 90.21 & 40.47 & 9.50 & 92.56 & 64.75 & 59.86 & 47.07 \\
 & PyramidKV & 29.14 & 44.13 & 55.22 & 61.92 & 48.18 & 34.81 & 33.10 & 24.68 & 24.59 & 40.50 & 89.81 & 40.59 & 9.50 & 90.93 & 64.63 & 59.74 & 46.97 \\
 & EA & 28.18 & 44.38 & 53.36 & 62.55 & 49.60 & 34.28 & 33.70 & 24.13 & 24.60 & 54.00 & 90.51 & 40.57 & 9.50 & 95.38 & 64.01 & 60.45 & 48.08 \\
 & KeyDiff & 25.81 & 38.37 & 51.53 & 53.98 & 47.03 & 28.16 & 34.14 & 23.76 & 25.17 & 63.00 & 86.74 & 40.29 & 7.69 & 100.00 & 67.71 & 55.82 & 46.83 \\
 & TOVA & 28.71 & 44.51 & 54.04 & 62.31 & 48.48 & 35.74 & 33.46 & 24.38 & 24.60 & 39.50 & 89.71 & 39.92 & 9.00 & 91.85 & 65.09 & 59.83 & 46.95 \\
 & PRKV & 28.79 & 44.06 & 53.82 & 62.52 & 48.26 & 35.86 & 33.66 & 24.45 & 24.89 & 45.00 & 90.46 & 40.59 & 11.00 & 96.29 & 64.87 & 59.30 & 47.74 \\
 & PRKV-GLB & 28.71 & 44.11 & 54.05 & 62.80 & 49.44 & 36.39 & 33.60 & 24.62 & 24.59 & 44.50 & 89.71 & 40.40 & 9.00 & 93.96 & 64.99 & 59.38 & 47.52 \\

\midrule

% ===== 25% =====
\multirow{7}{*}{\textbf{25\%}} 
 & SnapKV & 29.33 & 42.45 & 51.39 & 62.93 & 48.76 & 34.17 & 32.76 & 24.16 & 24.36 & 39.00 & 89.81 & 40.64 & 9.50 & 92.29 & 65.14 & 59.76 & 46.65 \\
 & PyramidKV & 29.18 & 42.40 & 50.60 & 61.21 & 48.19 & 33.58 & 32.80 & 23.82 & 24.19 & 37.00 & 89.81 & 40.58 & 8.50 & 91.33 & 64.98 & 59.38 & 46.10 \\
 & EA & 27.34 & 42.92 & 52.25 & 62.05 & 48.97 & 34.88 & 33.45 & 24.00 & 25.08 & 72.50 & 88.54 & 40.87 & 9.50 & 96.71 & 62.81 & 60.07 & 48.87 \\ 
 & KeyDiff & 25.17 & 35.48 & 48.57 & 48.01 & 43.00 & 22.16 & 33.04 & 23.11 & 24.91 & 63.00 & 84.88 & 38.24 & 6.39 & 91.50 & 59.91 & 54.21 & 43.85 \\
 & TOVA & 28.74 & 42.81 & 54.46 & 62.22 & 48.99 & 36.02 & 33.12 & 23.84 & 24.04 & 38.50 & 89.89 & 40.62 & 9.50 & 92.02 & 65.20 & 59.96 & 46.87 \\
 & PRKV & 28.40 & 45.30 & 51.90 & 62.72 & 47.72 & 35.17 & 33.69 & 24.03 & 24.99 & 63.50 & 89.66 & 40.15 & 9.50 & 96.29 & 64.64 & 59.26 & 48.56 \\
 & PRKV-GLB & 28.76 & 45.54 & 53.18 & 62.74 & 49.13 & 36.84 & 33.61 & 24.25 & 24.74 & 48.00 & 89.41 & 40.08 & 10.00 & 93.79 & 65.61 & 60.11 & 47.86 \\

\midrule

% ===== 50% =====
\multirow{7}{*}{\textbf{50\%}} 
 & SnapKV & 26.53 & 37.11 & 45.46 & 59.24 & 46.66 & 32.07 & 31.99 & 22.81 & 23.24 & 40.00 & 89.71 & 41.30 & 8.75 & 90.38 & 66.24 & 60.43 & 45.12 \\
 & PyramidKV & 26.11 & 30.19 & 39.05 & 52.04 & 40.62 & 29.61 & 29.64 & 22.06 & 21.88 & 32.00 & 89.24 & 41.15 & 8.50 & 92.29 & 65.56 & 59.99 & 42.50 \\
 & EA & 27.87 & 43.35 & 49.07 & 61.75 & 47.35 & 33.47 & 33.06 & 23.91 & 24.66 & 75.50 & 87.59 & 40.10 & 10.00 & 96.96 & 59.32 & 61.42 & 48.46 \\
 & KeyDiff & 23.54 & 28.85 & 40.31 & 36.49 & 33.94 & 16.10 & 29.51 & 22.21 & 22.04 & 54.50 & 85.62 & 38.45 & 5.15 & 86.08 & 43.78 & 50.61 & 38.57 \\
 & TOVA & 27.60 & 40.58 & 49.22 & 60.38 & 47.75 & 34.06 & 32.17 & 23.07 & 22.89 & 32.00 & 89.49 & 40.50 & 9.00 & 91.58 & 65.07 & 60.20 & 45.35 \\
 & PRKV & 29.92 & 43.86 & 50.18 & 62.46 & 48.05 & 35.69 & 33.36 & 24.15 & 24.25 & 68.50 & 90.24 & 39.04 & 12.00 & 95.62 & 65.07 & 60.28 & 48.92 \\
 & PRKV-GLB & 29.91 & 43.77 & 50.81 & 63.02 & 49.32 & 36.35 & 33.08 & 23.89 & 24.41 & 66.00 & 88.66 & 39.89 & 11.14 & 93.62 & 65.69 & 61.32 & 48.81 \\

\midrule

% ===== 75% =====
\multirow{7}{*}{\textbf{75\%}} 
 & SnapKV & 24.81 & 28.99 & 35.08 & 53.39 & 38.65 & 26.17 & 29.30 & 21.48 & 21.28 & 38.25 & 89.09 & 40.37 & 8.05 & 87.88 & 66.11 & 61.42 & 41.90 \\
 & PyramidKV & 23.21 & 26.39 & 31.85 & 45.56 & 36.25 & 23.45 & 27.58 & 20.62 & 21.00 & 33.75 & 88.99 & 40.87 & 7.00 & 83.46 & 65.99 & 60.53 & 39.78 \\
 & EA & 26.78 & 37.79 & 43.50 & 54.72 & 44.43 & 32.55 & 31.81 & 23.03 & 23.91 & 69.00 & 87.48 & 37.65 & 12.05 & 88.04 & 51.51 & 62.55 & 45.42 \\ 
 & KeyDiff & 14.94 & 18.81 & 29.38 & 21.14 & 28.55 & 9.37 & 21.77 & 21.03 & 13.99 & 40.50 & 81.92 & 35.81 & 6.31 & 46.08 & 25.85 & 51.33 & 29.17 \\
 & TOVA & 25.79 & 27.31 & 39.61 & 57.87 & 41.89 & 30.55 & 30.18 & 21.05 & 20.66 & 19.00 & 89.39 & 40.47 & 9.00 & 91.18 & 65.64 & 61.02 & 41.91 \\
 & PRKV & 27.52 & 37.11 & 43.31 & 55.96 & 44.27 & 30.82 & 31.66 & 23.36 & 22.84 & 67.50 & 90.04 & 37.07 & 13.25 & 89.29 & 64.00 & 62.09 & 46.26 \\
 & PRKV-GLB & 25.67 & 37.45 & 41.20 & 53.96 & 42.09 & 32.57 & 31.67 & 22.89 & 22.83 & 66.75 & 89.41 & 39.28 & 9.78 & 90.60 & 65.84 & 61.38 & 45.84 \\

\midrule

% ===== 90% =====
\multirow{7}{*}{\textbf{90\%}} 
 & SnapKV & 21.08 & 20.87 & 25.87 & 42.36 & 29.57 & 19.57 & 25.13 & 18.81 & 17.86 & 31.25 & 86.99 & 38.40 & 9.05 & 47.29 & 64.60 & 62.70 & 35.09 \\
 & PyramidKV & 16.97 & 20.81 & 26.35 & 38.23 & 28.65 & 14.55 & 24.91 & 18.82 & 17.78 & 30.50 & 87.49 & 38.62 & 8.55 & 45.29 & 64.88 & 62.26 & 34.04 \\
 & EA & 21.75 & 29.26 & 34.42 & 44.70 & 36.91 & 23.58 & 28.46 & 22.08 & 21.95 & 35.25 & 86.94 & 36.33 & 14.30 & 53.83 & 42.73 & 62.59 & 37.19 \\ 
 & KeyDiff & 10.47 & 7.96 & 21.82 & 15.17 & 23.66 & 6.90 & 16.03 & 19.69 & 8.18 & 7.50 & 78.83 & 30.96 & 7.21 & 14.08 & 12.18 & 49.45 & 20.63 \\
 & TOVA & 19.73 & 14.31 & 27.62 & 47.02 & 34.37 & 22.20 & 25.47 & 19.51 & 16.77 & 9.00 & 89.49 & 39.52 & 7.50 & 89.02 & 63.34 & 60.70 & 36.60 \\
 & PRKV & 22.86 & 23.98 & 32.12 & 46.24 & 34.52 & 23.88 & 27.47 & 21.85 & 19.52 & 30.50 & 86.83 & 35.47 & 15.05 & 59.98 & 59.57 & 61.57 & 37.59 \\
 & PRKV-GLB & 25.19 & 24.70 & 29.14 & 45.25 & 36.58 & 23.36 & 27.90 & 21.78 & 19.88 & 46.75 & 89.26 & 38.66 & 9.85 & 62.12 & 63.82 & 61.19 & 39.09 \\

\bottomrule
\end{tabular}}

\end{table*}

\begin{table*}[h]
\centering
\small
\caption{LongBench Full Result on Gemma-3-12B}
\label{tab:longbench-big_gemma}
\vspace{0.5em}
\setlength{\tabcolsep}{4.5pt}
\renewcommand{\arraystretch}{1.18}
\resizebox{\textwidth}{!}{
\begin{tabular}{@{} l |l|
*{3}{c} | % Single-Doc QA
*{3}{c} | % Multi-Doc QA
*{3}{c} | % Summarization
*{3}{c} | % Few-shot
*{2}{c} | % Synthetic
*{2}{c} | % Synthetic
c        % Avg
@{}}
\toprule
\multirow{2}{*}{\textbf{Ratio}} &
\multirow{2}{*}{\textbf{Method}} &
\multicolumn{3}{c|}{\textbf{Single-Document QA}} &
\multicolumn{3}{c|}{\textbf{Multi-Document QA}} &
\multicolumn{3}{c|}{\textbf{Summarization}} &
\multicolumn{3}{c|}{\textbf{Few-shot Learning}} &
\multicolumn{2}{c|}{\textbf{Synthetic}} &
\multicolumn{2}{c|}{\textbf{Code}} &
\multirow{2}{*}{\textbf{Avg}} \\
& &
\makecell{\texttt{NQA}} & \makecell{\texttt{QA}} & \makecell{\texttt{MQA}} & \makecell{\texttt{HQA}} &
\makecell{\texttt{2WM}} & \makecell{\texttt{MSQ}} & \makecell{\texttt{GR}} &
\makecell{\texttt{QM}} & \makecell{\texttt{MN}} & \makecell{\texttt{TR}} &
\makecell{\texttt{TQA}} & \makecell{\texttt{SS}} & \makecell{\texttt{PC}} &
\makecell{\texttt{PR}} & \makecell{\texttt{LCC}} & \makecell{\texttt{RB}} \\
\midrule

% ===== 100% =====
\textbf{0\%} & Full KV
& 25.59 & 41.18 & 58.17 & 61.42 & 55.95 & 37.83 & 29.89 & 23.01 & 23.14 & 70.50 & 90.42 & 43.92 & 4.50 & 98.00 & 68.11 & 64.94 & 49.79 \\
\midrule

% ===== 10% =====
\multirow{7}{*}{\textbf{10\%}}
 & SnapKV & 25.51 & 41.17 & 57.18 & 60.04 & 54.66 & 37.82 & 29.83 & 22.86 & 23.19 & 69.00 & 90.42 & 44.20 & 4.50 & 98.00 & 68.13 & 64.74 & 49.45 \\
 & PyramidKV & 26.04 & 41.11 & 57.66 & 61.17 & 55.82 & 36.89 & 29.77 & 22.66 & 23.04 & 70.50 & 90.42 & 43.74 & 4.50 & 98.00 & 67.99 & 65.03 & 49.65 \\
 & EA & 25.25 & 41.50 & 56.57 & 60.40 & 56.01 & 37.11 & 30.20 & 22.99 & 23.19 & 71.00 & 90.92 & 43.89 & 3.50 & 97.00 & 67.64 & 64.48 & 49.48 \\ 
 & KeyDiff & 27.68 & 35.55 & 57.48 & 54.80 & 51.35 & 32.74 & 29.77 & 23.43 & 23.10 & 68.50 & 89.96 & 43.92 & 6.00 & 98.50 & 67.41 & 63.02 & 48.33 \\
 & TOVA & 25.50 & 41.86 & 58.22 & 61.12 & 55.03 & 37.51 & 29.77 & 22.78 & 23.30 & 69.00 & 90.42 & 43.87 & 4.00 & 98.00 & 68.36 & 65.29 & 49.63 \\
 & PRKV & 25.62 & 41.48 & 57.47 & 60.80 & 55.36 & 36.81 & 30.36 & 23.01 & 23.30 & 72.00 & 90.92 & 43.91 & 3.00 & 97.00 & 67.91 & 64.13 & 49.57 \\
 & PRKV-GLB & 25.75 & 41.51 & 57.83 & 61.12 & 55.03 & 37.34 & 29.86 & 23.45 & 23.44 & 71.00 & 90.42 & 43.73 & 4.50 & 98.00 & 68.53 & 64.68 & 49.76 \\
\midrule

% ===== 25% =====
\multirow{7}{*}{\textbf{25\%}} 
 & SnapKV & 24.23 & 42.58 & 56.77 & 59.40 & 55.19 & 36.46 & 29.73 & 22.81 & 22.84 & 70.00 & 90.42 & 44.27 & 4.50 & 97.50 & 68.00 & 65.07 & 49.36 \\
 & PyramidKV & 24.82 & 42.15 & 54.92 & 59.85 & 53.23 & 34.92 & 29.57 & 22.73 & 22.63 & 68.50 & 90.42 & 43.89 & 4.50 & 98.50 & 67.77 & 65.48 & 48.99 \\
 & EA & 25.96 & 41.04 & 56.55 & 62.01 & 57.67 & 36.58 & 30.17 & 22.71 & 23.20 & 72.50 & 90.76 & 43.32 & 2.50 & 89.50 & 66.83 & 63.59 & 49.06 \\
 & KeyDiff & 27.82 & 27.13 & 53.47 & 50.23 & 50.84 & 28.80 & 29.55 & 23.17 & 23.06 & 63.00 & 91.10 & 42.49 & 4.81 & 95.50 & 66.82 & 58.57 & 46.02 \\
 & TOVA & 24.88 & 40.40 & 57.47 & 60.39 & 55.66 & 36.90 & 29.55 & 22.86 & 22.91 & 65.50 & 90.42 & 44.12 & 4.50 & 98.00 & 68.39 & 65.16 & 49.19 \\
 & PRKV & 24.96 & 40.56 & 55.67 & 60.67 & 56.61 & 37.46 & 30.28 & 22.60 & 23.00 & 73.00 & 90.26 & 43.95 & 2.50 & 89.50 & 66.99 & 64.17 & 48.89 \\
 & PRKV-GLB & 25.05 & 41.58 & 58.18 & 59.34 & 55.56 & 37.63 & 29.87 & 22.76 & 23.32 & 70.00 & 90.42 & 43.60 & 4.00 & 98.50 & 68.68 & 65.10 & 49.60 \\
\midrule

% ===== 50% =====
\multirow{7}{*}{\textbf{50\%}} 
 & SnapKV & 24.99 & 39.64 & 48.10 & 57.22 & 51.08 & 33.87 & 28.83 & 22.05 & 22.13 & 61.00 & 90.42 & 43.85 & 4.00 & 98.00 & 68.25 & 64.52 & 47.37 \\
 & PyramidKV & 26.92 & 37.71 & 44.49 & 56.02 & 49.72 & 32.27 & 27.99 & 21.43 & 21.73 & 58.00 & 90.42 & 44.73 & 7.00 & 96.50 & 67.94 & 64.91 & 46.74 \\
 & EA & 26.90 & 38.66 & 53.93 & 58.23 & 56.61 & 34.24 & 29.57 & 22.33 & 23.28 & 70.50 & 91.09 & 41.31 & 4.50 & 80.00 & 67.07 & 63.10 & 47.58 \\ 
 & KeyDiff & 20.56 & 22.82 & 43.76 & 43.53 & 40.01 & 22.64 & 27.66 & 23.10 & 22.17 & 55.50 & 89.01 & 42.12 & 1.99 & 83.50 & 54.99 & 52.26 & 40.35 \\
 & TOVA & 25.01 & 39.03 & 51.10 & 58.11 & 54.55 & 33.36 & 29.54 & 22.64 & 22.32 & 53.00 & 90.42 & 44.03 & 4.50 & 99.00 & 67.49 & 65.34 & 47.47 \\
 & PRKV & 25.32 & 38.75 & 54.10 & 57.16 & 56.66 & 34.16 & 29.48 & 22.02 & 22.96 & 70.50 & 90.67 & 41.79 & 6.00 & 81.00 & 67.10 & 62.94 & 47.54 \\
 & PRKV-GLB & 24.37 & 39.38 & 52.45 & 58.35 & 55.53 & 35.16 & 29.26 & 22.10 & 22.58 & 60.00 & 90.42 & 43.59 & 3.50 & 97.50 & 67.97 & 64.05 & 47.89 \\
\midrule

% ===== 75% =====
\multirow{7}{*}{\textbf{75\%}} 
 & SnapKV & 25.69 & 35.07 & 33.10 & 53.39 & 46.50 & 30.86 & 27.17 & 20.87 & 21.06 & 57.50 & 90.42 & 43.54 & 4.50 & 91.50 & 67.70 & 60.96 & 44.36 \\
 & PyramidKV & 22.51 & 34.02 & 32.63 & 49.51 & 43.50 & 27.04 & 26.96 & 20.22 & 20.98 & 53.00 & 89.92 & 43.94 & 5.00 & 89.00 & 67.92 & 62.06 & 43.01 \\
 & EA & 25.59 & 35.22 & 44.99 & 55.42 & 48.71 & 33.80 & 28.28 & 21.20 & 22.34 & 64.00 & 90.84 & 38.25 & 5.50 & 48.00 & 64.61 & 62.51 & 43.08 \\ 
 & KeyDiff & 15.24 & 20.56 & 30.58 & 30.56 & 28.75 & 9.29 & 23.16 & 21.81 & 19.07 & 47.00 & 88.39 & 38.65 & 1.79 & 47.50 & 37.78 & 42.15 & 31.39 \\
 & TOVA & 24.49 & 32.03 & 43.35 & 54.39 & 45.46 & 29.27 & 28.14 & 21.10 & 21.26 & 30.75 & 89.92 & 42.63 & 3.00 & 99.00 & 65.75 & 62.90 & 43.34 \\
 & PRKV & 26.35 & 35.80 & 43.35 & 55.98 & 48.77 & 32.49 & 27.76 & 21.17 & 21.75 & 61.00 & 89.05 & 38.46 & 6.00 & 55.50 & 66.61 & 62.18 & 43.26 \\
 & PRKV-GLB & 24.99 & 34.38 & 38.08 & 57.00 & 48.54 & 32.28 & 27.72 & 21.43 & 21.38 & 48.00 & 90.26 & 43.46 & 3.50 & 87.50 & 67.80 & 60.45 & 44.17 \\
\midrule

% ===== 90% =====
\multirow{7}{*}{\textbf{90\%}} 
 & SnapKV & 19.68 & 26.36 & 25.37 & 44.08 & 30.85 & 25.25 & 25.30 & 19.35 & 19.31 & 36.00 & 90.26 & 41.64 & 3.50 & 64.50 & 64.71 & 60.21 & 37.27 \\
 & PyramidKV & 16.06 & 26.99 & 25.39 & 41.62 & 30.85 & 20.92 & 25.03 & 19.29 & 19.36 & 36.00 & 90.92 & 41.74 & 4.50 & 64.50 & 64.99 & 59.57 & 36.73 \\
 & EA & 21.52 & 27.58 & 35.81 & 49.26 & 36.48 & 27.19 & 26.14 & 20.27 & 21.05 & 37.00 & 89.04 & 36.22 & 6.50 & 34.50 & 59.82 & 60.51 & 36.81 \\
 & KeyDiff & 8.69 & 14.57 & 17.28 & 20.98 & 26.47 & 2.76 & 18.66 & 20.80 & 17.57 & 25.00 & 82.11 & 33.77 & 4.02 & 19.00 & 18.71 & 34.63 & 22.81 \\
 & TOVA & 20.12 & 21.84 & 28.91 & 49.04 & 35.23 & 27.52 & 26.36 & 20.04 & 19.57 & 19.00 & 90.42 & 40.75 & 4.50 & 86.50 & 63.85 & 60.59 & 38.39 \\
 & PRKV & 23.64 & 25.33 & 31.92 & 44.20 & 38.93 & 26.80 & 25.50 & 20.23 & 19.90 & 35.25 & 86.85 & 36.25 & 6.00 & 28.50 & 62.66 & 61.23 & 35.82 \\
 & PRKV-GLB & 22.89 & 27.30 & 25.89 & 50.91 & 32.70 & 25.42 & 25.45 & 19.56 & 19.82 & 46.50 & 89.76 & 41.72 & 4.50 & 42.50 & 66.06 & 59.78 & 37.55 \\
\bottomrule
\end{tabular}}

\end{table*}

\clearpage

\section{Limitation and Future Work} \label{appendix:limitation}
PRKV has several limitations. First, like other attention-based KV eviction methods, PRKV relies on observed attention as a starting point for future token relevance. Although multi-hop propagation reduces sensitivity to transient local attention, its quality depends on the fidelity of the constructed transition operator. Second, while PRKV is training-free and model-agnostic, it introduces additional computation for importance propagation and score updates, whose overhead depends on eviction frequency and implementation efficiency. Finally, our theoretical analysis provides a qualitative decomposition of approximation error, but tighter bounds that predict downstream task performance remain an open direction. For future work, we plan to develop more efficient kernel-level implementations and further optimize PRKV for practical inference frameworks such as vLLM.

\section{More visualization}
\begin{figure*}[h]
    \centering

    % -------- Column headers --------
    \makebox[\linewidth][c]{
        \begin{minipage}{0.23\linewidth}\centering \small \textbf{Top-$K$ Overlap} \end{minipage}
        \begin{minipage}{0.23\linewidth}\centering \small \textbf{Hop Sweep} \end{minipage}
        \begin{minipage}{0.23\linewidth}\centering \small \textbf{AUC} \end{minipage}
        \begin{minipage}{0.23\linewidth}\centering \small \textbf{Logit KL} \end{minipage}
    }

    \vspace{0.3em}

    % -------- Layer 5 --------
    \subfigure[5\textsuperscript{th} layer]{
        \includegraphics[width=0.23\linewidth]{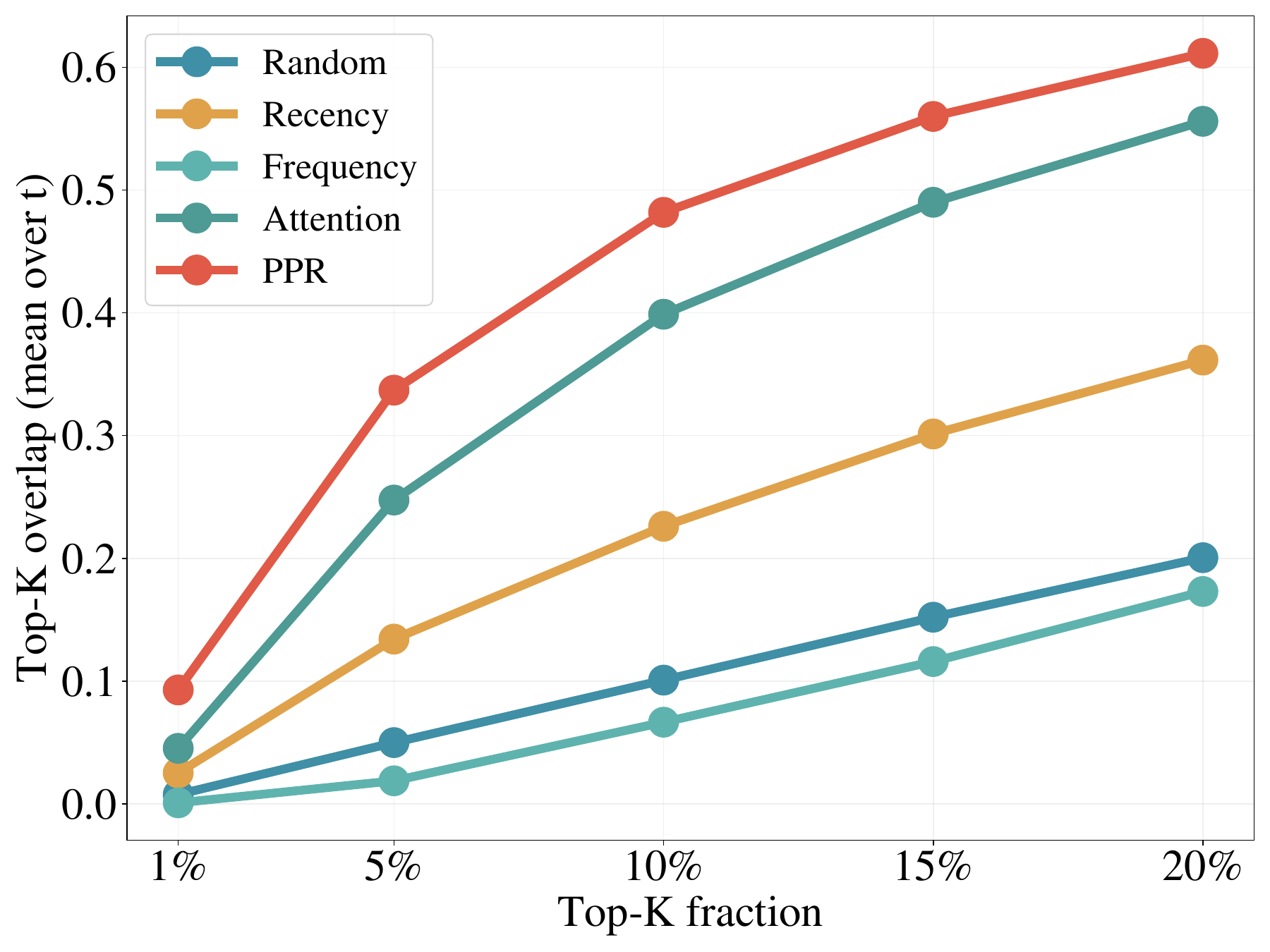}
        \includegraphics[width=0.23\linewidth]{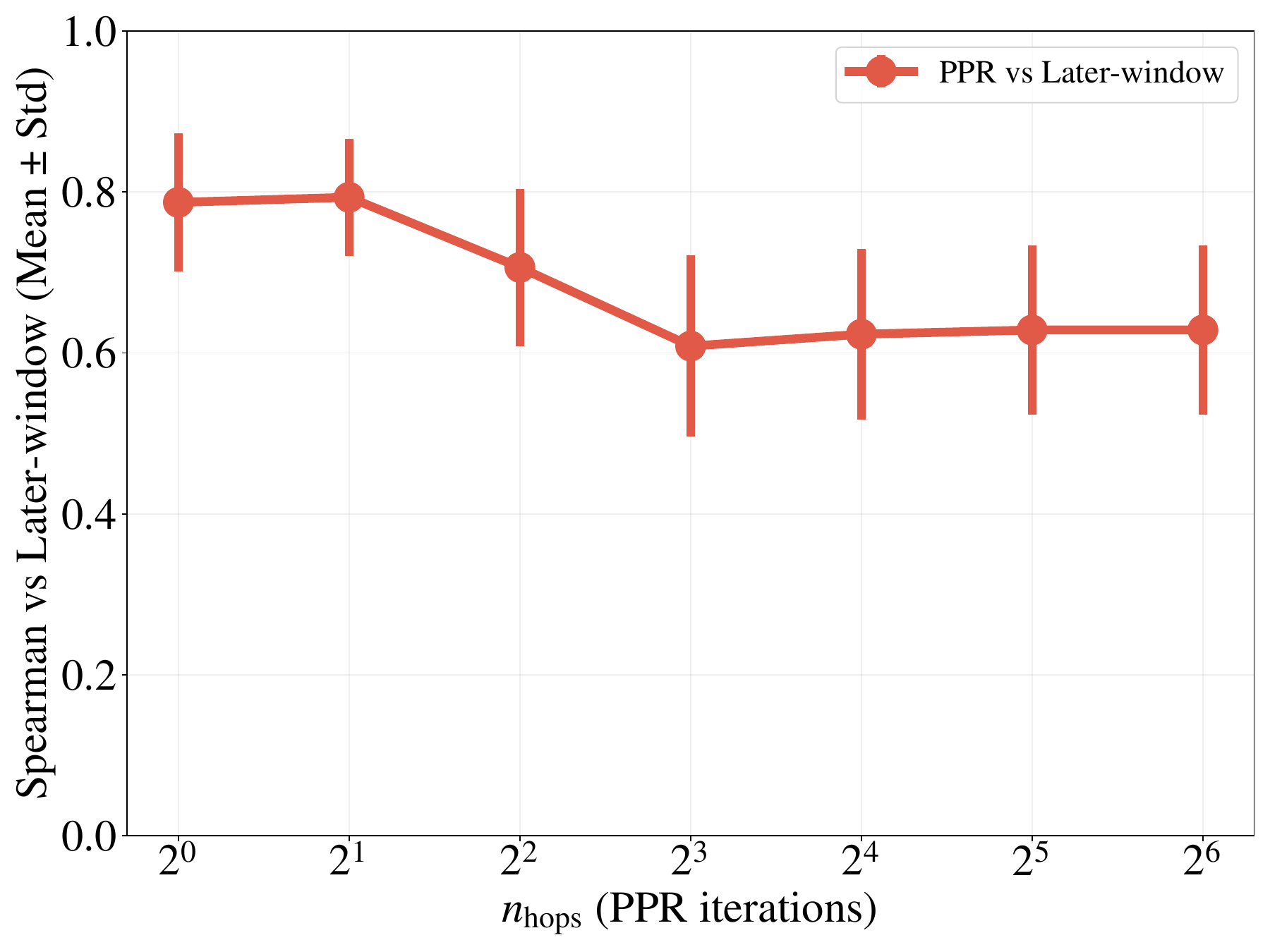}
        \includegraphics[width=0.23\linewidth]{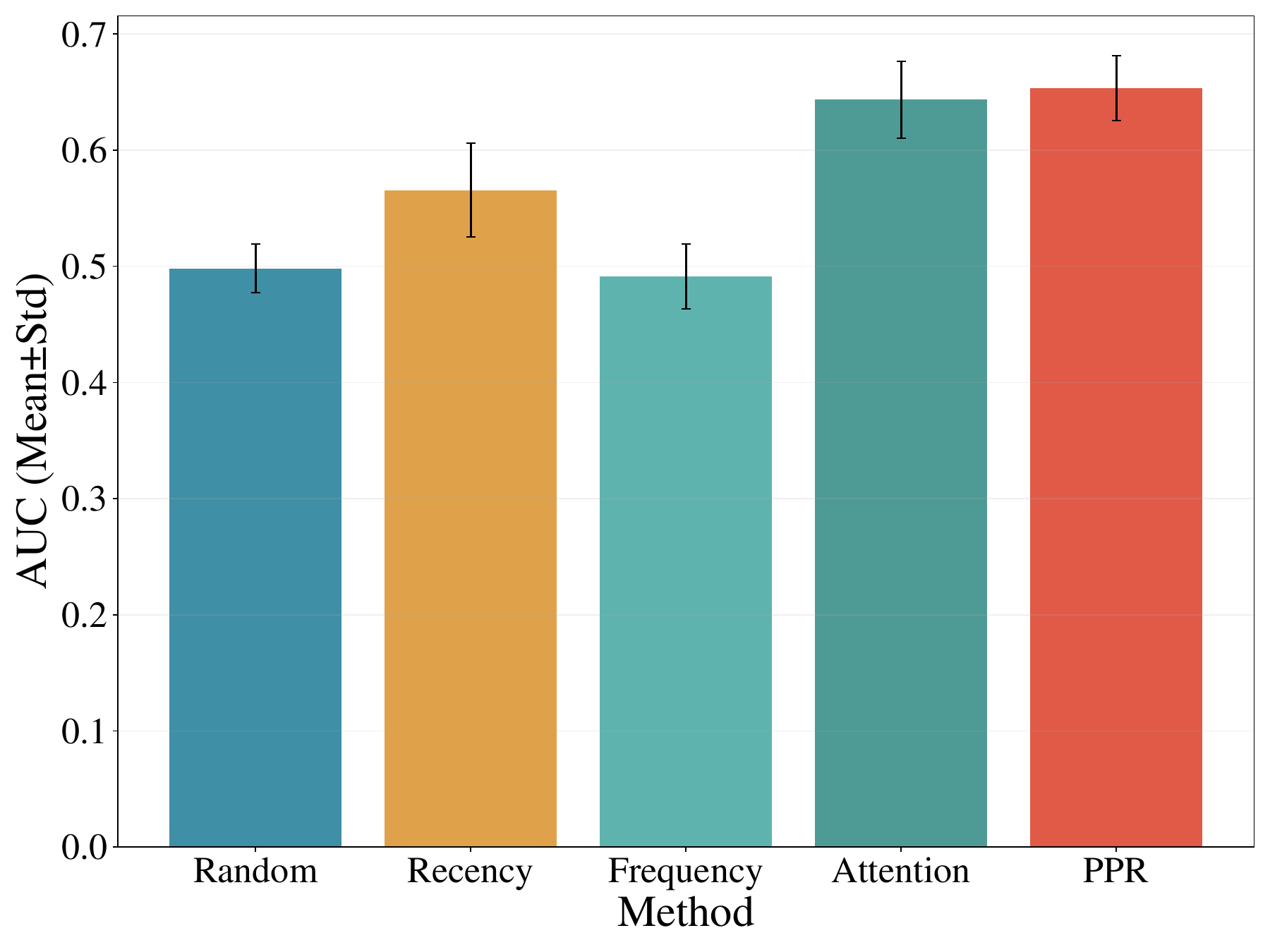}
        \includegraphics[width=0.23\linewidth]{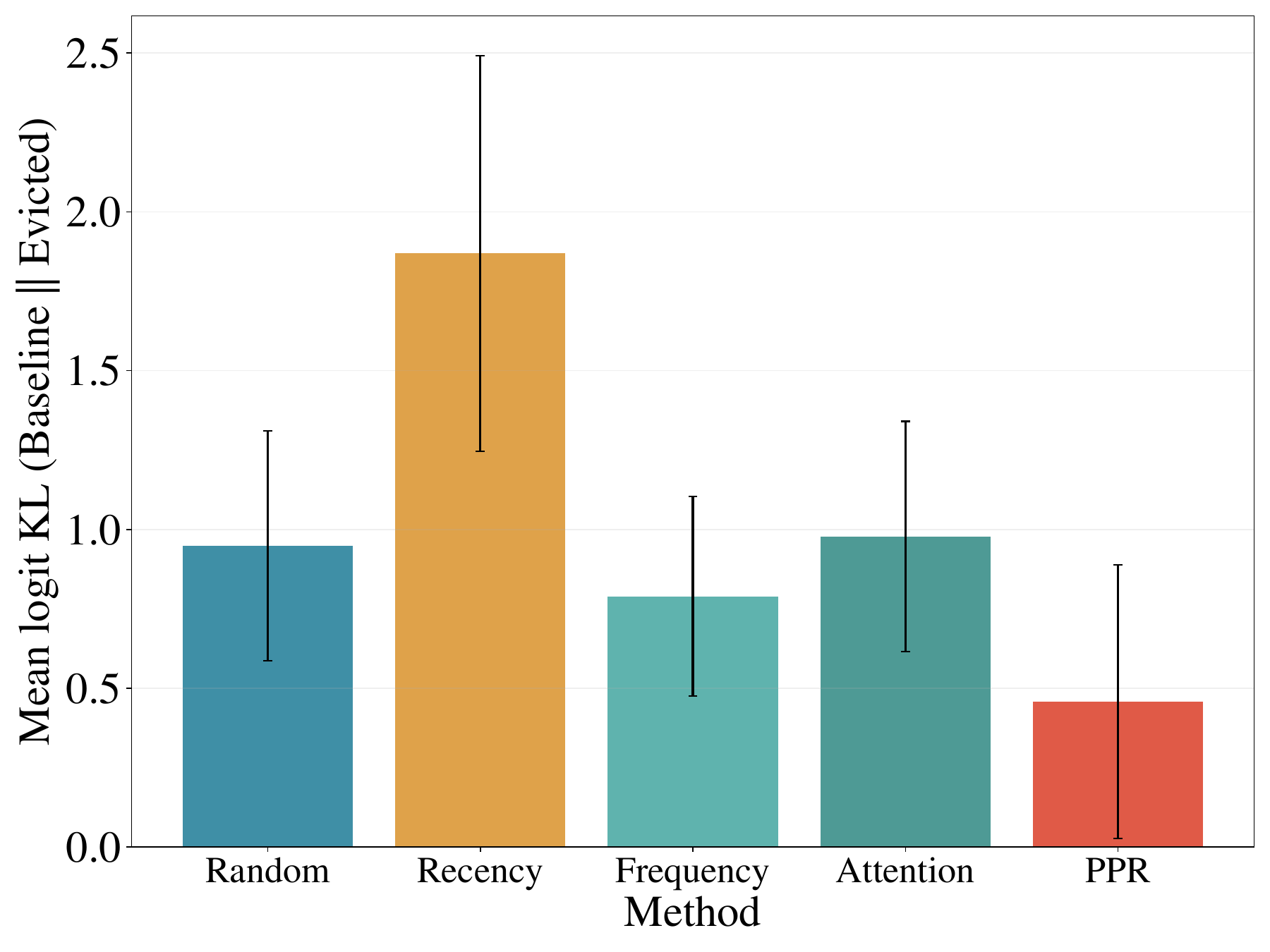}
    }

    % -------- Layer 20 --------
    \subfigure[20\textsuperscript{th} layer]{
        \includegraphics[width=0.23\linewidth]{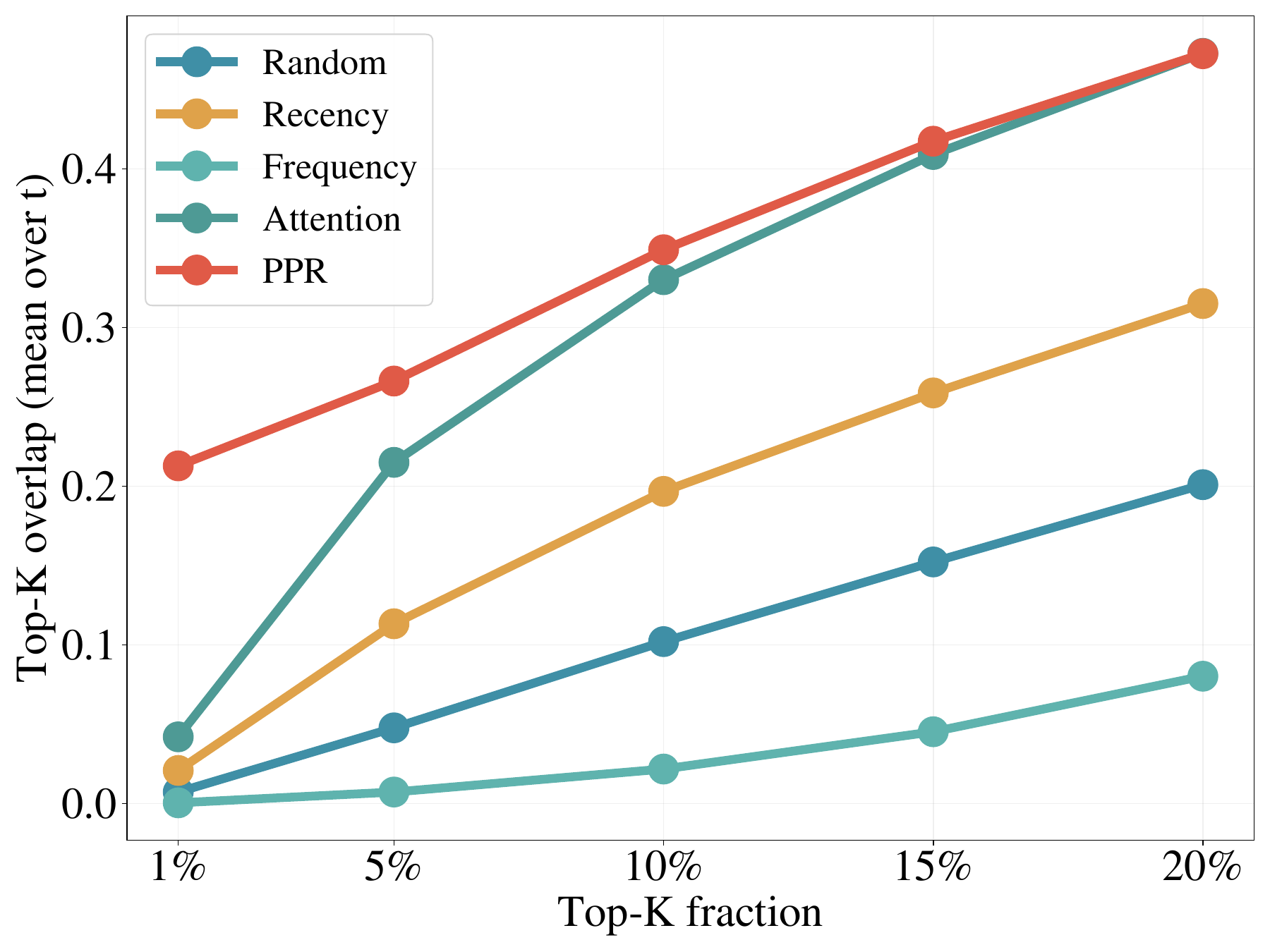}
        \includegraphics[width=0.23\linewidth]{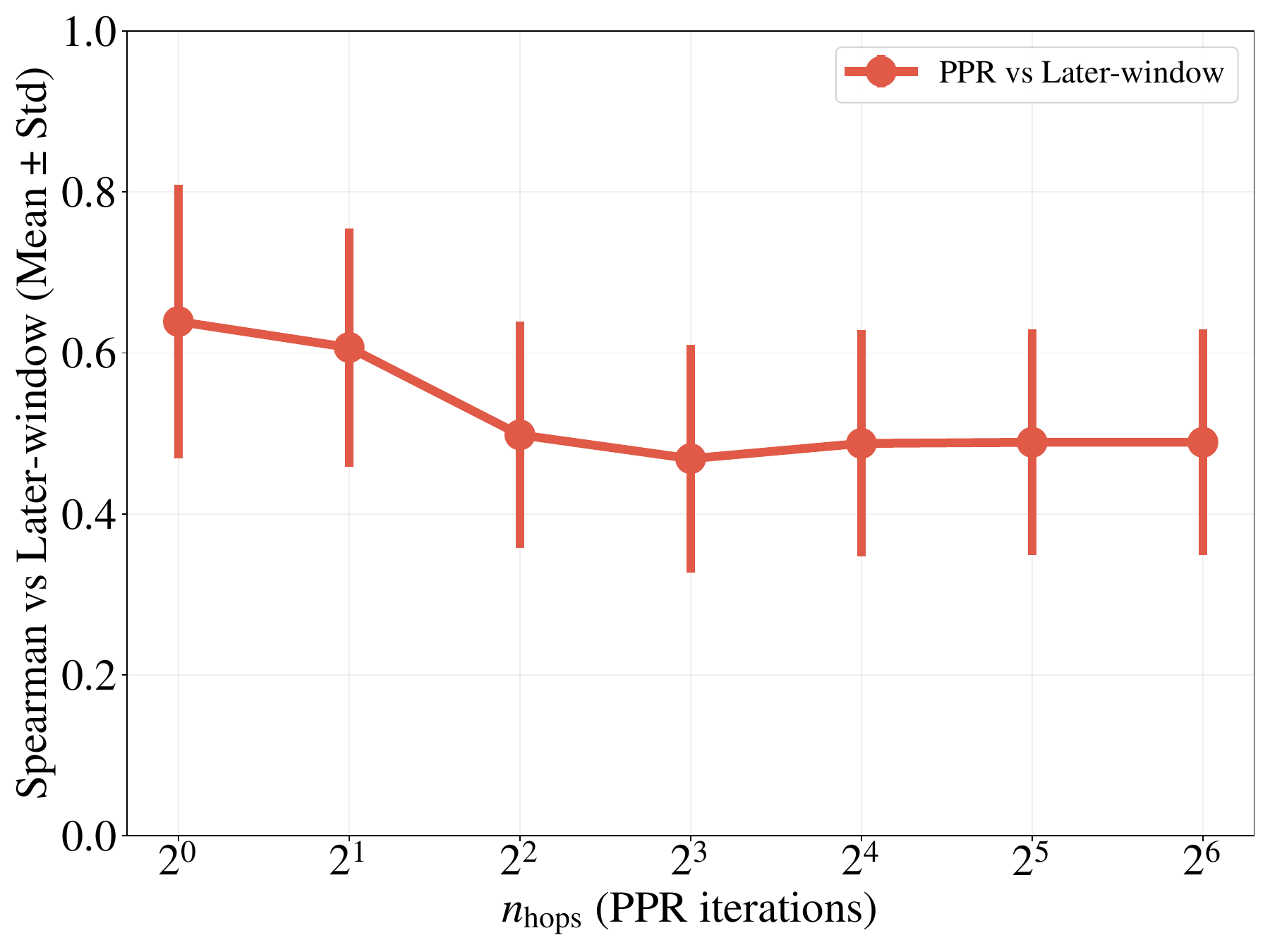}
        \includegraphics[width=0.23\linewidth]{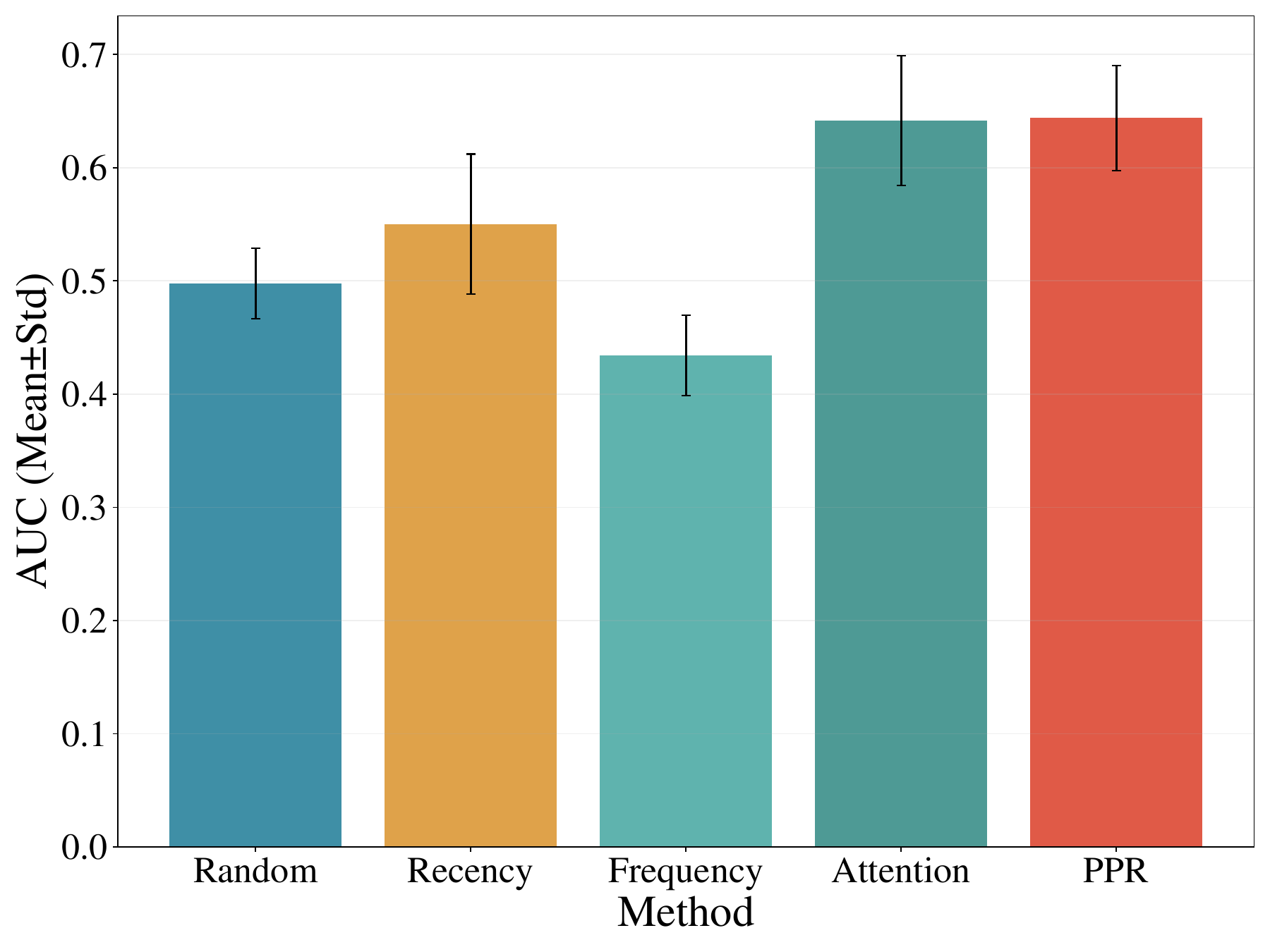}
        \includegraphics[width=0.23\linewidth]{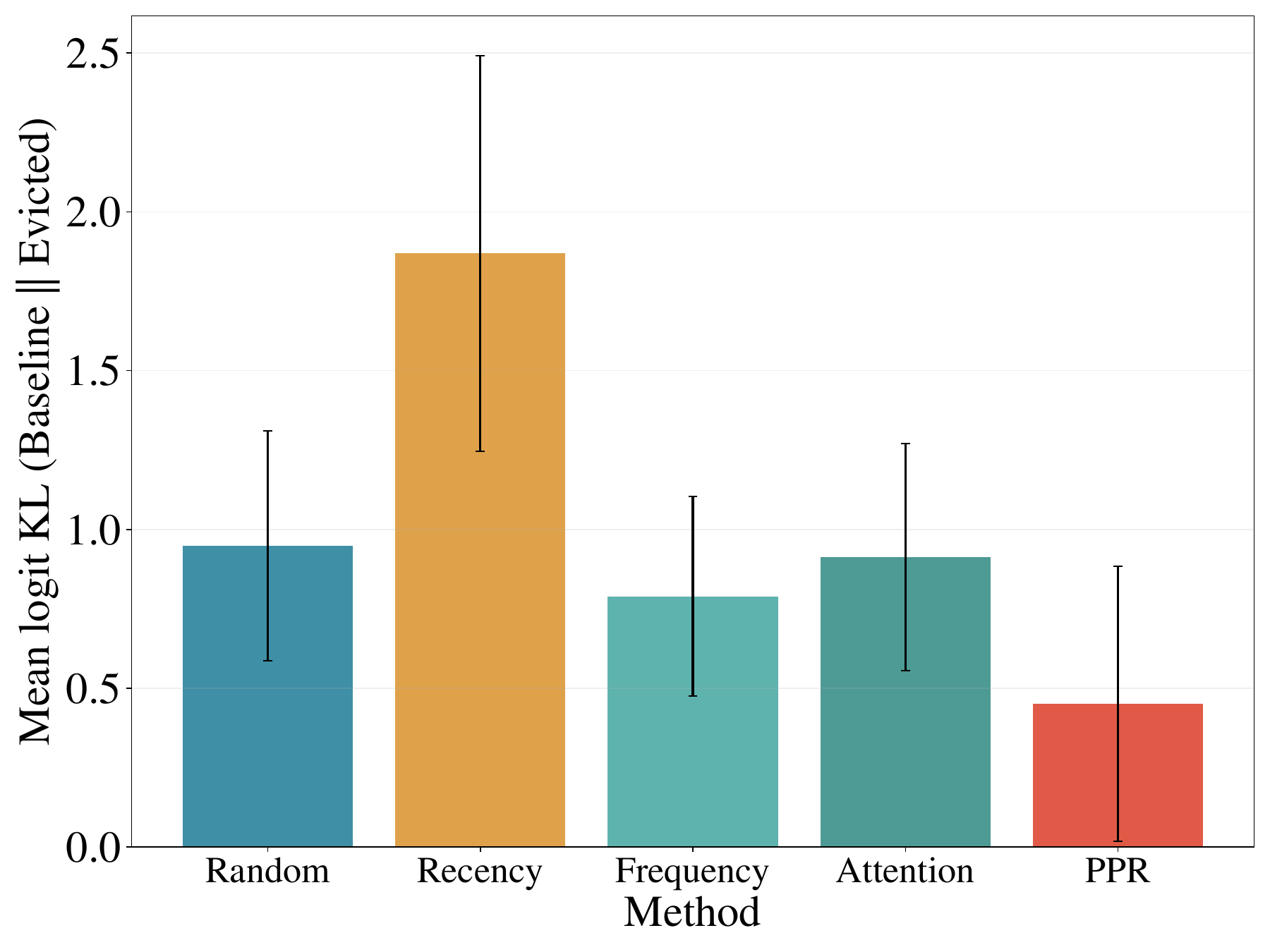}
    }

    % -------- Layer 32 --------
    \subfigure[32\textsuperscript{nd} layer]{
        \includegraphics[width=0.23\linewidth]{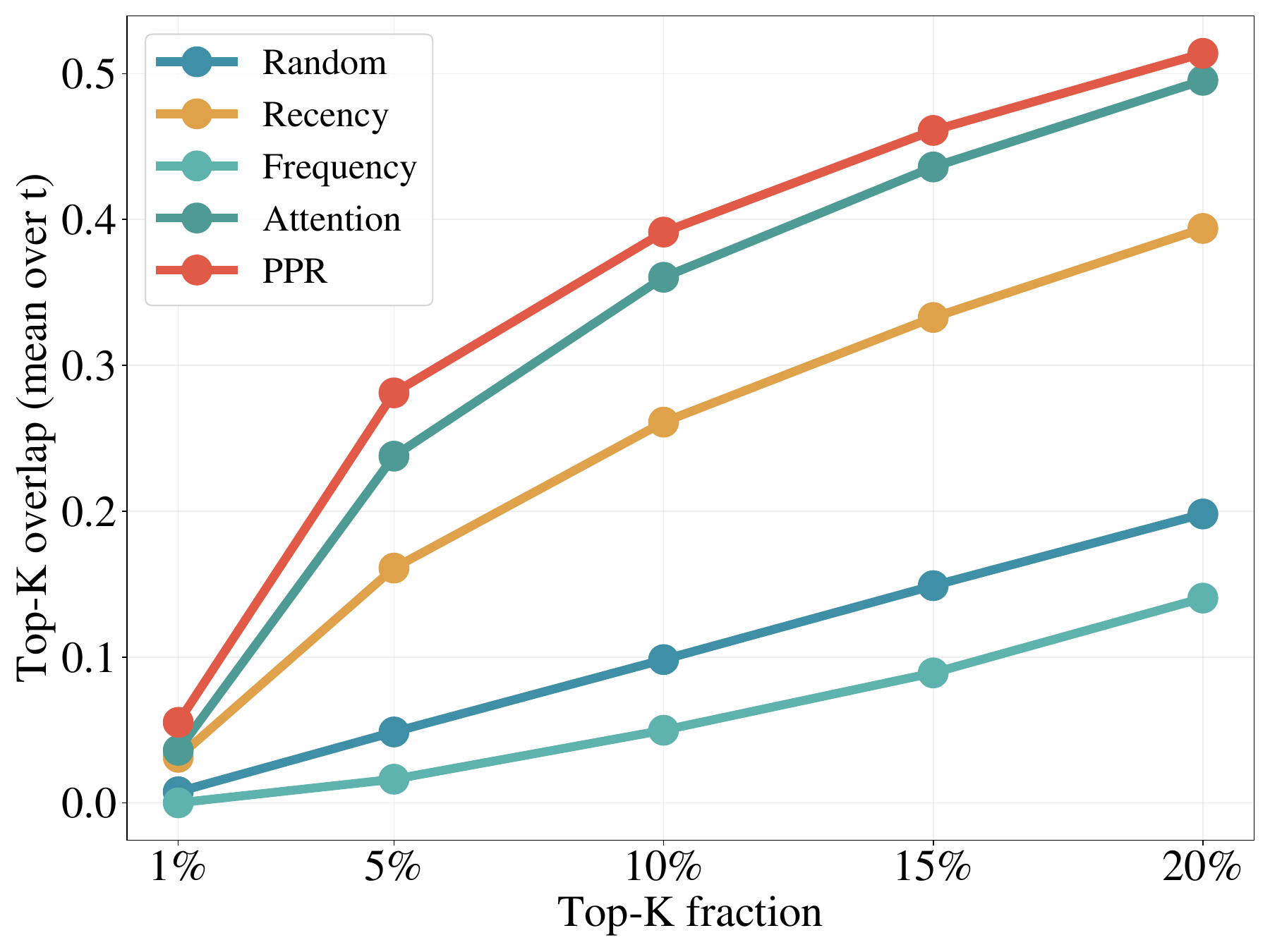}
        \includegraphics[width=0.23\linewidth]{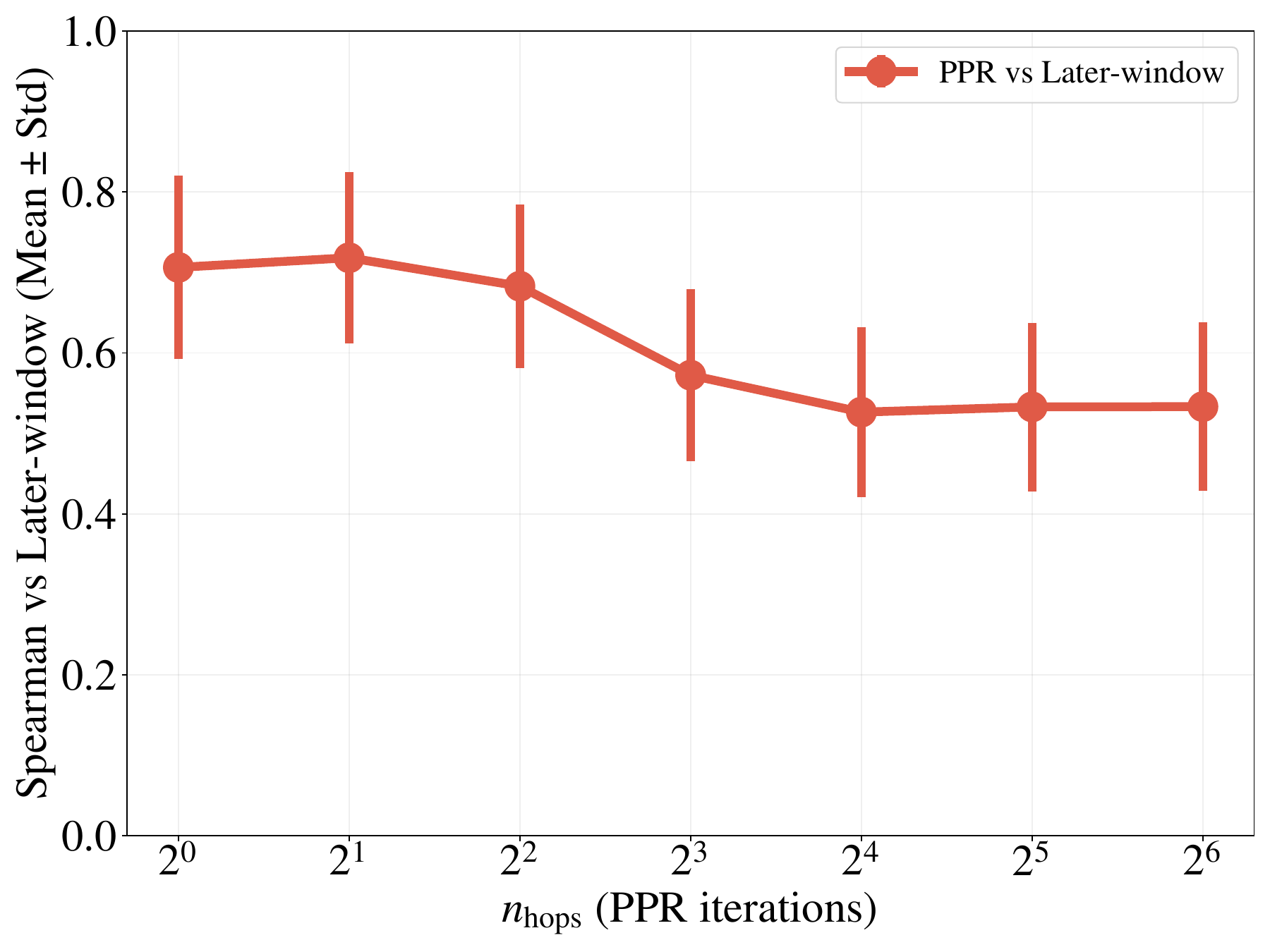}
        \includegraphics[width=0.23\linewidth]{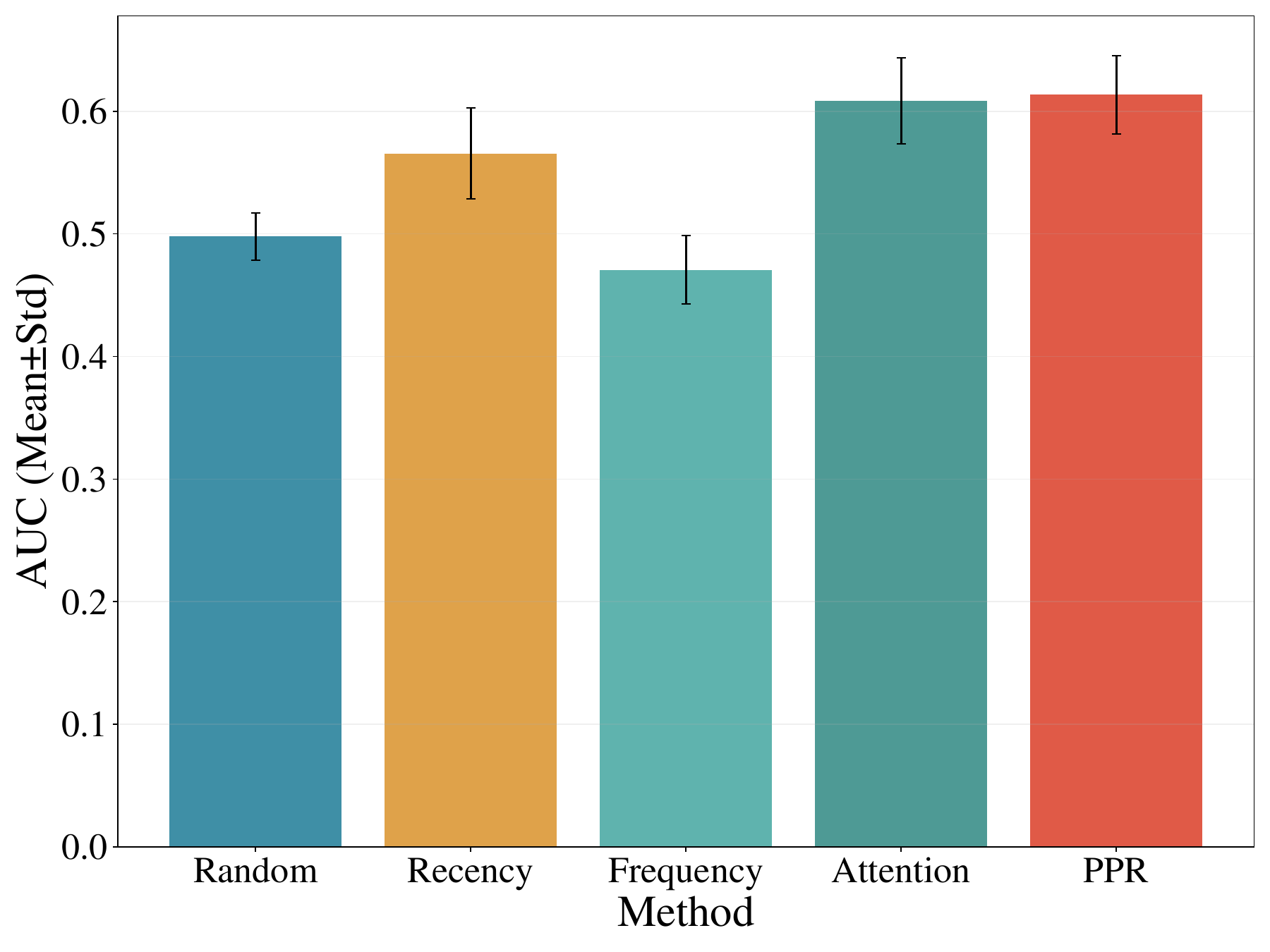}
        \includegraphics[width=0.23\linewidth]{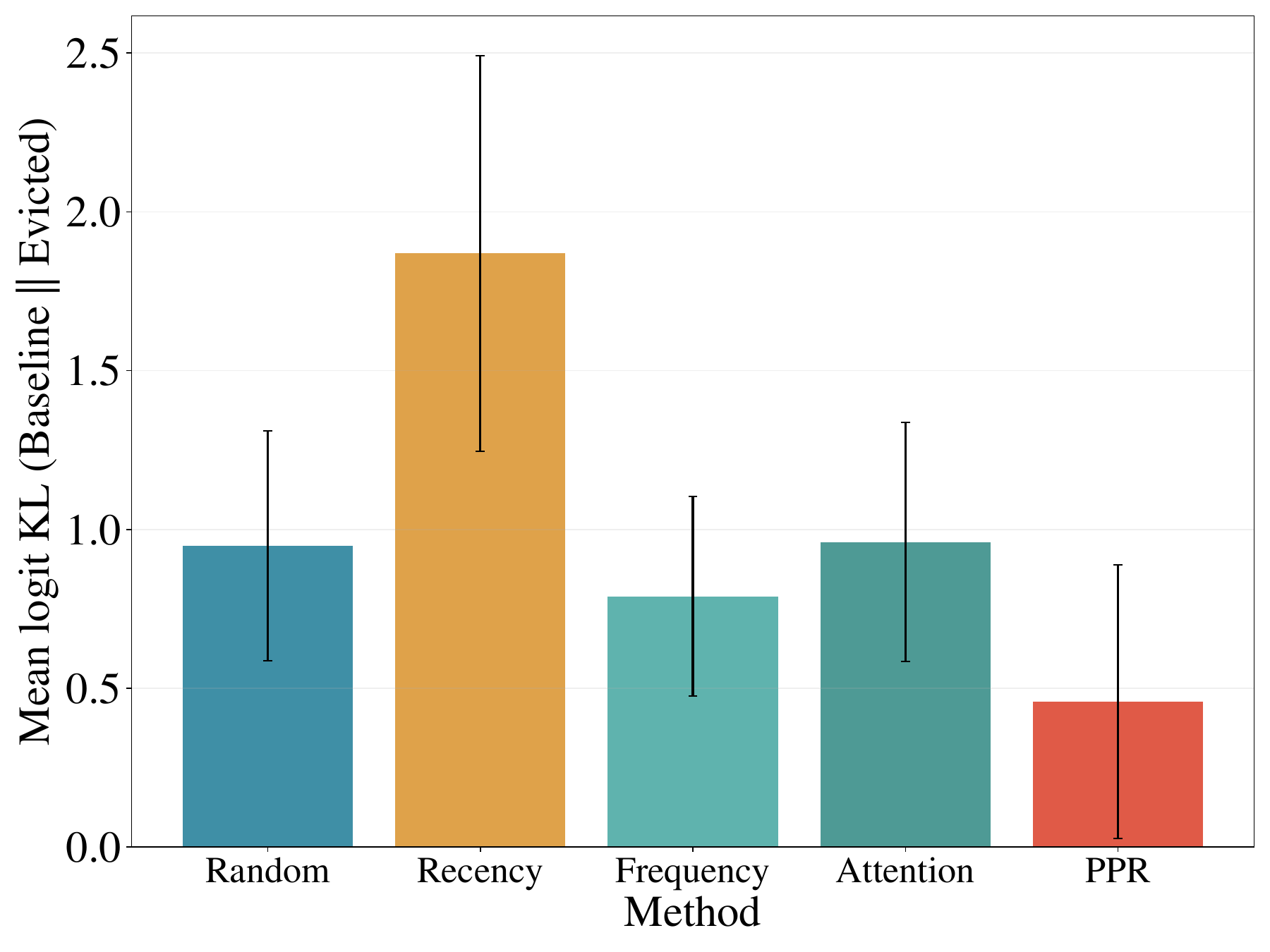}
    }

    \caption{
        \textbf{Layer-wise motivational analysis of PPR.}
        From left to right, the columns show:
        (1) Top-$K$ overlap with later-window attention under budgeted eviction,
        (2) predictive stability of importance scores across PPR hops,
        (3) alignment with future attention usage measured by AUC, and
        (4) eviction-induced output distortion measured by logit KL divergence.
        Results are shown for the 5\textsuperscript{th}, 20\textsuperscript{th}, and 32\textsuperscript{nd} layers of Llama-3.1-8B.
    }
    \label{fig:motivational_analysis_layers}
    \vspace{-1em}
\end{figure*}

\begin{figure*}[h]
    \centering

    % -------- Column headers --------
    \makebox[\linewidth][c]{
        \begin{minipage}{0.23\linewidth}\centering \small \textbf{PRKV} \\ \small Transition Matrix \end{minipage}
        \begin{minipage}{0.23\linewidth}\centering \small \textbf{PRKV-GLB} \\ \small Transition Matrix \end{minipage}
        \begin{minipage}{0.23\linewidth}\centering \small \textbf{PRKV} \\ \small Score Evolution \end{minipage}
        \begin{minipage}{0.23\linewidth}\centering \small \textbf{PRKV-GLB} \\ \small Score Evolution \end{minipage}
    }

    \vspace{0.3em}

    % -------- Layer 10 --------
    \subfigure[10\textsuperscript{th} layer]{
        \includegraphics[width=0.23\linewidth]{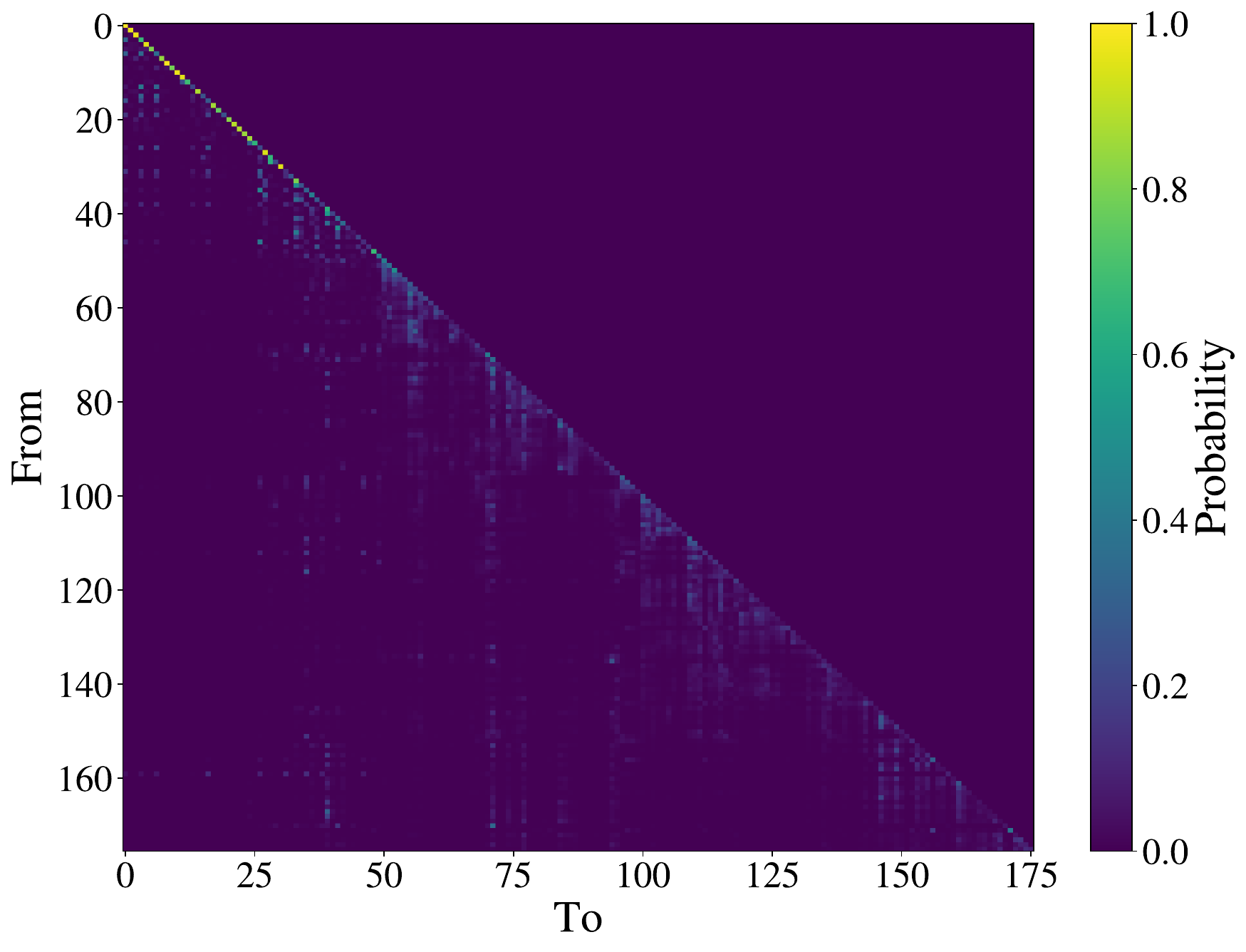}
        \includegraphics[width=0.23\linewidth]{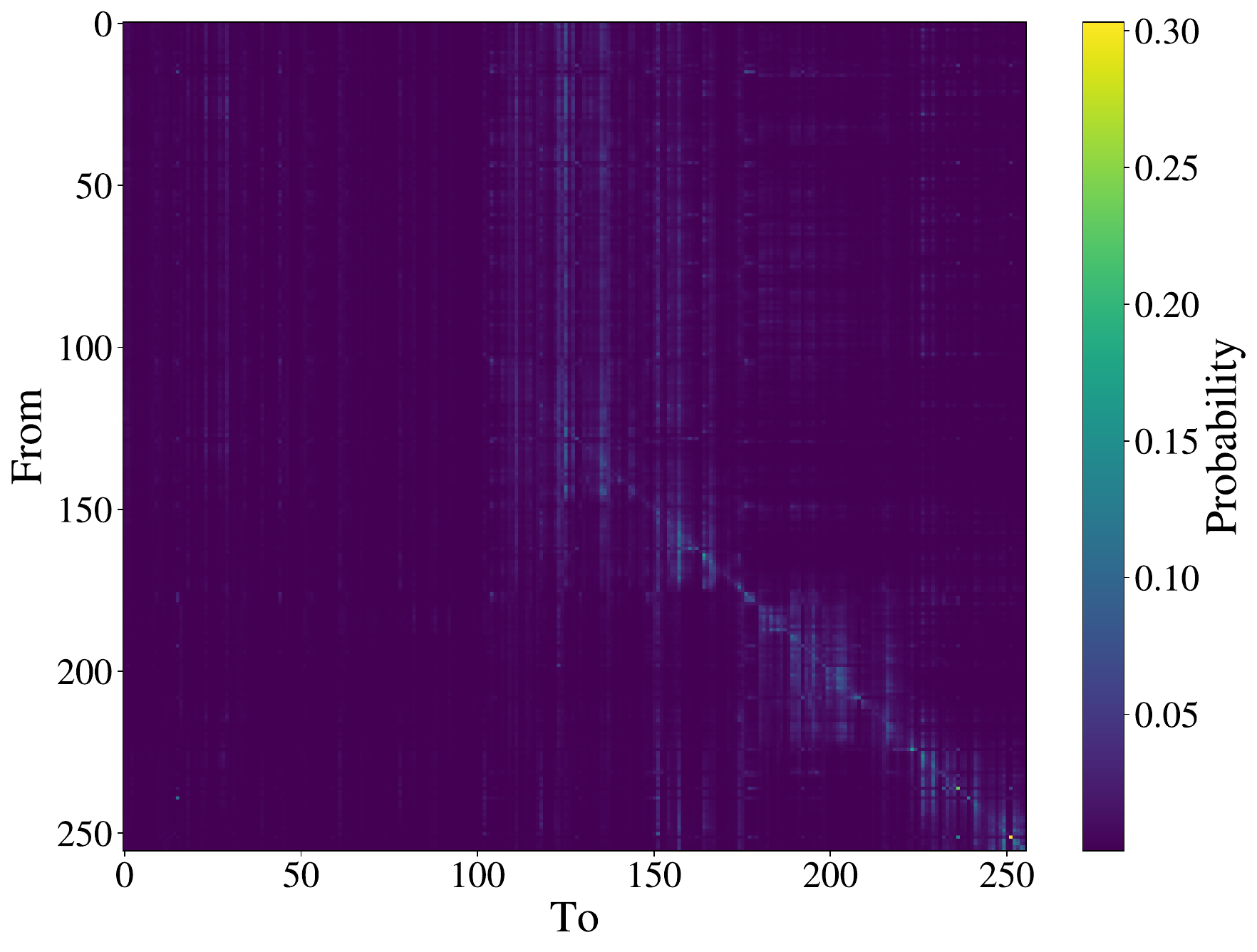}
        \includegraphics[width=0.23\linewidth]{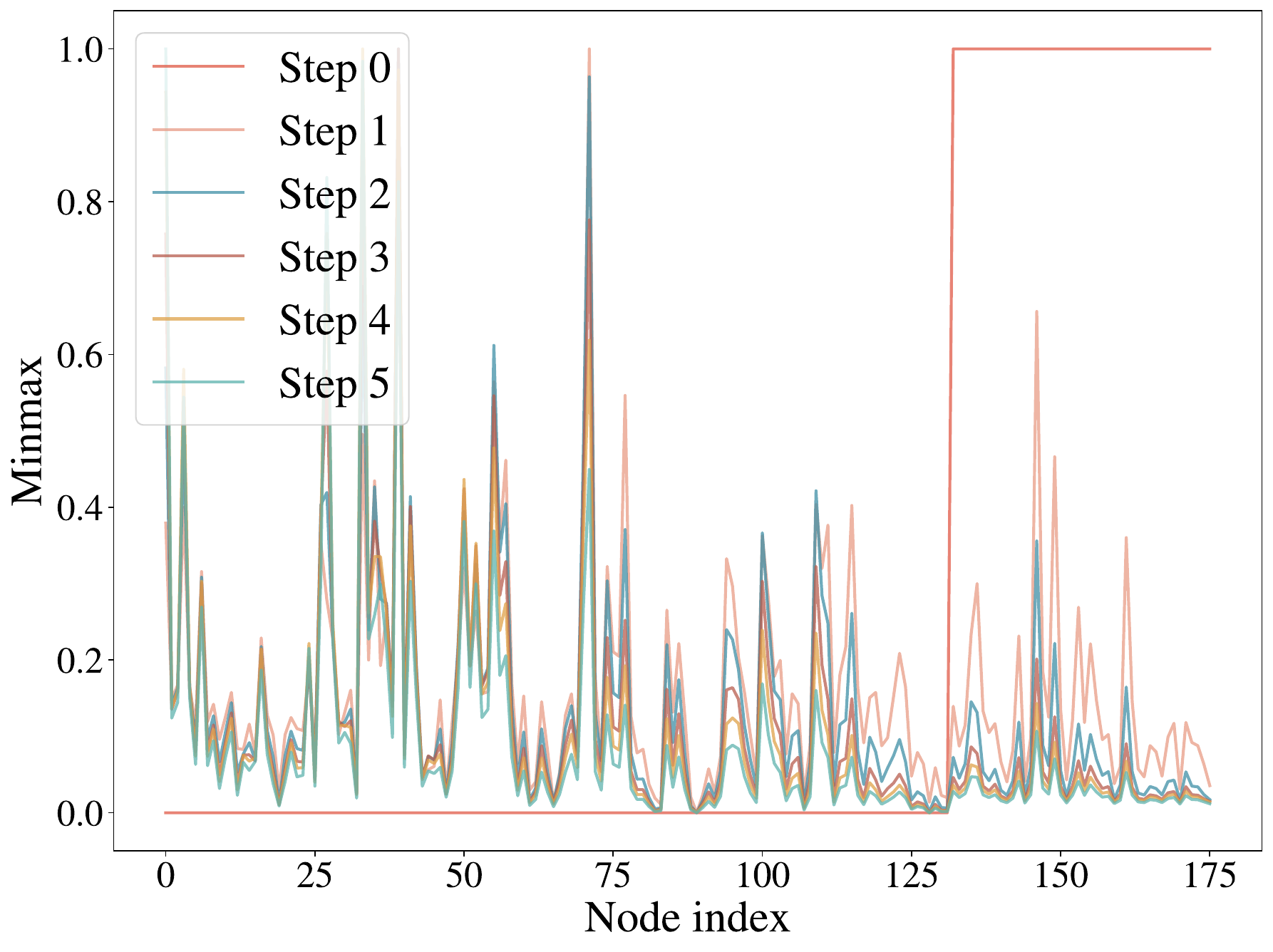}
        \includegraphics[width=0.23\linewidth]{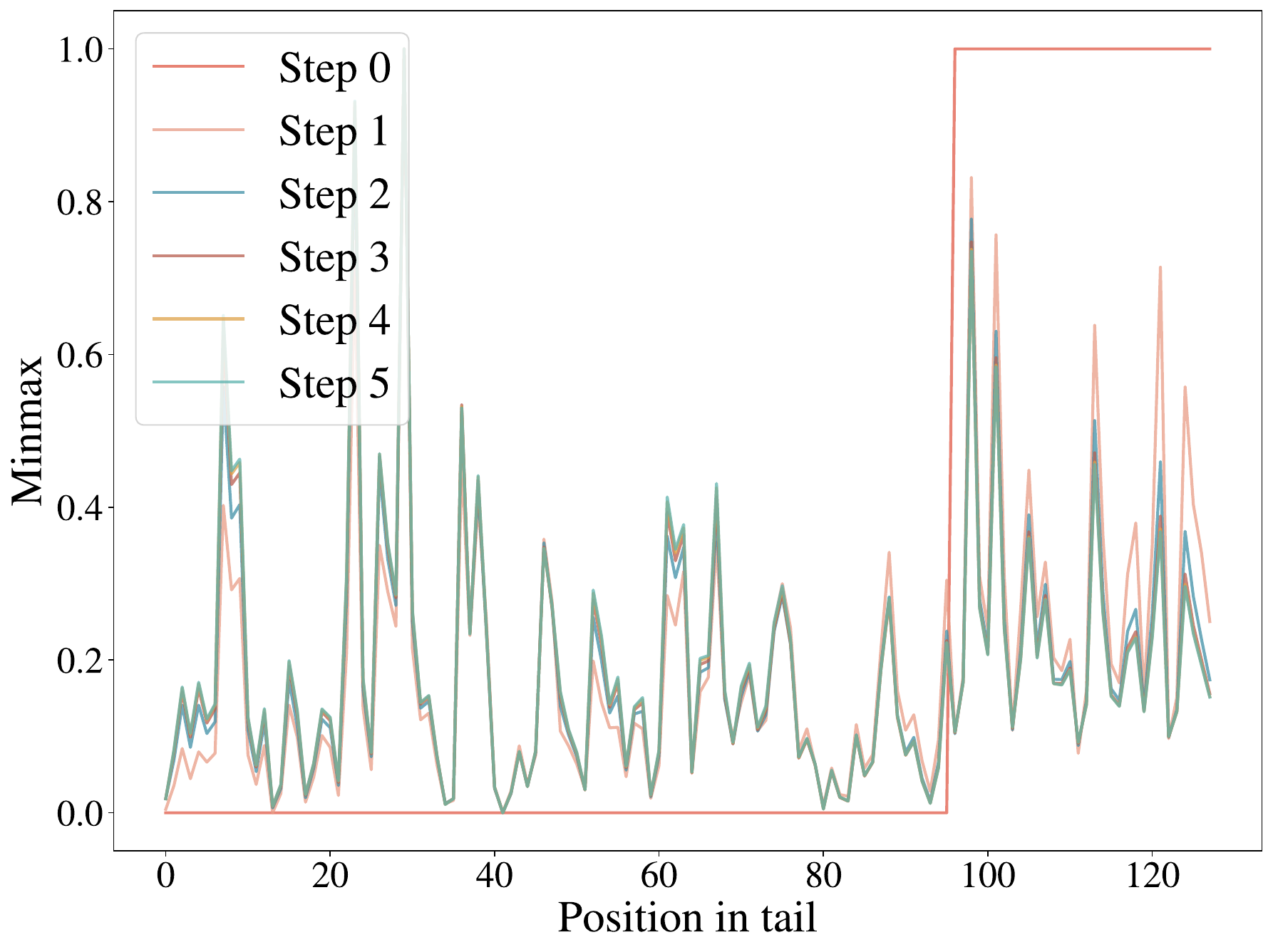}
    }

    % -------- Layer 20 --------
    \subfigure[20\textsuperscript{th} layer]{
        \includegraphics[width=0.23\linewidth]{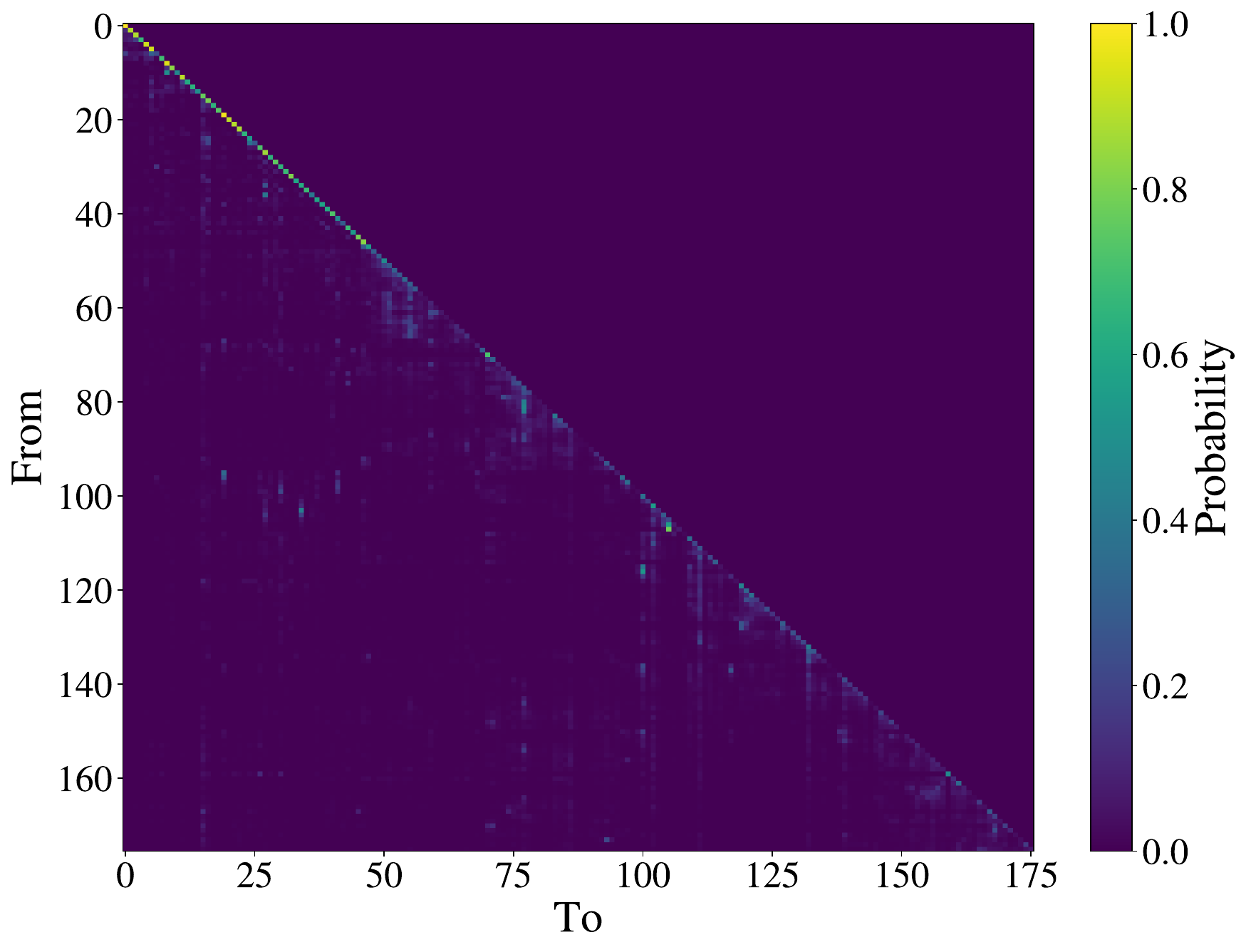}
        \includegraphics[width=0.23\linewidth]{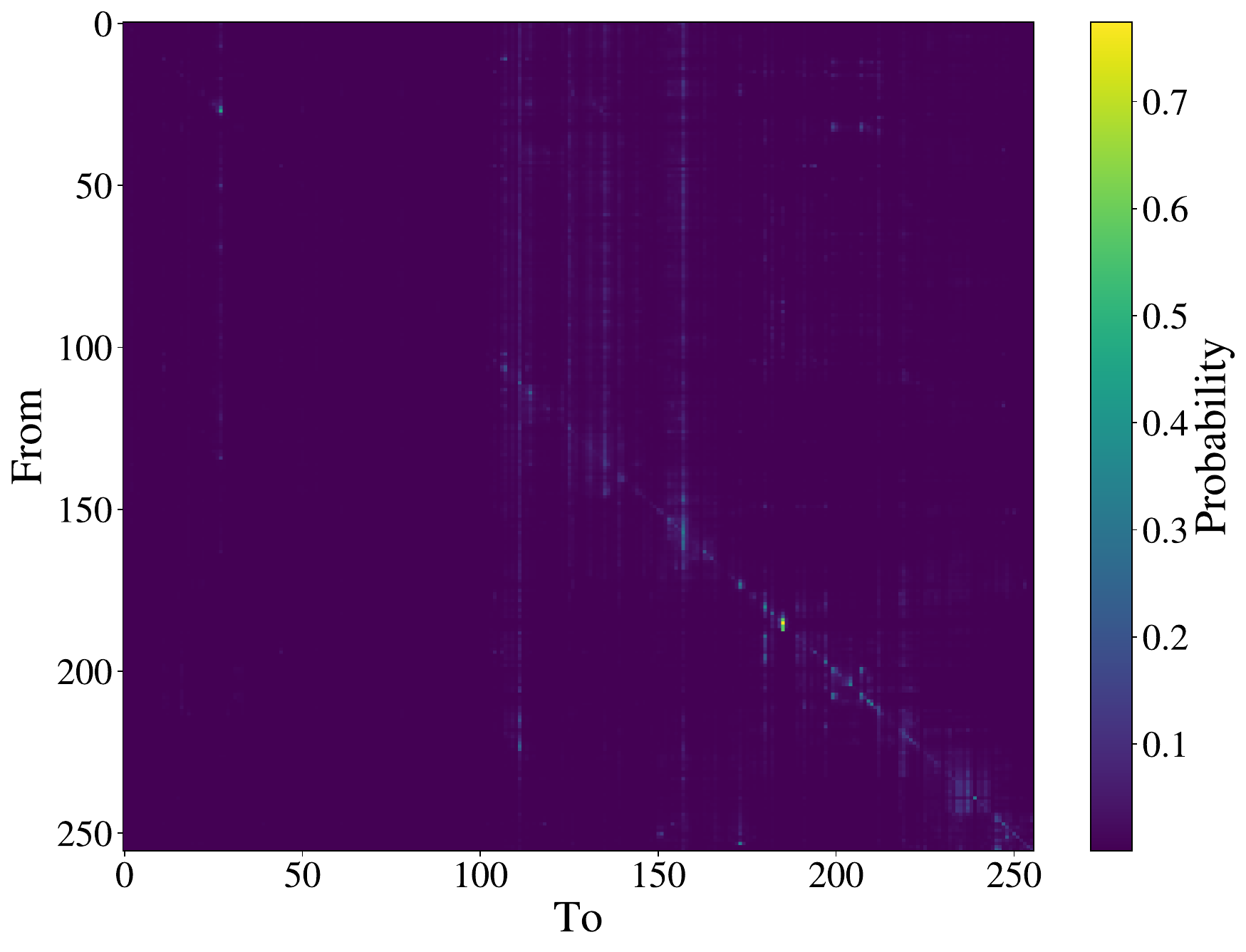}
        \includegraphics[width=0.23\linewidth]{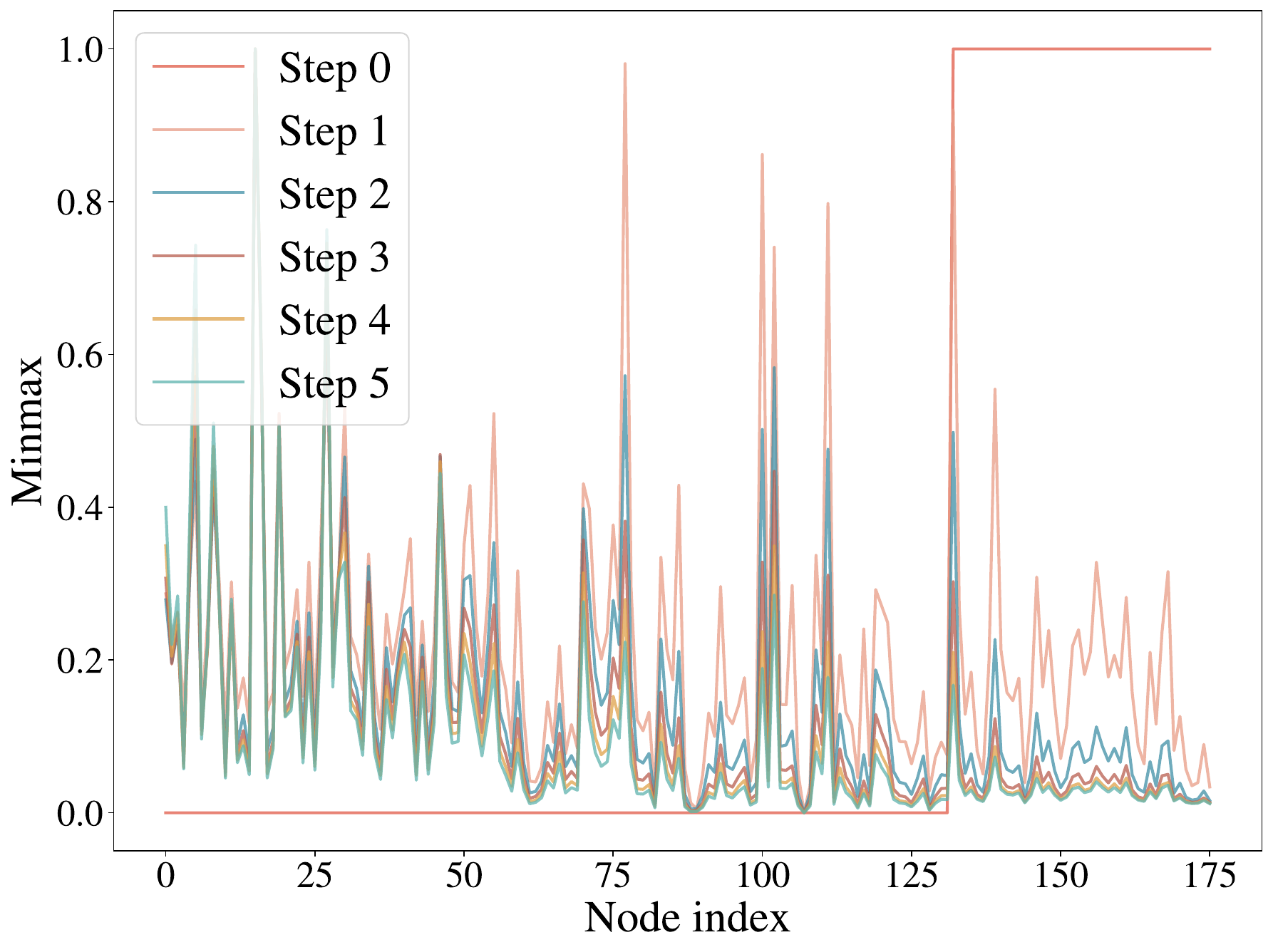}
        \includegraphics[width=0.23\linewidth]{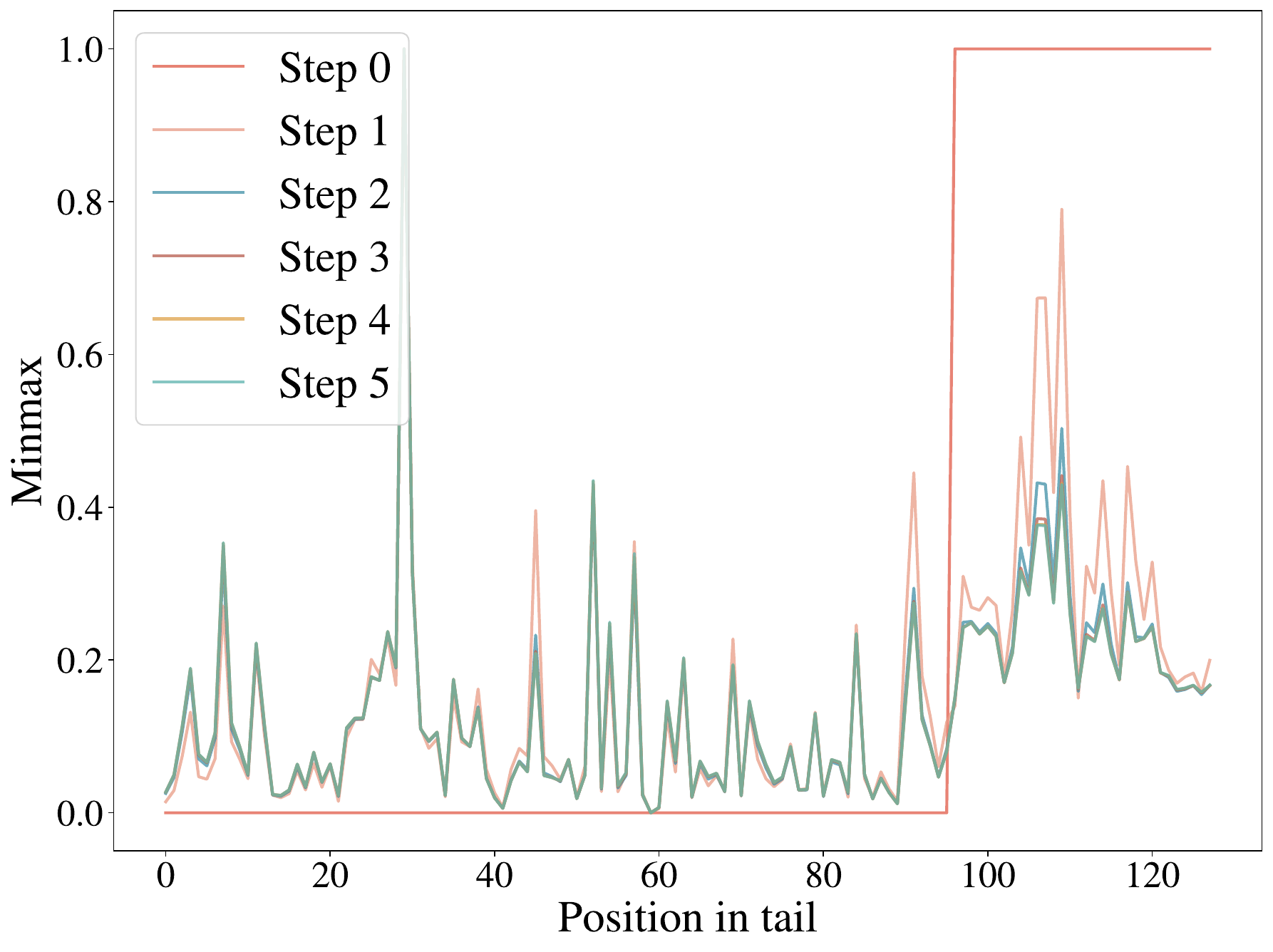}
    }

    % -------- Layer 30 --------
    \subfigure[30\textsuperscript{th} layer]{
        \includegraphics[width=0.23\linewidth]{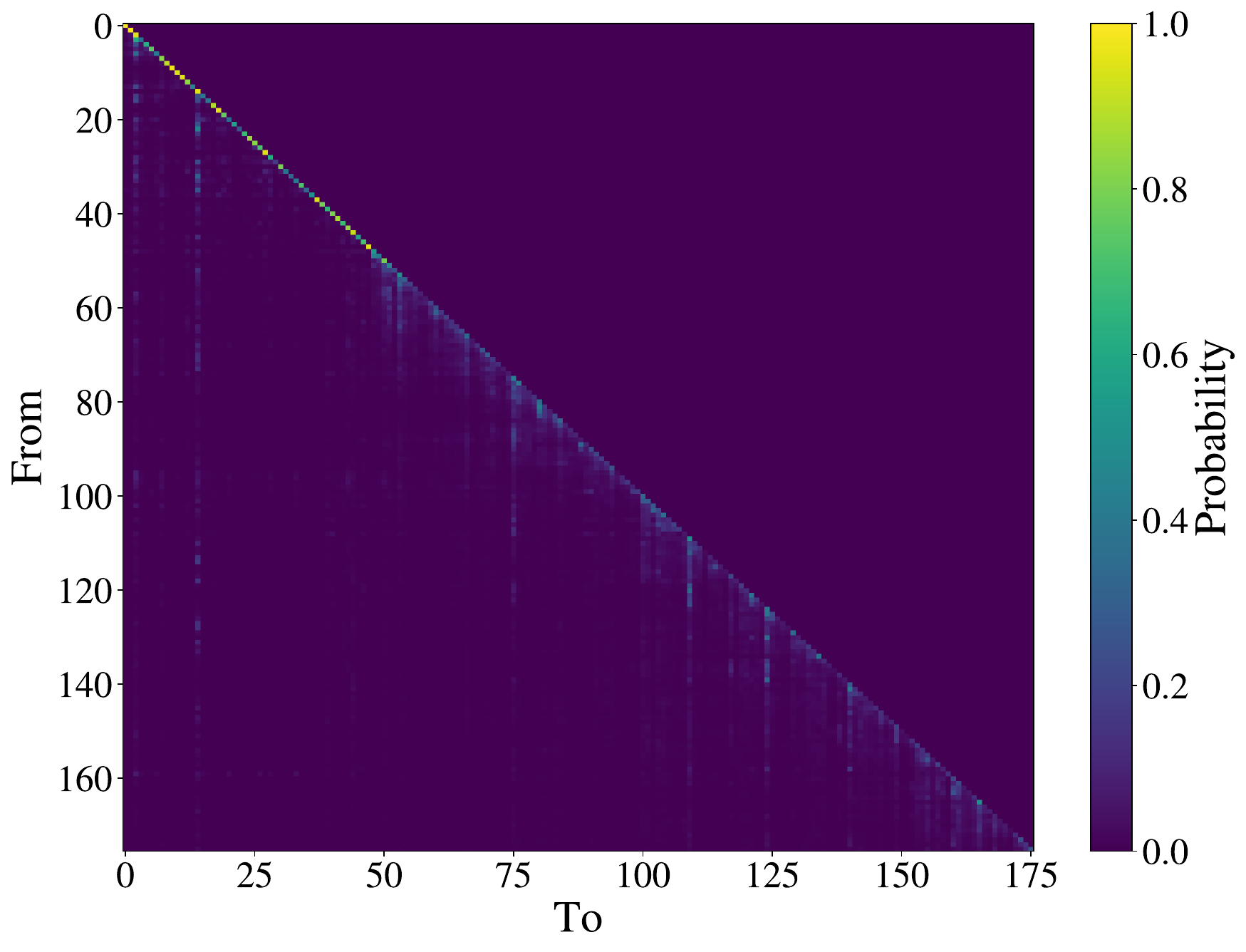}
        \includegraphics[width=0.23\linewidth]{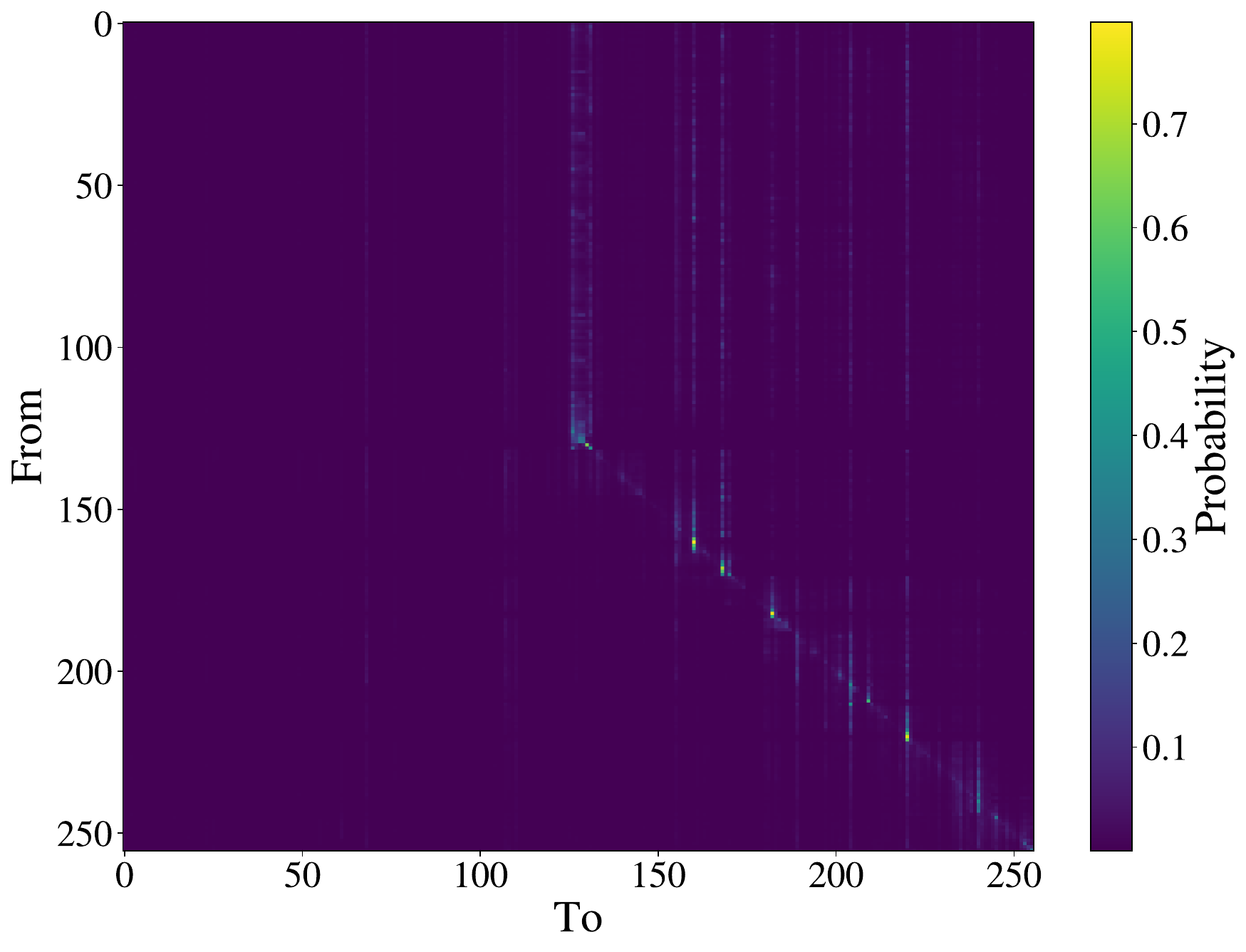}
        \includegraphics[width=0.23\linewidth]{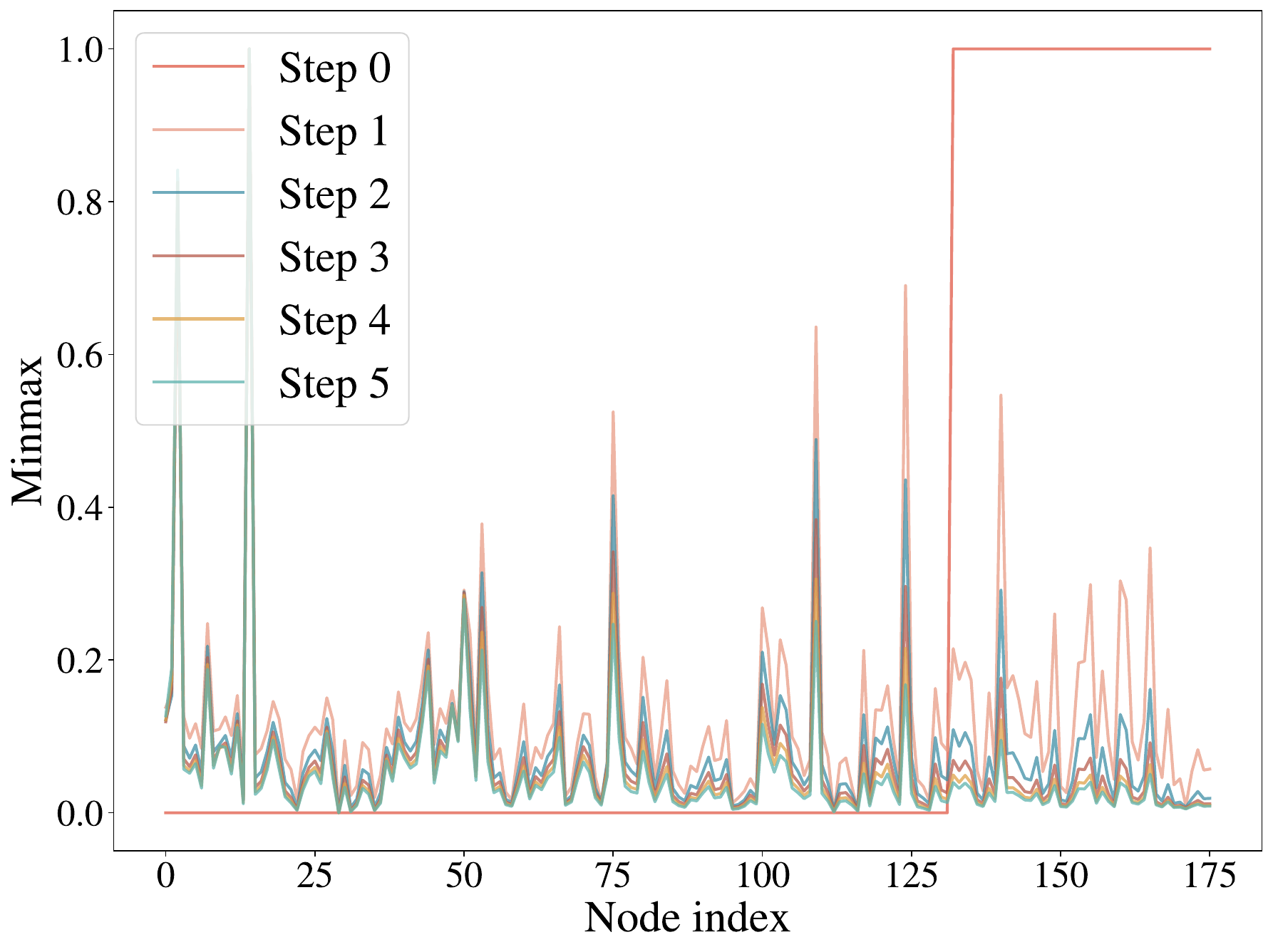}
        \includegraphics[width=0.23\linewidth]{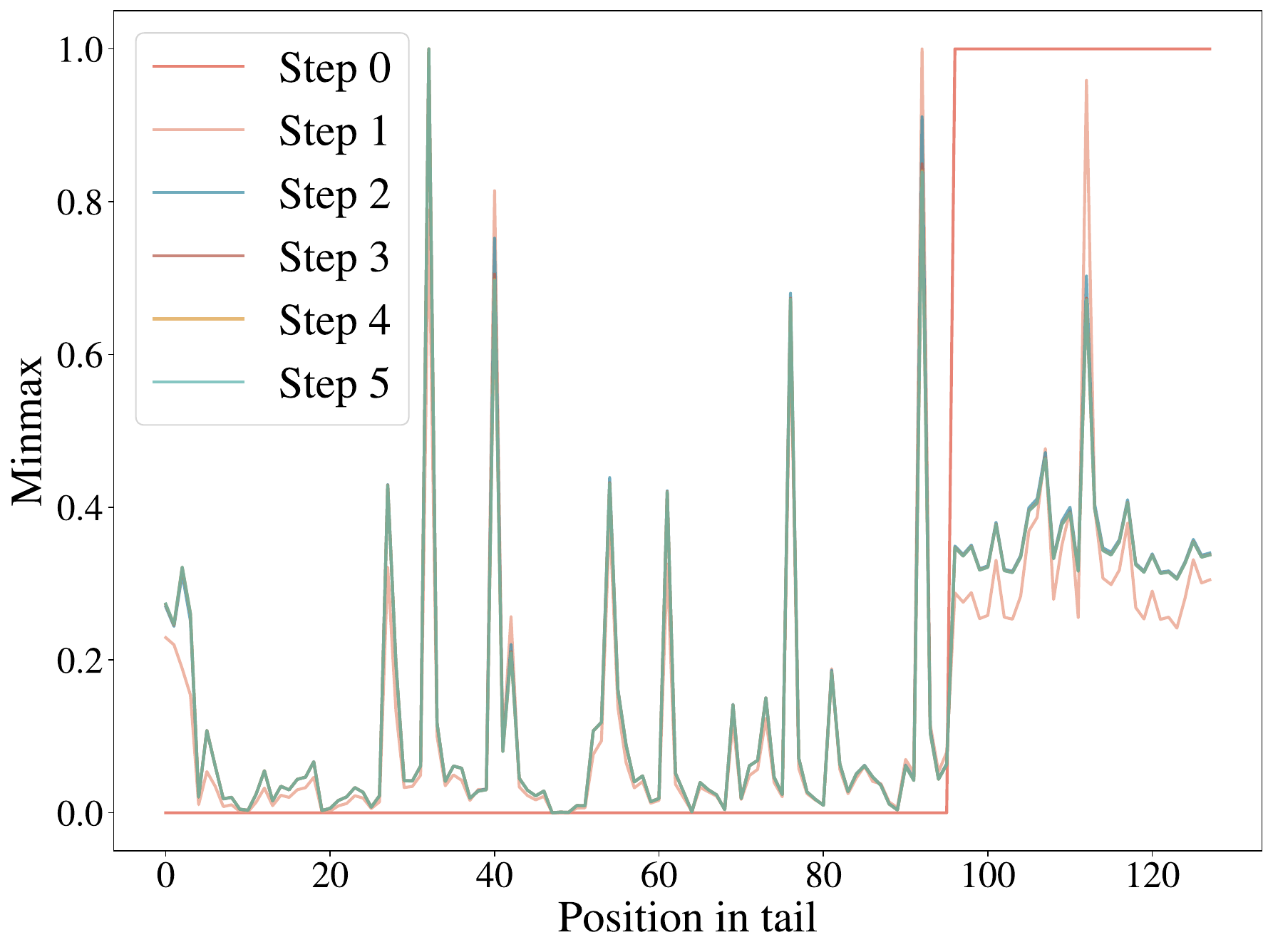}
    }

    \caption{
        \textbf{Behavior analysis of PRKV and PRKV-GLB across layers.}
        Each row corresponds to a different layer of Llama-3.1-8B.
        From left to right, the columns show:
        (1) transition matrix of PRKV,
        (2) transition matrix of PRKV-GLB,
        (3) score evolution of PRKV, and
        (4) score evolution of PRKV-GLB.
        Results are obtained with contexts from the PaulGrahamEssays dataset and shown for the 10\textsuperscript{th}, 20\textsuperscript{th}, and 30\textsuperscript{th} layers.
    }
    \label{fig:behavior_analysis_all}
\end{figure*}

\begin{figure*}[t]
    \centering
    \resizebox{0.9\textwidth}{!}{
    \begin{minipage}{\textwidth}
    % -------------------- 10th layer --------------------
    \subfigure[10\textsuperscript{th} Layer]{
        \includegraphics[width=0.3\linewidth]{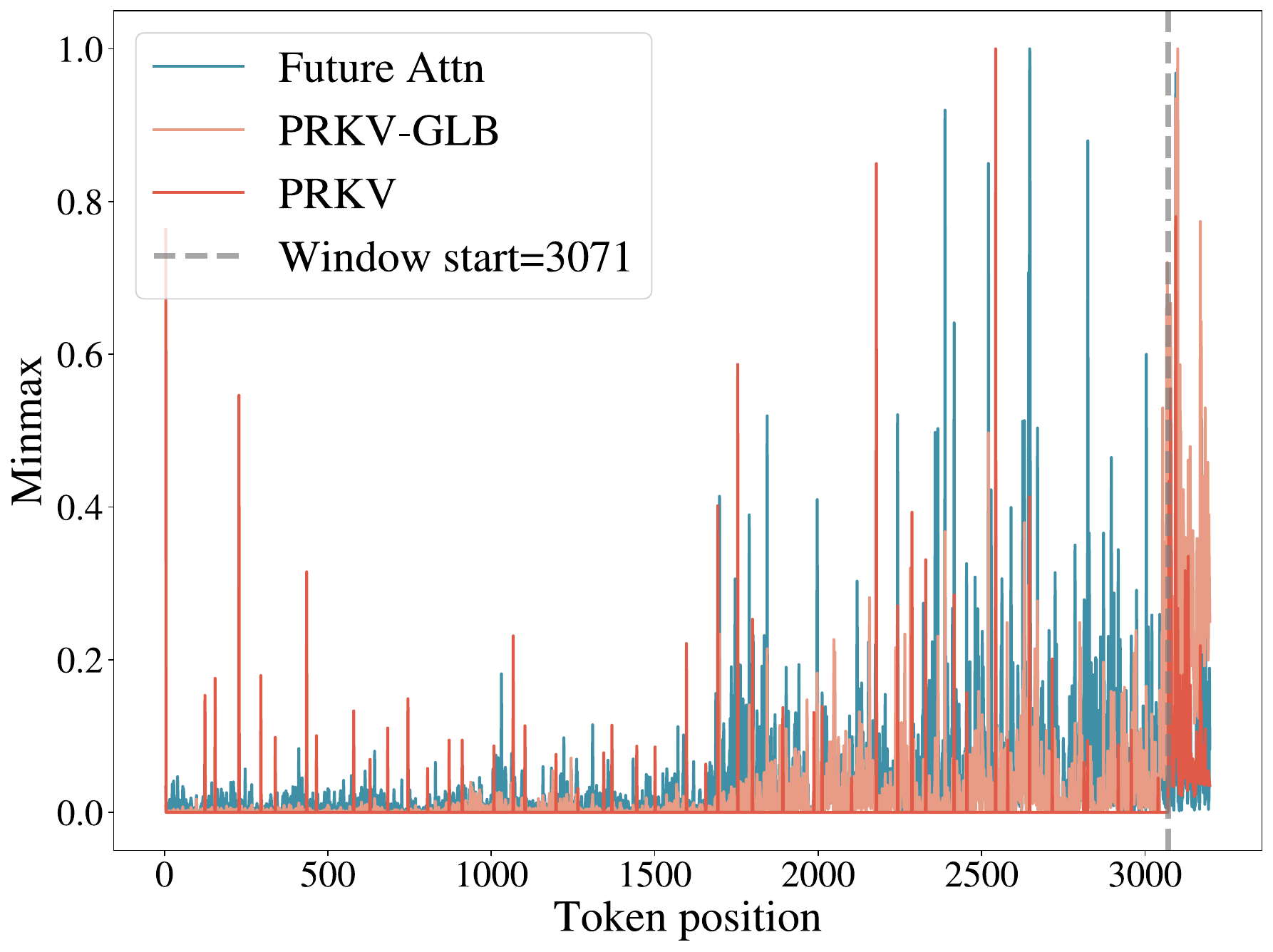}
        \includegraphics[width=0.3\linewidth]{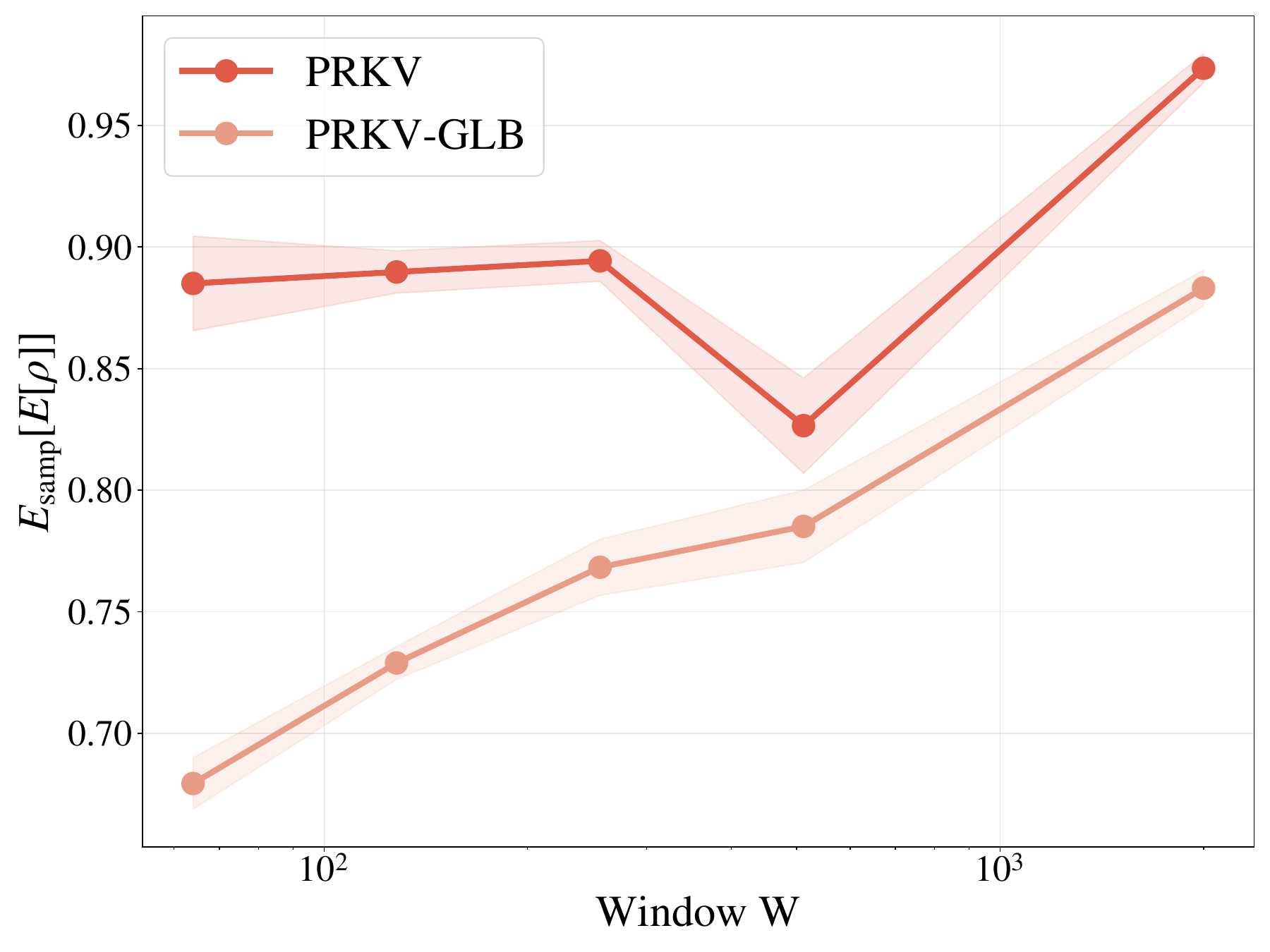}
        \includegraphics[width=0.3\linewidth]{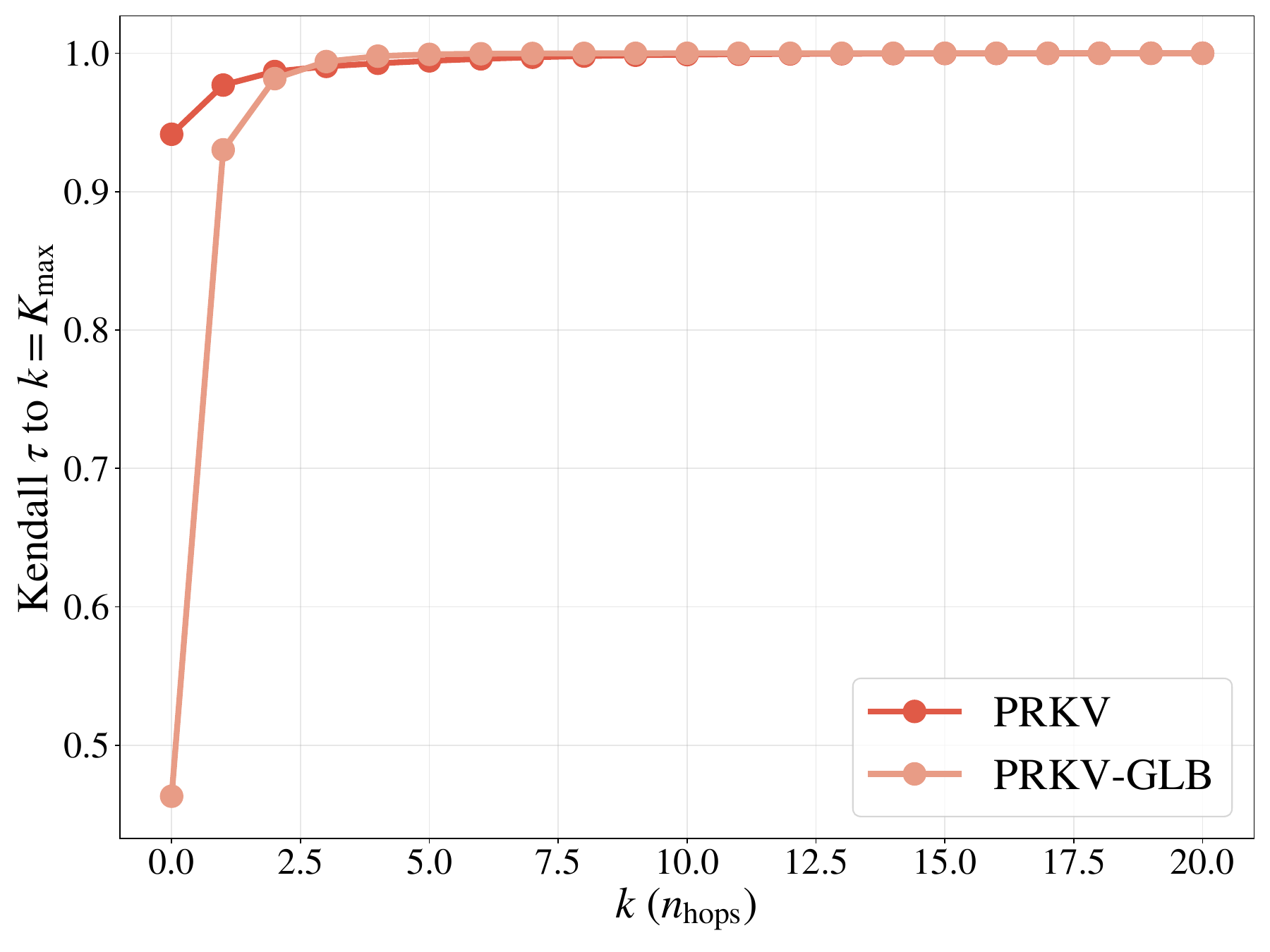}
    }

    % -------------------- 20th layer --------------------
    \subfigure[20\textsuperscript{th} Layer]{
        \includegraphics[width=0.3\linewidth]{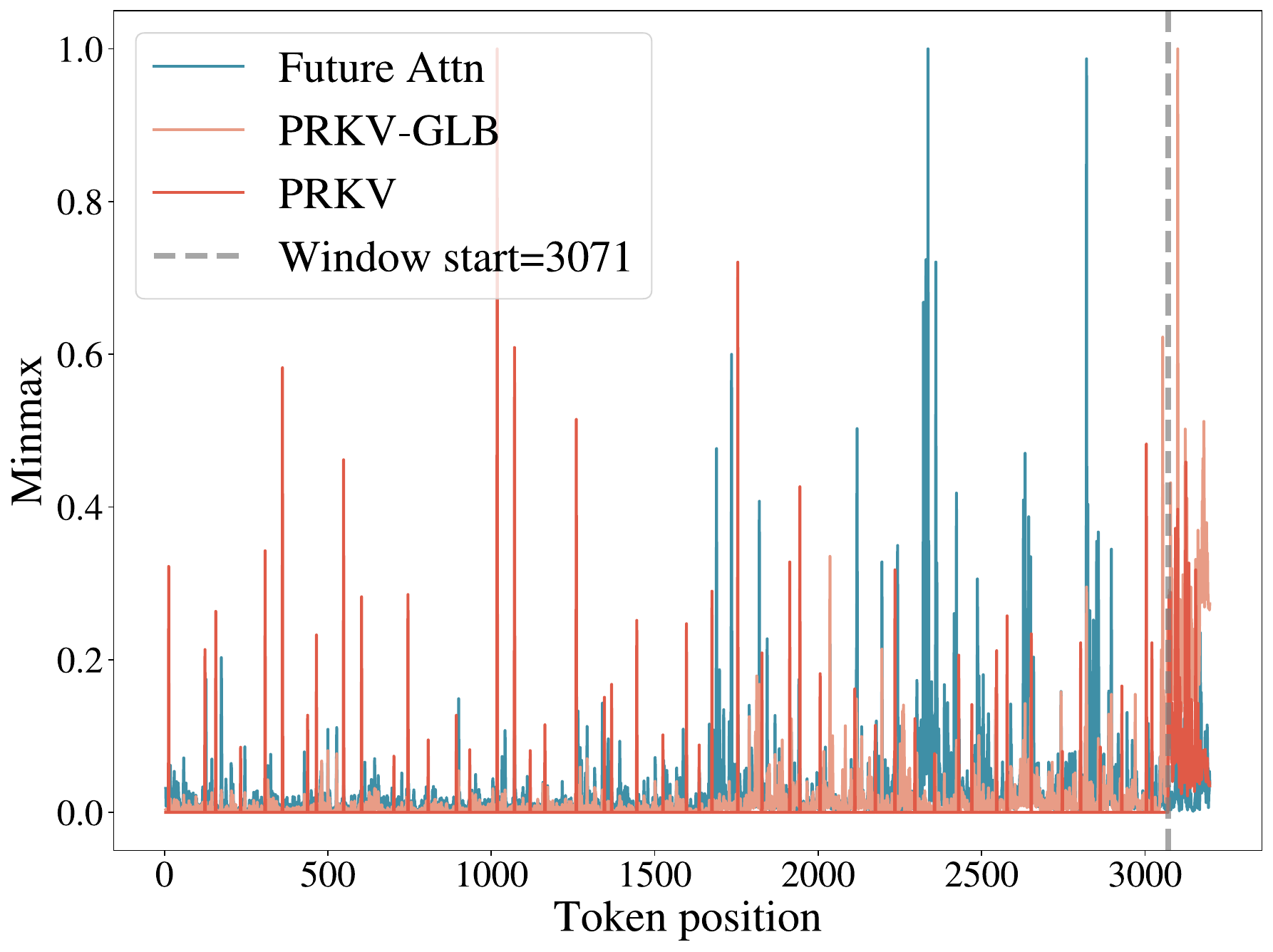}
        \includegraphics[width=0.3\linewidth]{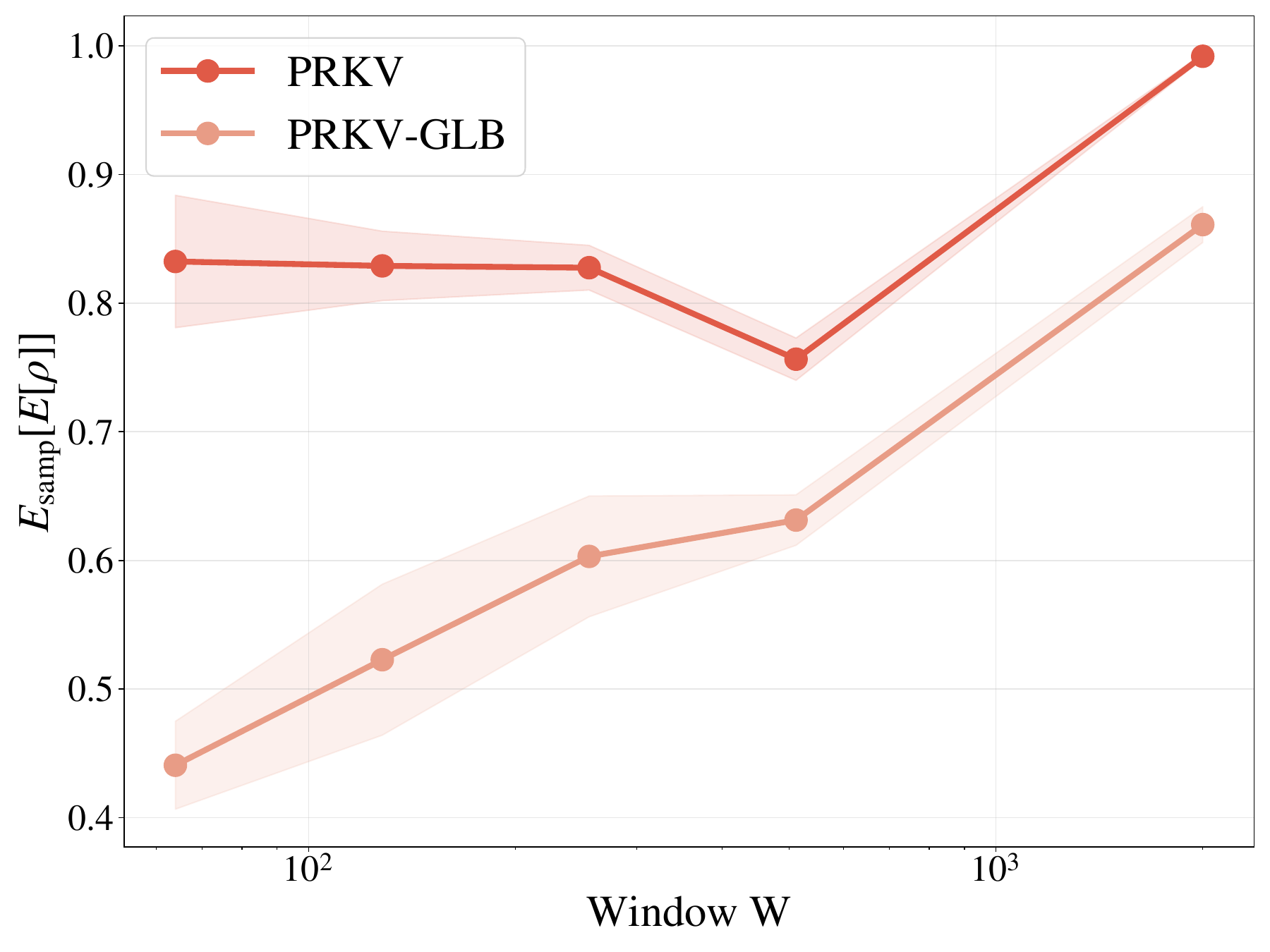}
        \includegraphics[width=0.3\linewidth]{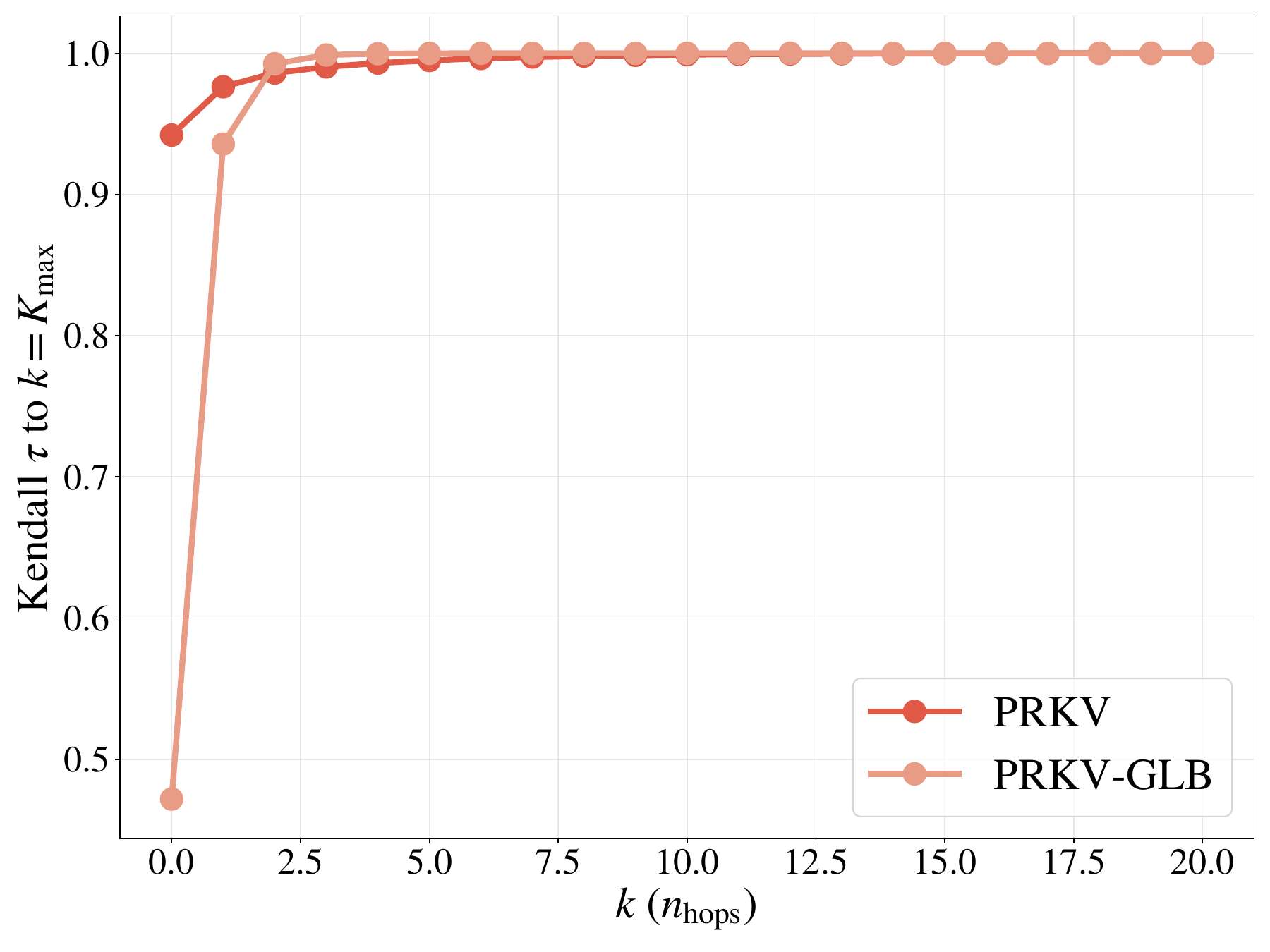}
    }
    % -------------------- 30th layer --------------------
    \subfigure[30\textsuperscript{th} Layer]{
        \includegraphics[width=0.3\linewidth]{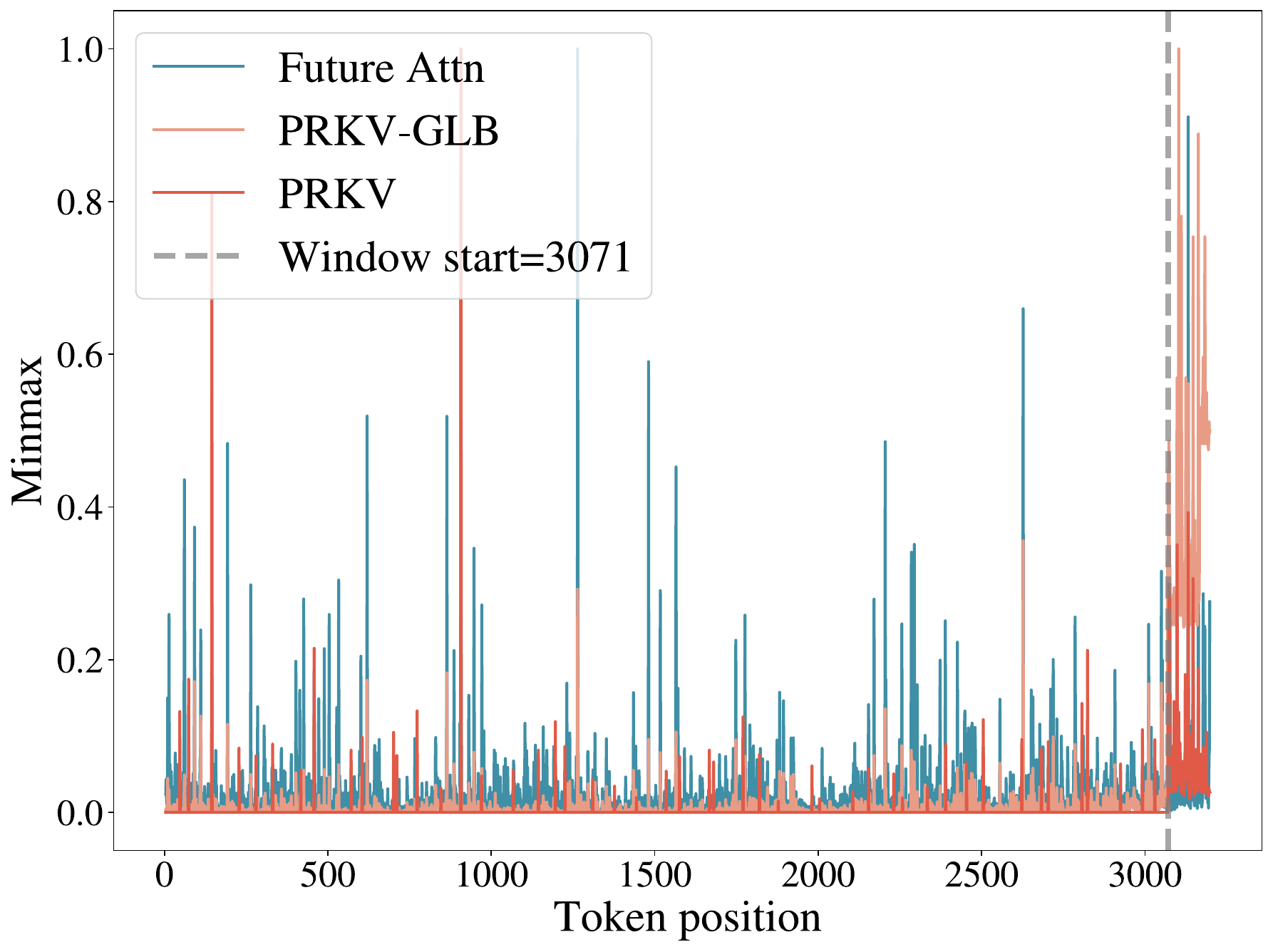}
        \includegraphics[width=0.3\linewidth]{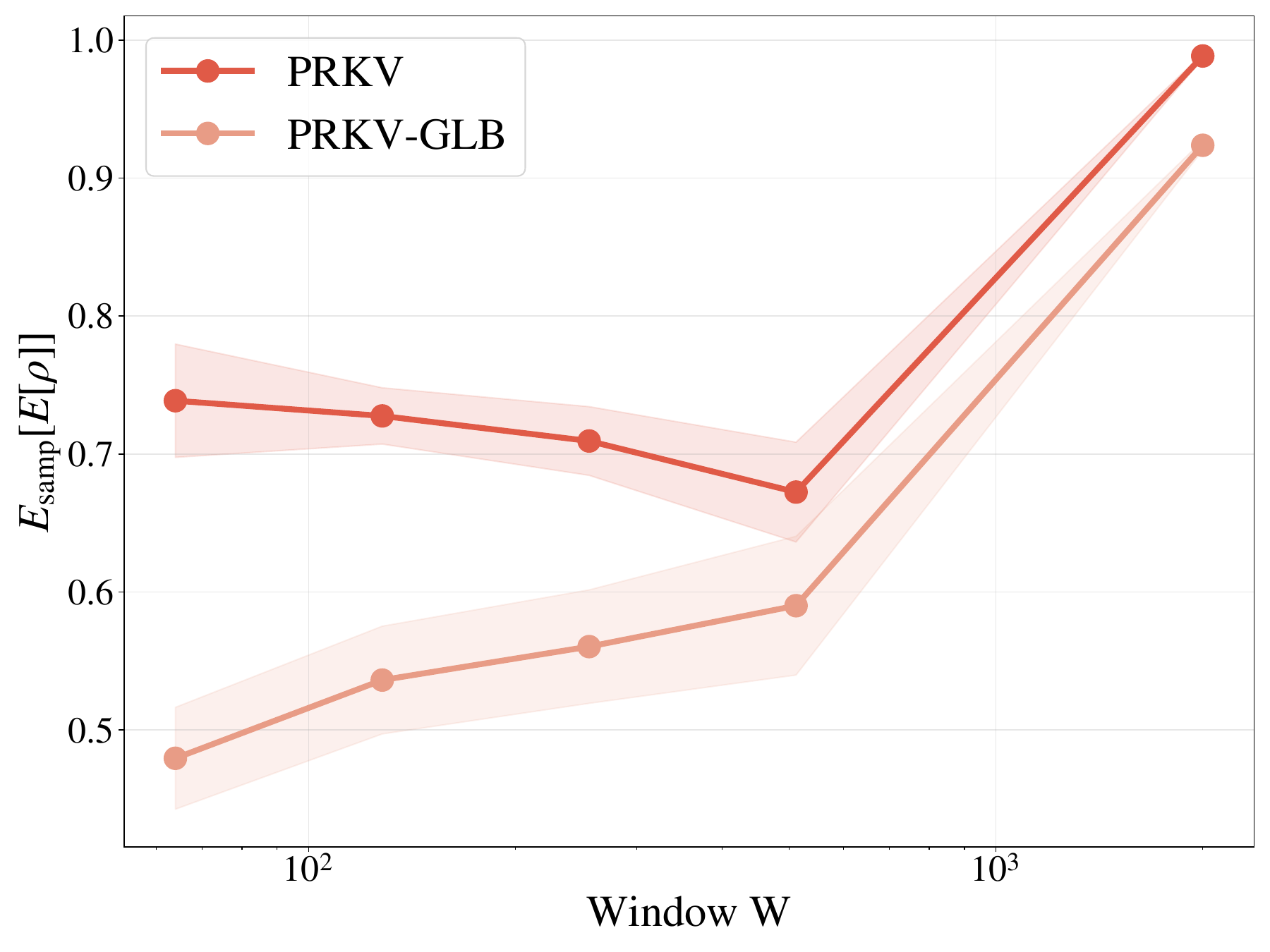}
        \includegraphics[width=0.3\linewidth]{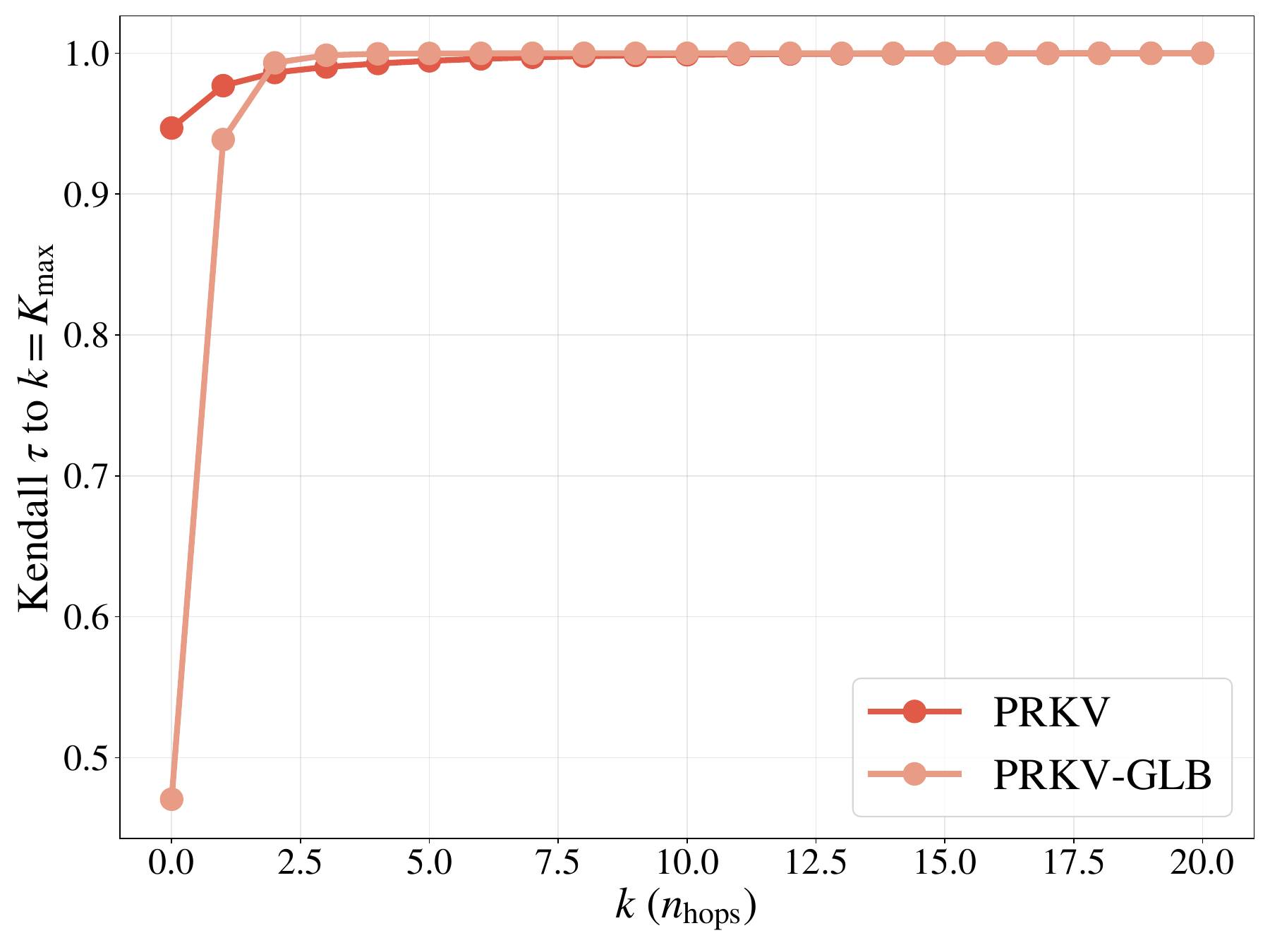}
    }
    \end{minipage}
    }
    \caption{
    \textbf{Layer-wise analysis of PRKV and PRKV-GLB.}
    (Left) Token-level scores produced by PRKV and PRKV-GLB compared to the future attention signal.
    (Center) Rank correlation between PRKV and $\pi^\ast$ as $W$ varies.
    (Right) Convergence of PPR scores to their converged values under finite iterations. Results are shown for the 10\textsuperscript{th}, 20\textsuperscript{th}, and 30\textsuperscript{th} layers of Llama-3.1-8B.
    }
    \vspace{-1em}
    \label{fig:ppr_convergence}
\end{figure*}

\begin{figure*}[h]
    \centering
    \subfigure[20\textsuperscript{th} Layer, 13\textsuperscript{th} Head]{
    \includegraphics[width=.9\linewidth]{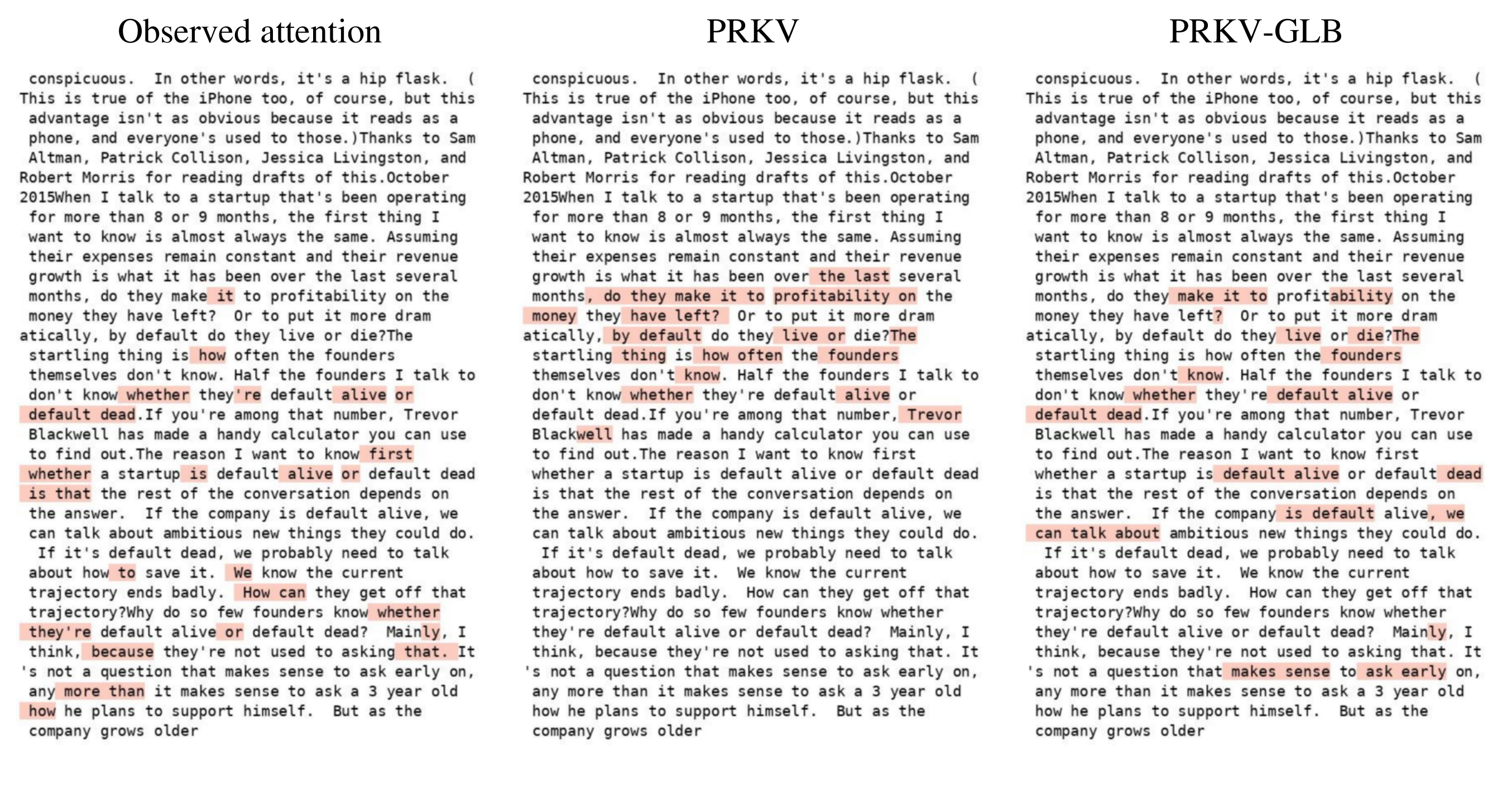}
    }
    \subfigure[30\textsuperscript{th} Layer, 30\textsuperscript{th} Head]{
    \includegraphics[width=.9\linewidth]{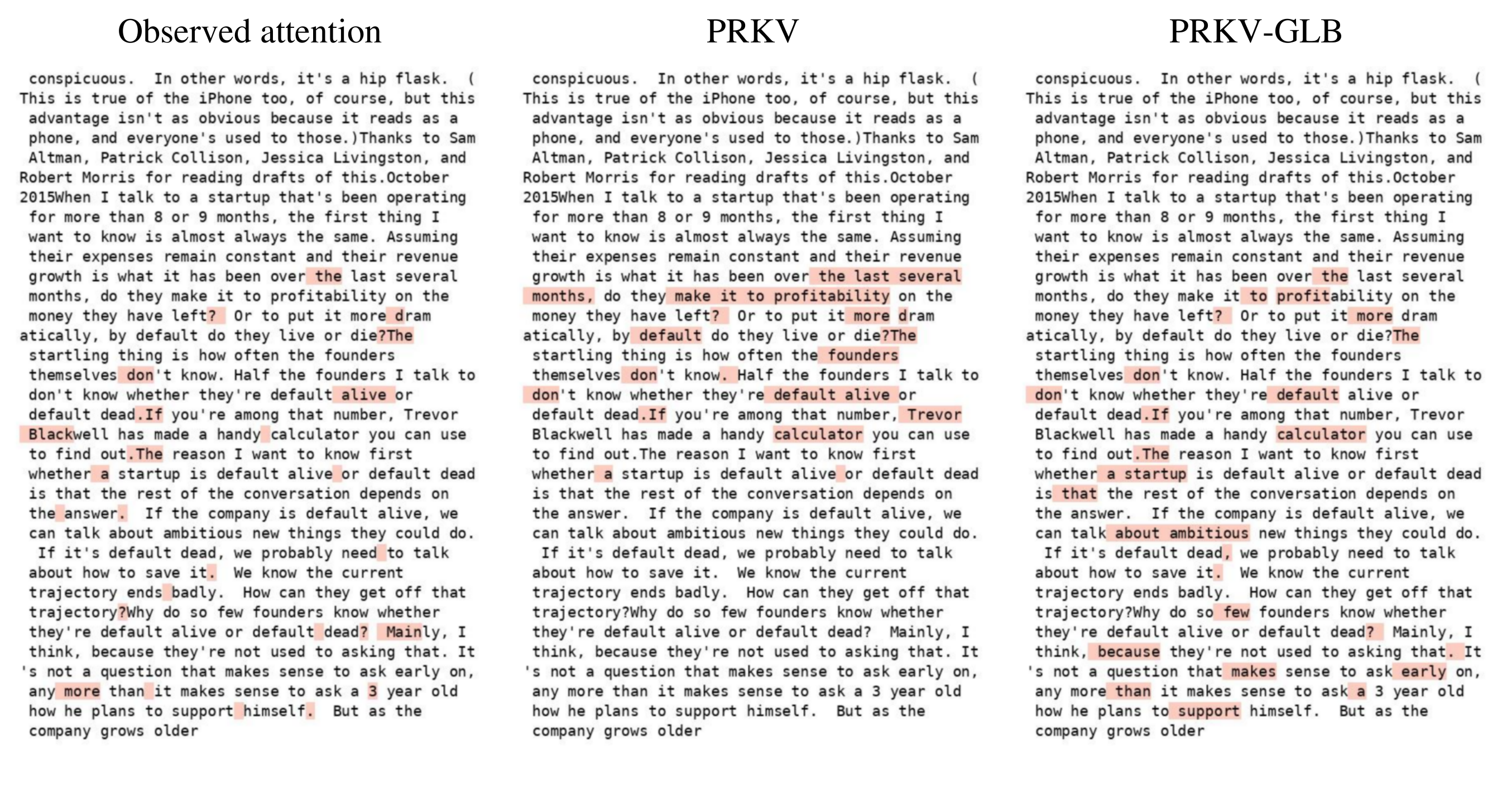}
    }
    \caption{\textbf{Layer-wise case study of token selection.}
    We report token selection result of (a) observed attention, (b) PRKV, and (c) PRKV-GLB. All measurements are conducted using the \texttt{bfloat16} data type on Llama-3.1-8B with PaulGrahamEssays dataset.}
    \label{fig:topk_selection_layer}
\end{figure*}

In this section, we provide layer-wise visualization of each analysis in Figures~\ref{fig:motivational_analysis_layers}--\ref{fig:topk_selection_layer}.

\clearpage

%%%%%%%%%%%%%%%%%%%%%%%%%%%%%%%%%%%%%%%%%%%%%%%%%%%%%%%%%%%%

\end{document}